%% file: top.tex
\documentclass[10pt,journal,letterpaper,compsoc]{IEEEtran}

\usepackage{times}
\usepackage{epsfig}
\usepackage{graphicx}
\usepackage{amsmath}
\usepackage{amssymb}
\usepackage{subfigure}
\usepackage{rotating}
\usepackage{multirow}
\usepackage{ragged2e}
\usepackage{cancel}
\usepackage[ruled,lined,linesnumbered]{algorithm2e}
\usepackage{bbm}
\usepackage{textcomp}

\usepackage[pagebackref=false,breaklinks=true,colorlinks,bookmarks=false]{hyperref}
\usepackage[font=small,labelfont=bf]{caption}

\input{defs}

\begin{document}

\title{Template-based Monocular 3D Shape Recovery using Laplacian Meshes}
\author{Dat Tien Ngo, Jonas \"Ostlund, and~Pascal~Fua,~\IEEEmembership{Fellow,~IEEE}%
\IEEEcompsocitemizethanks{
\IEEEcompsocthanksitem The authors are with the \'Ecole Polytechnique F\'ed\'erale de Lausanne, CH-1015 Lausanne, 
Switzerland.\protect\\
\IEEEcompsocthanksitem E-mail: \{firstname.lastname\}@epfl.ch\protect\\
\IEEEcompsocthanksitem This work was supported in part by the Swiss National Science Foundation
}}
\markboth{IEEE Transactions On Pattern Analysis And Machine Intelligence}%
{Ngo~\MakeLowercase{\textit{et~al.}}: Template-based Monocular 3D Shape Recovery using Laplacian Meshes}

\IEEEcompsoctitleabstractindextext
{
 \input{abstract.tex}

}

\maketitle

\IEEEdisplaynotcompsoctitleabstractindextext
\IEEEpeerreviewmaketitle

\input{intro.tex}
\input{related.tex}

\input{problem.tex}
\input{laplacian.tex}
\input{results.tex}
\input{conc.tex}

\bibliographystyle{IEEEtran}
\bibliography{string,vision,graphics,misc,optim}

\input{bio.tex}

\clearpage
\input{appendix}

\checknbdrafts
\end{document}

%% file: defs.tex
\newcounter{nbdrafts}
\makeatletter
\newcommand{\checknbdrafts}{
\ifnum \thenbdrafts > 0
\@latex@warning@no@line{**********************************************************************}
\@latex@warning@no@line{* The document contains \thenbdrafts \space draft note(s)}
\@latex@warning@no@line{**********************************************************************}
\fi}

\makeatother

\newcommand{\kronecker}{\raisebox{1pt}{\ensuremath{\:\otimes\:}}}

\def\x1new{\mathbf{{\bx}_1^\prime}}

\def\y1new{\mathbf{{\by}_1^\prime}}

\newcommand{\trademark}{$^{\text{tm}}$}

\newcommand{\vt}[1]{{\bf v}_{#1}}

\newcommand{\norm}[1]{\left\| #1 \right\|}
\newcommand{\sqnorm}[1]{\norm{#1}^2}
\newcommand{\any}{^{\prime}}
\newcommand{\sref}{_{\textrm{ref}}}

\newcommand{\inv}{^{-1}}
\newcommand{\transpose}{^T}
\newcommand{\matrixbrackets}[1]{\left[ #1 \right]}

\newcommand{\realnumbers}{\mathbb{R}}

\newcommand{\vertexcount}{N_v}
\newcommand{\ctrlcount}{N_c}

\newcommand{\bA}{\mathbf{A}}

\newcommand{\bc}{\mathbf{c}}

\newcommand{\bH}{\mathbf{H}}

\newcommand{\bI}{\mathbf{I}}

\newcommand{\bK}{\mathbf{K}}

\newcommand{\bM}{\mathbf{M}}

\newcommand{\bn}{\mathbf{n}}

\newcommand{\bP}{\mathbf{P}}

\newcommand{\bp}{\mathbf{p}}

\newcommand{\bR}{\mathbf{R}}

\newcommand{\bt}{\mathbf{t}}

\newcommand{\bw}{\mathbf{w}}
\newcommand{\bx}{\mathbf{x}}

\newcommand{\by}{\mathbf{y}}

\newcommand{\vtr}[1]{{\vt{#1}}\sref}
\newcommand{\cst}{_\textrm{c}}

\newcommand{\pctab}{\hspace{0.2in}}

\newcommand{\comment}[1]{}

\newcommand{\freeparam}{\boldsymbol{\lambda}}

%% file: abstract.tex

\begin{abstract}

We show that by extending the Laplacian formalism, which was first introduced in
the Graphics  community to regularize  3D meshes, we  can turn the  monocular 3D
shape  reconstruction  of a  deformable  surface  given  correspondences with  a
reference image  into a much better-posed  problem. This allows us  to quickly and
reliably eliminate  outliers by simply  solving a linear least  squares problem.
This yields  an initial 3D shape  estimate, which is not necessarily accurate, but
whose 2D  projections are.  The initial  shape is then refined  by a constrained
optimization problem to output the final surface reconstruction.

Our  approach   allows  us   to  reduce  the   dimensionality  of   the  surface
reconstruction problem without sacrificing accuracy, thus allowing for real-time
implementations.

\end{abstract}

%% file: intro.tex

\section{Introduction}
\label{sec:intro}

Shape  recovery  of  deformable   surfaces  from  single  images  is  inherently
ambiguous,  given  that  many  different  configurations can  produce  the  same
projection.  In particular,  this is true of template-based  methods, in which a
{\it reference} image  of the surface in a known  configuration is available and
point correspondences between  this reference image and an  {\it input} image in
which the shape is to be recovered are given.

One approach  is to compute a  2D warp between the  images and infer a  3D shape
from it, which  can be done pointwise and in  closed form~\cite{Bartoli12b}. The
quality of the  recovered 3D shape then  depends on the quality of  the 2D warp,
which does not necessarily account for  the 3D nature of the deformations
  and the constraints it imposes.  This problem can be avoided by computing the
3D  shape  directly  from  the  correspondences,  which  amounts  to  solving  a
degenerate linear system  and requires either reducing the number  of degrees of
freedom or imposing additional constraints~\cite{Salzmann10b}.  The first can be
achieved          by          various          dimensionality          reduction
techniques~\cite{Blanz99,Salzmann11a} while  the second often  involves assuming
the  surface to  be either  developable~\cite{Gumerov04,Liang05,Perriollat12} or
inextensible~\cite{Ecker08,Shen09,Brunet10,Salzmann11a}.   These two  approaches
are  sometimes  combined and  augmented  by  introducing additional  sources  of
information  such  as  shading or  textural  clues~\cite{White06b,Moreno09b}  or
physics-based  constraints~\cite{Malti13}.   The  resulting  algorithms  usually
require solving a fairly large optimization problem and, even though it is often
well                     behaved                     or                     even
convex~\cite{Brunet10,Ecker08,Moreno10,Perriollat11,Salzmann11a},   it   remains
computationally demanding.  Closed-form approaches  to directly computing the 3D
shape from  the correspondences have  been proposed~\cite{Salzmann08b,Moreno09b}
but they also involve  solving very large systems of equations  and making more
restrictive  assumptions  than  the  optimization-based ones,  which  can  lower
performance~\cite{Salzmann11a}.

Here,  we  show  that,  by  extending  the  Laplacian  formalism  first
  introduced  in the  Graphics Community~\cite{Sorkine04,Summer04,Botsch06}  and
  introducing a novel regularization term designed to minimize curvature changes
  from  the  potentially non-planar  reference  shape,  we  can turn  the  large
  degenerate linear system mentioned above into a non-degenerate one. We further
  show that the resulting least-squares problem can be reduced to a much smaller
  one by expressing all vertex coordinates  as linear combinations of those of a
  small number  of control vertices, which  considerably increases computational
  efficiency at no loss in reconstruction accuracy.

In other words,  instead of performing a minimization  involving many degrees of
freedom or solving a large system of  equations, we end up simply solving a very
compact linear system.   This yields an initial 3D shape  estimate, which is not
necessarily  accurate, but  whose  2D  projections are.   In  practice, this  is
extremely useful  to quickly  and reliably eliminate  erroneous correspondences.
We     can    then     achieve    better     accuracy    than     the    methods
of~\cite{Salzmann11a,Brunet10,Bartoli12b} on both planar and non-planar surfaces
by  enforcing simple inextensibility  constraints.  This  is done  by optimizing
over a small number of variables, thereby lowering the computational complexity and
allowing for a real-time implementation.


In short, our contribution is a  novel approach to regularizing and reducing the
dimensionality  of the  surface  reconstruction problem.   It  does not  require
either an  estimate of the  rigid transformation  with respect to  the reference
shape  or  access  to  training  data  or  material  properties,  which  may  be
unavailable   or  unknown.    Furthermore,  this   is  achieved   without
  sacrificing accuracy and can handle arbitrarily large deformations and generic
  objective functions,  which is beyond  what earlier Laplacian  formalisms were
  designed to do. Each one of these  features can be found in isolation in other
  approaches but ours brings them all  together in a unified framework.  It was
first introduced  in a conference  paper~\cite{Ostlund12} and is  validated more
thoroughly here.

In the remainder  of this paper, we first review  existing approaches and remind
the reader  how the problem  can be  formulated as one  of solving a  linear but
degenerate system  of equations, as  we did in  earlier work~\cite{Salzmann10b}.
We then introduce our extended  Laplacian formalism, which lets us transform the
degenerate  linear system into  a non-degenerate  one and  if necessary  make it
smaller for better computational efficiency. Finally, we present our results and
compare                them               against               state-of-the-art
methods~\cite{Salzmann11a,Brunet10,Bartoli12b}.


%% file: related.tex

\section{Related Work}
\label{sec:related}

Reconstructing the 3D shape of a  non-rigid surface from a single input image is
a severely under-constrained problem, even when a reference image of the surface
in a  different but  known configuration  is available. This  is the  problem we
address here, as opposed to recovering the shape from sequences as in many recent
monocular Non-Rigid Structure from Motion methods such as~\cite{Fayad10,Garg13}.

When point  correspondences can be  established between the reference  and input
images, one can compute  a 2D warp between the images and infer  a 3D shape from
it,      which      can      be      done      in      closed      form      and
pointwise~\cite{Bartoli12b}. However, the accuracy of the recovered shape
  can be  affected by the  fact that the  2D warp may  not take into  account he
  constraints  that   arise  from  the  3D   nature  of  the  surface   and  its
  deformations.

An alternative is  therefore to go directly from correspondences  to 3D shape by
solving an ill-conditioned  linear-system~\cite{Salzmann10b}, which requires the
introduction of additional constraints to  make it well-posed.  The most popular
ones involve preserving  Euclidean or Geodesic distances as  the surface deforms
and    are    enforced    either    by    solving    a    convex    optimization
problem~\cite{Brunet10,Ecker08,Salzmann07b,Shen09,Moreno10,Perriollat11,Salzmann11a,Alcantarilla12}
or     by      solving     in     closed     form      sets     of     quadratic
equations~\cite{Salzmann08b,Moreno09b}.   The   latter  is  typically   done  by
linearization,  which  results in  very  large systems  and  is  no faster  than
minimizing a  convex objective function, as is  done in~\cite{Salzmann11a} which
has been  shown to be an  excellent representative of this  class of techniques.
The  results  can  then   be  improved  by  imposing  appropriate  physics-based
constraints~\cite{Malti13} via  non-linear minimization, but  that requires some
knowledge of the physical properties that may not be available.

The complexity  of the problem can  be reduced using a  dimensionality reduction
technique  such  as  Principal  Component Analysis  (PCA)  to  create  morphable
models~\cite{Cootes98,Blanz99,Dimitrijevic04},                             modal
analysis~\cite{Moreno09b,Moreno10},        Free         Form        Deformations
(FFDs)~\cite{Brunet10}, or  3D warps~\cite{DelBue11}.   One drawback of  PCA and
modal analysis is that it requires  either training data or sufficient knowledge
of the surface properties to compute a stiffness matrix, neither of which may be
forthcoming.  Another is that the  modal deformations are expressed with respect
to  a reference  shape,  which  must be  correctly  positioned.   This makes  it
necessary to introduce  additional rotation and translation  parameters into the
computation. This complicates  the computations because the  rotations cannot be
treated   as   being   linear   unless   they   are   very   small.    The   FFD
approach~\cite{Brunet10}  avoids these  difficulties and,  like ours,  relies on
parameterizing the  surface in  terms of control  points.  However,  its
  affine-invariant curvature-based quadratic regularization  term is designed to
  preserve local structures and planarity.  For non-planar surfaces, it tends to
  flatten  the  surface in  areas  with  only  limited image  information.   By
contrast, our approach naturally handles non-planar reference surfaces, tends to
preserve curvature,  and its  control vertices can  be arbitrarily  placed. This
makes it closer to earlier ones to fitting 3D surfaces to point clouds that also
allow arbitrary  placement of the  control points  by using Dirichlet  Free Form
Deformations~\cite{Ilic02}    or,     more    recently,    sets     of    affine
transforms~\cite{Jacobson11}.    These   approaches,   however,   also   require
regularization of the control  points if they are to be used  to fit surfaces to
noisy data.

In  short,   none  of  these   dimensionality  reduction  methods   allows  both
orientation-invariant  and  curvature-preserving  regularization.   To  do  both
simultaneously, we took  our inspiration from the  Laplacian formalism presented
in~\cite{Sorkine04} and  the rotation invariant  formulation of~\cite{Summer04},
which like  ours involves introducing  virtual vertices.  In both  these papers,
the  mesh Laplacian  is  used to  define  a regularization  term  that tends  to
preserve  the  shape  of  a non-planar  surface. 

In~\cite{Botsch06},  it  is  shown  that explicitly  adding  the  virtual
  vertices is not  necessary to achieve similar or even  better results.  In our
  work, we  go one  step further  by not only  introducing a  rotation invariant
  regularization  term  expressed   as  a  linear  combination   of  the  vertex
  coordinates but showing that these coordinates  can themselves be written as a
  linear function  of those  of a  subset of  control vertices  while preserving
  rotation invariance.  Furthermore, in~\cite{Summer04}, the regularization term
  involves  minimizing the  magnitude of  the local  deformations, which  favors
  small deformations. By contrast, our approach only penalizes curvature changes
  and can accommodate arbitrarily large deformations.


%% file: problem.tex

\section{Linear Problem Formulation}
\label{sec:problem}

As shown in~\cite{Salzmann10b}, but for the sake of completeness we show again
that given point correspondences between a reference image in  which the 3D 
shape is known and an  input image, recovering  the new shape in this image
amounts to solving a linear system. 

Let $\vt{i}$ be  the 3D coordinates  of the $i^{\rm th}$ vertex  of the
$N_v$-vertex triangulated mesh representing the surface, $\bK$ be the intrinsic
camera matrix, $\bp$ be a 3D point lying on facet $f$. One can represent $\bp$
in the barycentric coordinates of $f$:
$\bp=\sum _{i=1}^{3}b_{i}\vt{f,i}$, where $\{\vt{f,i}\}_{i = 1,2,3}$ are the
three vertices of $f$. The fact that $\bp$ projects to the 2D image point
$(u,v)$ can be expressed by
\begin{equation}
  \bK(b_1 \vt{f,1} + b_2 \vt{f,2} + b_3 \vt{f,3}) = k \begin{bmatrix} u\\ v\\1 \end{bmatrix} \; ,  
  \label{eq:projection}  
\end{equation}
where $k$ is the homogeneous component. Since $k$ can be expressed in terms of
the vertex coordinates using the last row of the above equation, we can rewrite
Eq.~\ref{eq:projection} to be
\begin{equation}
[b_1 \bH \quad b_2 \bH \quad b_3 \bH] \begin{bmatrix} \vt{f,1}\\ \vt{f,2} \\ \vt{f,3} \end{bmatrix} = 0 \; ,
\end{equation}
with
\begin{align}
 \bH = \bK_{2\times3} - \begin{bmatrix} u \\ {v} \end{bmatrix} \bK_3 \; ,
\end{align}
where $\bK_{2\times3}$ are the first two rows, and $\bK_3$ is the third one of
$\bK$. Given $n$ correspondences between 3D reference surface locations and 2D
image points, we obtain $2n$ linear equations which can be jointly expressed by
a linear system
\begin{equation}
{\bf M}  \bx =  {\bf 0}\;,  \mbox{where } \bx  = \left[  \begin{array}{c} \vt{1}
    \\ \vdots \\ \vt{N_v} \end{array} \right] \;.
\label{eq:LinProj}
\end{equation}
and ${\bf M}$ is a matrix obtained by concatenating the $[b_1 \bH ~ b_2 \bH
~ b_3 \bH]$ matrices. In practice, the sum of squares  of successive  elements
of  the vector ${\bf  M} \bx$  are squared distances  in   the  direction 
parallel   to  the  image-plane   between  the line-of-sight defined by a
feature point and its corresponding 3D point on the surface.  A  solution of
this system  defines a surface such  that 3D feature points  that project to 
specific locations  in the  reference image  project at corresponding  locations
 in  the  input  image.  Solving  this  system  in  the least-squares sense
therefore  yields surfaces, up to a  scale factor, for which the overall
reprojection error is small.

The difficulty comes  from the fact that, for all  practical purposes, ${\bf M}$
is  rank  deficient as shown in Fig.~\ref{fig:singular_values}(a), with  at 
least one third  of  its  singular values  being extremely small  with respect
to the other  two thirds even when  there are many correspondences. This  is why
the inextensibility constraints, as mentioned in Section~\ref{sec:related}, will be
introduced.

A seemingly natural way to address  this issue is to introduce a linear subspace
model and to write surface deformations as linear combinations of relatively few
basis vectors. This can be expressed as
\begin{equation}
\bx = \bx_0 + \sum_{i=1}^{N_s} w_i {\bf b}_i = \bx_0 + {\bf B}\bw \; ,
\label{eq:PcaModes}
\end{equation}
where $\bx$ is  the coordinate vector of Eq.~\ref{eq:LinProj},  ${\bf B}$ is the
matrix whose columns are the ${\bf b}_i$ basis vectors typically taken to be the
eigenvectors  of a  stiffness matrix,  and $\bw$  is the  associated  vectors of
weights $w_i$. Injecting this  expression into Eq.~\ref{eq:LinProj} and adding a
regularization term yields a new system
\begin{equation}
\left[  \begin{array}{cc} {\bf  MB} &  {\bf Mx}_0  \\ \lambda_r  {\bf L}  & {\bf
      0}  \end{array} \right]  \left[  \begin{array}{c} {\bw}  \\ 1  \end{array}
  \right] = {\bf 0} \; ,
\label{eq:PcaLinSys}
\end{equation}
which is to be solved in the  least squares sense, where ${\bf L}$ is a diagonal
matrix whose  elements are the eigenvalues associated to the  basis vectors, 
and $\lambda_r$  is a  regularization weight.   This favors basis vectors that
correspond to the lowest-frequency deformations and therefore enforces
smoothness.

In practice,  the linear system  of Eq.~\ref{eq:PcaLinSys} is better  posed than
the one  of Eq.~\ref{eq:LinProj}.  But, because there  are usually  several {\it
  smooth} shapes that all yield  virtually the same projection, its matrix still
has  a  number of  near  zero singular  values.   As  a consequence,  additional
constraints still need  to be imposed for the problem  to become well-posed.  An
additional difficulty  is that, because rotations are  strongly non-linear, this
linear formulation can only handle small  ones. As a result, the reference shape
defined by ${\bf  x}_0$ must be roughly aligned with the  shape to be recovered,
which means that a global rotation must be computed before shape recovery can be
attempted.

In  the  remainder  of  this  paper,  we  will  show  that  we  can  reduce  the
dimensionality in a different and rotation-invariant way.

%% file: laplacian.tex

\section{Laplacian Formulation}
\label{sec:Laplacian}

In   the    previous   section,   we    introduced   the   linear    system of
Eq.~\ref{eq:LinProj}, which  is so ill-conditioned  that we cannot  minimize the
reprojection error by simply solving it. In this section, we show how to turn it
into a well-conditioned system using a novel regularization term and how to
reduce the size of the problem for better computational efficiency.

\newcommand{\kpctrlw}{0.4\linewidth}
\begin{figure}[htbp]
\begin{center}
\begin{tabular}{cc}
 \includegraphics[width=\kpctrlw]{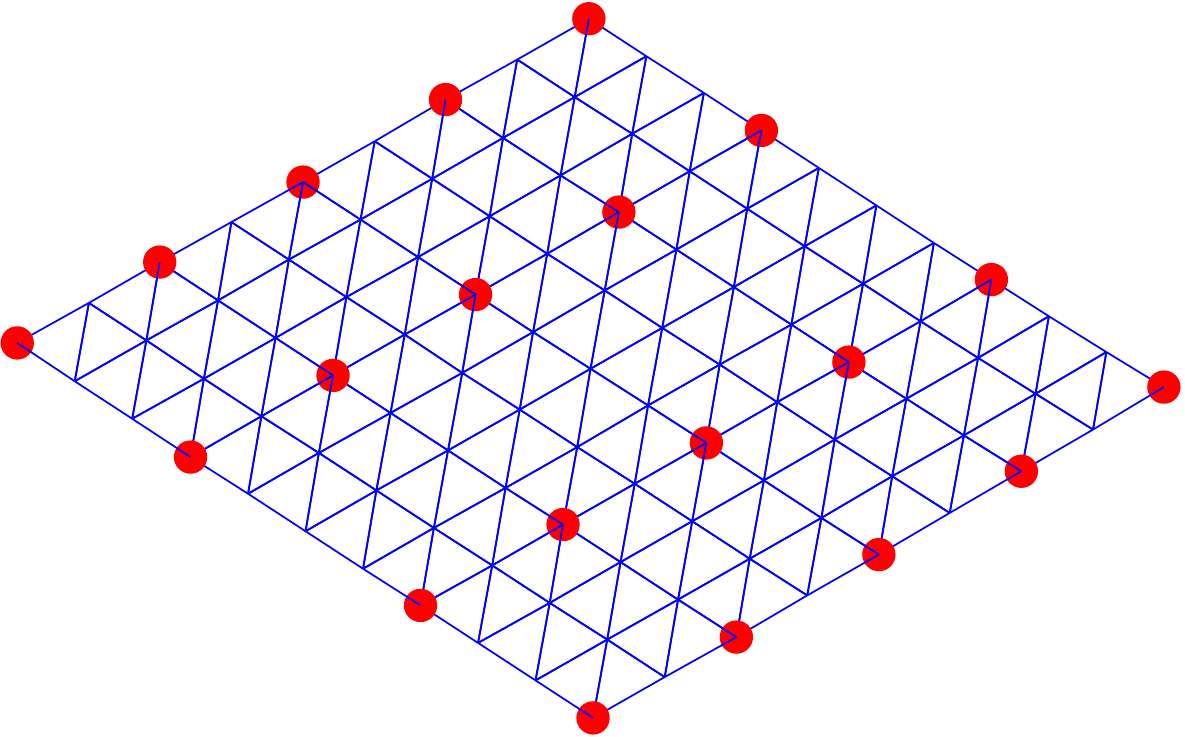} \hspace{0.3cm} &
 \includegraphics[width=0.35\linewidth]{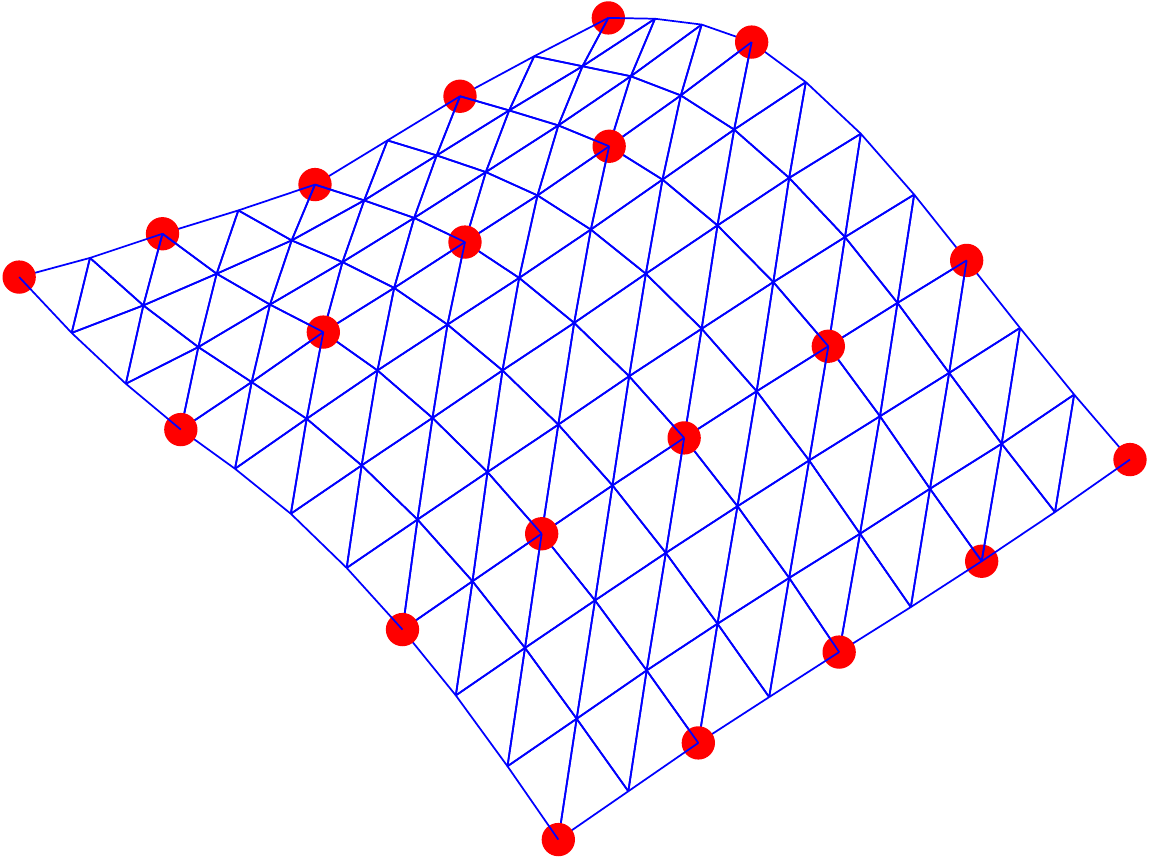} \\ 
 (a)&(b)
\end{tabular}
\end{center}
\vspace{-.3cm}
\caption{{\bf Linear parameterization  of the mesh using control  vertices.}  (a) Reference
  shape and (b) Deformed shape. Every vertex is a linear combination of the control
  vertices, shown in red. In this case, the reference shape is planar.}
\label{fig:ctrlpts}
\vspace{-.1cm}
\end{figure}

\begin{figure*}[htbp]
\centering
\begin{tabular}{cc}
 \includegraphics[height=4.4cm]{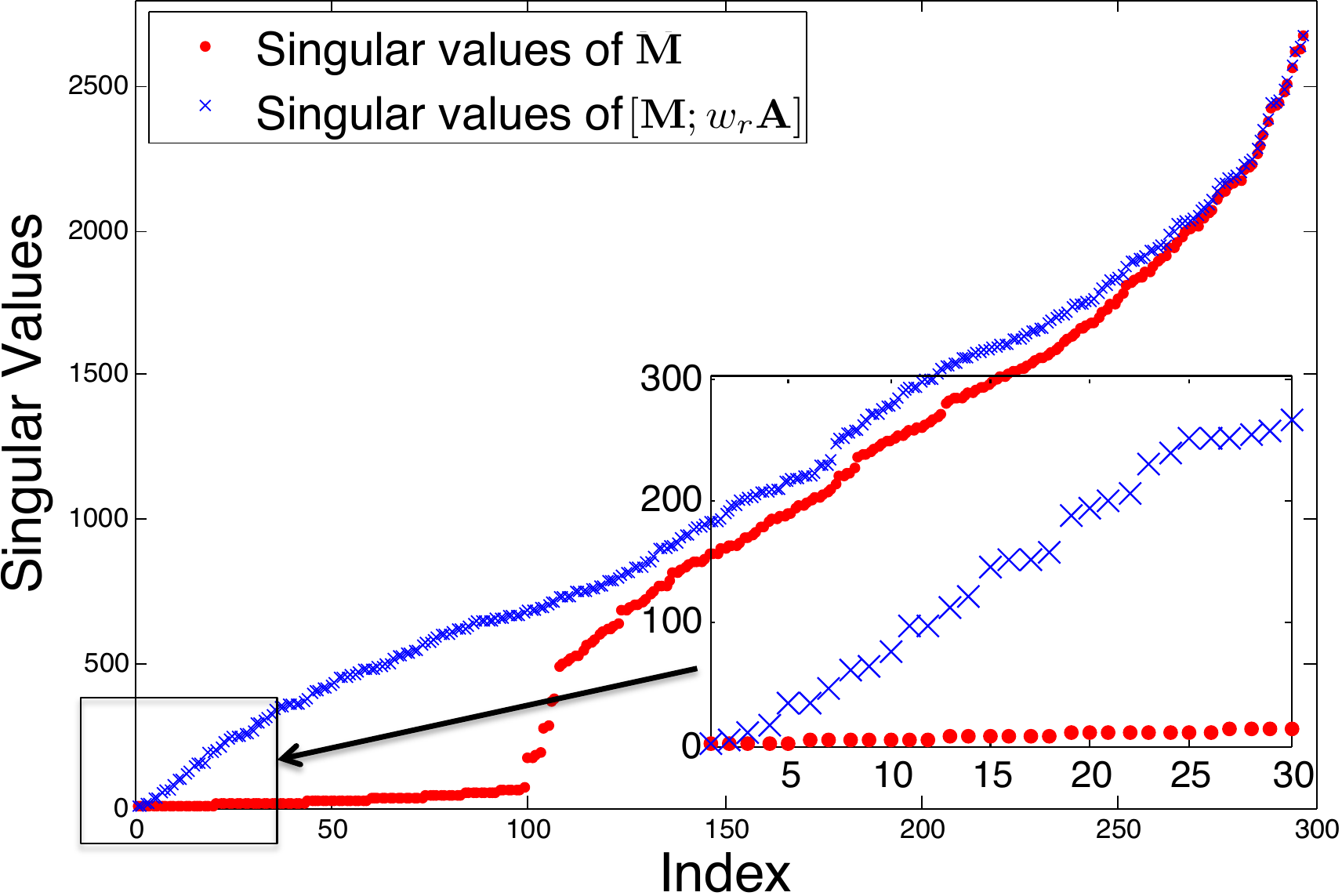} \hspace{1cm} &
 \includegraphics[height=4.4cm]{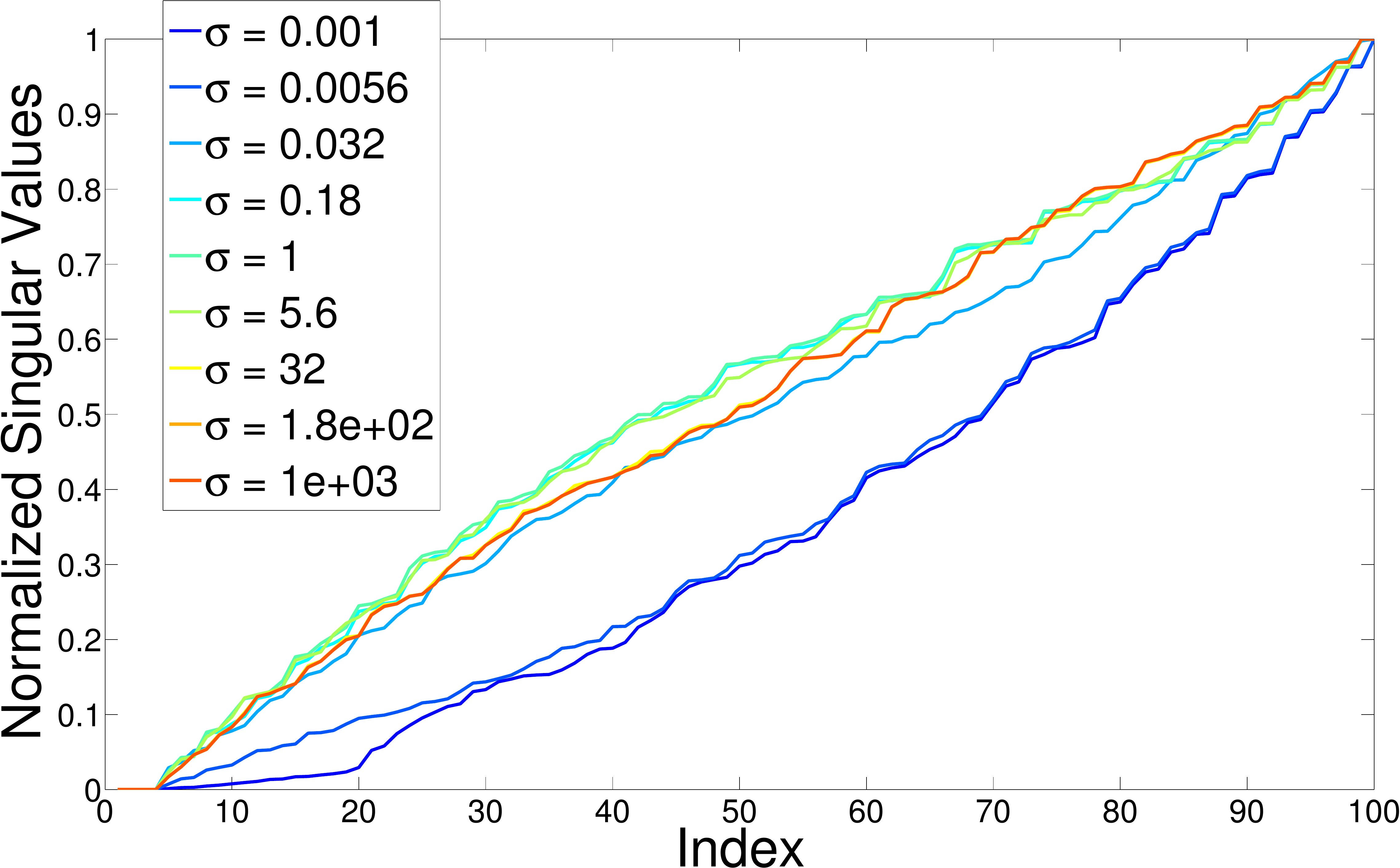}\\
 (a) \hspace{1cm} & (b)
\end{tabular}
\vspace{-.1cm}
\caption{{\bf Conditioning of the regularization  matrix.} (a) Singular values
    of ${\bf M}$ and $\bM_{w_r} = [\bM; w_r \bA]$, in red and blue respectively,
    for a planar  model and a given set of  correspondences.  $\bM_{w_r}$ has no
    zero singular values and  only a few that are small  whereas the first $N_v$
    singular values of ${\bf M}$ are much closer to zero than the rest.  (b)
    Singular  values  of regularization  matrix  $\bA$  for a  non-planar  model
    corresponding to one specific coordinate divided  by the largest one.  Each
    curve corresponds  to a different value  $\sigma$ introduced in
    Section~\ref{sec:NonPlanarReg}. Note that the  curves are almost superposed
    for $\sigma$ values greater than one. The first 4 singular values being 0
    indicates that affine transformations are not penalized.}
\label{fig:singular_values}
\vspace{-.3cm}
\end{figure*}

To  this   end,  let  us   assume  we  are  given   a  reference  shape   as  in
Fig.~\ref{fig:ctrlpts}(a), which may or may not  be planar and let $\bx\sref$ be
the coordinate  vector of  its vertices.   We first  show that  we can  define a
regularization matrix $\bA$ such that $\sqnorm{\bA \bx}=0$, with $\bx$ being the
coordinate vector  of Eq.~\ref{eq:LinProj},  when $\bx\sref=\bx$  up to  a rigid
transformation.   In   other  words,  $\sqnorm{\bA  \bx}$   penalizes  non-rigid
deformations  away   from  the   reference  shape  but   not  rigid   ones.  The
ill-conditioned  linear system  of  Eq.~\ref{eq:LinProj} is  augmented with  the
regularization  term  $\sqnorm{\bA \bx}$  to  obtain  a much  better-conditioned
linear system 
\begin{eqnarray}
\underset{{\bx}}{\text{min}}  & \hspace{0.5cm} \|  {{\bf M}  \bx} \|^2  + {w_r^2}
\|{\bA \bx}\|^{2},    \;   \text{s.~t.}    \;   \norm{\bx} = 1 \; ,
\label{eq:UnconstrOptFull} 
\end{eqnarray}
where  $w_r$  is a  scalar  coefficient  defining  how  much we  regularize  the
solution. Fig.~\ref{fig:singular_values}(a)  illustrates   this  for  a
  specific  mesh. In  the result  section,  we will  see that  this behavior  is
  generic and that we can solve this  linear system for all the reference meshes
  we use in our experiments.

We then show that, given a subset of $N_c$ mesh vertices whose coordinates are
\begin{equation}
\bc =  \left[ \begin{array}{c} \vt{i_1}  \\ \vdots \\ \vt{i_{N_c}}  \end{array} \right]
\;\; ,
\end{equation}
if  we force  the mesh  both to  go through  these coordinates  and  to minimize
$\sqnorm{\bA \bx}$,  we can  define a  matrix $\bP$ that  is independent  of the
control vertex coordinates $\bc$ and such that
\begin{equation}
\bx = \bP \bc \; \; .
\label{eq:xpc}
\end{equation}
In  other words, we  can linearly  parameterize the  mesh as  a function  of the
control vertices' coordinates, as illustrated in Fig.~\ref{fig:ctrlpts}, where
the control vertices are shown in red and (a) is the reference shape and (b) is
a deformed shape according to Eq.~\ref{eq:xpc}. Injecting this parameterization
into Eq.~\ref{eq:UnconstrOptFull} yields a more compact linear system
\begin{eqnarray}
\underset{{\bc}}{\text{min}}  & \hspace{0.5cm} \|  {{\bf MP}  \bc} \|^2  + {w_r^2}
\|{\bA \bP \bc}\|^{2},    \;   \text{s.~t.}    \;   \norm{\bc} = 1 \; ,
\label{eq:UnconstrOpt} 
\end{eqnarray}
which similarly can be solved in the least-square sense up to a scale factor by
finding the eigenvector corresponding to the smallest eigenvalue of the matrix
$\bM_{w_r}^T\bM_{w_r}$, in which
\begin{equation} 
\bM_{w_r} = \left[
\begin{array}{cc}
\bM\bP \\ w_r \bA\bP
\end{array}
\right] \; \; . 
\label{eq:Mwrcat} 
\end{equation}
We  fix the  scale by  making the  average  edge-length be  the same  as in  the
reference shape. This  problem is usually sufficiently  well-conditioned as will
be shown in Section~\ref{sec:analysis}.  Its solution is a mesh whose projection
is very accurate but  whose 3D shape may not be  because our regularization does
not penalize affine deformations away from the reference shape.  In practice, we
use this initial mesh to eliminate  erroneous correspondences. We then refine it
by solving
\begin{equation}
\underset{{\bc}}{\text{min}}  \hspace{0.5cm} \|  {{\bf MP}  \bc} \|^2  + {w_r^2}
\|{\bA \bP \bc   }\|^{2},    \;   \text{s.    t.}    \;   C\left({\bP
  \bc}\right)\leq0 \; ,
\label{eq:ConstrOpt} 
\end{equation}
where  $C\left({\bP \bc}\right)$  are inextensibility  constraints that  prevent
Euclidean distances between neighboring vertices to grow beyond a bound, such as
their geodesic distance  in the reference shape.  We  use inequality constraints
because, in  high-curvature areas, the  Euclidean distance becomes  smaller than
the geodesic  distance. These  are exactly  the same  constraints as  those used
in~\cite{Salzmann11a}, against which we compare ourselves below.  The inequality
constraints  are  reformulated as  equality  constraints  with additional  slack
variables whose norm is penalized to prevent lengths from becoming too small and
the  solution  from   shrinking  to  the  origin~\cite{Fua10}.   This  makes  it
unnecessary to introduce the  depth constraints of~\cite{Salzmann11a}. As
  shown in Appendix~B, this  non-linear minimization step is
  important to guarantee that not only  are the projections correct but also the
  actual 3D shape.

\subsection{Regularization Matrix}
\label{sec:regmatrix}
\input{regmatrix.tex}

\subsection{Linear Parameterization}
\label{sec:parameterization}
\input{parameter.tex}



\subsection{Rejecting Outliers}
\label{sec:robust}
\input{complexity.tex}


%% file: regmatrix.tex

We  now  turn to  building  the  matrix $\bA$  such  that  $\bA \bx\sref=
\mathbf{0}$  and $\sqnorm{\bA   \bx}=\sqnorm{\bA  \bx^\prime}$  when   $\bx\any$
 is   a  rigidly transformed version  of $\bx$.  We  first propose a  very
simple scheme  for the case  when  the  reference  shape  is  planar  and  then
a  similar,  but  more sophisticated one, when it is not.


\begin{figure*}[htbp]
\centering
\begin{tabular}{ccc}
 \includegraphics[width=0.22\textwidth]{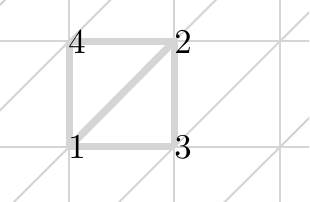} \hspace{1cm} &
 \includegraphics[width=0.20\textwidth]{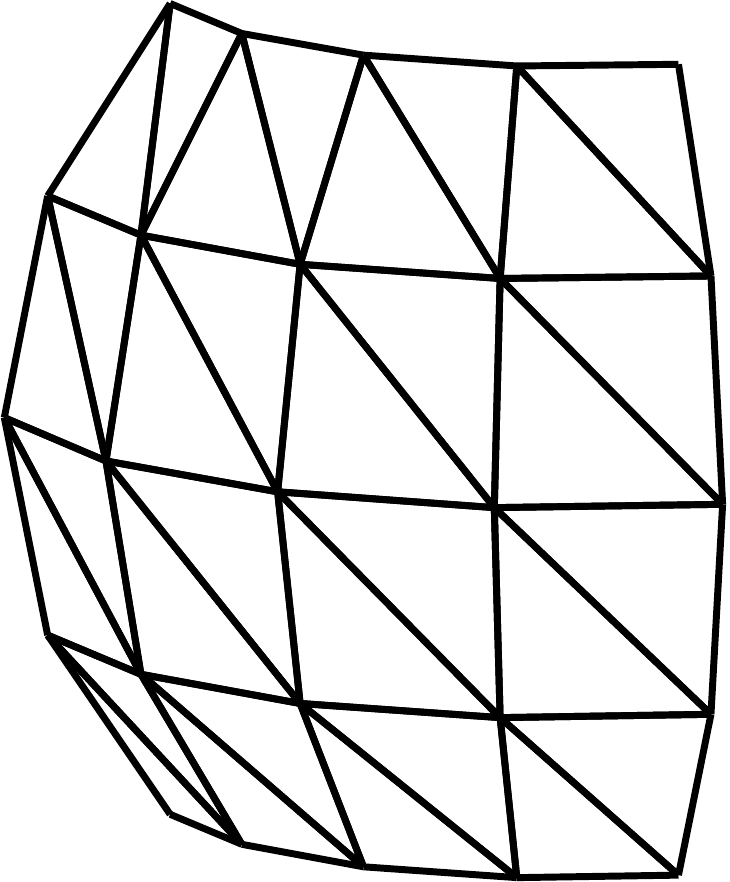} \hspace{1cm} &
 \includegraphics[width=0.20\textwidth]{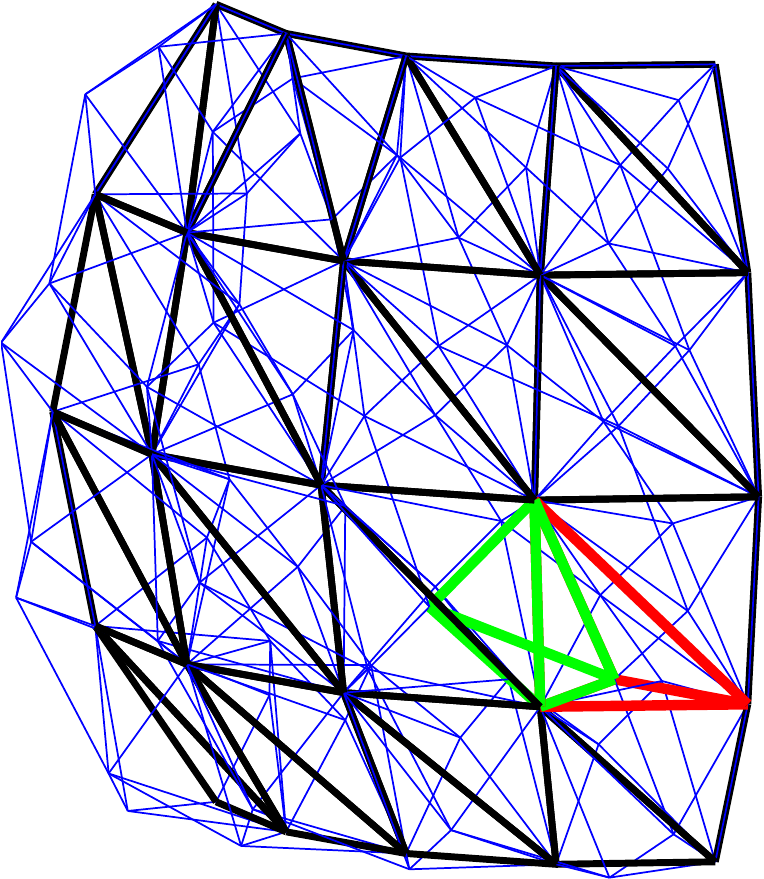} \\ (a) \hspace{1cm} & (b) \hspace{1cm} & (c)
\end{tabular}
\vspace{-.1cm}
\caption{Building       the      $\bA$       regularization      matrix       of
  Section~\ref{sec:regmatrix}.   (a)  Two  facets   that  share  an  edge.   (b)
  Non-planar  reference  mesh.   (c)  Non-planar  reference  mesh  with  virtual
  vertices and edges added. The virtual vertices are located above and below the
  center of  each facet. They are  connected to vertices of  their corresponding
  mesh facet and  virtual vertices of the neighbouring facets  on the same side.
This produces pairs of  blue tetrahedra, one of those is shown in green  and the
other in red.}
\label{fig:regim}
\vspace{-.4cm}
\end{figure*}

\subsubsection{Planar Reference Shape}
\label{sec:PlanarReg}

Given a planar triangular mesh that represents  a reference surface in its reference shape,
consider every  pair of  facets that share  an edge, such  as those  depicted by
Fig.~\ref{fig:regim}(a).   They   jointly  have   four  vertices   $\vtr{1}$  to
${\vtr{4}}$.  Since they lie on a plane, whether or not the mesh is regular, one
can always find a unique set of weights $w_1$, $w_2$, $w_3$ and $w_4$ such that
\begin{eqnarray}
\mathbf{0}  &=  & w_1\vtr{1}  +  w_2\vtr{2} +  w_3\vtr{3}  +  w_4{\vtr{4}} \;  ,
\nonumber\\ 0 &= & w_1 + w_2 + w_3  + w_4 \; , \label{eq:planar} \\ 1 &= & w_1^2
+ w_2^2 + w_3^2 + w_4^2 \; ,\nonumber
\end{eqnarray}
up to  a sign  ambiguity, which can  be resolved  by simply requiring  the first
weight to be positive. The  first equation  in Eq.~\ref{eq:planar}  applies for 
each of  three spatial coordinates $x,y,z$. To form the $\bA$ matrix, we first
generate a matrix $\bA^\prime$ for a single coordinate component. For every
facet pair $k$, let $i_1, i_2, i_3, i_4$ be the indices of four corresponding
vertices. The values at row $k$, columns $i_1, i_2, i_3, i_4$ of $\bA^\prime$
are taken to be $w_1, w_2, w_3, w_4$. All remaining elements are set to zero and
the matrix $\bA$ is $\bA = \bI_{3} \kronecker \bA^\prime$, that is the Kronecker
product with a $3 \times 3$ identity matrix.

We show in Appendix~A that $\mathbf{Ax=0}$ when $\bx$ is an affine
transformed  version  of the  reference  mesh  and  that $\sqnorm{\bA  \bx}$  is
invariant   to   rotations   and    translations.    The   first   equality   of
Eq.~\ref{eq:planar} is designed to enforce planarity while the second guarantees
invariance. The  third equality is there  to prevent the weights  from all being
zero.  The more  the mesh  deviates from  an affine  transformed version  of the
reference mesh, the more the regularization term penalizes.

\subsubsection{Non-Planar Reference Shape}
\label{sec:NonPlanarReg}

\newcommand{\all}{^{\cup}} 
\newcommand{\real}{}                              \newcommand{\Real}{^\textrm{R}}
\newcommand{\virtual}{^\textrm{V}}      
\newcommand{\realverts}{\bx} \newcommand{\virtverts}{\bx\virtual}
\newcommand{\Areal}{\hat{\bA}}
\newcommand{\Avirtual}{\tilde{\bA}}
\newcommand{\virtualparam}{\left( {\Avirtual} \transpose \Avirtual \right)\inv{\Avirtual}\transpose\Areal}

When the  reference shape  is non-planar, there  will be  facets for which  we
cannot solve Eq.~\ref{eq:planar}. As in~\cite{Summer04}, we extend the scheme
described above by introducing virtual  vertices.  As shown in
Fig.~\ref{fig:regim}(c), we create  virtual vertices  at  above and  below the
center  of each  facet as  a distance controlled  by the scale  parameter
$\sigma$. Formally, for  each facet $\vt{i}$, $\vt{j}$ and $\vt{k}$, its center
$\vt{c} = \frac{1}{3} \left(\vt{i} + \vt{j} +  \vt{k} \right)$ and  its normal
$\bn  = \left(\vt{j} -  \vt{i} \right) \times \left(\vt{k} - \vt{i} \right)$, we
take the virtual vertices to be
\begin{equation}
\vt{ijk}^{+} = \vt{c} + \sigma \frac{\bn}{\sqrt{\norm{\bn}}} \quad \textrm{and}
\quad \vt{ijk}^{-} = \vt{c} - \sigma \frac{\bn}{\sqrt{\norm{\bn}}} \; , 
\label{eq:VirtualVerts}
\end{equation}
where  the norm  of  $\bn /  \sqrt{\norm{\bn}}$  is approximately  equal to  the
average edge-length  of the facet's edges. These virtual vertices are connected
to vertices of their corresponding mesh facet and virtual vertices of the
neighbouring facets on the same side, making up the blue tetrahedra of
Fig.~\ref{fig:regim}(c).

Let  $\virtverts$ be the  coordinate vector  of the  virtual vertices  only, and
$\bx\all=\left[\realverts;{\virtverts}\right]$  the  coordinate vector  of
both real and virtual vertices.  Let $\bx\sref\all$ be similarly defined for the
reference shape. Given  two tetrahedra that share a facet, such  as
the red and green ones
in   Fig.~\ref{fig:regim}(c),   and   the   five   vertices   $\vtr{1}\all$   to
${\vtr{5}\all}$ they share, we can now find weights $w_1$ to $w_5$ such that
\begin{eqnarray}
\mathbf{0} &=& w_1\vtr{1}\all + w_2\vtr{2}\all + w_3\vtr{3}\all + w_4\vtr{4}\all
+ w_5\vtr{5}\all \; , \nonumber \\ 0 & = & w_1 + w_2 + w_3 + w_4 + w_5 \; ,
\label{eq:nonplanar} \\ 1 & = & w_1^2 + w_2^2 + w_3^2 + w_4^2 + w_5^2 \; . \nonumber
\end{eqnarray}
The three equalities  of Eq.~\ref{eq:nonplanar} serve the same  purpose as those
of Eq.~\ref{eq:planar}.  We form a  matrix $\bA\all$ by considering all pairs of
tetrahedra that share a facet, computing  the $w_1$ to $w_5$ weights that encode
local linear dependencies  between real and virtual vertices,  and using them to
add successive rows  to the matrix, as we  did to build the $\bA$  matrix in the
planar case.   One can again verify that  $\bA\all \bx\sref\all=\mathbf{0}$
and  that $\|\bA\all  \bx\all\|$ is  invariant to  rigid  transformations of
  $\bx\all$.  In our scheme, the regularization term can be computed as
\begin{equation}
C=\sqnorm{\bA\all \bx\all}=\sqnorm{\Areal \bx\real + \Avirtual \bx\virtual} \; ,
\label{eq:RegVirtual}
\end{equation} 
if we  write $\bA\all$ as $\matrixbrackets{\Areal \:  \Avirtual}$ where $\Areal$
has three times as many columns as there are real vertices and $\Avirtual$ as
there are virtual  ones. Given the real vertices  $\bx\real$, the virtual vertex
coordinates that minimize the $C$ term of Eq.~\ref{eq:RegVirtual} is
\begin{eqnarray}
\virtverts & = & -\virtualparam \realverts \label{eq:regNonplanar} \\ 
\Rightarrow  C &  = &
\sqnorm{\Areal\realverts           -           \Avirtual           \virtualparam
  \realverts} \label{eq:virtualparam} = \sqnorm{\bA \realverts} \; , \nonumber
\end{eqnarray}
where  $\bA = \Areal  - \Avirtual  \virtualparam$. In  other words,  this matrix
$\bA$ is  the regularization matrix we  are looking for and  its elements depend
only on the  coordinates of the reference vertices and  on the scale parameter $\sigma$
chosen to build the virtual vertices.  

To    study   the    influence   of    the   scale    parameter    $\sigma$   of
Eq.~\ref{eq:VirtualVerts}, which  controls the distance of  the virtual vertices
from the mesh, we  computed the $\bA$ matrix and its singular  values for a sail
shaped   mesh    and   many    $\sigma$   values. For each $\sigma$, we
normalized the singular values to be in the range $[0,1]$. As can be seen in
Fig.~\ref{fig:singular_values}(b), there is a wide range of $\sigma$ for which
the distribution of singular values remains practically identical. This suggests
that  $\sigma$  has  limited  influence   on  the  numerical  character  of  the
regularization matrix.  In all our experiments,  we set $\sigma$ to $1$ and were
nevertheless able to obtain  accurate reconstructions of many different surfaces
with very different physical properties.


%% file: parameter.tex

We now show how to use our regularization matrix $\bA$ to linearly parameterize
the mesh using a few control vertices. As before, let $N_v$ be the total number
of vertices and $N_c<N_v$ the number of control points  used to parameterize 
the mesh. Given  the coordinate vector $\bc$  of these control vertices,  the
coordinates of  the minimum energy mesh that  goes through  the control vertices
 and minimizes  the regularization energy can be found by minimizing
\begin{equation}
\sqnorm{\bA \bx} \; \textrm{subject to} \; \bP\cst \bx = \bc \; ,
\label{eq:MinControl}
\end{equation}
where $\bA$  is the regularization matrix  introduced above and $\bP_c$  is a $3
\ctrlcount   \times  3   \vertexcount$   matrix  containing   ones  in   columns
corresponding to the indices of  control vertices and zeros elsewhere.  We prove
below that,  there exists a matrix  $\bP$ whose components can  be computed from
$\bA$ and $\bP\cst$ such that
\begin{equation}
\forall \bc \; , \; \bx = \bP \bc \;,
\label{eq:PMatrix}
\end{equation}
if $\bx$ is a solution to the minimization problem of Eq.~\ref{eq:MinControl}. 

To prove  this without loss  of generality, we  can order the  coordinate vector
$\bx$ so that those of the control  vertices come first. This allows us to write
all coordinate vectors that satisfy the constraint as
\begin{equation}
\bx = \left[ \begin{array}{c} \bc \\ \freeparam \end{array} \right] \; ,
\end{equation} 
where   $\freeparam   \in   \realnumbers^{3*(\vertexcount  -   \ctrlcount)}$.
Let us rewrite the matrix $\bA$ as
\begin{equation}
\bA = \left[\bA_{\bc} | \bA_{\freeparam} \right] \; , \\
\end{equation}
where  $\bA_{\bc}$  and  $\bA_{\freeparam}$ have  $  3  \ctrlcount$  and $  3  (
\vertexcount - \ctrlcount)$ columns,
respectively. Solving the problem of Eq.~\ref{eq:MinControl}, becomes equivalent
to minimizing
\begin{eqnarray}
 \sqnorm{\bA_{\bc} \bc + \bA_{\freeparam} \freeparam} &
\Rightarrow & \freeparam = - (\bA_{\freeparam}^T \bA_{\freeparam})^{-1}
\bA_{\freeparam}^T \bA_{\bc} \bc \\
& \Rightarrow & \bx =
\left[
\begin{array}{c}
\bI \\
-(\bA_{\freeparam}^T \bA_{\freeparam})^{-1} \bA_{\freeparam}^T \bA_{\bc}
\end{array}
\right] \bc \; . \nonumber
\end{eqnarray} 
Therefore,
\begin{equation}
\bP = \left[
\begin{array}{c}
   \bI   \\
   -(\bA_{\freeparam}^T   \bA_{\freeparam})^{-1}
   \bA_{\freeparam}^T    \bA_{\bc}
\end{array}
  \right] \; ,
\end{equation}
is the matrix of Eq.~\ref{eq:PMatrix} under our previous assumption
that the vertices were ordered such that the control vertices come first.


%% file: complexity.tex

To  handle   outliers  in  the  correspondences,  we   iteratively  perform  the
unconstrained  optimization   of  Eq.~\ref{eq:UnconstrOpt}  starting   with  a
relatively  high regularization weight  $w_r$ and  reducing it  by half  at each
iteration. Given a current shape estimate,  we project it on the input image and
disregard  the correspondences  with higher  reprojection error  than  a pre-set
radius and reduce it by half  for the next iteration. Repeating this procedure a
fixed number of  times results in an initial shape  estimate and provides inlier
correspondences for the  more computationally demanding constrained optimization
that follows.  

Even   though  the   minimization  problem   we   solve  is   similar  to   that
of~\cite{Salzmann11a},  this  two-step  outlier  rejection scheme  brings  us  a
computational  advantage.   In  the   earlier  approach,  convex  but  nonlinear
minimization had to  be performed several times at each  iteration to reject the
outliers  whereas  here we  repeatedly  solve  a  linear least  squares  problem
instead.      Furthermore,    the     final    non-linear     minimization    of
Eq.~\ref{eq:ConstrOpt} involves  far fewer  variables since the  control vertices
typically form a small subset of all vertices, that is, $ N_c \ll N_v$.

%% file: results.tex

\section{Results}
\label{sec:results}

In this section, we first compare  our approach to surface reconstruction from a
single  input image  given correspondences  with a  reference image  against the
recent    methods     of~\cite{Brunet10,Salzmann11a,Bartoli12b},    which    are
representative  of the  current state-of-the-art.   We then  present a  concrete
application scenario and  discuss our real-time implementation  of the approach.
We  supply the  corresponding videos  as supplementary  material. They  are also
available  on   the  web~\footnote{\texttt{http://cvlabwww.epfl.ch/\texttildelow
    fua/tmp/laplac}}   along  with   the   Matlab  code   that  implements   the
method.~\footnote{\texttt{http://cvlab.epfl.ch/page-108952-en.html}}

\subsection{Quantitative Results on Real Data}
\label{sec:RealQuant}

To provide quantitative  results both in the planar and  non-planar cases we use
five different sequences for which we have other means besides our single-camera
approach to  estimate 3D  shape.  We  first describe them  and then  present our
results.  Note that we process every  single image {\it individually} and do not
enforce  any kind  of temporal  consistency  to truly  compare the  single-frame
performance of all four approaches.

\begin{figure*}
  \centering
  \input{figs_capaperseq_table.tex}
  \input{figs_apronseq_table.tex}
  \caption{{\bf Paper  and Apron  using a  planar template.}   In the  first and
    third rows, we project the 3D surfaces reconstructed using different methods
    into  the image  used to  perform  the reconstruction.   The overlay  colors
    correspond to those of the  graphs in Fig.~\ref{fig:PlanarAccuracy}.  In the
    second and  fourth rows,  we show  the same surfaces  seen from  a different
    viewpoint.  The gray  dots denote the projections of the  mesh vertices onto
    the ground-truth surface.   The closer they are to the  mesh, the better the
    reconstruction.  (a)  Our method  using regularly sampled  control vertices.
    (b) Brunet  et al.~\cite{Brunet10} (c) Bartoli  et al.~\cite{Bartoli12b} (d)
    Salzmann et al.~\cite{Salzmann11a}. }
\label{fig:paper_apron_seq}
\vspace{-.2cm}
\end{figure*}
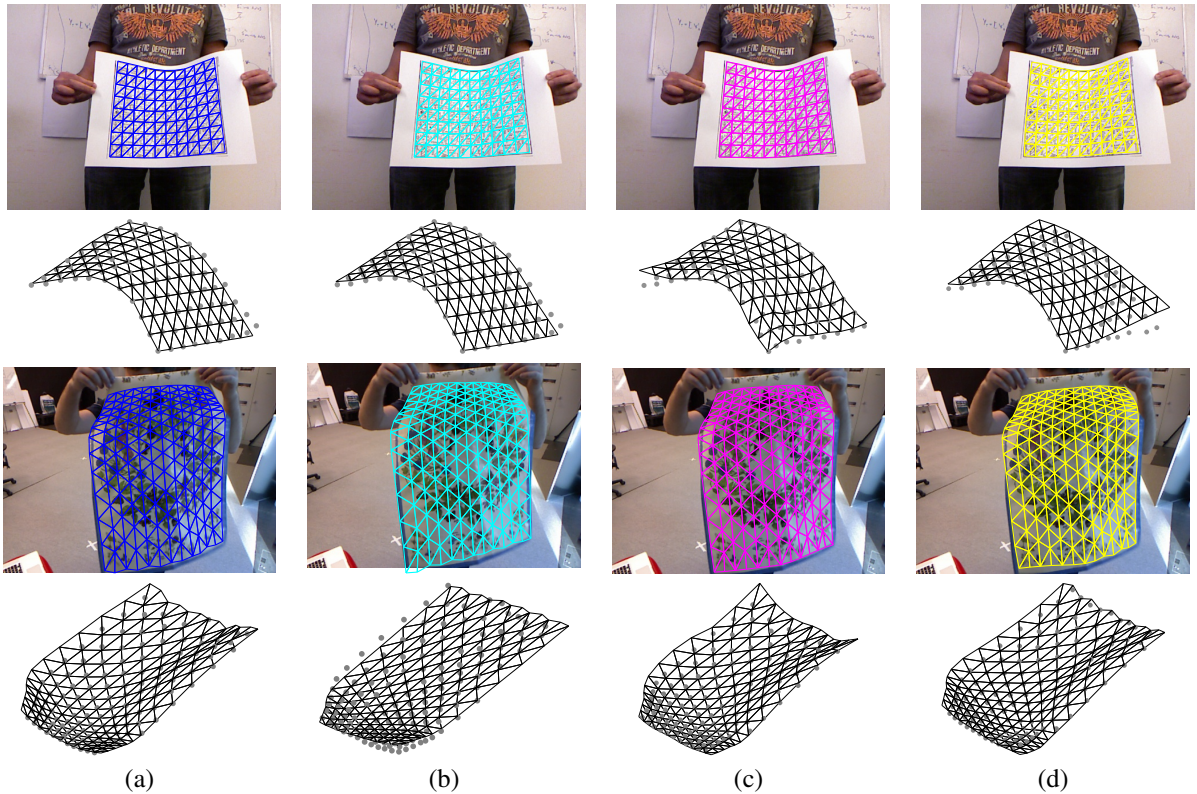

\begin{figure*}
\centering
  \centering
  \input{figs_cushionseq_table.tex}
  \input{figs_bananaseq_table.tex}
  \vspace{-.1cm}
  \caption{{\bf Cushion  and banana  leaf using a  non-planar template.}   As in
    Fig.~\ref{fig:paper_apron_seq}, we  overlay the  surfaces reconstructed
    using the  different methods on  the corresponding  images and also show them as
    seen  from a  different  viewpoint.  The  gray  dots denote  the
    projections  of the  mesh vertices  onto  the ground-truth  surface and  the
    overlay    colors    correspond    to     those    of    the    graphs    in
    Fig.~\ref{fig:NonPlanarAccuracy}(a)  Our  method   using  regularly  sampled
    control  vertices.  (b)  Bartoli  et al.~\cite{Bartoli12b}  (c) Salzmann  et
    al.~\cite{Salzmann11a}.}
  \label{fig:cushion_banana_seq}
  \vspace{-.3cm}
\end{figure*}

\subsubsection{Sequences, Templates, and Validation}
\label{sec:datasets}

We acquired  four sequences using a  Kinect\trademark, whose RGB camera focal length is about 528 pixels. We  show one representative
image      from       each dataset     in      Figs.~\ref{fig:paper_apron_seq}      and
\ref{fig:cushion_banana_seq}. We  performed   the  3D   reconstruction  using
individual  RGB images  and ignoring  the corresponding  depth images.   We used
these {\it  only} to  build the initial  template from  the first frame  of each
sequence  and then  for validation  purposes.  The  specifics for  each  one are
provided in Table~\ref{tab:Datasets}.

For the  paper and apron  of Fig.~\ref{fig:paper_apron_seq}, we used  the planar
templates shown  in the  first two rows  of Fig.~\ref{fig:ctrlsels} and  for the
cushion and banana leaf of Fig.~\ref{fig:cushion_banana_seq} the non-planar ones
shown in next two rows of the  same figure. For the paper, cushion, and leaf, we
defined the  template from the  first frame, which  we took to be  our reference
image. For the apron, we acquired a  separate image in which the apron lies flat
on the floor so that the template is  flat as well, which it is not in the first
frame  of the  sequence,  so that  we  could use  the method  of~\cite{Brunet10}
in the conditions it was designed for.

\input{figs_datasets.tex}

To  create the required  correspondences, we  used either  SIFT~\cite{Lowe04} or
Ferns~\cite{Ozuysal10} to  establish correspondences between the  template
and the input image.     We    ran    our    method    and    three
others~\cite{Brunet10,Bartoli12b,Salzmann11a}  using  these  correspondences  as
input     to     produce    results     such     as     those    depicted     by
Figs.~\ref{fig:paper_apron_seq} and \ref{fig:cushion_banana_seq}.

As stated above, we did not use the depth-maps for reconstruction purposes, only
for validation purposes.  We take  the reconstruction accuracy to be the average
distance of  the reconstructed 3D vertices  to their projections  onto the depth
images,  provided  they fall  within  the  target  object. Otherwise,  they  are
ignored. Since we do the same  for all four methods we compare, comparing their
performance in terms of these averages and corresponding variances is meaningful
and does not introduce a bias.

The last  dataset is depicted by  Fig.~\ref{fig:sail}. It consists of 6 pairs of
images of a sail,  which is a much larger surface than  the other four and would
be hard to capture using a Kinect-style sensor.  The images were acquired by two
synchronized cameras whose focal lengths are about 6682 pixels,  which let us  accurately reconstruct the 3D  positions of
the circular black  markers by performing bundle-adjustment on  each image pair.
As  before we  took the  first image  captured  by the  first camera  to be  our
reference image and established  correspondences between the markers it contains
and those seen in the  other images. We also established correspondences between
markers within the stereo pairs and checked them manually. We then estimated the
accuracy  of the  monocular reconstructions  in terms  of the  distances  of the
estimated markers' 3D positions to those obtained from stereo.

\input{figs_sail.tex}

The 3D  positions of the markers  used to evaluate the  reconstruction error are
estimated from the stereo images up  to an unknown scale factor: This stems from
the  fact  that both  the  relative  pose of  one  stereo  camera  to the  other
\emph{and} the 3D positions of the  markers are unknown and have to be estimated
simultaneously.  However, once  a stereo reconstruction of the  markers has been
obtained, fitting a mesh to those  markers lets us estimate an approximate scale
by assuming the total edge length of the fitted mesh to be that of the reference
mesh. However it  is not accurate and, for evaluation purposes,  we find a scale
factor within the range $[0.8, 1.25]$ that we apply to each one of the monocular
reconstructions so  as to  minimize the mean  distance between the  predicted 3D
marker positions of the monocular  reconstructions and the 3D positions obtained
from stereo.

\subsubsection{Robustness to Erroneous Correspondences}
\label{sec:evalrobustness}
\input{figs_robustness_comparison1.tex}

We  demonstrate the  robustness of  our  outlier rejection  scheme described  in
Section~\ref{sec:robust}  for  both planar  and  non-planar  surfaces.  We  also
compare  our method  against  two  recent approaches  to  rejecting outliers  in
deformable surface reconstruction  \cite{Pizarro12,Tran12} whose implementations
are  publicly  available.   Alternatives  include the  older  M-estimator  style
approaches   of~\cite{Zhu07,Pilet08a},   which   are  comparable   in   matching
performance on the kind of data  we use but slower~\cite{Tran12}. A more
  recent method~\cite{Collins14b}  involves detecting  incorrect correspondences
  using  local isometry  deformation constraints  by locally performing  intensity-based
  dense  registration.  While  potentially  effective,  this algorithm  requires
  several seconds per  frame, which is prohibitive  for real-time implementation
  purpose.

We used  one image from  both the  paper and the  cushion datasets in  which the
surfaces  undergo  large   deformations,  as  depicted  by  the   first  row  of
Figs.~\ref{fig:paper_apron_seq} and~\ref{fig:cushion_banana_seq}.   We used both
the corresponding Kinect  point cloud and SIFT correspondences  to reconstruct a
mesh that  we treated  as the  ground truth.  We then  synthesized approximately
600'000 sets  of correspondences  by synthetically generating  10 to  400 inlier
correspondences spread uniformly across the surface, adding a zero mean Gaussian
noise with standard deviation of 1 pixel to the corresponding pixel coordinates,
and  introducing proportions  varying from  $0$  to $100\%$  of randomly  spread
outlier correspondences.

We ran our algorithm with $25$ regularly sampled control vertices on each one of
these sets of correspondences. To ensure  a fair comparison of outlier rejection
capability in  the context  of deformable  surface reconstruction,  also because
\cite{Tran12} was designed only for outlier rejection, and the implementation of
\cite{Pizarro12}  only performs  outlier rejection,  we used  the set  of inlier
correspondences returned by these methods  to perform surface registration using
our  method  assuming  that  all input  correspondences  are  inliers.
  For~\cite{Pizarro12,Tran12}, we  used the published parameters  except for the
  threshold used to determine if a correspondence is an inlier or an outlier and
  the number  of RANSAC iterations. We  tuned the threshold manually  and used 5
  times  more RANSAC  iterations than  the suggested  default value for~\cite{Tran12}
  to avoid  a performance drop when  dealing with a large
  proportion of  outliers. Note that choosing the  threshold parameter
  for~\cite{Tran12} is  non trivial because the manifold of  inlier
  correspondences is not exactly a  2-D affine plane, as
  assumed in~\cite{Tran12}.

  We  evaluated the  success of  these competing methods  in terms  of how
  often at least $90\%$ of the reconstructed 3D mesh vertices project within $2$
  pixels  of where  they should.   Fig.~\ref{fig:robustnesscompare} depicts  the
  success rates  according to this criterion  over many trials as  a function of
  the total  number of  inliers and  the proportion of  outliers. Our  method is
  comparable to~\cite{Tran12}  on the  cushion dataset and  better on  the paper
  dataset.   It is  much better  than  \cite{Pizarro12} on  both datasets.   Our
  algorithm requires  approximately 200 inlier correspondences  to guarantee the
  algorithm    will   tolerate    high    ratios   of    outliers   with    0.99
  probability. Rejecting  outliers and registering images on  approximately 600'000 set  of synthesized
  correspondences  on the  paper  and cushion  dataset,  respectively, took  our
  method  11.73  (14.80)  hours  i.e.   0.070 (0.089)  second  per  frame,  took
  \cite{Tran12} 44.02 (48.83) hours i.e.  0.26 (0.29) second per frame, and took
  \cite{Pizarro12} 197.46 (201.23) hours i.e. 1.18 (1.21) second per frame.

\comment{
We used two  separate criteria to assess the effectiveness  of the three outlier
rejection methods.  The first criterion for  success is that at  least $90\%$ of
the  reconstructed 3D  mesh vertices  project within  $2$ pixels  of where  they
should. The second is  that at least $90\%$ of the  inlier matches are correctly
labeled  and  retrieved.  Fig.~\ref{fig:robustnesscompare} depicts  the  success
rates according  to the  first criteria  as a  function of  the total  number of
inliers  and the  proportion of  outliers. The  success rates  according to  the
second  criteria  is  given  in  Appendix~\ref{app:inlierrate}.  Our  method  is
comparable to \cite{Tran12} on the cushion dataset, is better than \cite{Tran12}
on the paper dataset, and is much better than \cite{Pizarro12} on both datasets.
It  takes  ours  approximately  200  inlier  correspondences  to  guarantee  the
algorithm will tolerate high ratios of outliers with 0.99 probability. Rejecting
outliers  on approximately  600'000 set  of synthesized  correspondences on  the
paper and  cushion dataset,  respectively, took our  method 11.73  (14.80) hours
i.e. 0.070 (0.089) second per frame, took \cite{Tran12} 44.02 (48.83) hours i.e.
0.26 (0.29)  second per frame,  and took \cite{Pizarro12} 197.46  (201.23) hours
1.18 (1.21) second per frame.
}

\subsubsection{Control Vertex Configurations}

\input{figs_controlpts.tex}

One  of the  important parameters  of  our algorithm  is the  number of  control
vertices to be used.  For each one  of the five datasets, we selected four such
numbers and hand-picked  regularly spaced sets of control vertices  of the right
cardinality.  We also  report the  results using  all mesh  vertices as  control
vertices.  The  left  side  of  Fig.~\ref{fig:ctrlsels}  depicts  the  resulting
configurations.

To  test the  influence of  the positioning  of the  control vertices,  for each
sequence, we  also selected  control vertex sets of the  same size  but randomly
located, such as those depicted on the right side of Fig.~\ref{fig:ctrlsels}. We
did this six times  for each one of the four templates  and each possible number
of control vertices.

In other  words, we tested 29 different control vertex  configurations for each
sequence. 

\newcommand{\allrealerrorgraphs}{\ref{fig:PlanarAccuracy} and~\ref{fig:NonPlanarAccuracy}}

\subsubsection{Comparative Accuracy of the Four Methods}
\label{sec:comparative}

We   fed  the   same   correspondences,  which   contain   erroneous  ones,   to
implementations,  including  the outlier  rejection  strategy,  provided by  the
authors of~\cite{Brunet10,Salzmann11a,Bartoli12b} to compare against our method.
In the case of~\cite{Bartoli12b}, we computed the 2D image warp using the algorithm
of~\cite{Pizarro12} as described in the section {\em Experimental Results} of
~\cite{Bartoli12b}.

As discussed  above, for  the
kinect sequences,  we characterize the  accuracy of their respective  outputs in
terms of the mean and variance of  the distance of the 3D vertices reconstructed
from the  RGB images to  their projections onto the  depth images. For  the sail
images,  we report  the mean  and  variance of  the distance  between 3D  marker
positions computed either monocularly or using stereo, under the assumption that
the latter are more reliable.

In  Fig.~\ref{fig:PlanarAccuracy}, we  plot these  accuracies when  using planar
templates      to      reconstruct     the      paper      and     apron      of
Fig.~\ref{fig:paper_apron_seq}. The graphs are labeled as
\begin{itemize}

 \item {\bf  Ours Regular}  and {\bf  Ours Random} when  using our  method with
   either   the  regular   or  randomized   control  vertex   configurations  of
   Fig.~\ref{fig:ctrlsels}. They  are shown in  blue and green  respectively. In
   the case of  the randomized configurations, we plot  the average results over
   six random trials. We denote our results as {\bf  Ours All} when all mesh
   vertices are used as control vertices.

 \item {\bf Brunet  10}, {\bf Bartoli 12}, and {\bf Salzmann  11} when using the
   methods of~\cite{Brunet10},~\cite{Bartoli12b}, and~\cite{Salzmann11a}.  They
   are shown in cyan, purple, and yellow respectively.

\end{itemize}
The numbers below the graphs denote the  number of control vertices we used. For a
fair comparison,  we used the  same number of  control vertices to run  the method
of~\cite{Brunet10}  and  to   compute  the  2D  warp  required   by  the  method
of~\cite{Bartoli12b}.   The  algorithm of~\cite{Salzmann11a}  does  not rely  on
control  vertices since  it operates  directly on  the vertex  coordinates. This
means it depends on far more degrees  of freedom than all the other methods and,
for comparison  purposes, we depict its accuracy  by the same yellow  bar in all
plots of any given row.

Fig.~\ref{fig:NonPlanarAccuracy} depicts the results similarly in the non planar
case of the cushion, banana leaf, and sail of Figs.~\ref{fig:cushion_banana_seq}
and~\ref{fig:sail}. The only difference is that we do not show the ``Brunet 10''
results because this method was not designed to handle non-planar surfaces.

\subsubsection{Analysis}
\label{sec:analysis}

Our approach performs consistently as well or better than all the others when we
use enough control vertices arranged in one of the regular configurations shown
on the left side of Fig.~\ref{fig:ctrlsels}. This is depicted by the blue bars
in  the four right-most columns of Figs.~\ref{fig:PlanarAccuracy}
and~\ref{fig:NonPlanarAccuracy}. Even  when we use too few  control vertices, we
remain comparable to  the other methods, as shown in the  first columns of these
figures.

More  specifically, in the  planar case,  our closest  competitor is  the method
of~\cite{Brunet10}. We do approximately the same on the paper but much better on
the apron. In the non-planar case that~\cite{Brunet10} is not designed to handle
and when using  enough control vertices, we outperform the  other two methods on
the  cushion and  the leaf  and perform  similarly to~\cite{Salzmann11a}  on the
sail.

As  stated above,  these  results  were obtained  using  regular control  vertex
configurations and  are depicted by blue  bars in Figs.~\ref{fig:PlanarAccuracy}
and~\ref{fig:NonPlanarAccuracy}.   The  green bars  in  these  plots depict  the
results  obtained using  randomized configurations  such as  those shown  on the
right  side of  Fig.~\ref{fig:ctrlsels}.   Interestingly, they  are often  quite
similar, especially for larger numbers  of control vertices.  This indicates the
relative insensitivity  of our  approach to the  exact placement of  the control
vertices.  In fact, we  attempted to  design an  automated strategy  for optimal
placement of the control vertices but found it very difficult to consistently do
better than simply spacing them regularly, which is also much simpler.

Note that our linear subspace parameterization approach to reducing the
dimensionality of the problem does not decrease the reconstruction accuracy when
enough control vertices are used. We interpret this to mean that the true
surface deformations lie in a lower dimensional space than the one spanned by
all the vertices' coordinates.

\begin{figure*}
\centering
\input{figs_ctest022results_table.tex}
\input{figs_ctest030results_table.tex}
\vspace{-.1cm}
\caption{{\bf  Accuracy  results  when  using planar  templates}  and  different
  numbers  of control  vertices for  the paper  (top row)  and apron  (bottom row)  sequences of Fig.~\ref{fig:paper_apron_seq}. }
\label{fig:PlanarAccuracy}
\end{figure*}

\begin{figure*}
\centering
\input{figs_ctest026results_table.tex} 
\input{figs_ctest031results_table.tex} 
\input{figs_ctest019results_table.tex} 
\vspace{-.1cm}
\caption{{\bf Accuracy  results when  using non-planar templates}  and different
  numbers of control vertices for the cushion (top row), banana leaf (middle row),
  and   sail  (bottom   row)  sequences   of  Figs.~\ref{fig:cushion_banana_seq}
  and~\ref{fig:sail}.}
\label{fig:NonPlanarAccuracy}
\vspace{-.4cm}
\end{figure*}

\comment{
\newcommand{\recerrgraph}[4]{
\begin{figure*}
  \centering
\input{figs/ctest#1results/table.tex} 
  \caption{Reconstruction
    errors on {\bf #2} dataset for different methods and number of
    degrees of freedom. #3}
  \label{fig:#4}

}
\recerrgraph{022}{paper}{ }{recerrpaper}
\recerrgraph{025}{apron}{ }{recerrapron}
\recerrgraph{026}{cushion}{ }{recerrcushion}
\recerrgraph{028}{banana leaf}{ }{recerrbanana}
\recerrgraph{019}{sail}{ }{recerrsail}
}

\begin{figure}
\centering
\includegraphics[width=0.85\linewidth, height=0.65\linewidth]{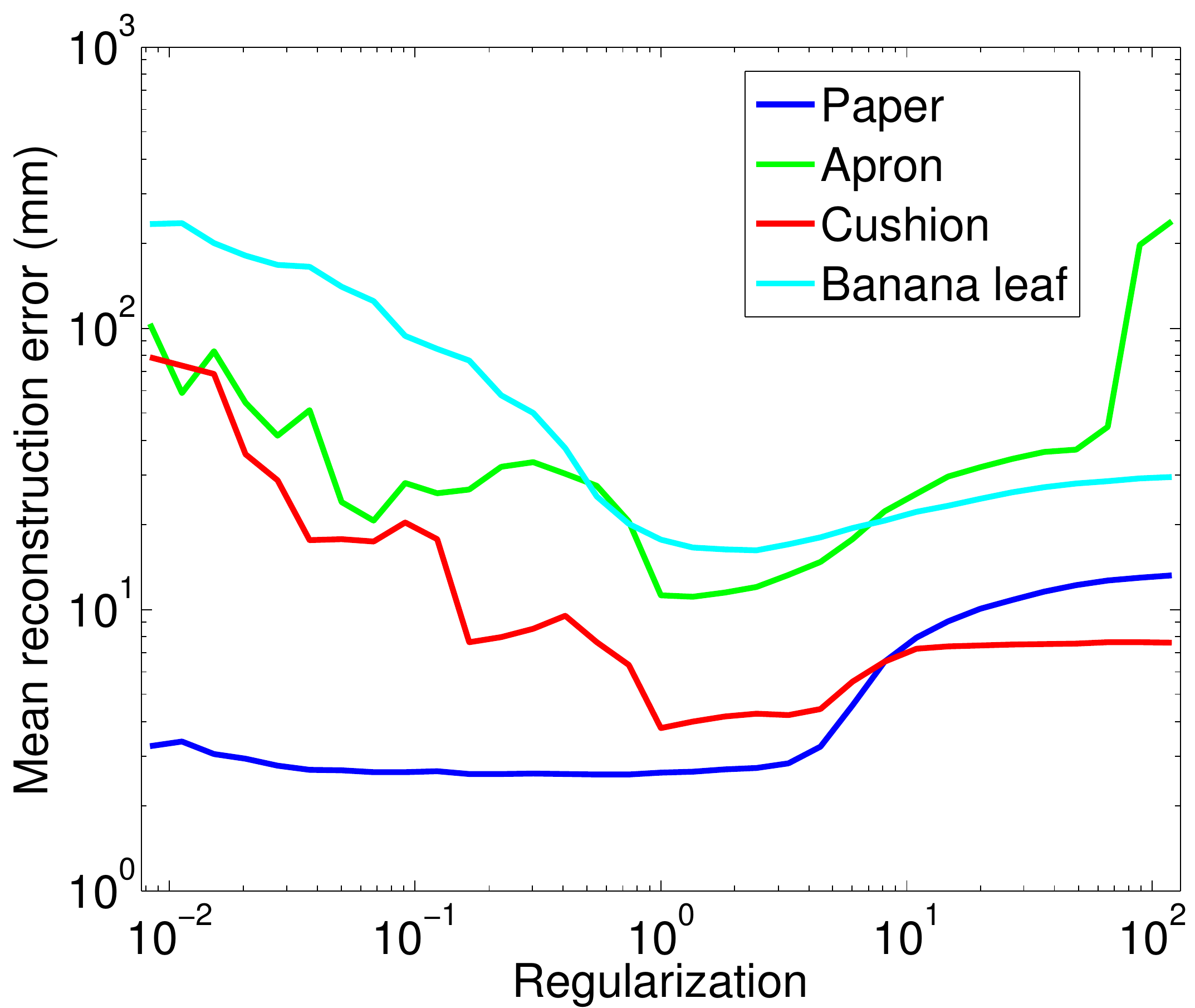}
\caption{{\bf Influence  of the regularization  parameter.}  Mean reconstruction
  errors for the four kinect sequences as a function of the regularization
  weight. For each sequence, we  used the control vertex configuration that gave
  the best result in the experiment of Section~\ref{sec:comparative}.}
\label{fig:regscale}
\vspace{-0.4cm}
\end{figure}

All these experiments were carried out with a regularization weight of $w_r = 1$
in Eq.~\ref{eq:ConstrOpt}.  To characterize  the influence of this parameter, we
used  the control vertex  configuration that  yielded the  best result  for each
dataset and re-ran the algorithm with the same input correspondences but with 33
different logarithmically distributed values of $w_r$ ranging from $0.00833=1 /
  120$  to $120$.   Fig.~\ref{fig:regscale} depicts  the results  and indicates
that values ranging from $1$ to $10$ all yield similar results.

\input{figs_wellposed.tex}

\newcommand{\mcat}{\bM_{w_r}}
Fig.~\ref{fig:wellposed} illustrates how well- or ill-conditioned the linear
system of Eq.~\ref{eq:UnconstrOpt} is depending  on the value of $w_r$. We plot
condition  numbers of $\mcat$ in Eq.~\ref{eq:Mwrcat}, averaged over  several
examples, as a function of $w_r$  for all the regular control-vertex
configurations we used to perform the  experiments described above on  the four
kinect  sequences. We also show the  singular values of $\mcat$, also averaged
over several examples,  for two different configurations  of the  control
vertices  in the  apron case.
Note that  the best condition numbers  are obtained for values  of $w_r$ between
$1$ and $10$, which is consistent with the observation made above that these are
good  default values  to  use for  the  regularization parameter.   This can  be
interpreted  as follows:  For  very small  values  of $w_r$,  the $\bM\bP$  term
dominates and  there are many shapes that  have virtually the  same projections.
For very large values of $w_r$, the $\bA\bP$ term dominates in
Eq.~\ref{eq:UnconstrOpt} and many shapes that are rigid  transformations
of  each  other  have virtually  the  same  deformation energy. The best
compromise is therefore for intermediate values of $w_r$.

\subsection{Application Scenario}
\label{sec:baseball}

In this section, we  show that our approach can be used  to address a real-world
problem, namely modeling  the deformations of a baseball as it  hits a bat. We
obtained from Washington State University's Sport  Science Laboratory 7000
frames-per-second videos such as the one  of Fig.~\ref{fig:baseball}, which
shows the  very strong deformation that occurs at the moment of impact.  The
camera focal length is about 2853 pixels. In this specific case, the ball was
thrown at 140 mph by a launcher against a half cylinder that serves as the bat
to study its behavior.

\begin{figure*}
\newcommand{\baseballwidth}{0.179\linewidth}
\centering
\begin{tabular}{ccccc}
\includegraphics[width=\baseballwidth]{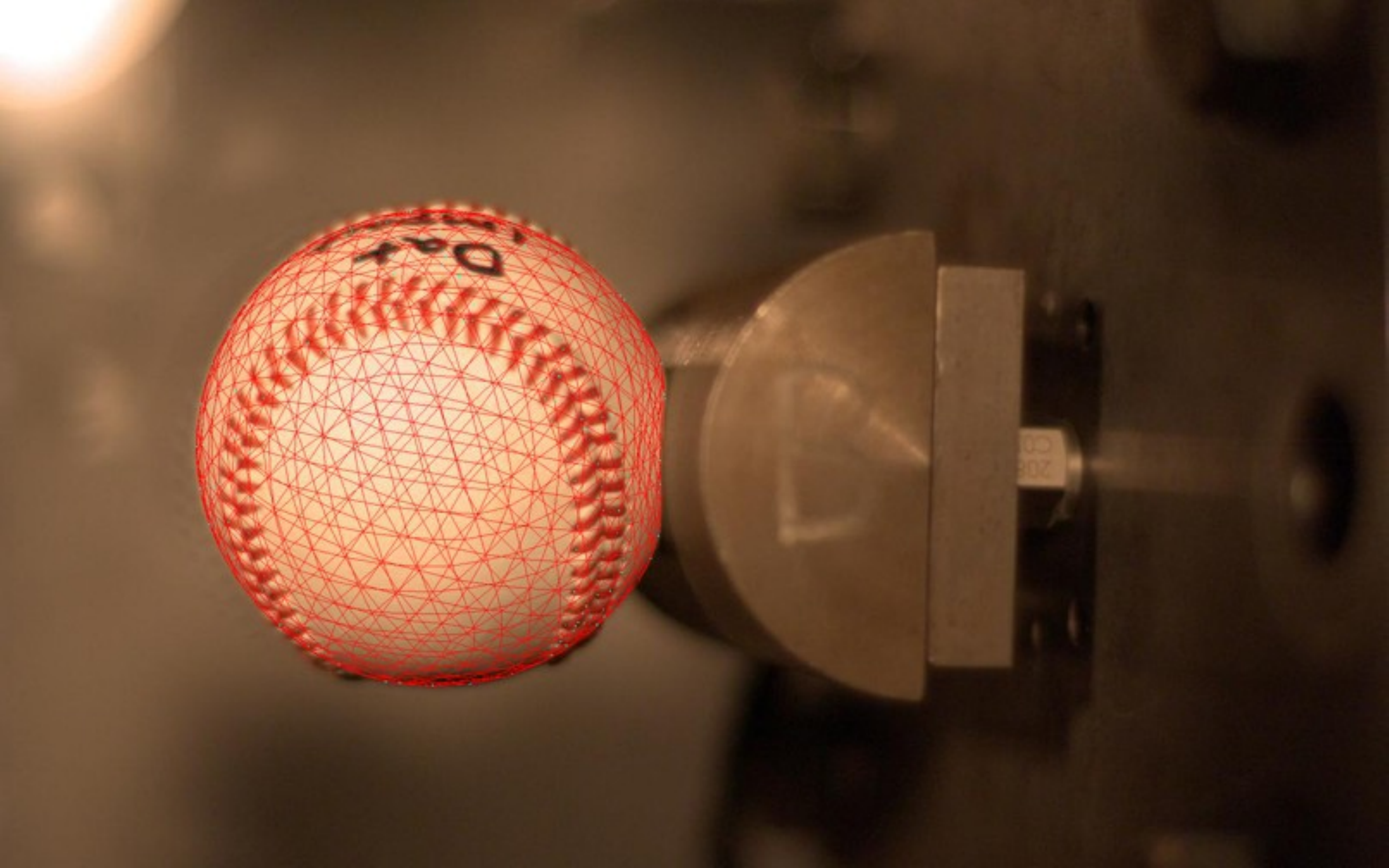} & 
\includegraphics[width=\baseballwidth]{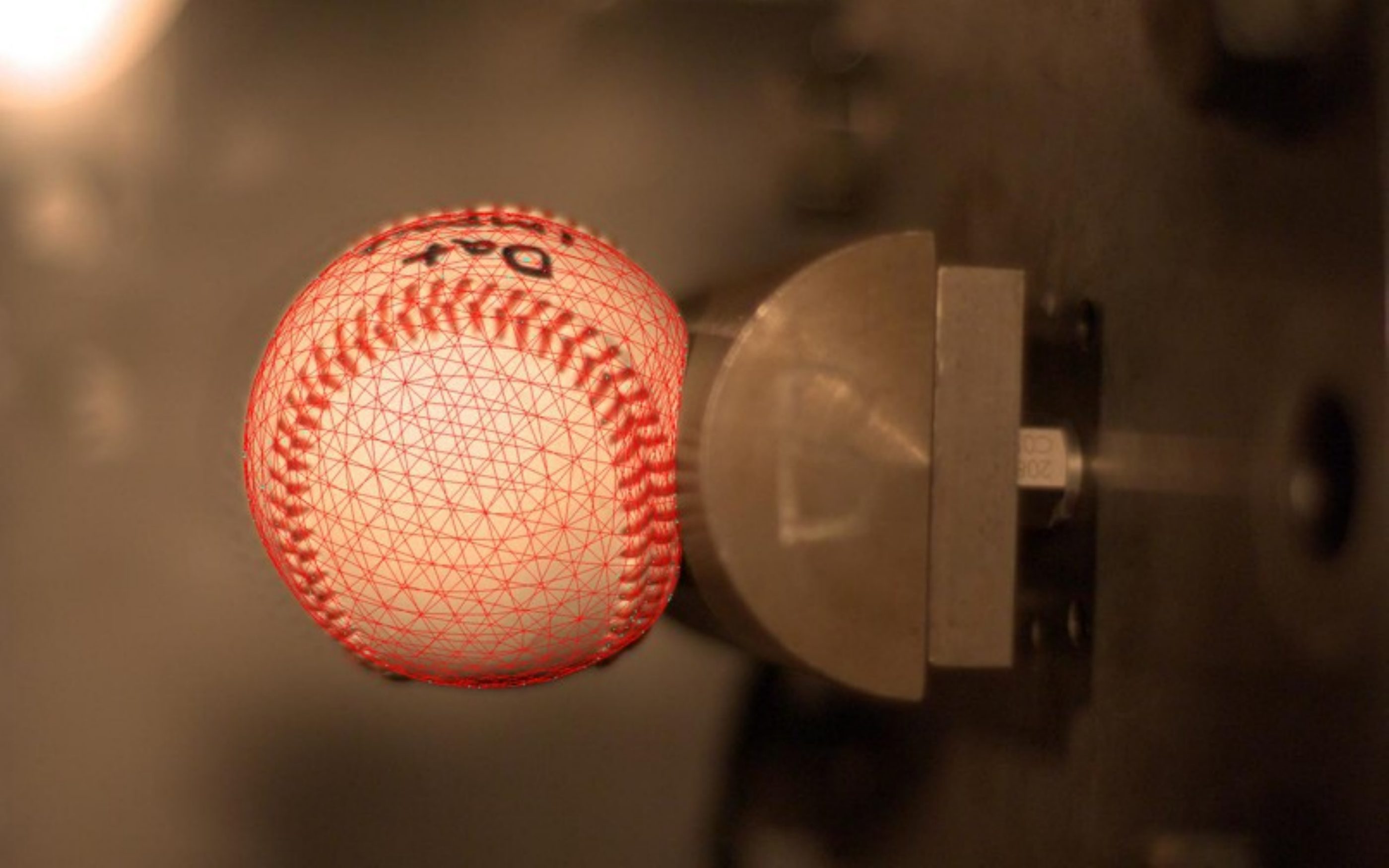} & 
\includegraphics[width=\baseballwidth]{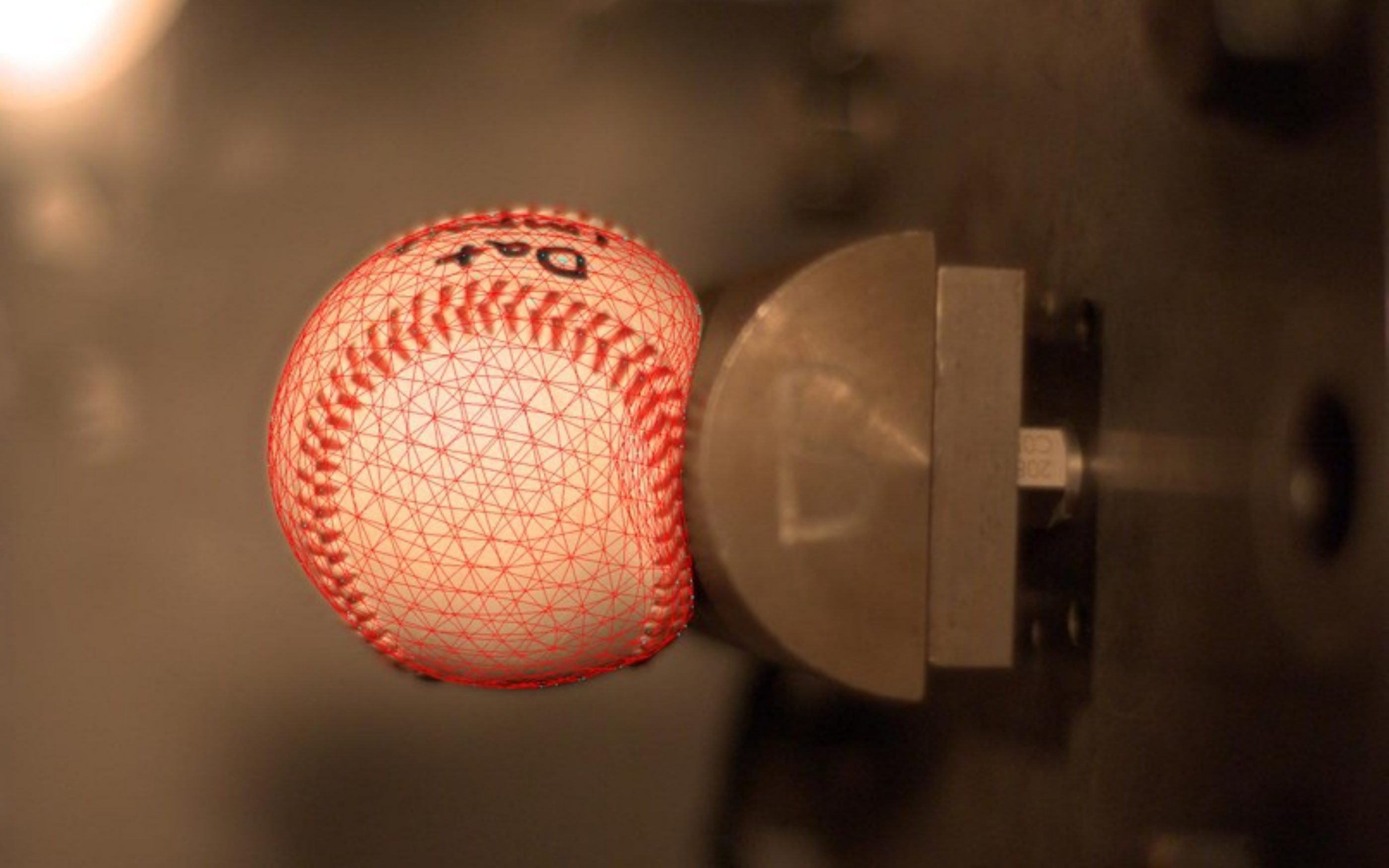} & 
\includegraphics[width=\baseballwidth]{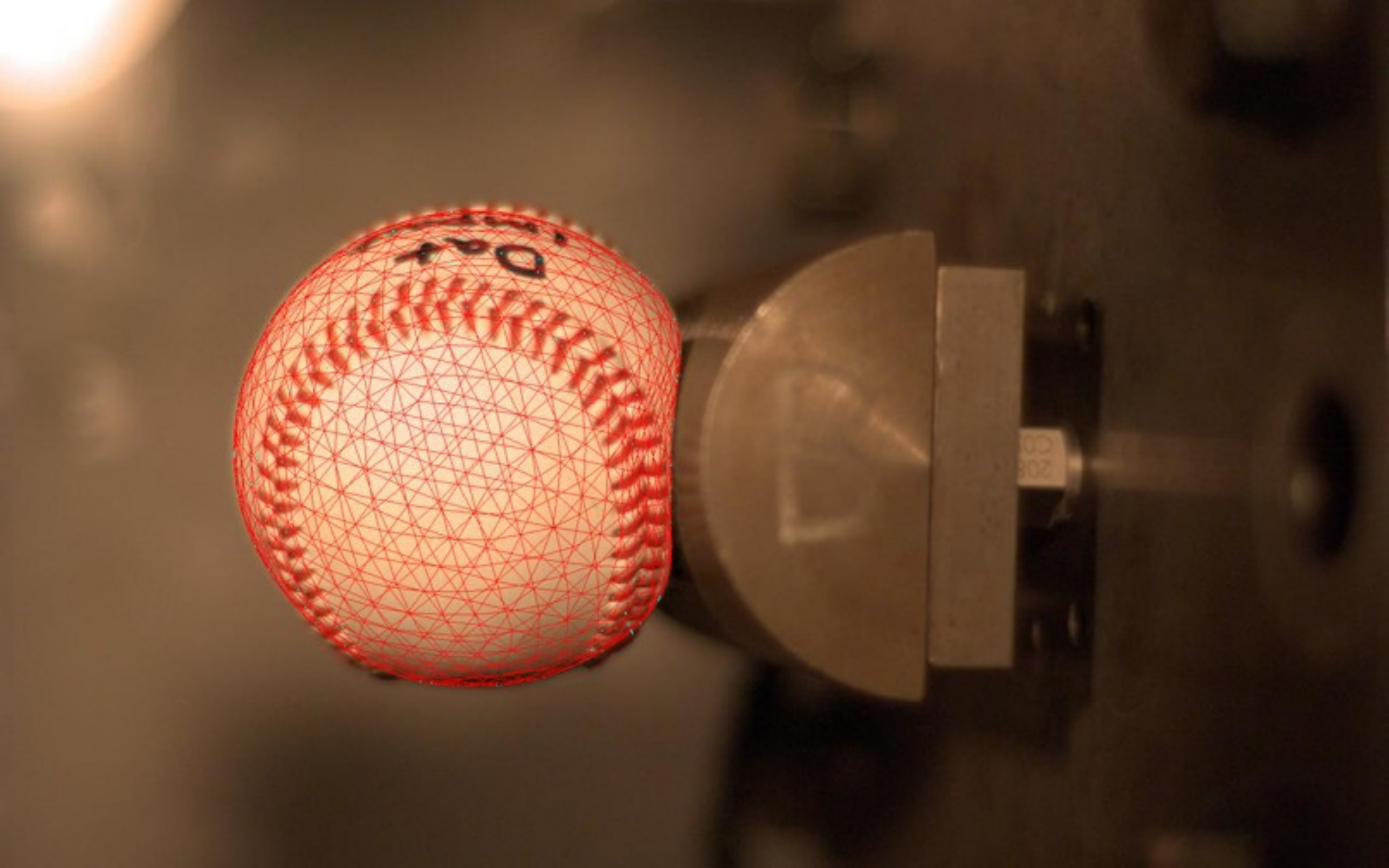} & 
\includegraphics[width=\baseballwidth]{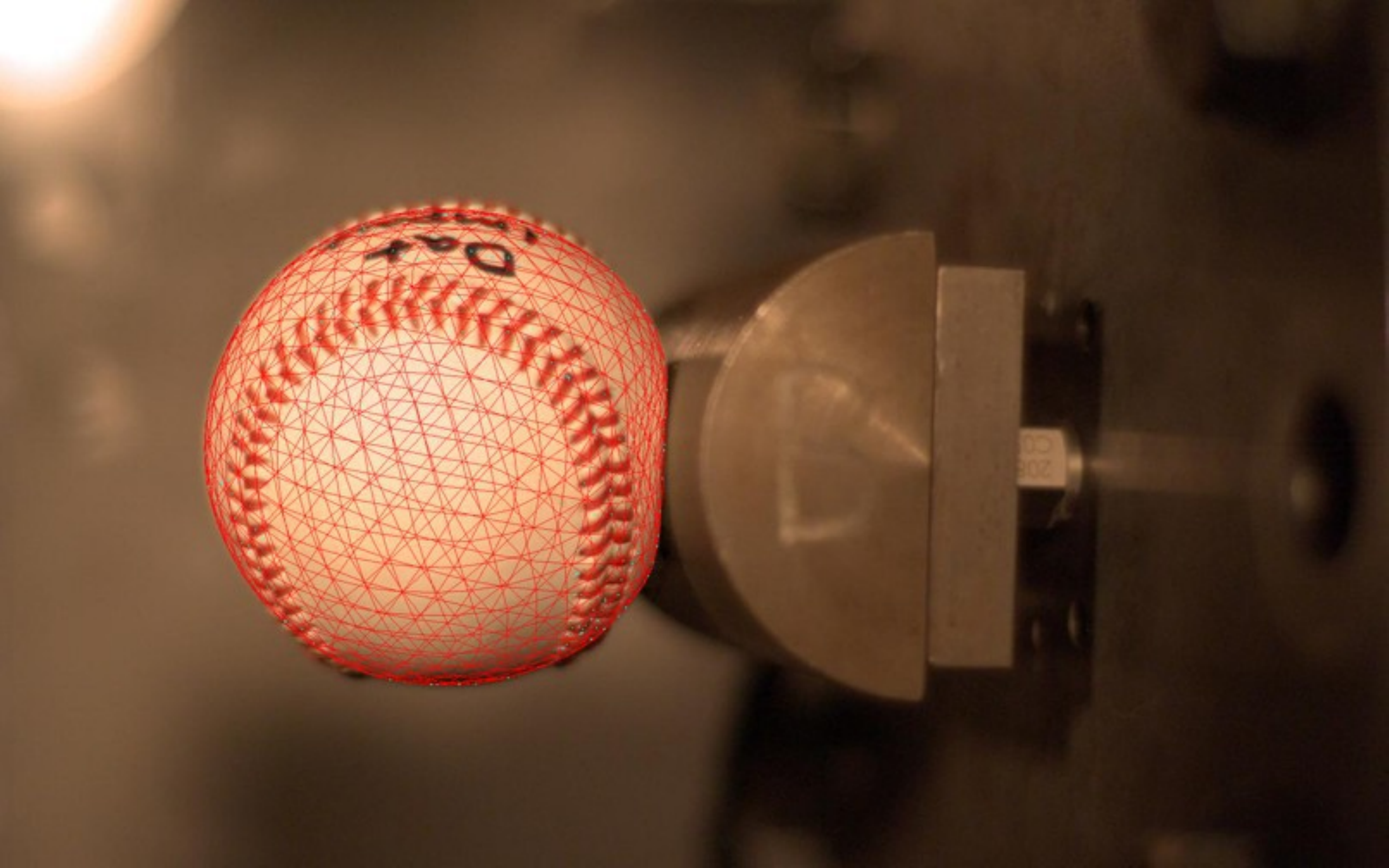} \\

\includegraphics[width=\baseballwidth]{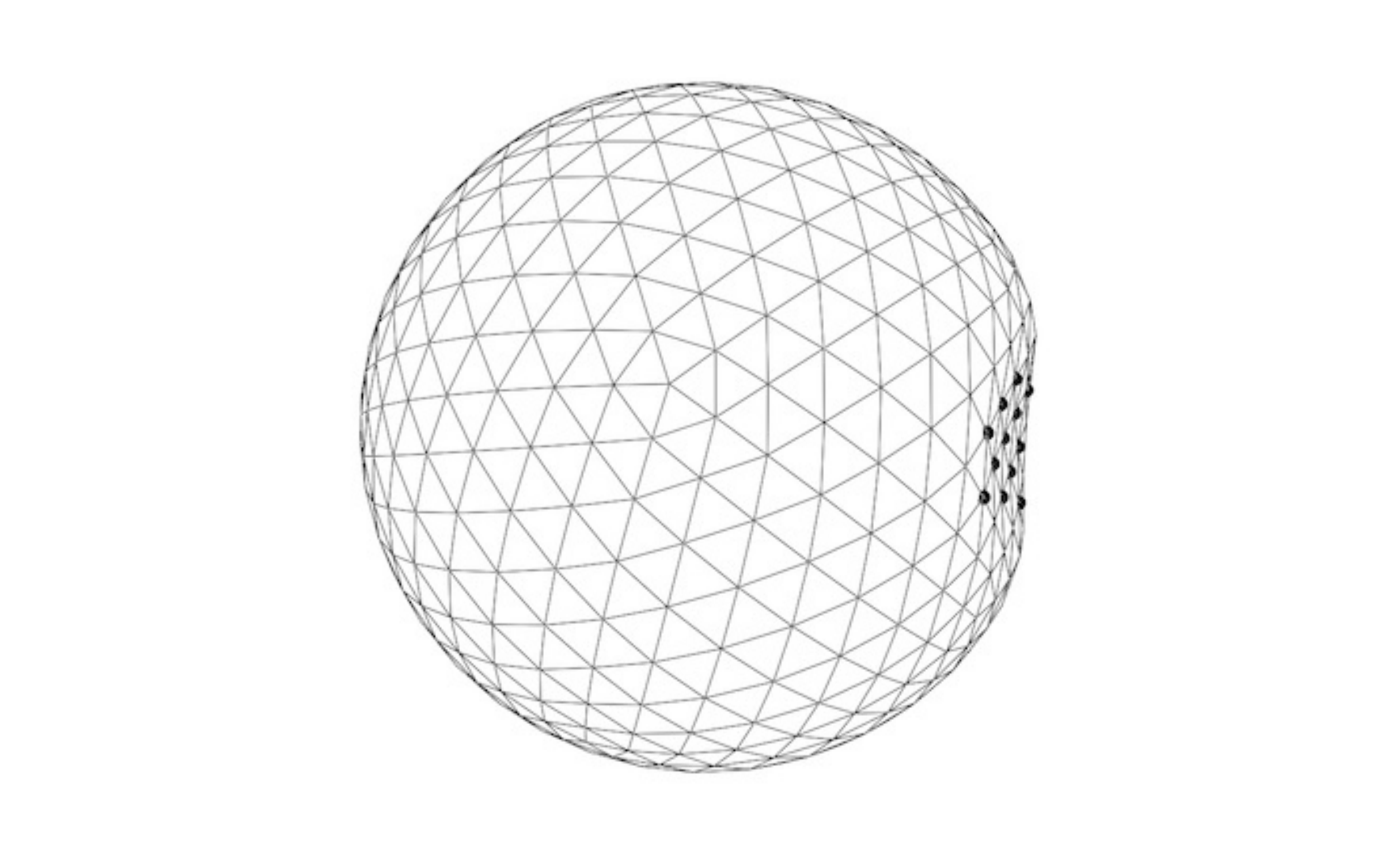} & 
\includegraphics[width=\baseballwidth]{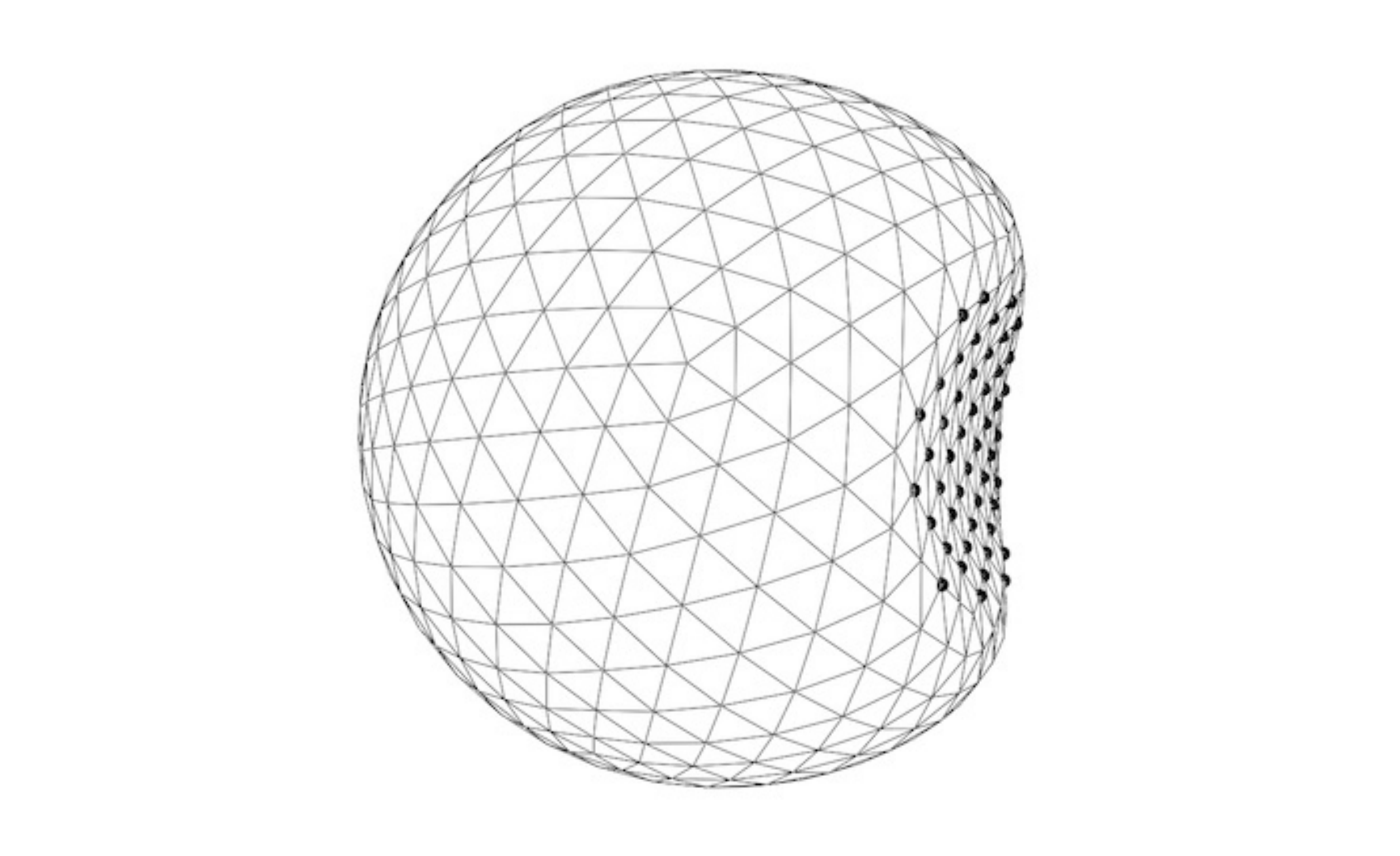} & 
\includegraphics[width=\baseballwidth]{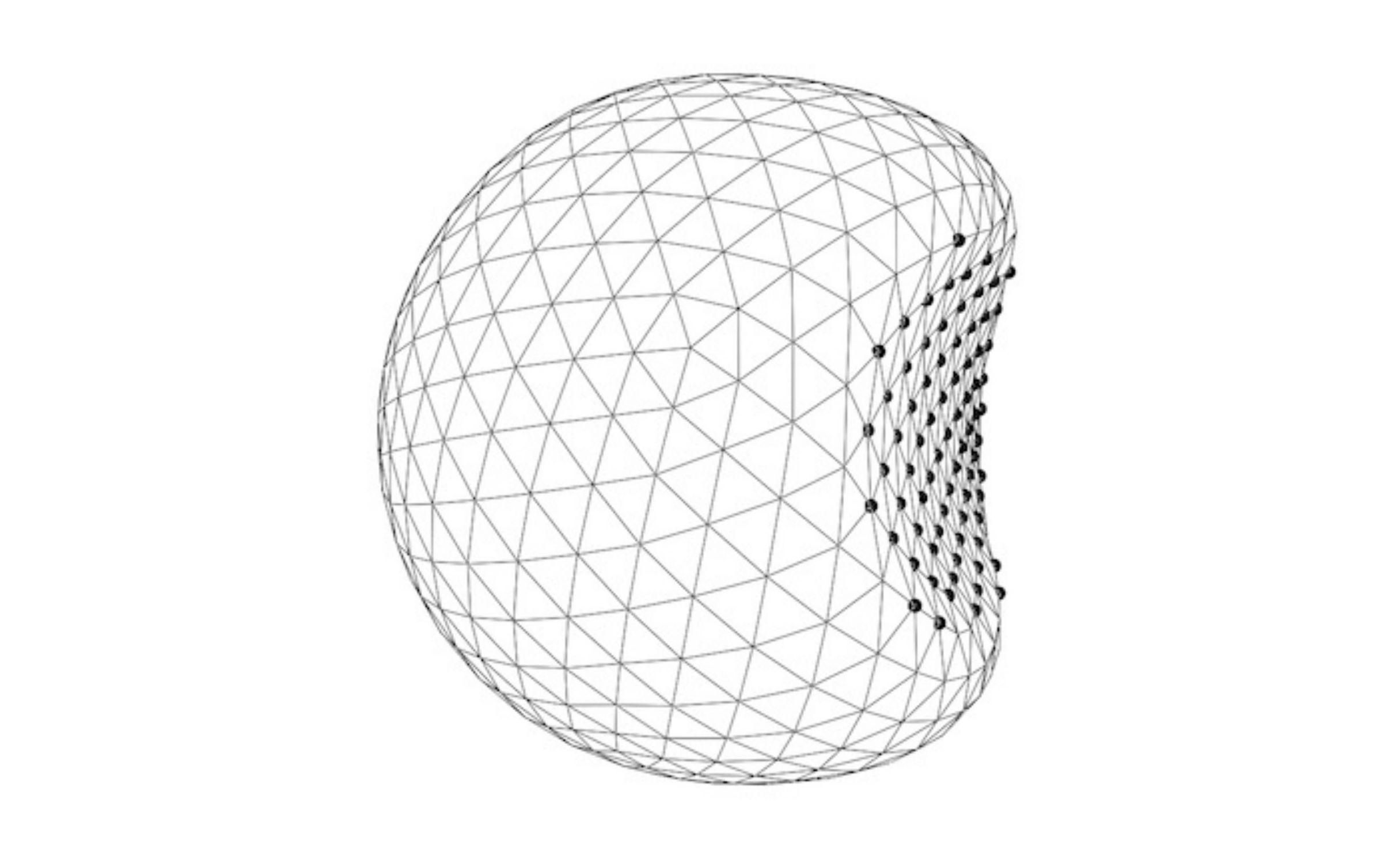} & 
\includegraphics[width=\baseballwidth]{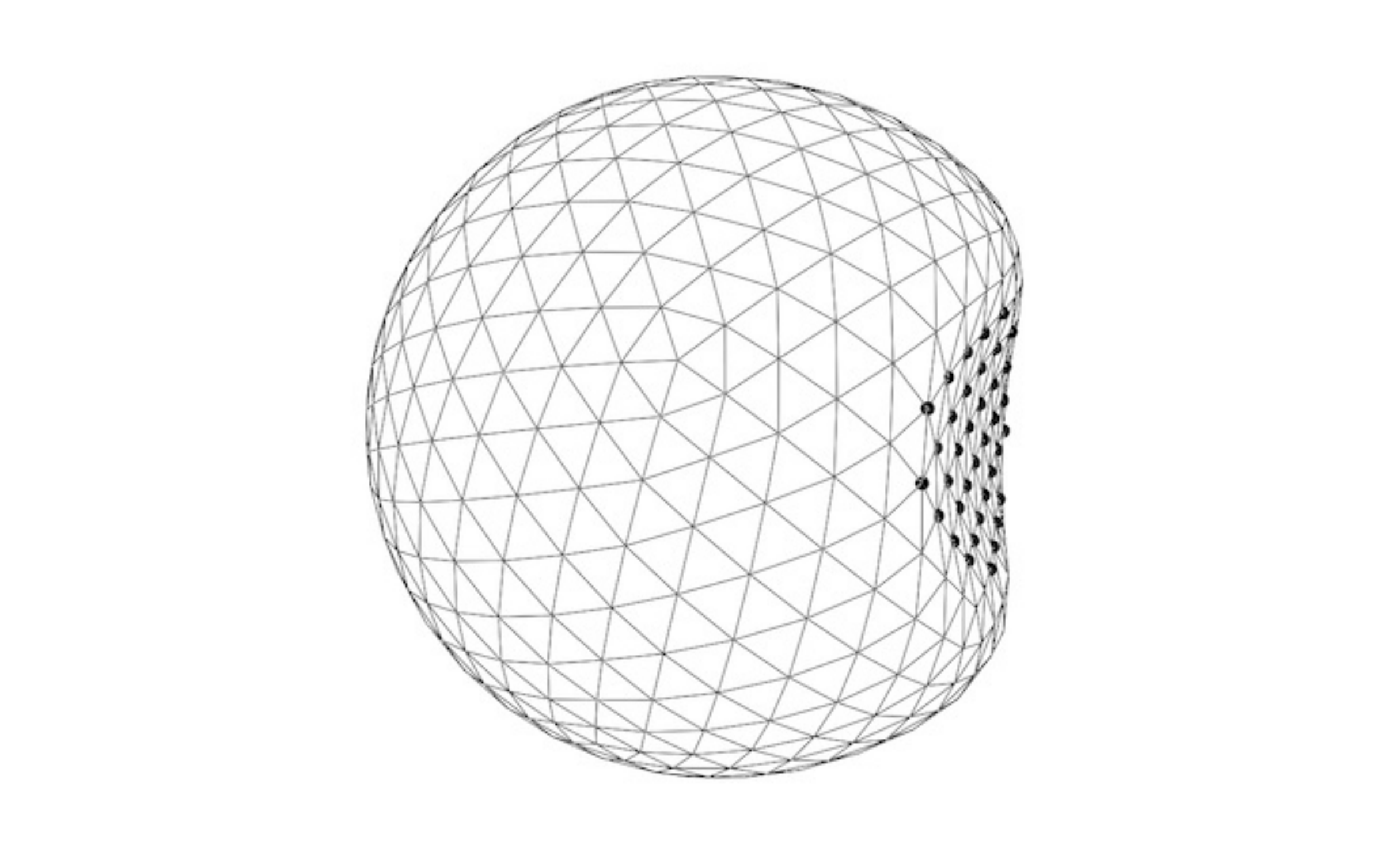} & 
\includegraphics[width=\baseballwidth]{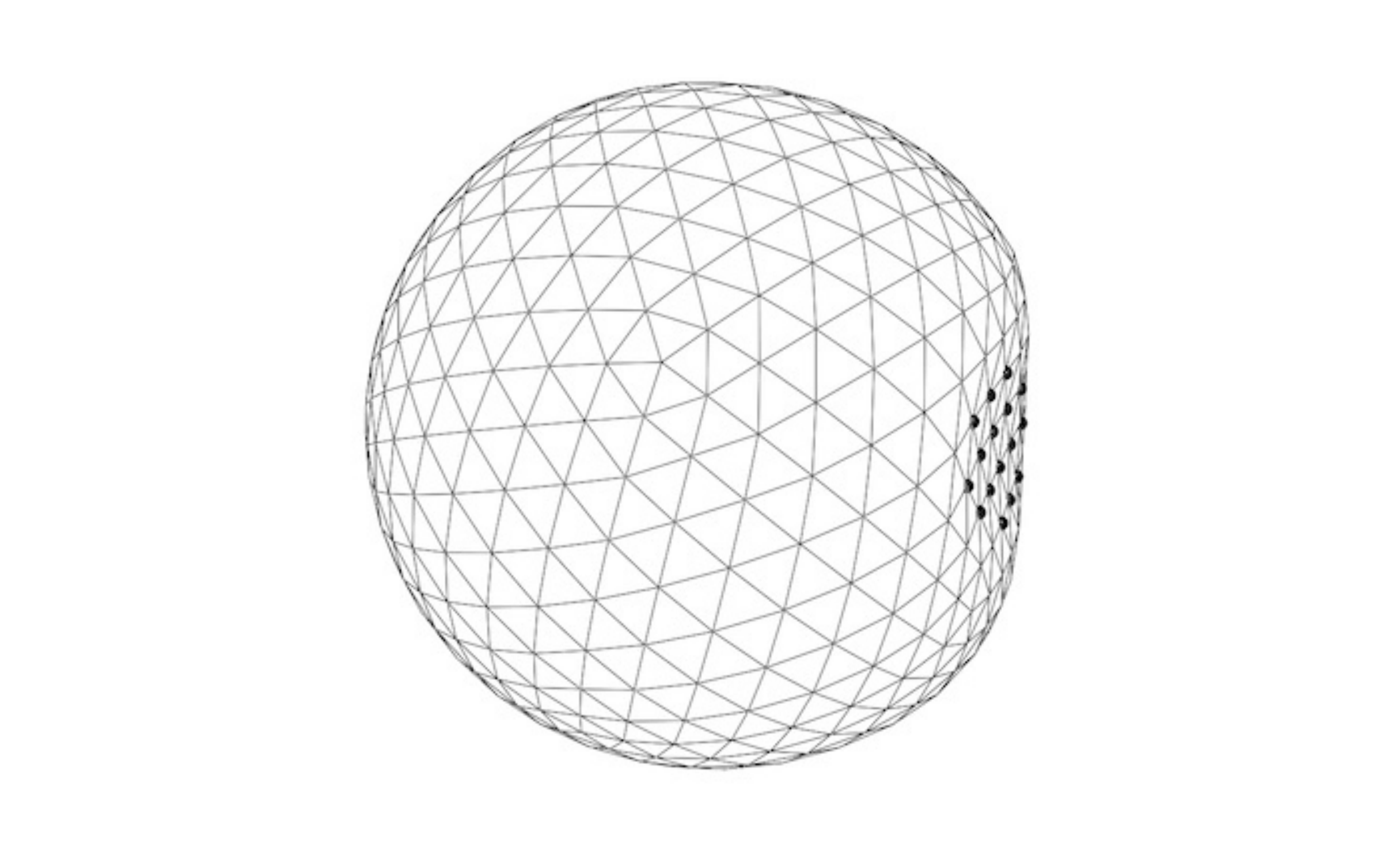} \\

\end{tabular}
\vspace{-0.2cm}
\caption{{\bf Reconstruction of a baseball colliding with a cylinder in a video sequence.}
{\bf Top row:} Reprojection of the reconstructed meshes for different frames in
the sequence.
{\bf Bottom row:} The reconstructed meshes seen from a different view point.
Black dots represent on-contact vertices.}
\label{fig:baseball}
\vspace{-0.4cm}
\end{figure*}

We take the reference frame to be the first one where the ball is undeformed and
can be represented by a spherical  triangulation of diameter 73.52 mm. In this
case, the physics is slightly different  from the examples shown in the previous
section  because the  surface does  stretch but  cannot penetrate  the cylinder,
which is  securely fastened to the  wall.  We therefore  solve the unconstrained
minimization problem of Eq.~\ref{eq:UnconstrOpt} as before and use its result to
initialize a  slightly modified version of the  constrained minimization problem
of Eq.~\ref{eq:ConstrOpt}. We solve
\begin{eqnarray}
\underset{{\bc}}{\text{min}} \hspace{0.5cm} \|  {{\bf MP}  \bc} \|^2 +  {w_r^2} \|{\bA  \bP \bc
}\|^{2} + \nonumber \\ 
{w_l^2}    \sum_{ij}    ||d(V_i,V_j)-l_{ij}||^2    +    {w_t^2}~\text{Traj}(\bc)
&,& \label{eq:ConstrOptBaseBall} \\
\text{s.  t.}  \; C\left({\bP
  \bc}\right)\leq0 \quad , \quad \quad \quad & & \nonumber
\end{eqnarray}
where  $C\left({\bP \bc}\right)$  now stands  for constraints  that  prevent the
vertices from being inside  the cylinder, the $\text{Traj}(\bc)$ term encourages
the ball to move in a straight line computed from frames in which
the  ball is  undeformed,  the $d(V_i,V_j)$  are  distances between  neighboring
vertices, and the $l_{ij}$ are distances in the reference shape. In other words,
the  additional  term   $\sum_{ij}  ||d(V_i,V_j)-l_{ij}||^2$  in  the  objective
function penalizes changes in edge-length but does not prevent them. 
Since the horizontal speed of the ball is approximately 140 mph before impact and
still 70 mph  afterwards, we  neglect gravity over  this very short  sequence and
assume the ball keeps moving in a straight line.  We therefore fit a line to the
center  of  the  ball  in  the  frames  in  which  it  is  undeformed  and  take
$\text{Traj}(\bc)$ to be the distance of the center of gravity from this line.

The results are shown in Fig.~\ref{fig:baseball}.  As in the case of the sail of
Fig.~\ref{fig:sail},  we  had  a  second  sequence  that  we  did  not  use  for
reconstruction purposes but used instead  for validation purposes.  To this end,
we  performed   dense  stereo  reconstruction~\cite{Furukawa09b}   and  show  in
Fig.~\ref{fig:BaseballAccuracy}  the median  distance between  the  resulting 3D
point cloud and  our monocular result. The undeformed  ball diameter
is 73.52 mm
and  the median distance  hovers around  1\% of  that value  except at  the very
beginning of the sequence when the ball is still undeformed but on the very left
side of the frame.   In fact, in those first few frames, it is the stereo
  algorithm that is slightly imprecise,  presumably because the line of sight is
  more slanted.  This  indicates that it might not be  perfectly accurate in the
  remaining  frames   either,  thus  contributing   to  some  of   the
  disagreement between the two algorithms.

\begin{figure}
\centering
\includegraphics[width=0.48\textwidth]{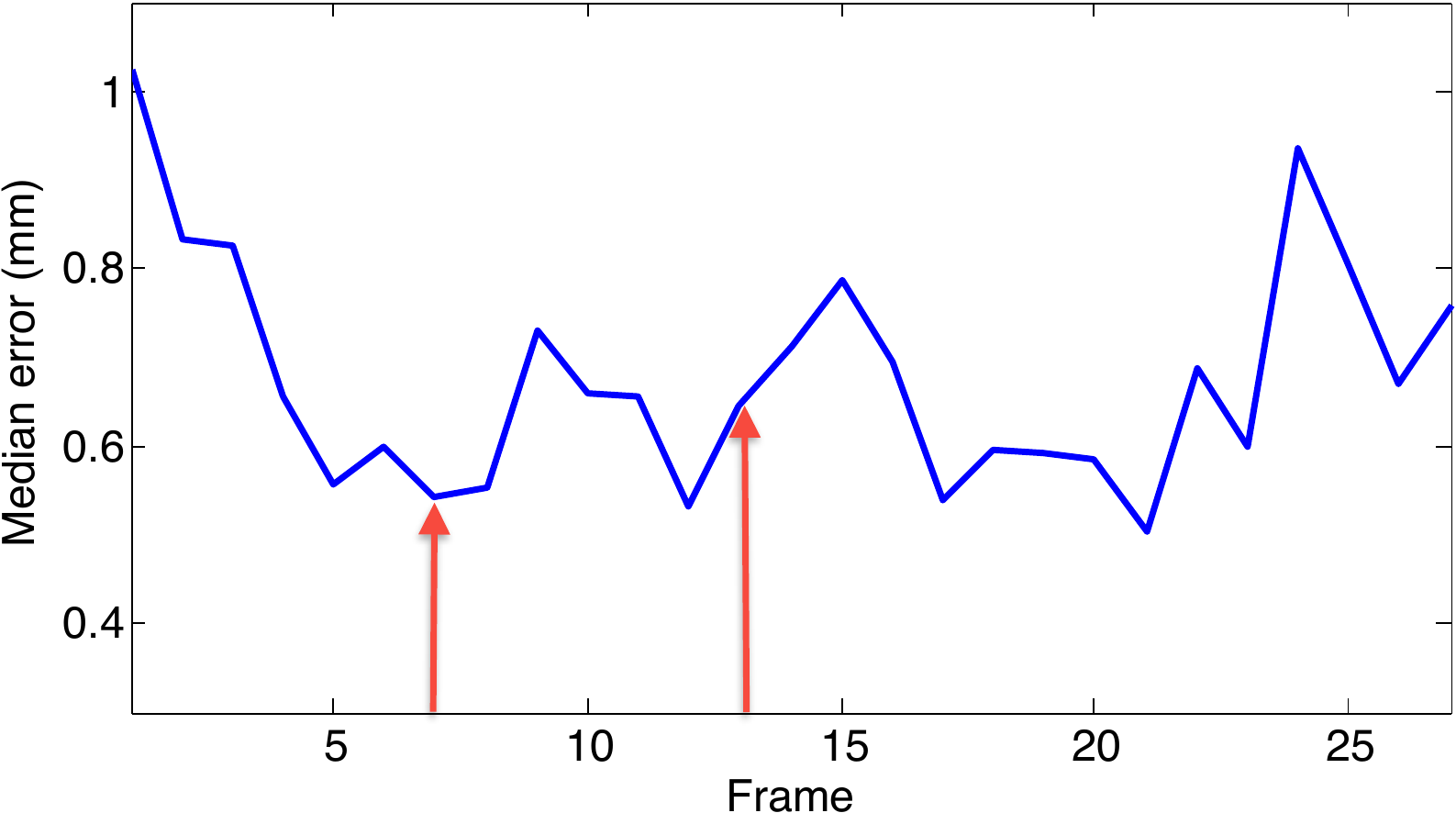}
\caption{Median  distance in  each frame  between our  monocularly reconstructed
  surface and a 3D point  cloud obtained from stereo~\cite{Furukawa09b}. The two
  red arrows  indicate the frames in which  the ball first and  last touches the
  bat. }
\label{fig:BaseballAccuracy}
\vspace{-.5cm}
\end{figure}

\subsection{Real Time Implementation}
\label{sec:RealTime}

\begin{figure*}
\centering
\begin{tabular}{ccccc}
\includegraphics[height=2.6cm]{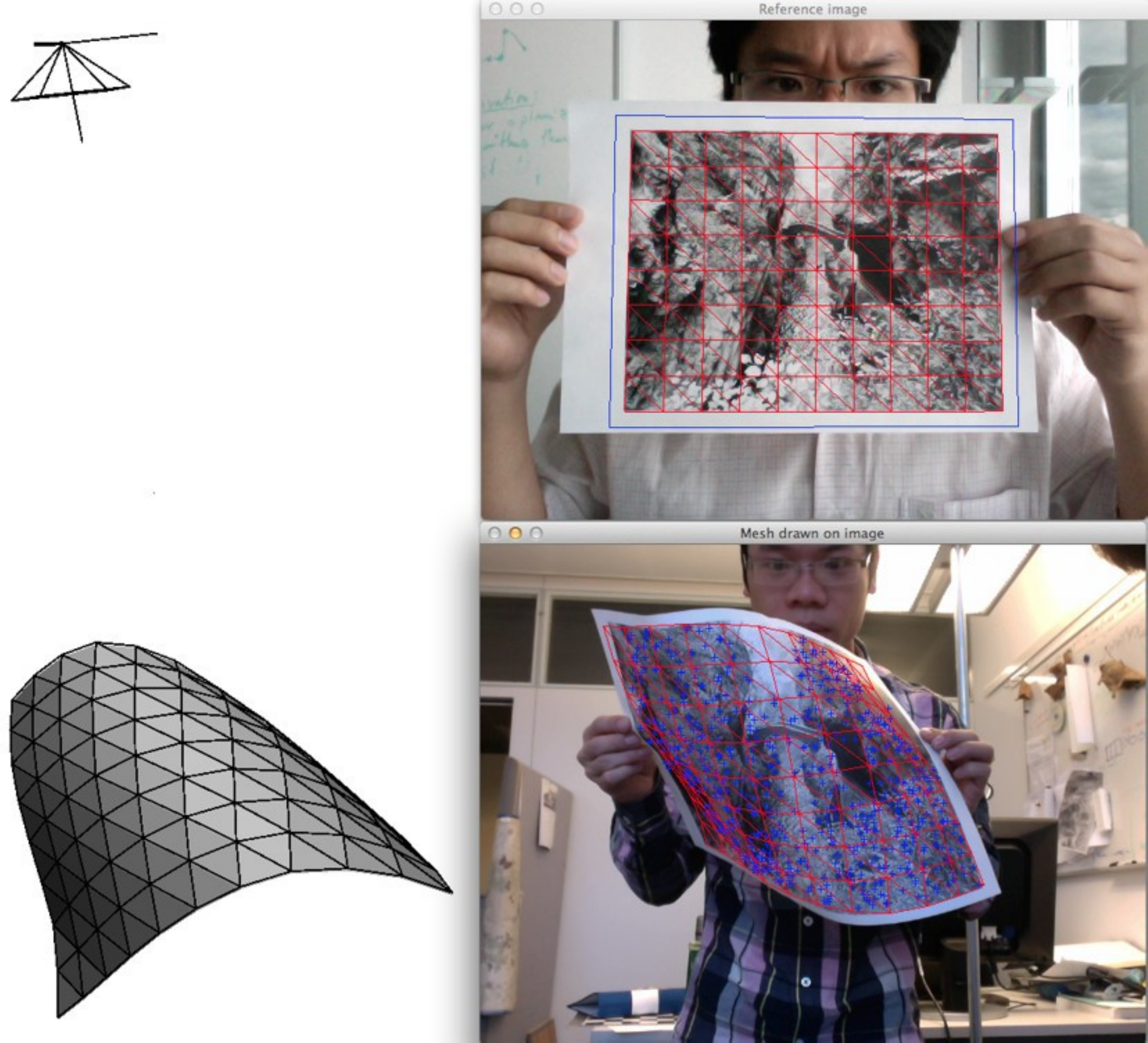} & 
\includegraphics[height=2.6cm]{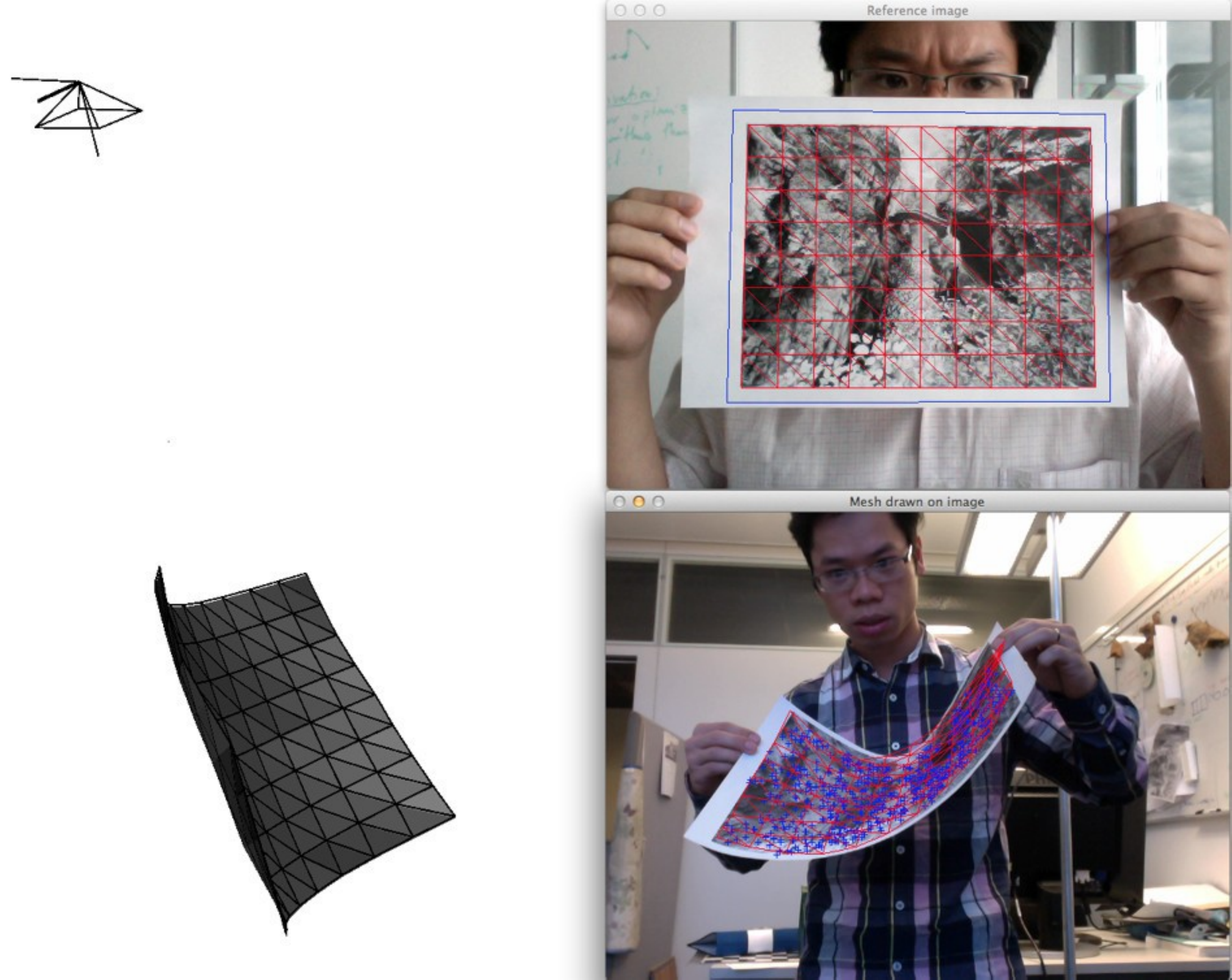} & 
\includegraphics[height=2.6cm]{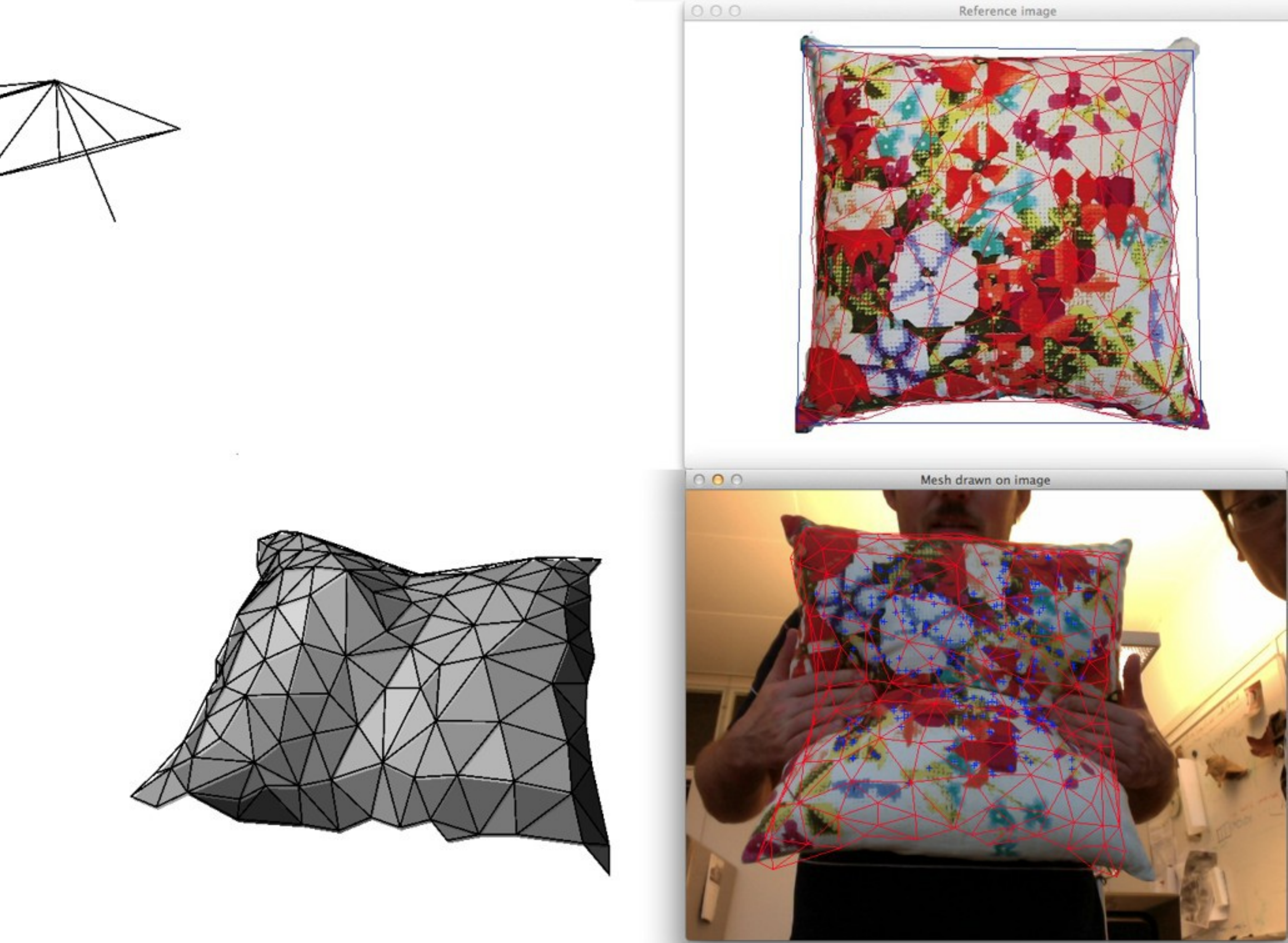} & 
\includegraphics[height=2.6cm]{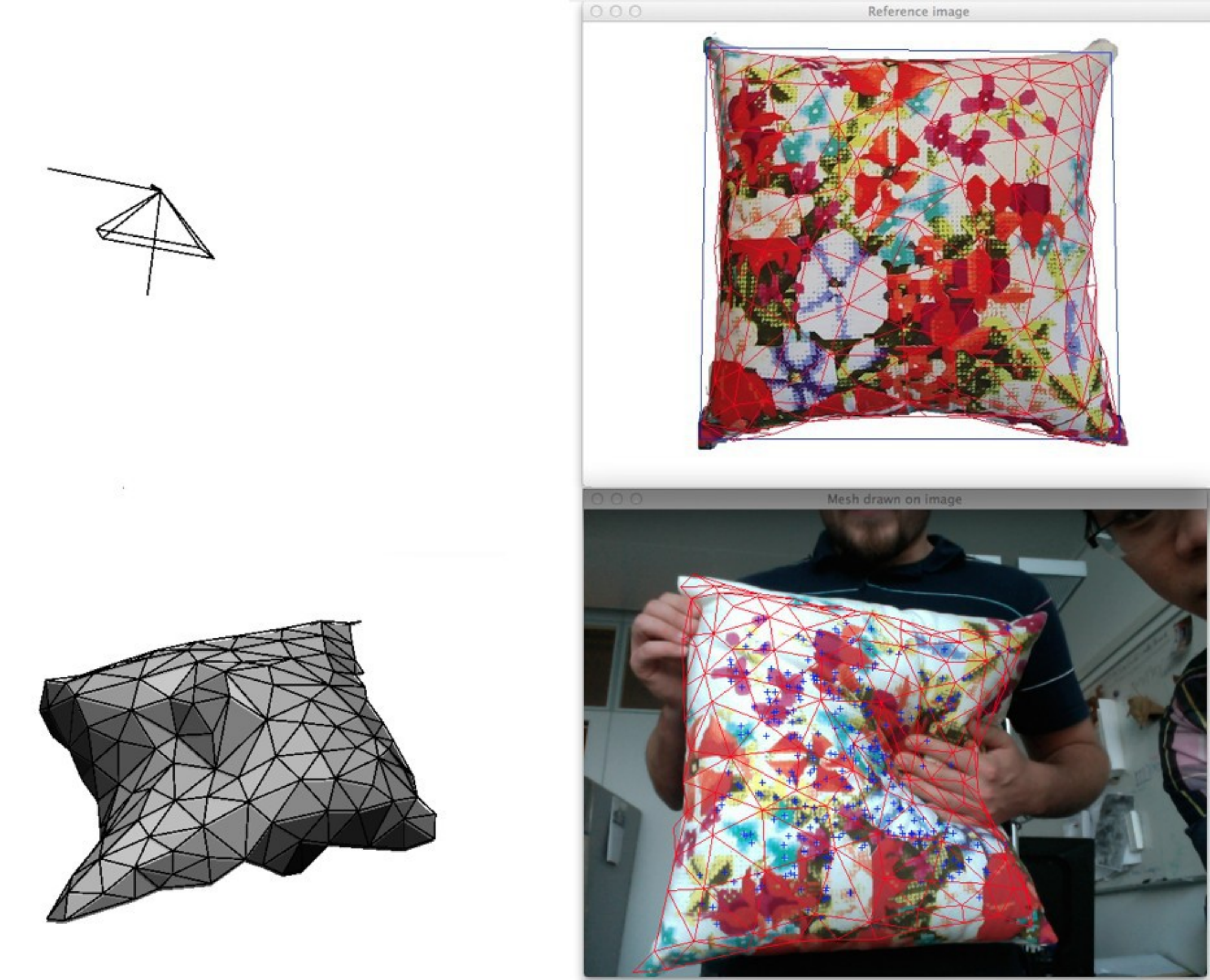} & 
\includegraphics[height=2.6cm]{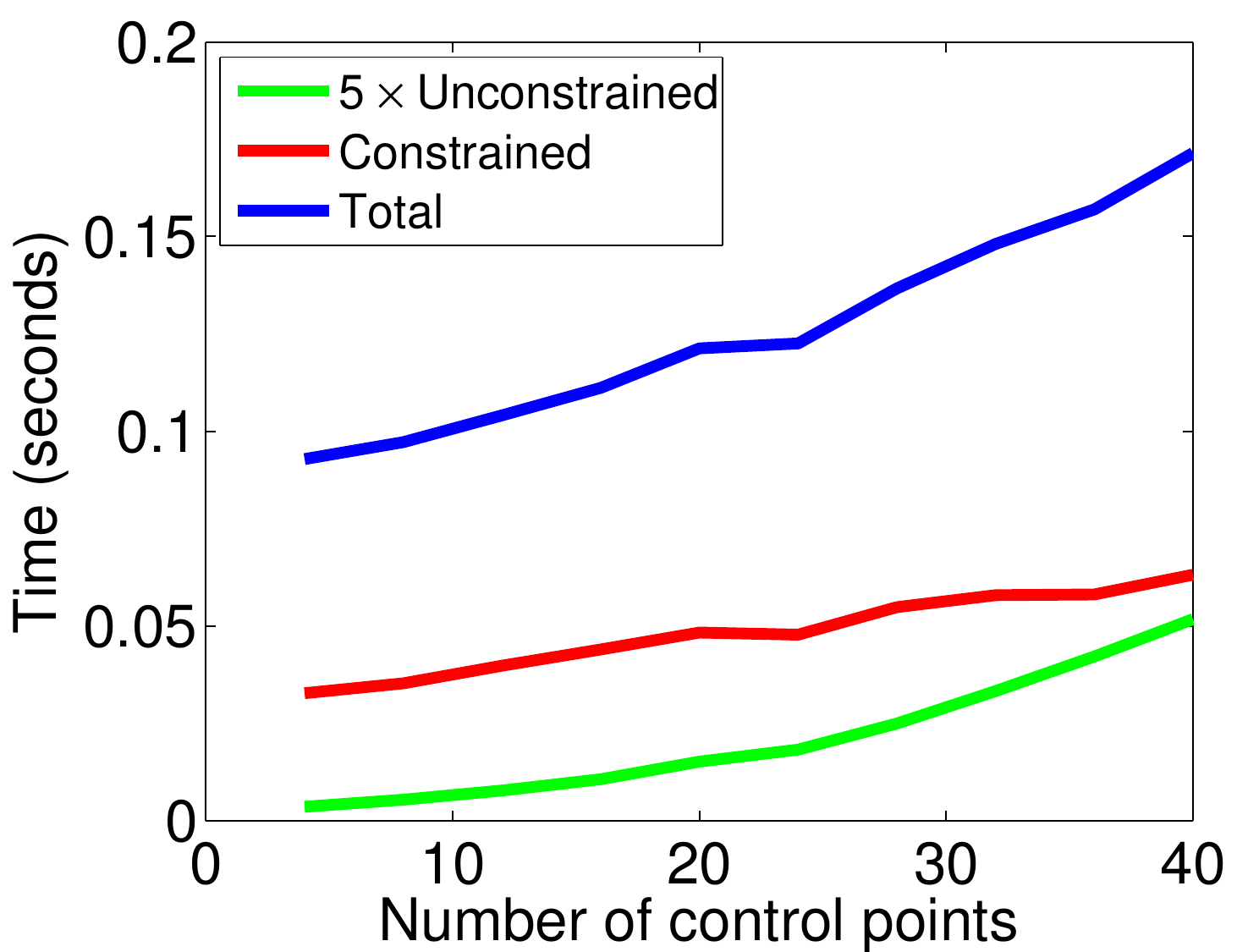} \\ 
(a) & (b) & (c) & (d) & (e)
\end{tabular}
\caption{{\bf Real-Time  Implementation.}  (a,b) Screen-shots  for two different
  configurations  of a  piece of  paper.  (c,d)  Screen-shots for  two different
  configurations  of a  cushion.  (e)  Average  computation times  to solve  the
  linear  system of  Eq.~\ref{eq:UnconstrOpt} five  times to  eliminate outliers
  (green),  to  do  this  and  solve the  constrained  minimization  problem  of
  Eq.~\ref{eq:ConstrOpt} (red),  and to  perform the full  computation including
  keypoint  extraction and  matching (blue).   These  times are  expressed as  a
  function of the number of control vertices. }
\label{fig:RealTime}
\end{figure*}

We have  incorporated our  approach to surface  reconstruction into  a real-time
demonstrator  that we showed  at the  CVPR'12 and  ECCV'12 conferences.   It was
running at speeds of  up to 10 frame-per-second on a MacBook  Pro on $640 \times
480$  images  acquired by  the  computer's webcam,  such  as  those depicted  by
Fig.~\ref{fig:RealTime}  that feature  a planar  sheet of  paper  and non-planar
cushion such as those we used  in our earlier experiments.  The videos we supply
as supplementary material  demonstrate both the strengths and  the limits of the
approach: When there  are enough correspondences, such as on  the sheet of paper
or in the  center of the cushion,  the 3D reconstruction is solid  even when the
deformations are severe.  By contrast, on the edges of the cushion, we sometimes
lose track  because we  cannot find enough  correspondences anymore.   To remedy
this, a natural research direction would  be to look into using additional image
clues such as those provide  by shading and contours, while preserving real-time
performance.

Our  C++  implementation relies  on  a  feature-point  detector to  extract  500
interest points from the reference and  2000 ones from input images as maxima of
the Laplacian and then uses the Ferns classifier~\cite{Ozuysal10} to match them.
Additionally,  since  our algorithm  can  separate  inlier correspondences  from
spurious ones, we track the former from images to images to increase our pool of
potential correspondences,  which turns  out to be  important when  the surface
deformations are large.

Given the  correspondences in the  first input image  of a sequence, we  use the
outlier rejection  scheme of Section~\ref{sec:robust} to  eliminate the spurious
one and obtain a 3D shape estimate. As long as the deforming surface is properly
tracked, we can then simply estimate  the 3D shape in subsequent frames by using
the  3D  shape  estimate in  a  previous  frame  to initialize  the  constrained
minimization of Eq.~\ref{eq:ConstrOpt}, without having to solve again the linear
system of Eq.~\ref{eq:UnconstrOpt}.  However, if the system loses  track, it can
reinitialize itself by running the complete procedure again.

Real-time performance is made possible by  the fact that the 3D shape estimation
itself  is fast  enough  to represent  only  a small  fraction  of the  required
computation,  as shown  in  Fig.~\ref{fig:RealTime}(e).  A  further increase  in
robustness and accuracy  could be achieved by enforcing  temporal consistency of
the          reconstructed         shape          from          frame         to
frame~\cite{Torresani08a,Russell11,Salzmann11a}.

We will release the real-time code on the lab's website so  that fellow
researchers can  try it for themselves.


%% file: figs_capaperseq_table.tex
\newcommand{\capaperseqwidth}{0.20\linewidth}
\newcommand{\capaperseqheight}{1.8cm}
\begin{tabular}{cccc}
\includegraphics[width=\capaperseqwidth]{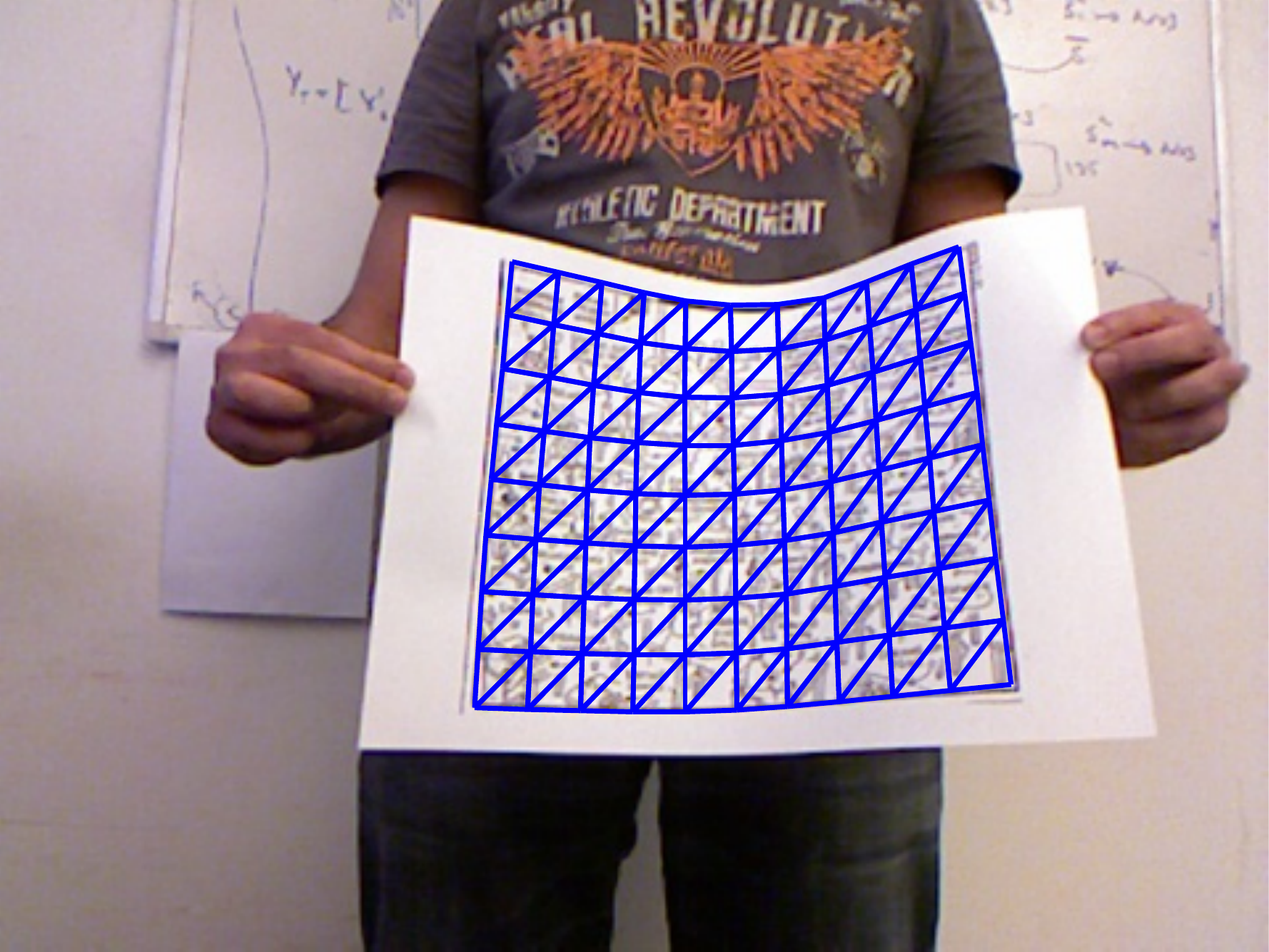} & \includegraphics[width=\capaperseqwidth]{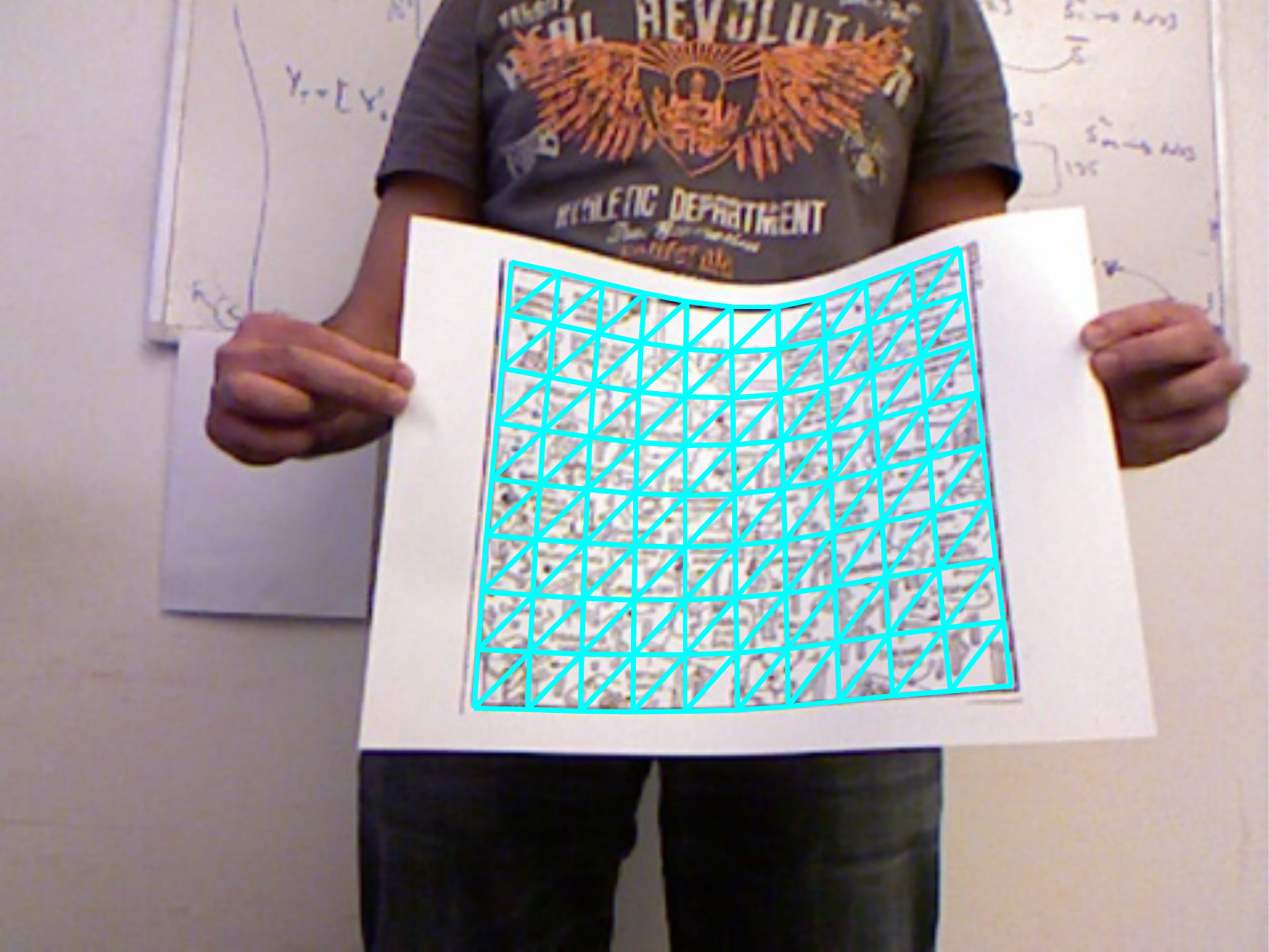} & \includegraphics[width=\capaperseqwidth]{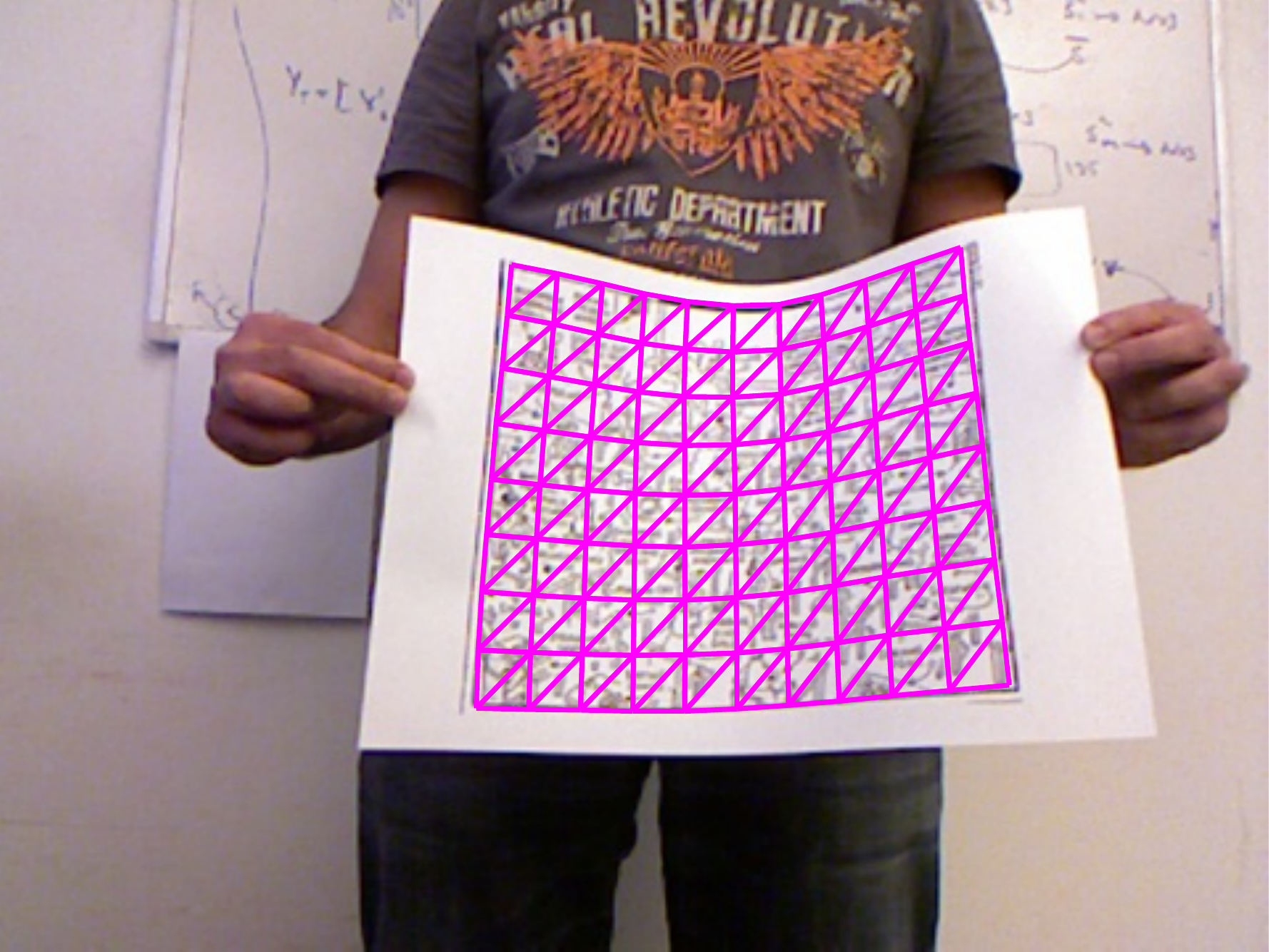} & \includegraphics[width=\capaperseqwidth]{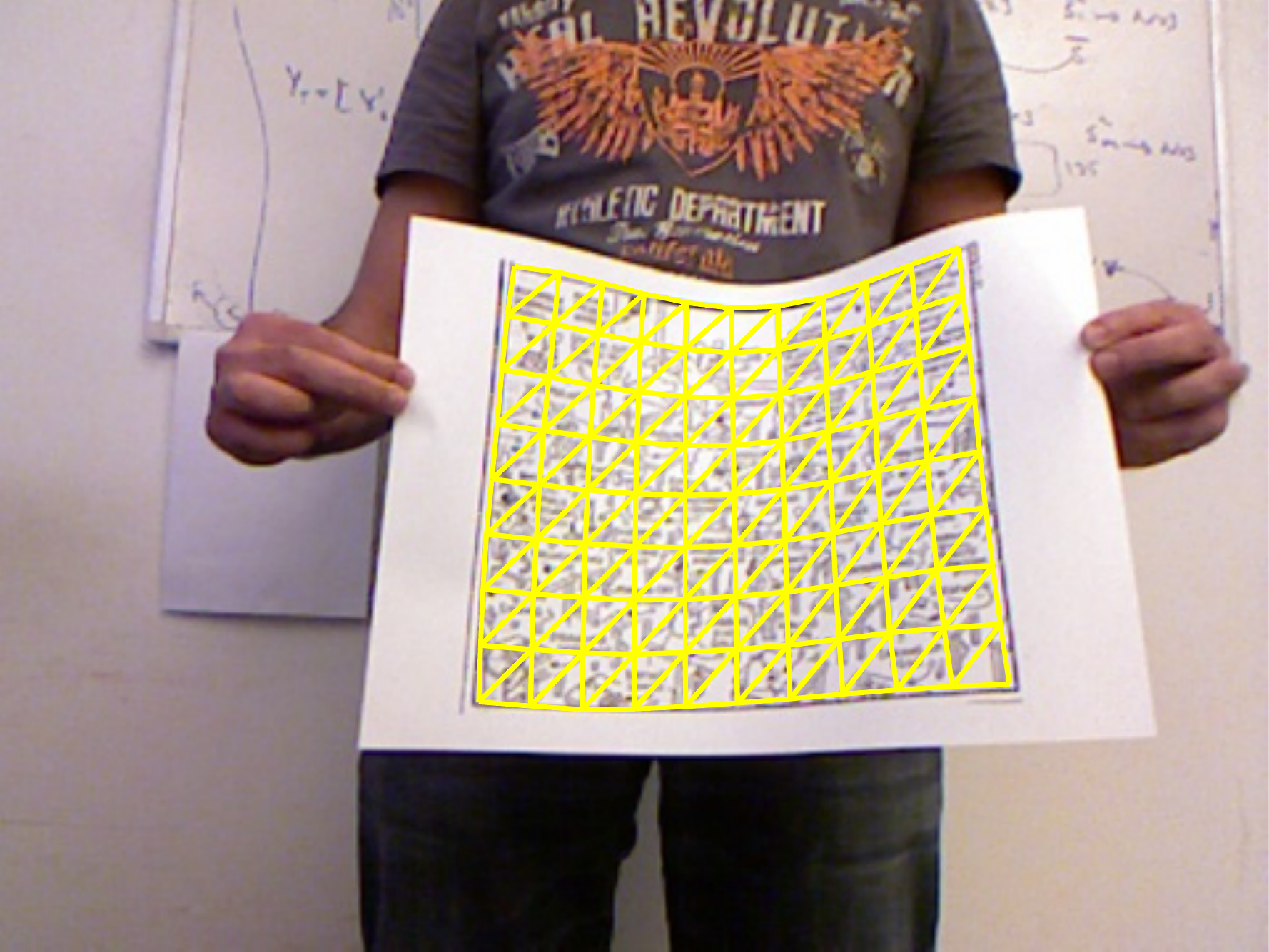}\\
\includegraphics[height=\capaperseqheight]{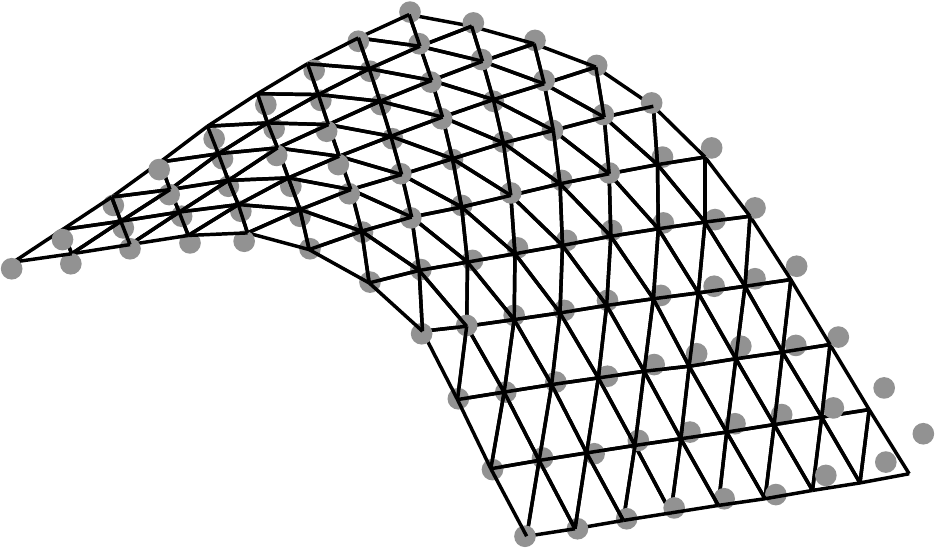} & \includegraphics[height=\capaperseqheight]{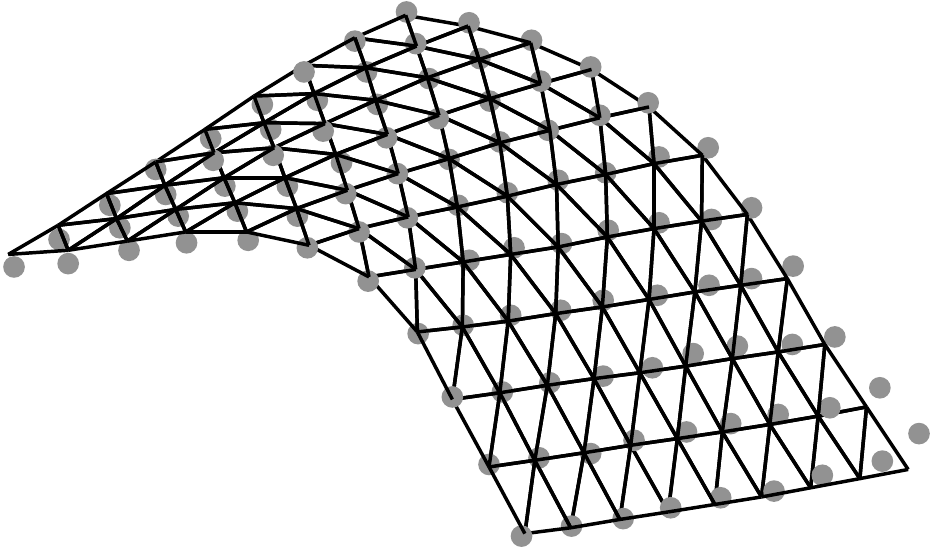} & \includegraphics[height=\capaperseqheight]{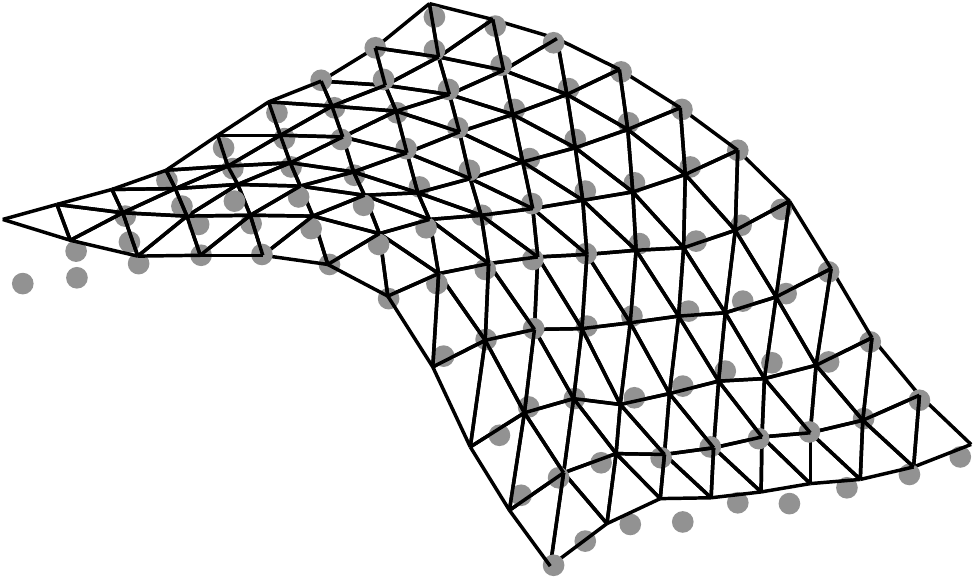} & \includegraphics[height=\capaperseqheight]{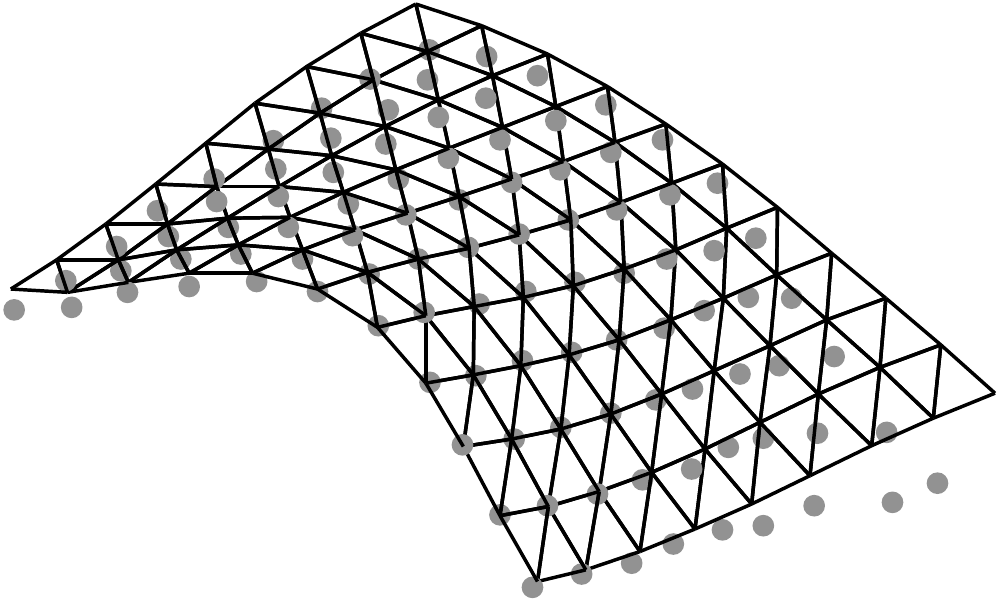}\\
\end{tabular}

%% file: figs_apronseq_table.tex
\newcommand{\apronseqwidth}{0.20\linewidth}
\newcommand{\apronseqheight}{2.3cm}
\begin{tabular}{cccc}
\includegraphics[width=\apronseqwidth]{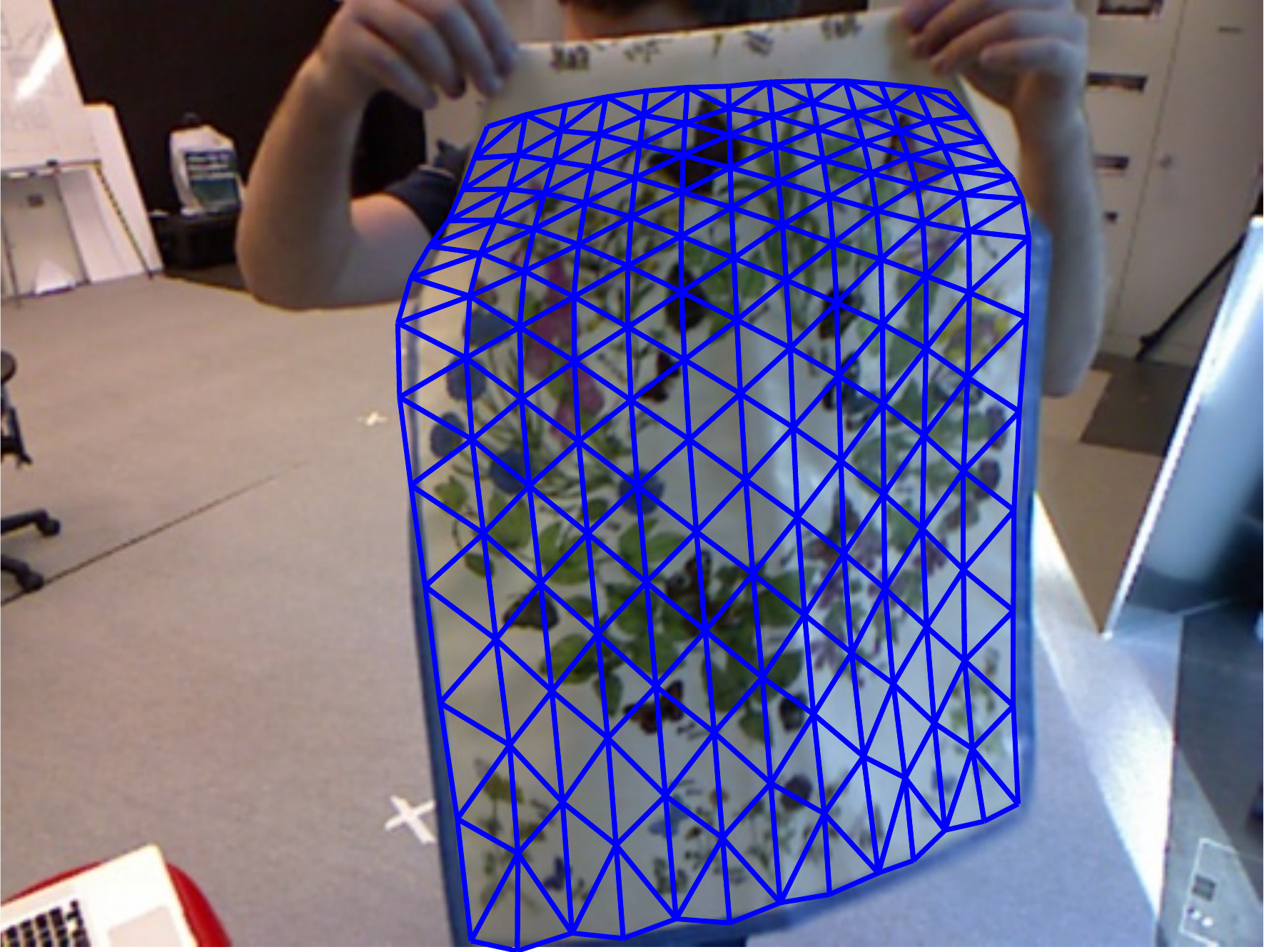} & \includegraphics[width=\apronseqwidth]{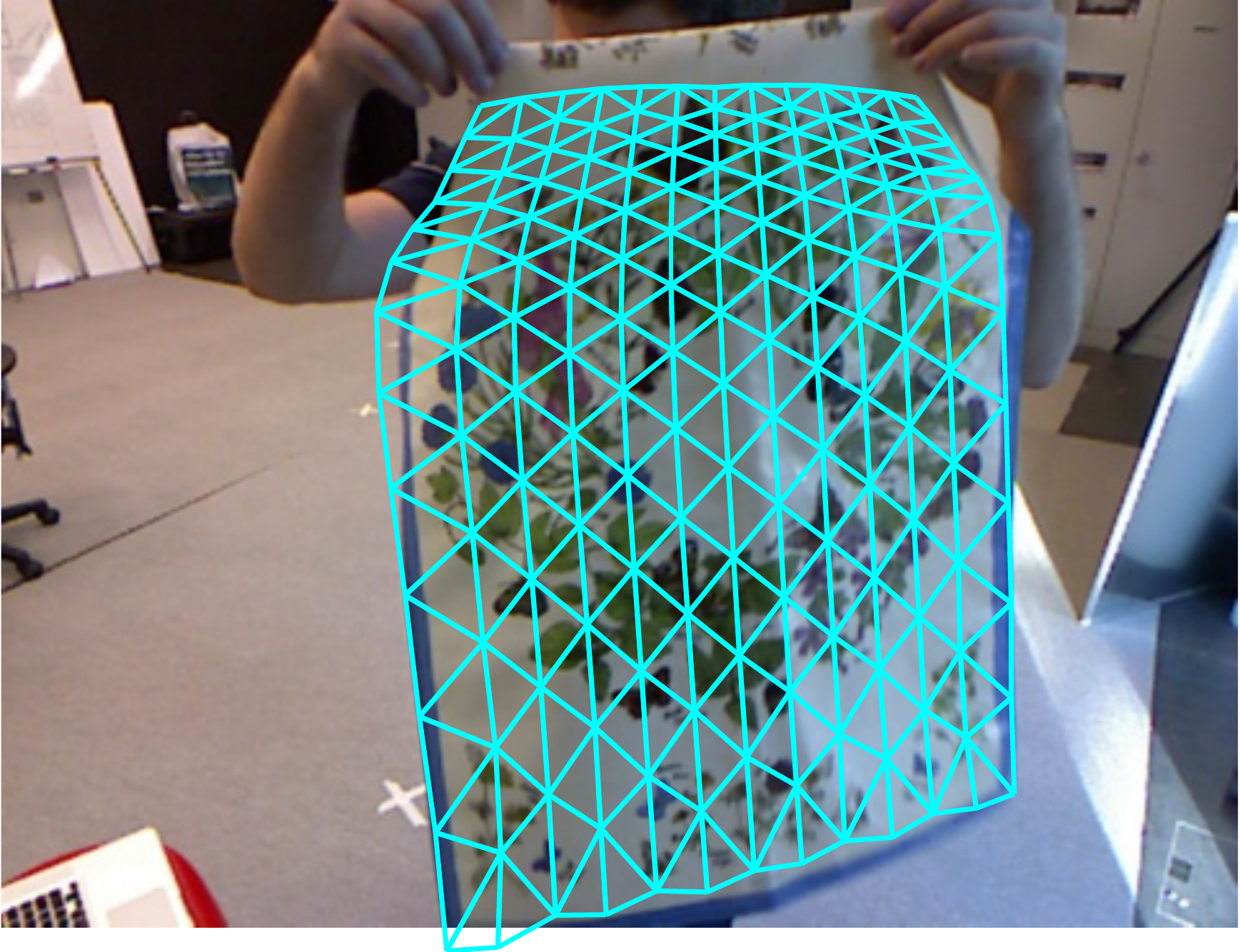} & \includegraphics[width=\apronseqwidth]{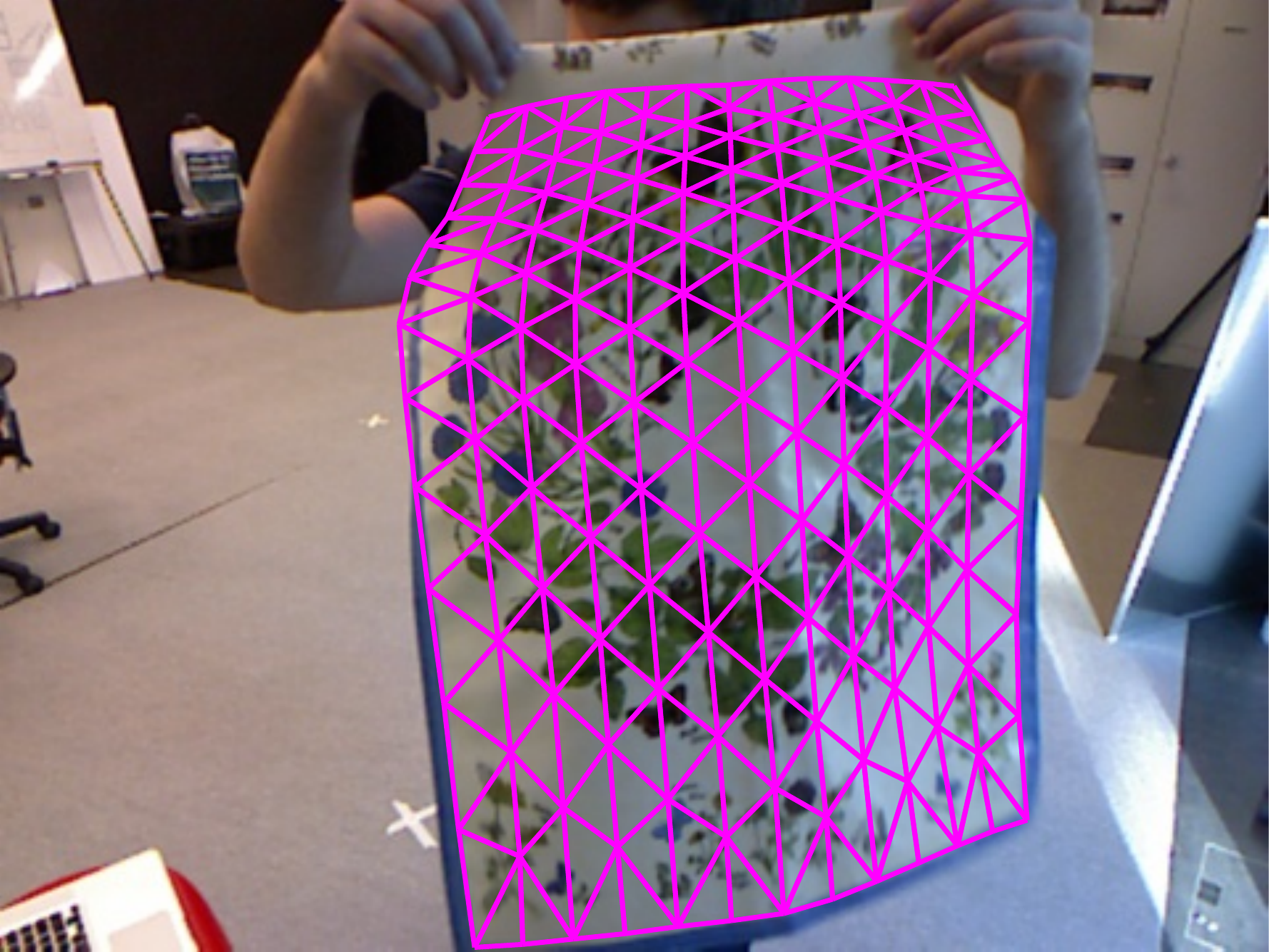} & \includegraphics[width=\apronseqwidth]{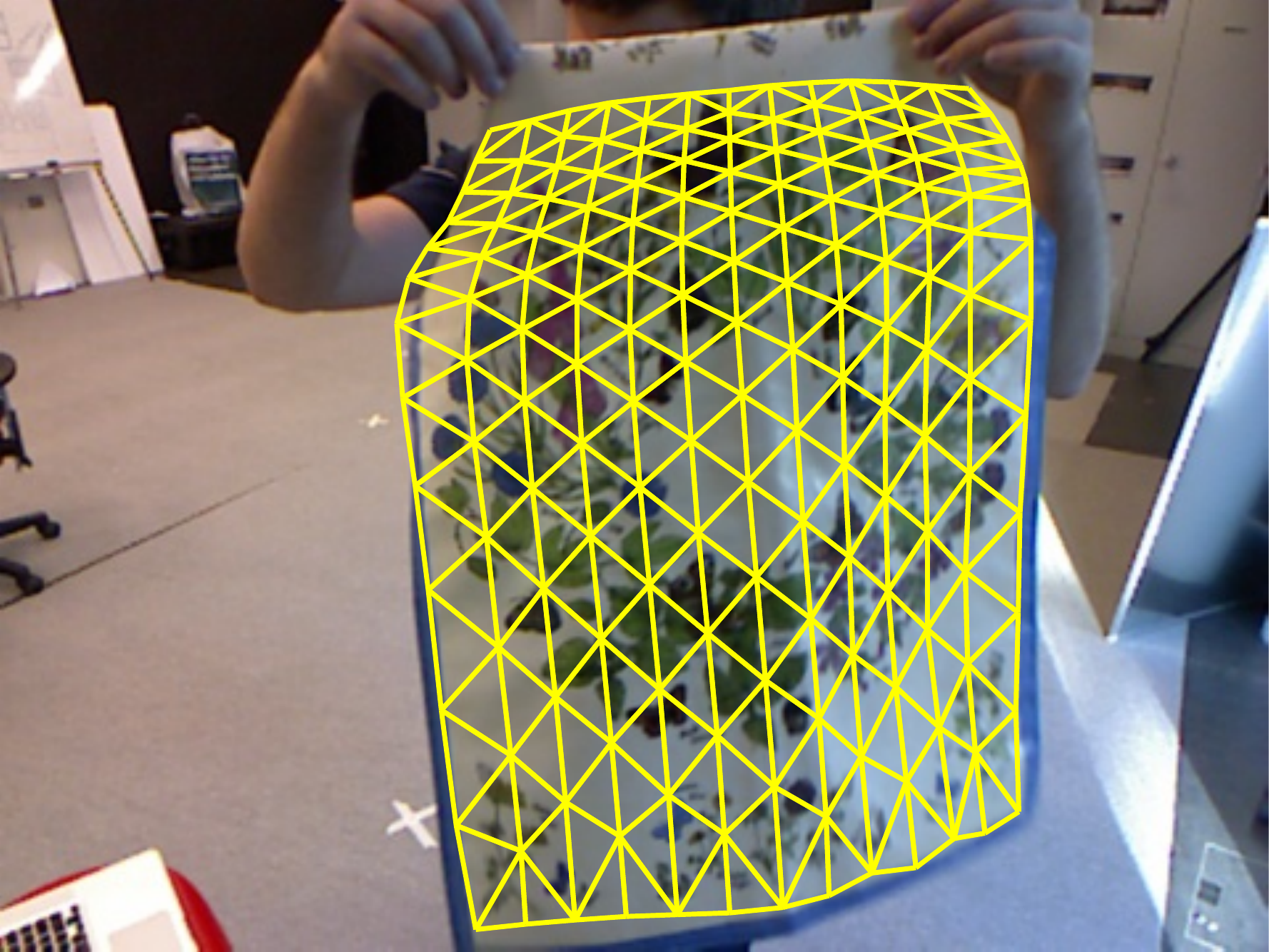}\\
\includegraphics[height=\apronseqheight]{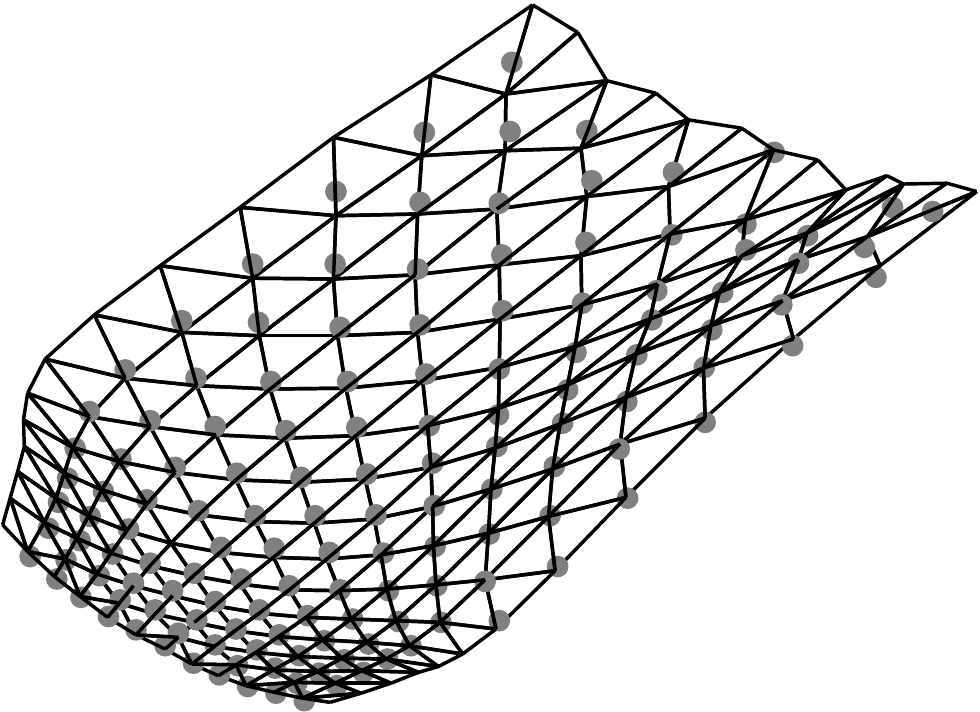} & \includegraphics[height=\apronseqheight]{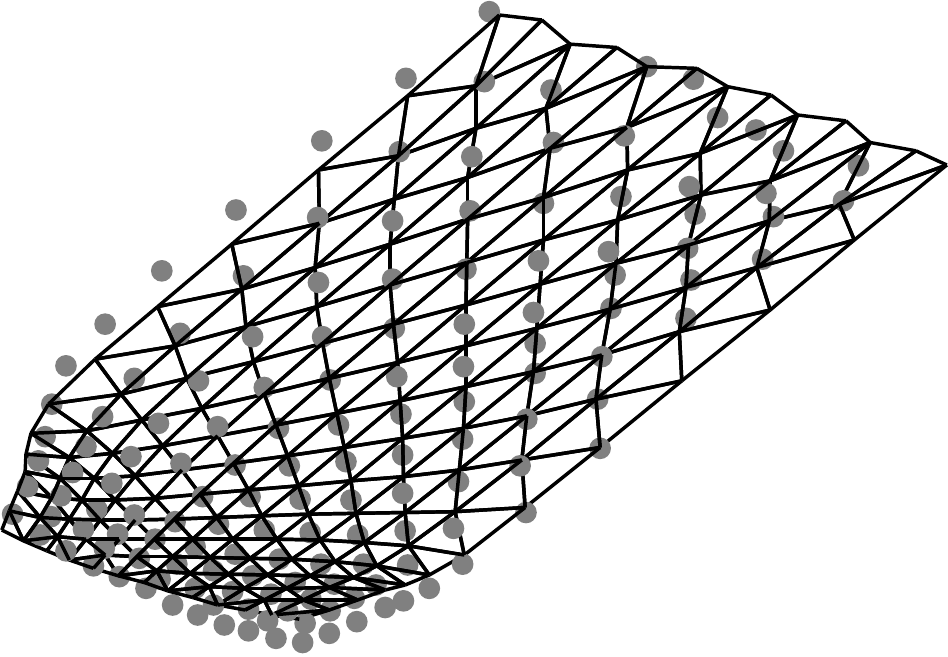} & \includegraphics[height=\apronseqheight]{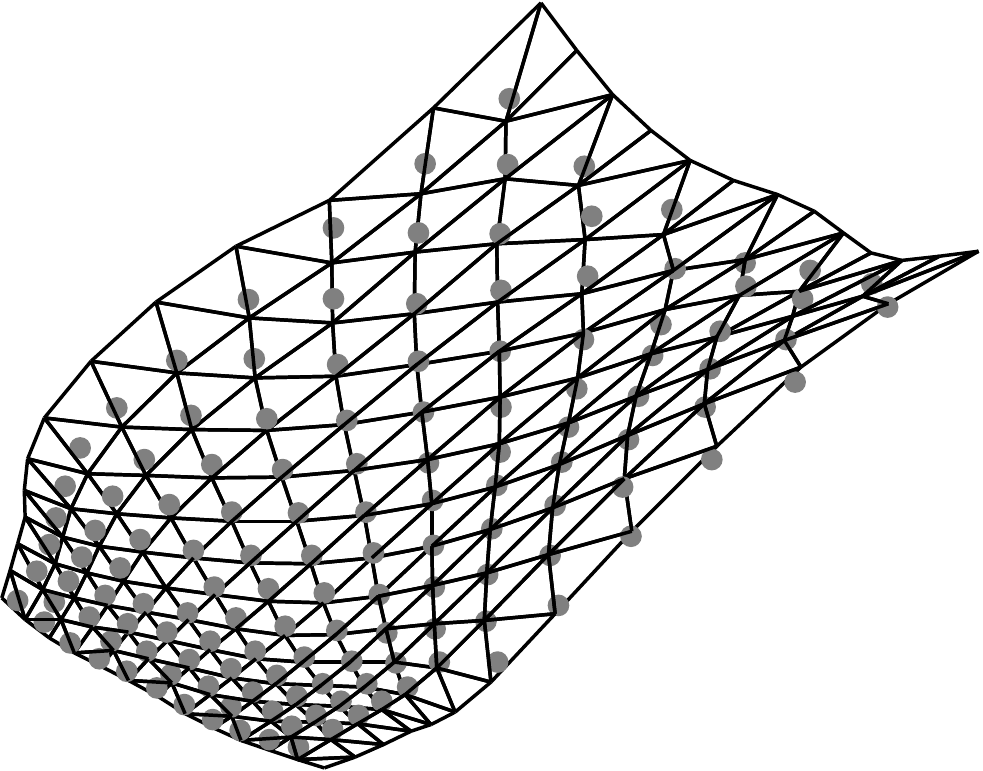} & \includegraphics[height=\apronseqheight]{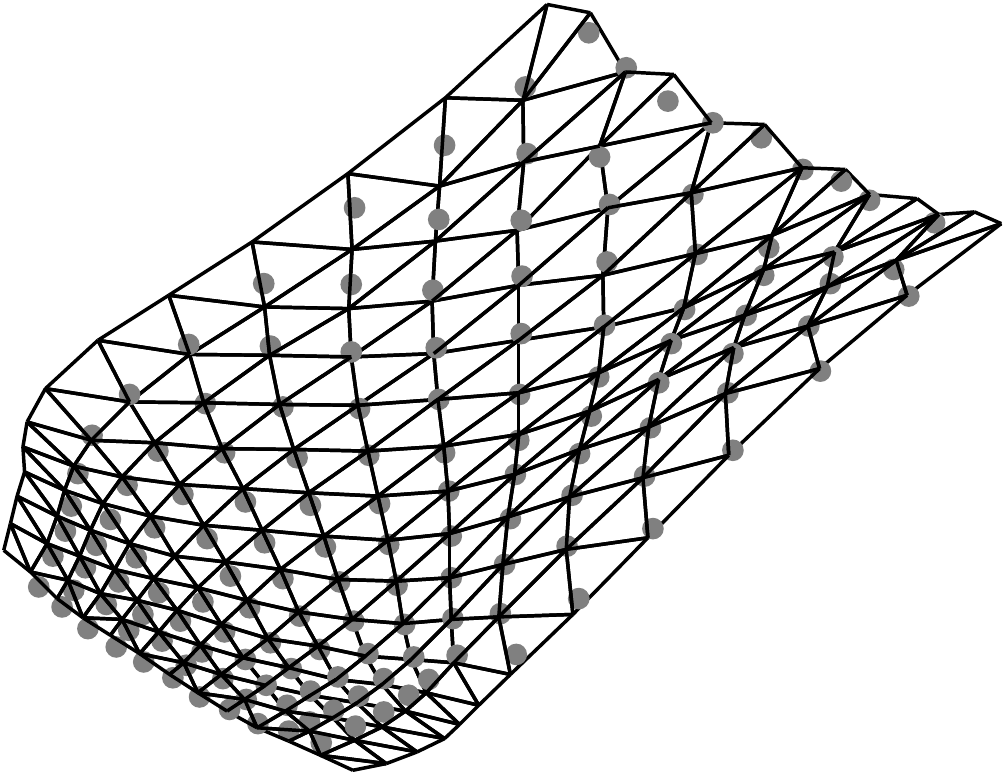}\\
(a) & (b) & (c) & (d)
\end{tabular}

%% file: figs_cushionseq_table.tex
\newcommand{\cushionseqwidth}{0.21\linewidth}
\newcommand{\cushionseqheight}{2.0cm}
\newcommand{\cuspaddingwidth}{0.05\linewidth}
\begin{tabular}{ccc}
\includegraphics[width=\cushionseqwidth]{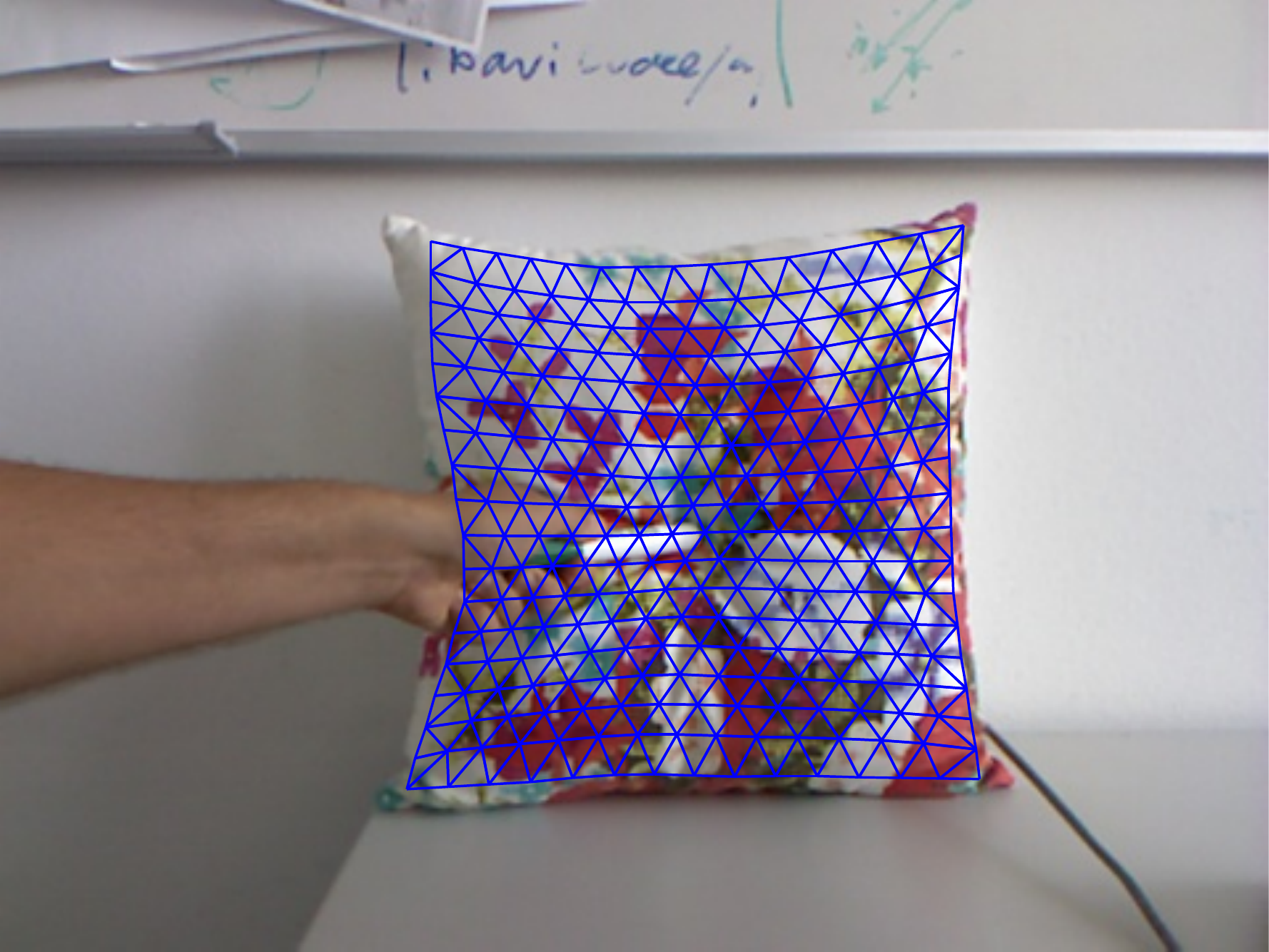} \hspace{\cuspaddingwidth} & \includegraphics[width=\cushionseqwidth]{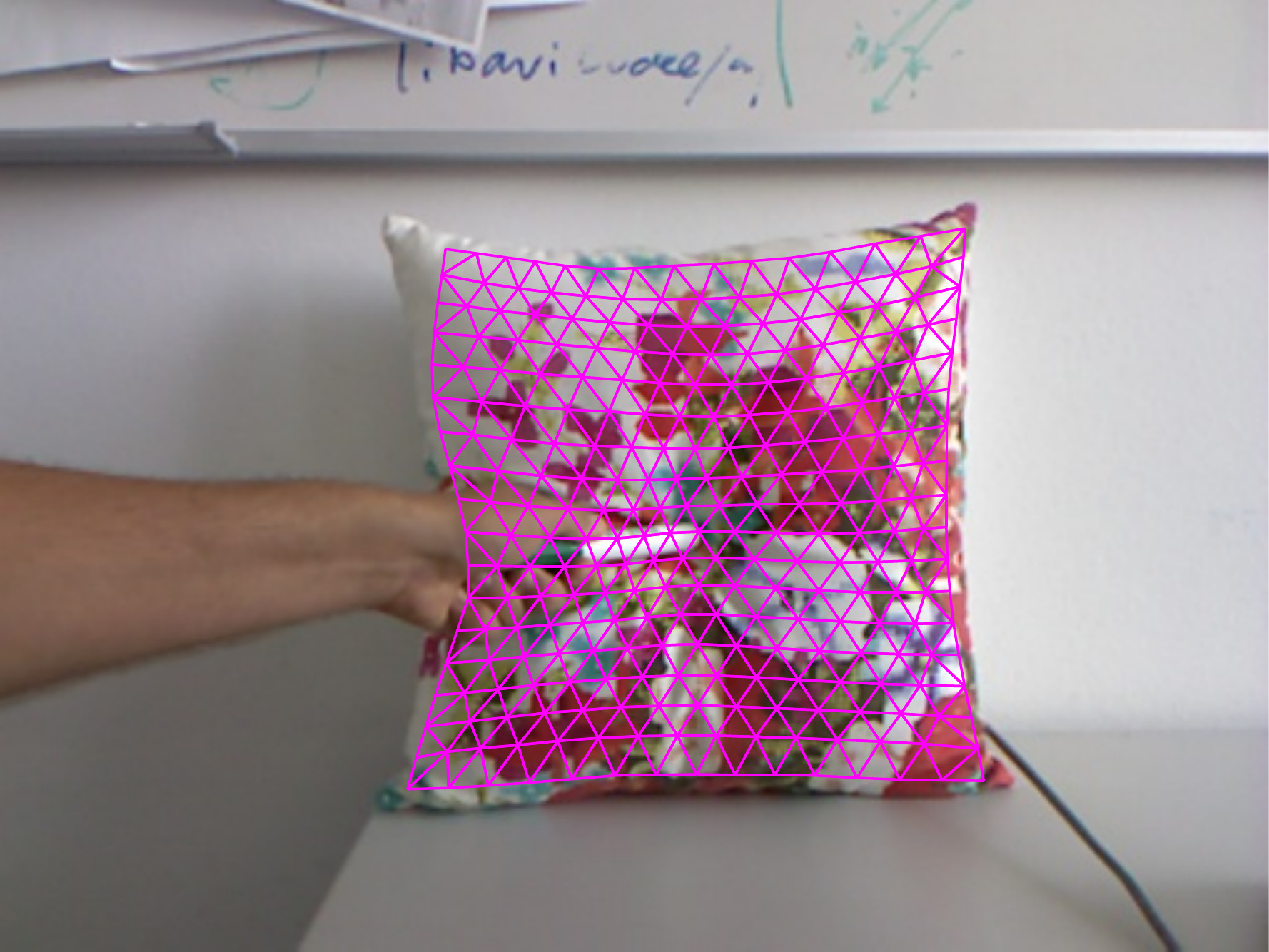} \hspace{\cuspaddingwidth} & \includegraphics[width=\cushionseqwidth]{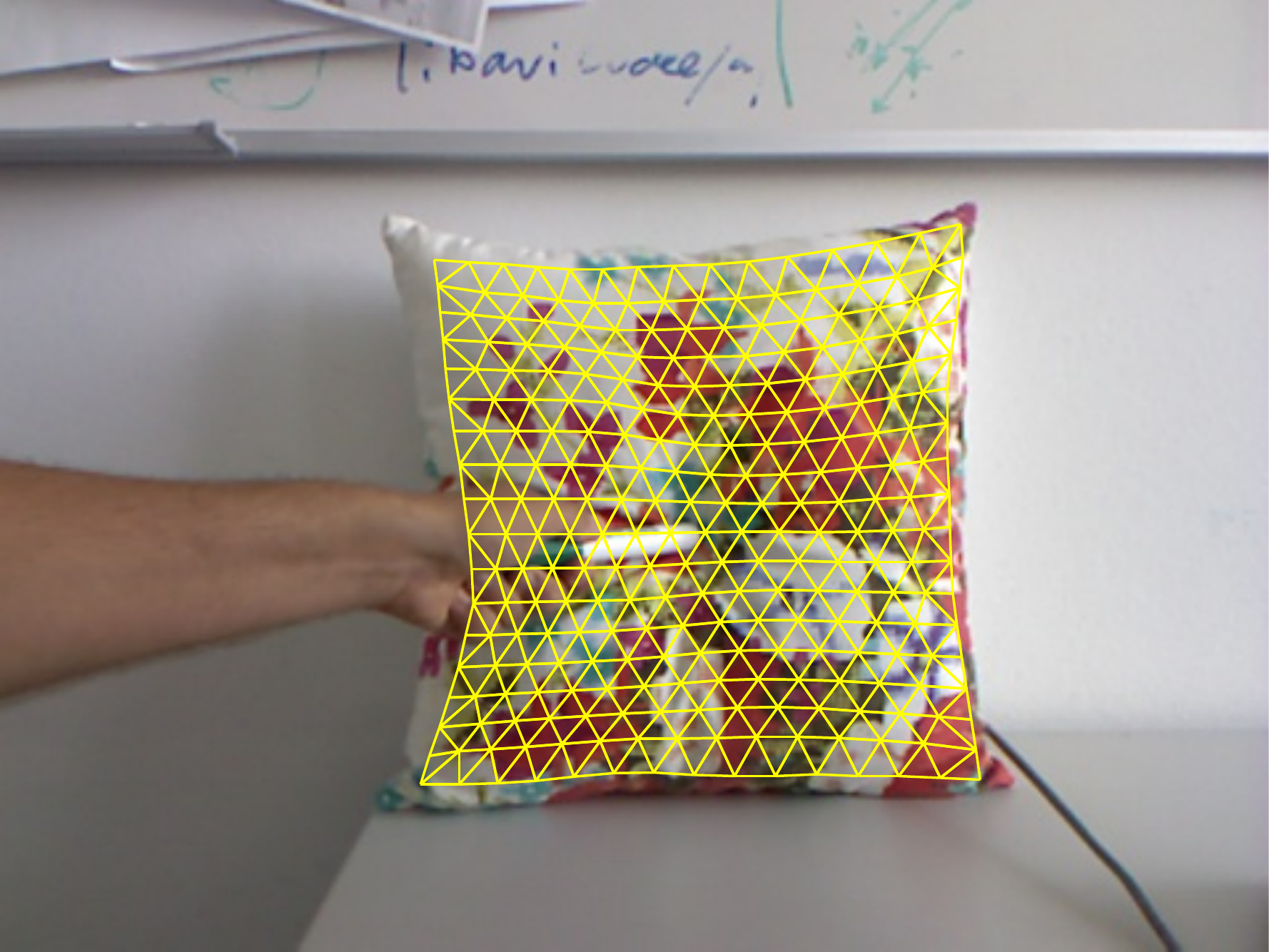}\\
\includegraphics[height=\cushionseqheight]{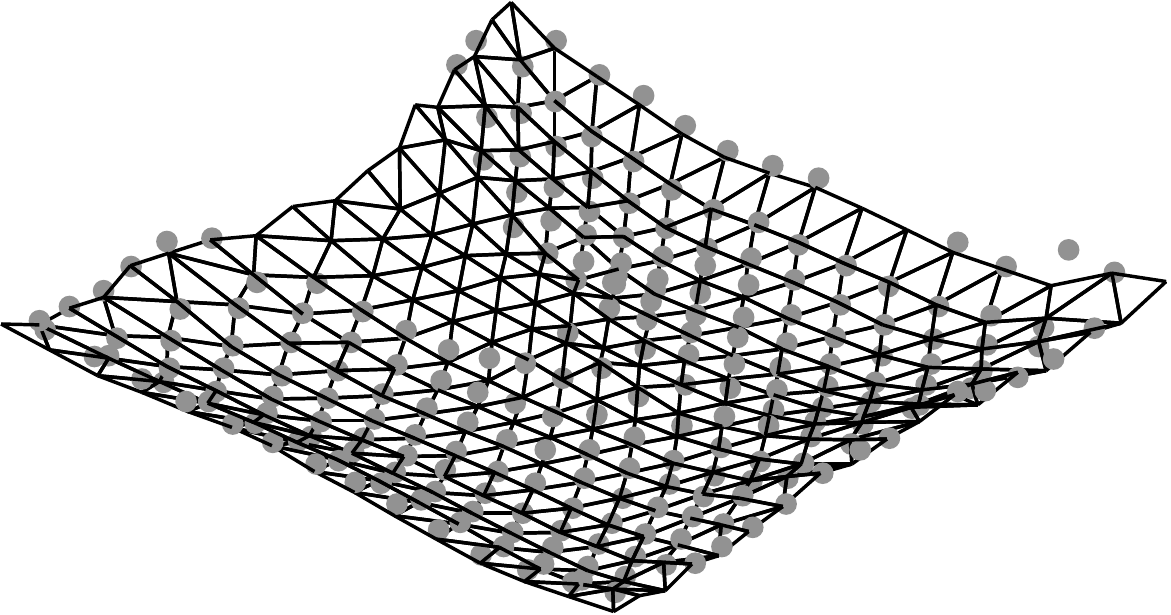} \hspace{\cuspaddingwidth} & \includegraphics[height=\cushionseqheight]{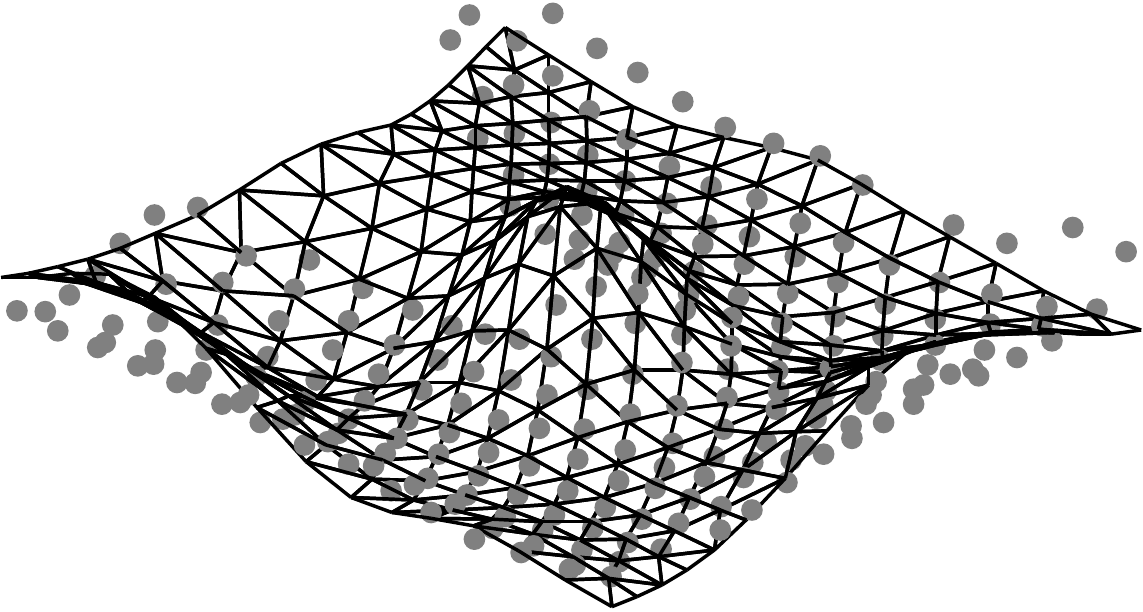} \hspace{\cuspaddingwidth} & \includegraphics[height=\cushionseqheight]{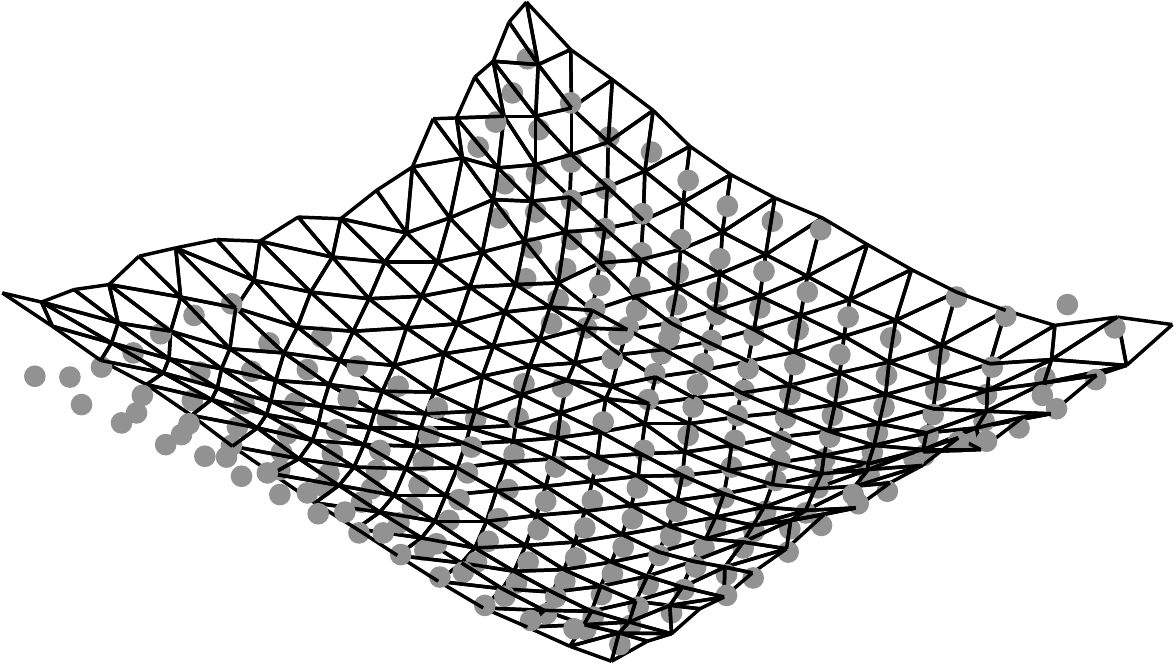}\\
\end{tabular}

%% file: figs_bananaseq_table.tex
\newcommand{\bananaseqwidth}{0.21\linewidth}
\newcommand{\bananaseqheight}{2.7cm}
\newcommand{\bananapaddingwidth}{0.05\linewidth}
\begin{tabular}{ccc}
\includegraphics[width=\bananaseqwidth]{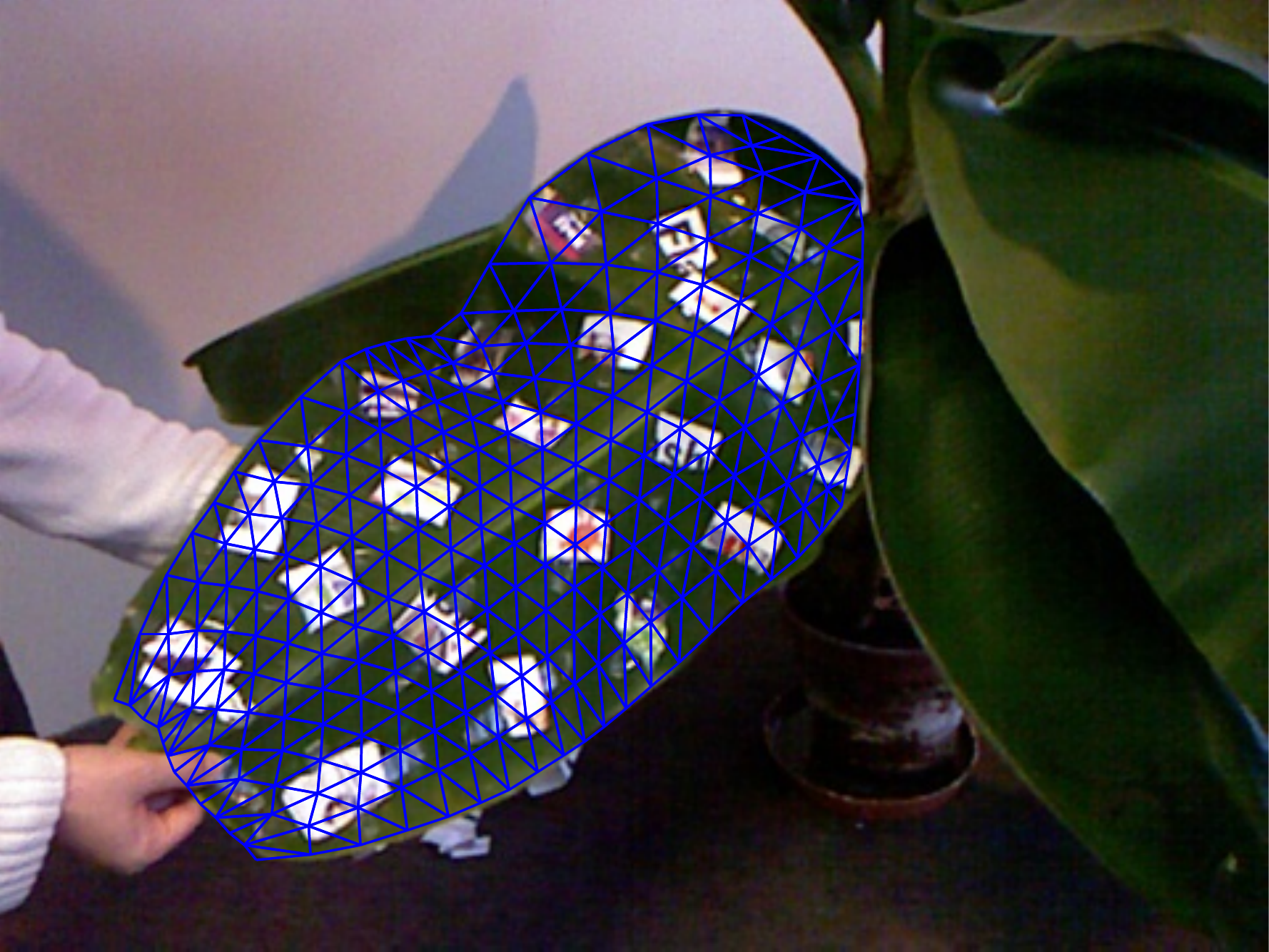} \hspace{\bananapaddingwidth} & \includegraphics[width=\bananaseqwidth]{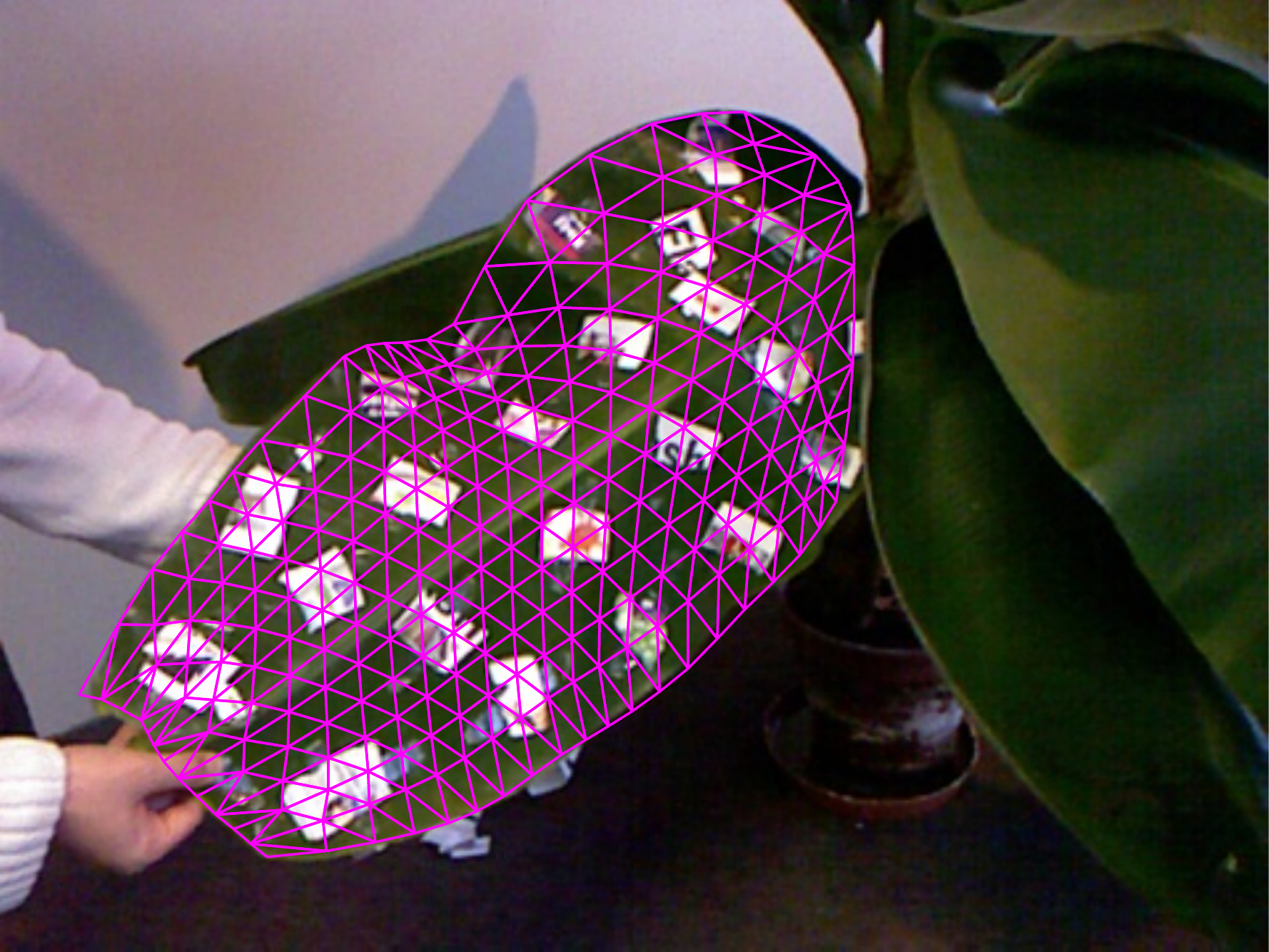} \hspace{\bananapaddingwidth} & \includegraphics[width=\bananaseqwidth]{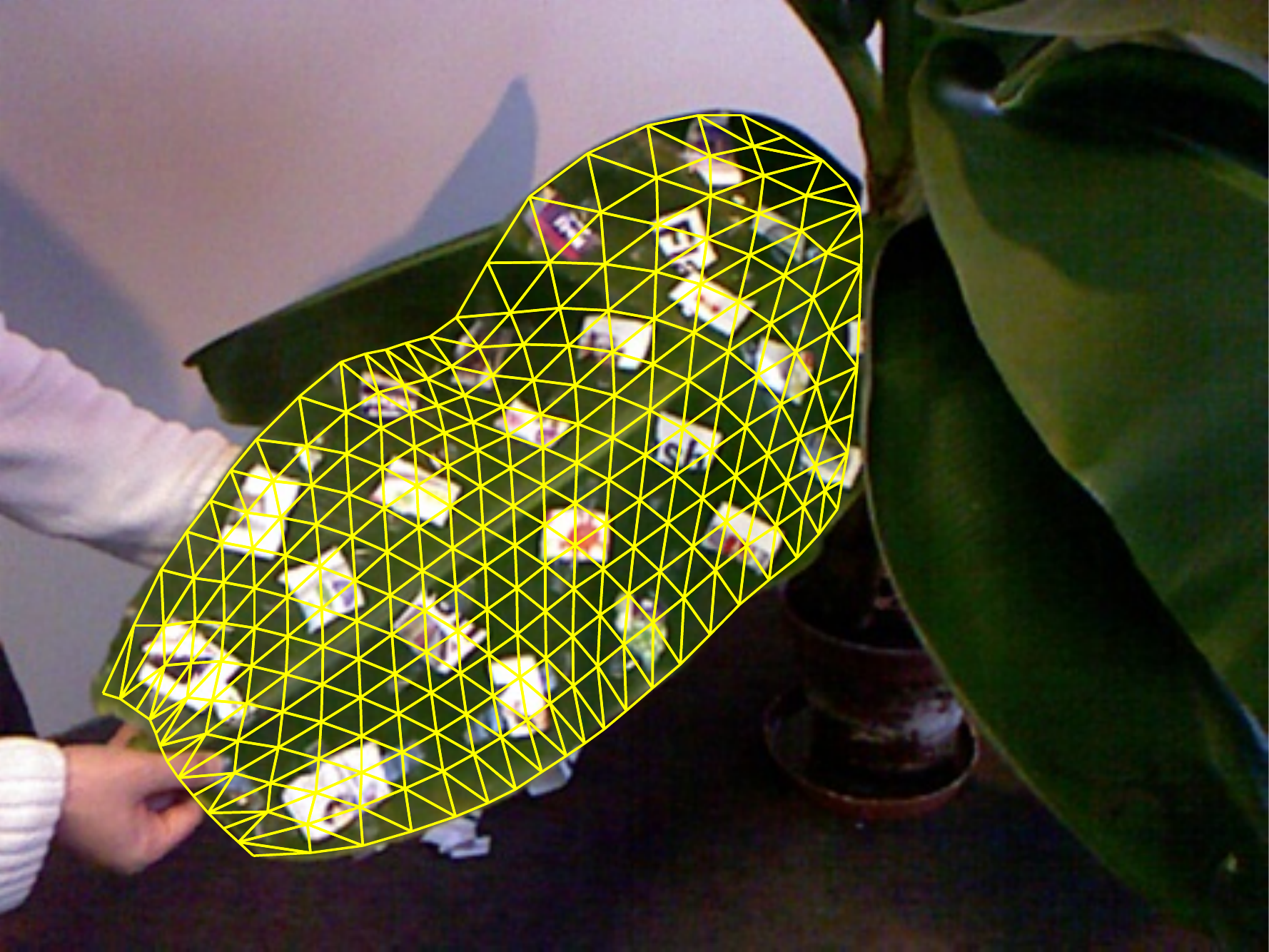}\\
\includegraphics[height=\bananaseqheight]{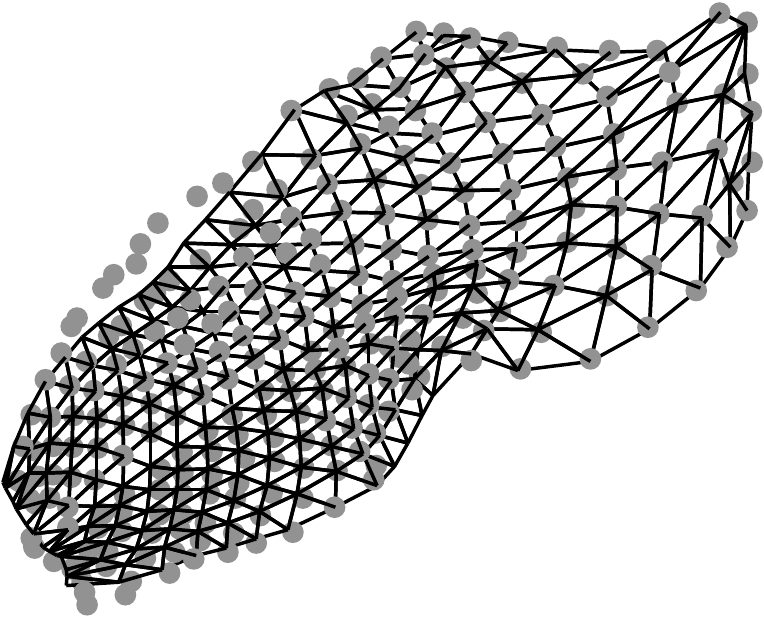} \hspace{\bananapaddingwidth} & \includegraphics[height=\bananaseqheight]{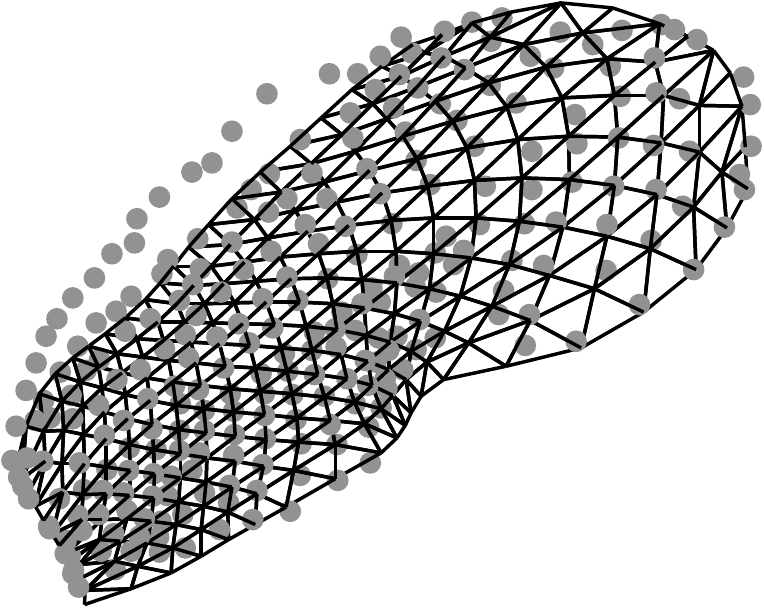} \hspace{\bananapaddingwidth} & \includegraphics[height=\bananaseqheight]{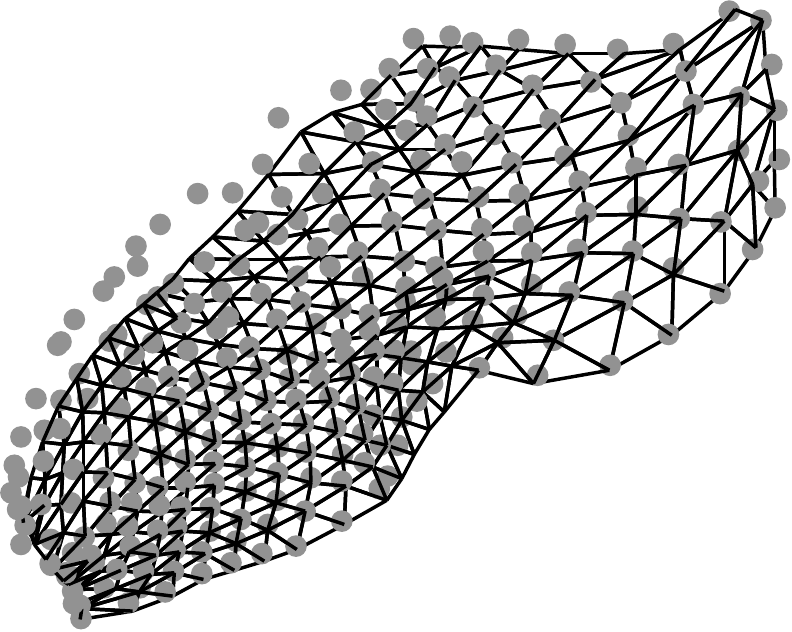}\\
(a) & (b) & (c)
\end{tabular}

%% file: figs_datasets.tex

\begin{table}
\caption{The five datasets used for quantitative
evaluation. Number of frames in each sequence and specifics of the
corresponding reference shape.}
\label{tab:Datasets}
\begin{center}
\begin{tabular}{rcccc}
{\bf Dataset} & {\bf \# frames} & {\bf \# vertices} & {\bf \# facets} &
{\bf Planar} \\ \hline
Paper   & 192 & 99 & 160 & Yes \\
Apron   & 160 & 169 & 288  & Yes \\
Cushion & 46 & 270 & 476  & No \\
Leaf    & 283 & 260 & 456  & No \\
Sail    & 6 & 66 & 100  & No
\end{tabular}
\end{center}
\vspace{-.6cm}
\end{table}

%% file: figs_sail.tex
\begin{figure}
  \centering
  \input{figs_sailrecs_table.tex}
  \caption{{\bf    Sail     using    a    non-planar     template.}     As    in
    Figs.~\ref{fig:paper_apron_seq}  and ~\ref{fig:cushion_banana_seq},  we both
    overlay the surfaces reconstructed using  the different methods on the image
    used to  perform the reconstruction and  show them as seen  from a different
    viewpoint.   (a) Our  method using  regularly sampled  control  points.  (b)
    Bartoli et al.~\cite{Bartoli12b} (c) Salzmann et al.~\cite{Salzmann11a}.}
  \label{fig:sail}
\end{figure}

%% file: figs_sailrecs_table.tex
\newcommand{\sailrecswidth}{0.3\linewidth}
\newcommand{\sailrecswidthtwo}{0.25\linewidth}
\begin{tabular}{ccc}
\includegraphics[width=\sailrecswidth]{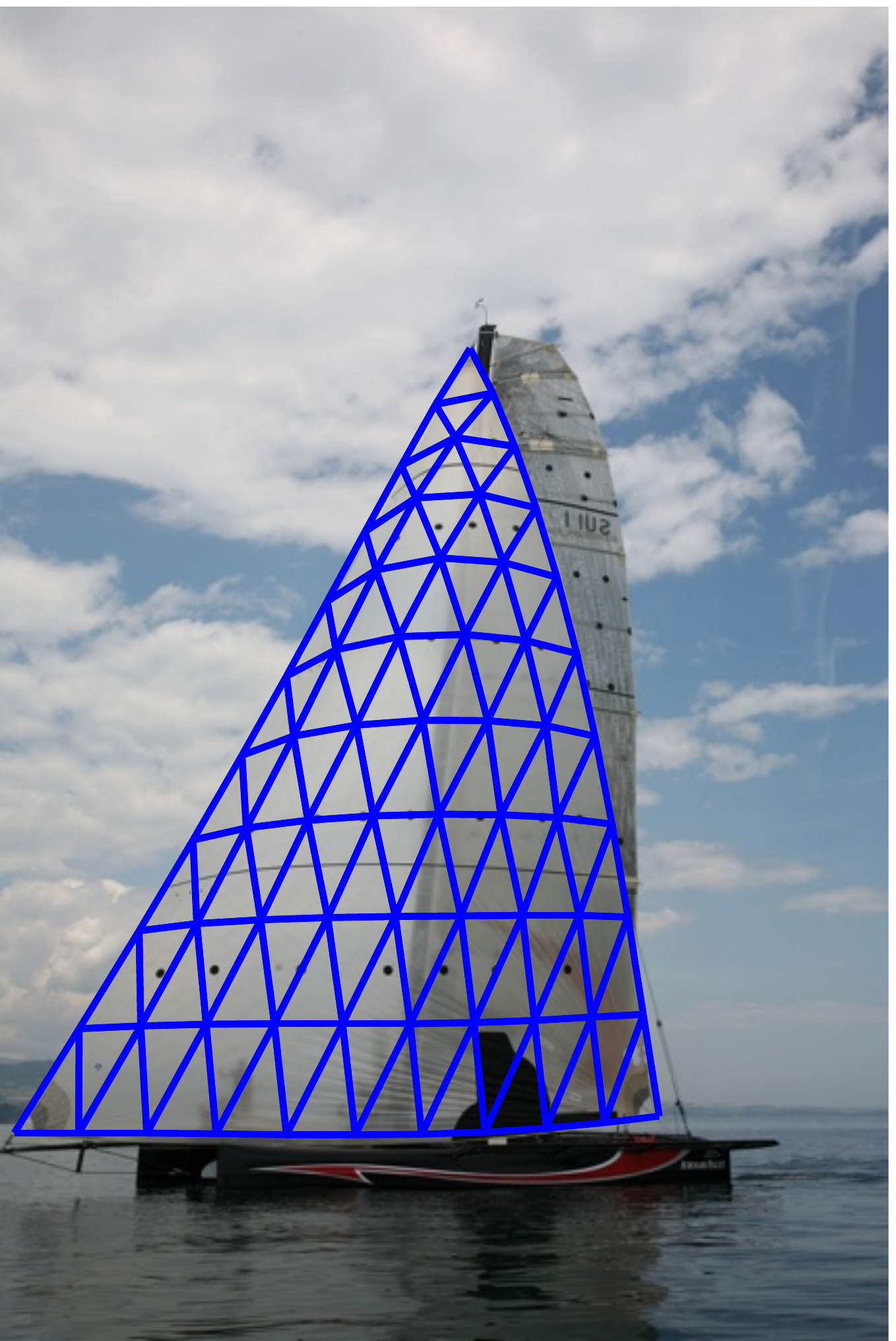} & \includegraphics[width=\sailrecswidth]{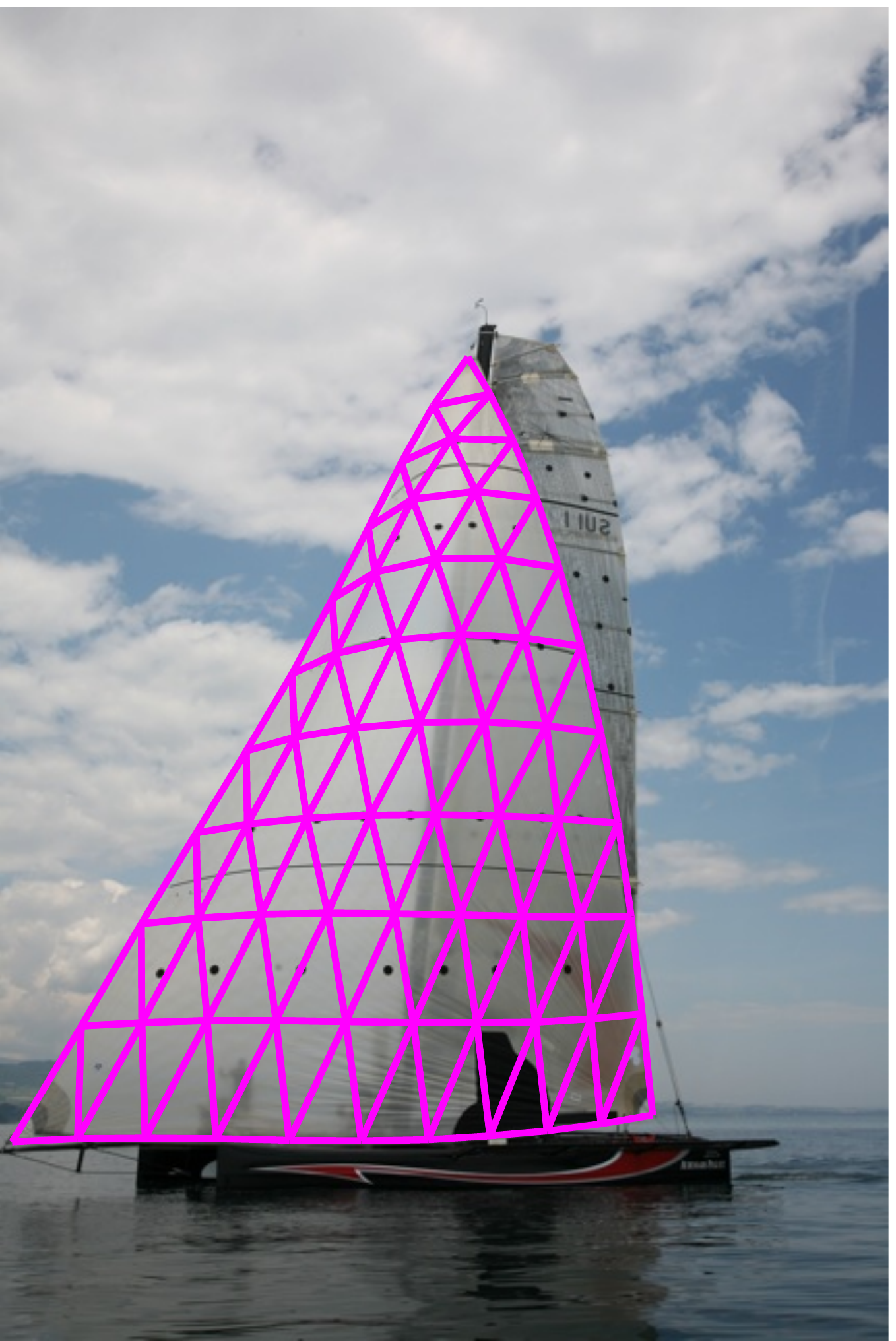} & \includegraphics[width=\sailrecswidth]{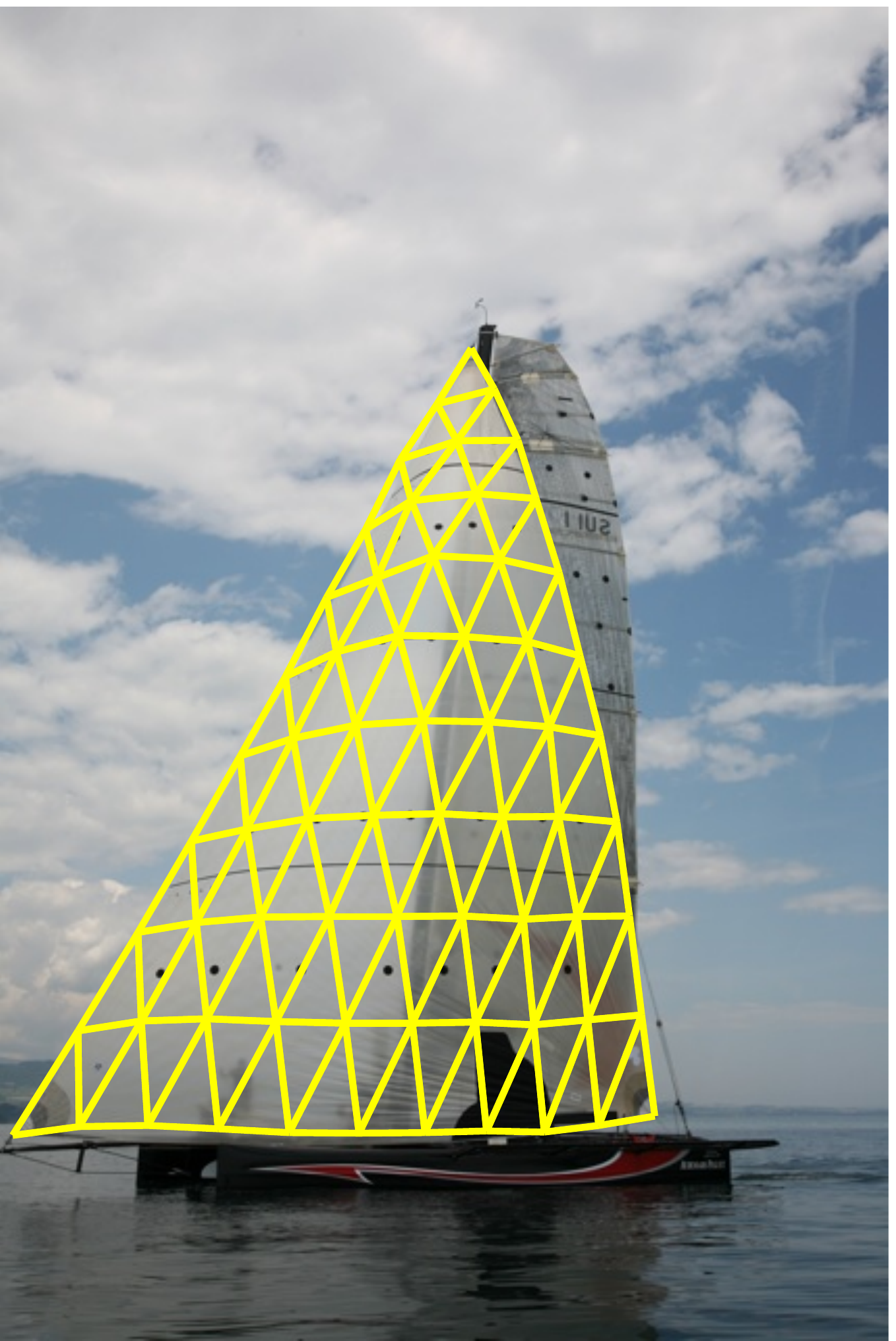}\\
\includegraphics[width=\sailrecswidthtwo]{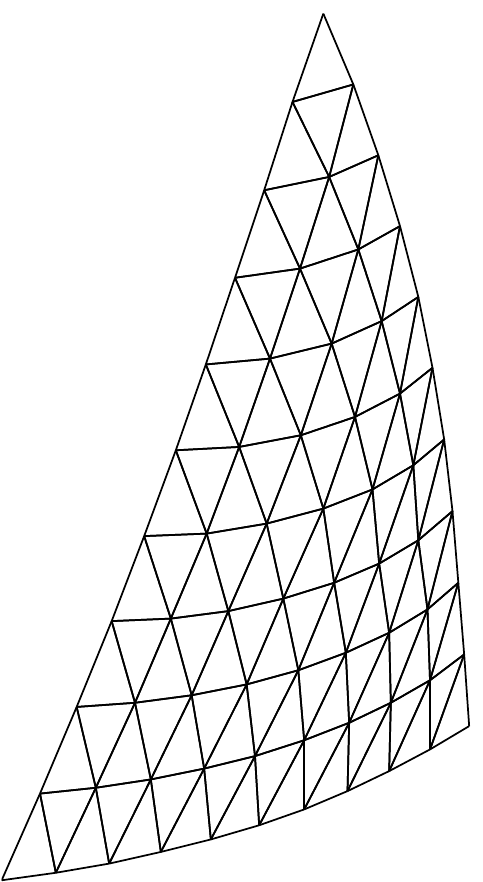} & \includegraphics[width=\sailrecswidthtwo]{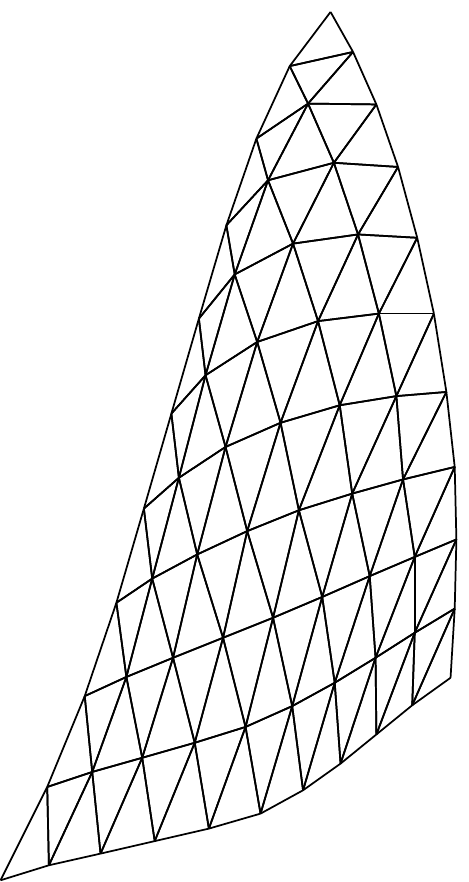} & \includegraphics[width=\sailrecswidthtwo]{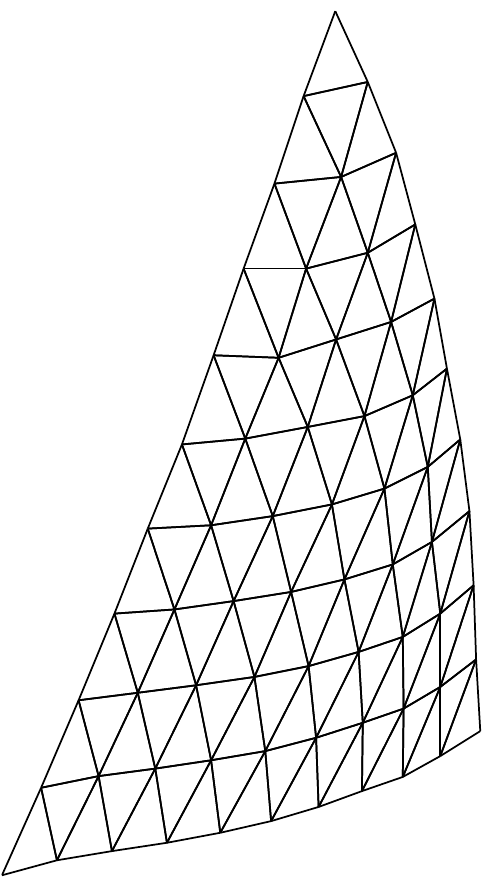}\\
(a) & (b) & (c)
\end{tabular}

%% file: figs_robustness_comparison1.tex
\newcommand{\robustcmpw}{0.40\linewidth}
\newcommand{\robustcmph}{0.258\linewidth}
\begin{figure*}[t]
\begin{center}
\begin{tabular}{ccc}
 \includegraphics[height=\robustcmph]{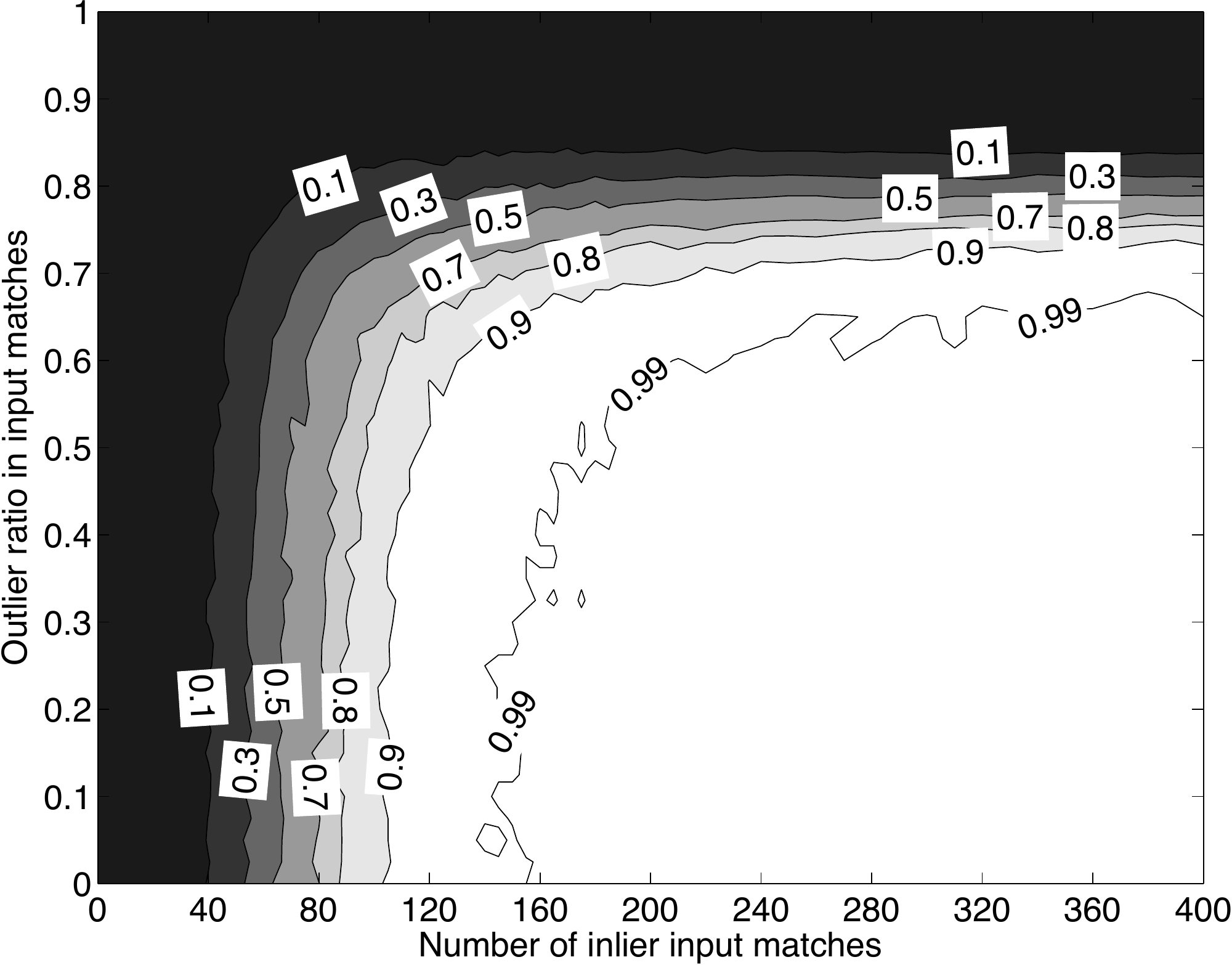} \hspace{-0.5cm} &
 \includegraphics[height=\robustcmph]{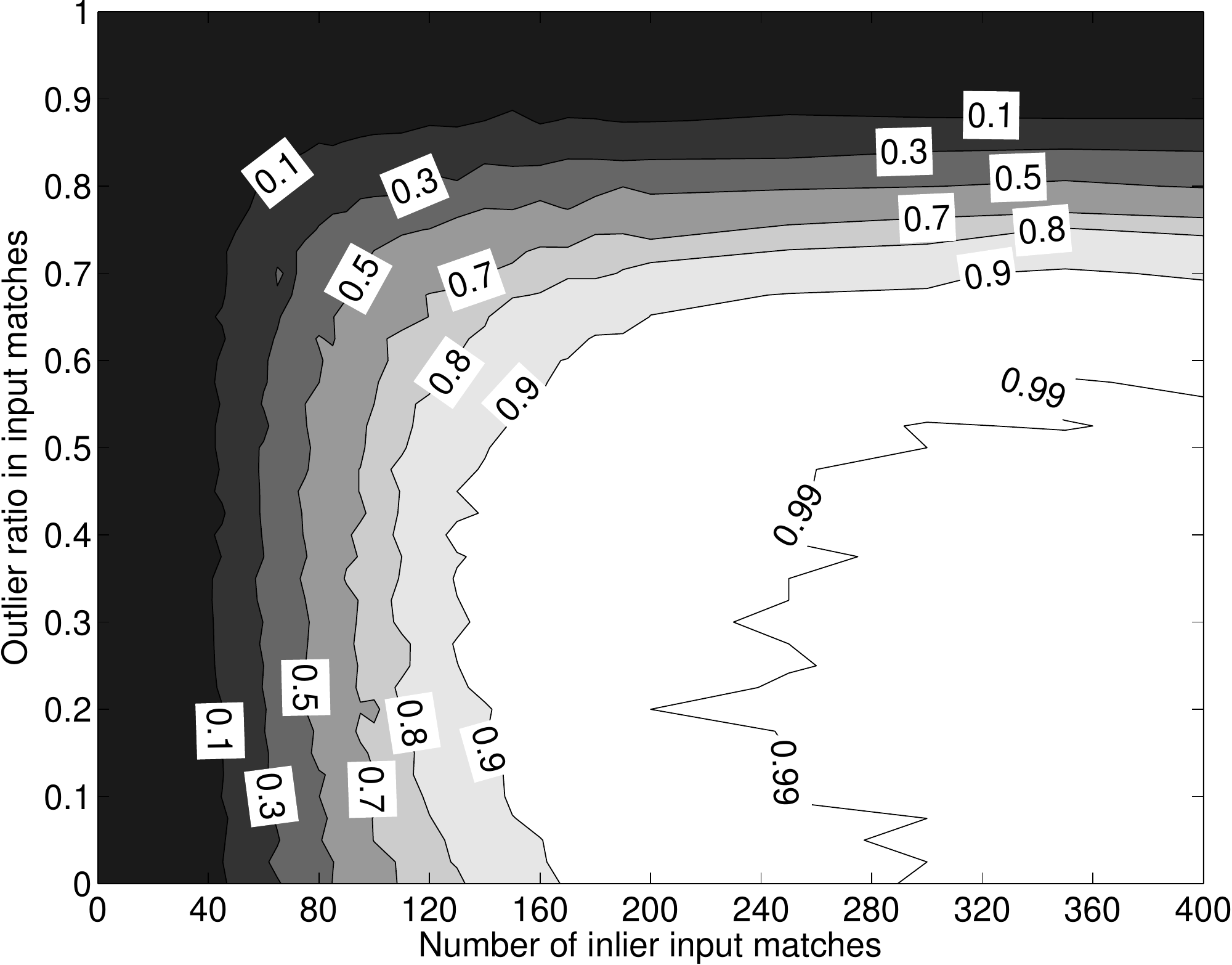} \hspace{-0.5cm} &
 \includegraphics[height=\robustcmph]{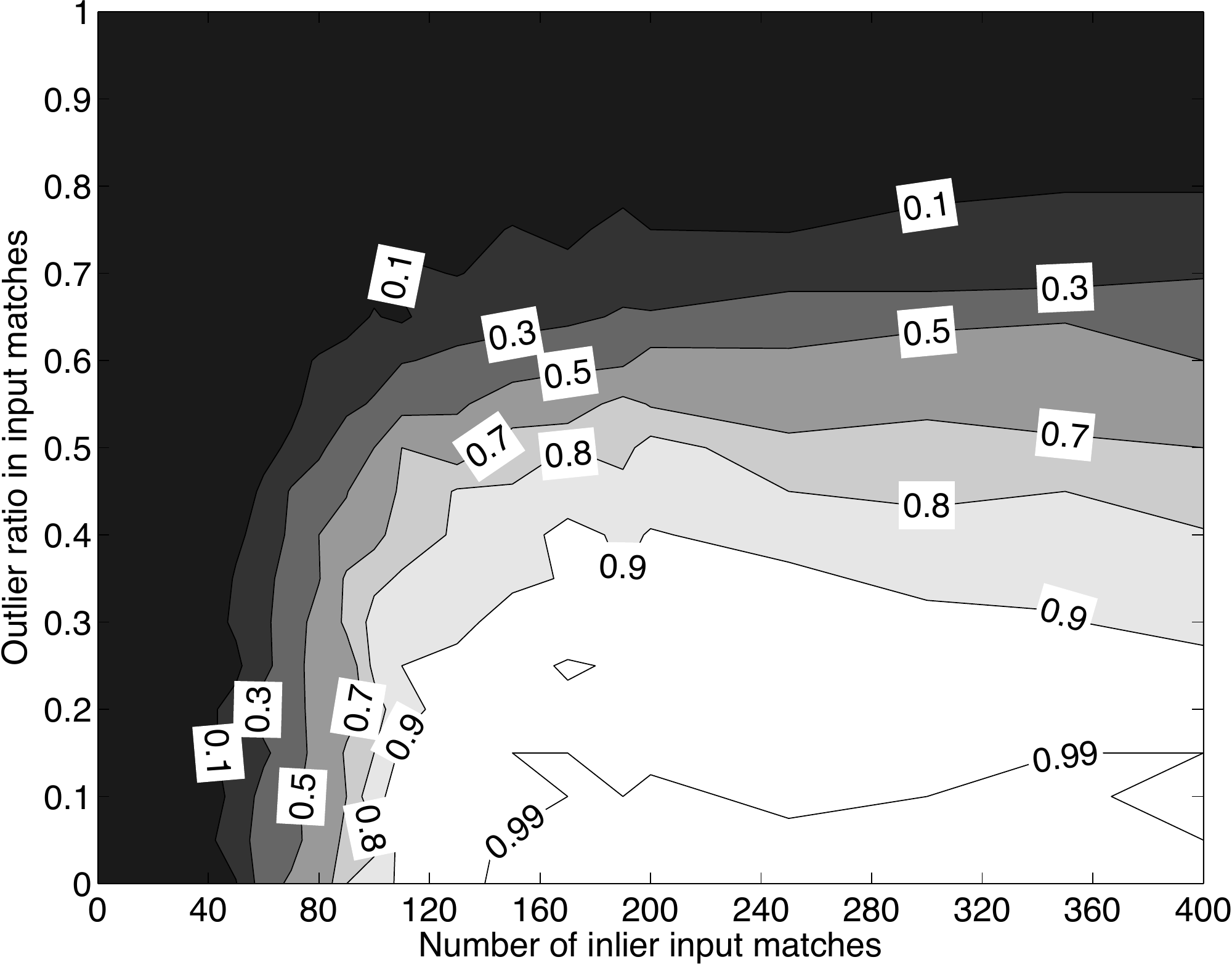} \\ 
 \includegraphics[height=\robustcmph]{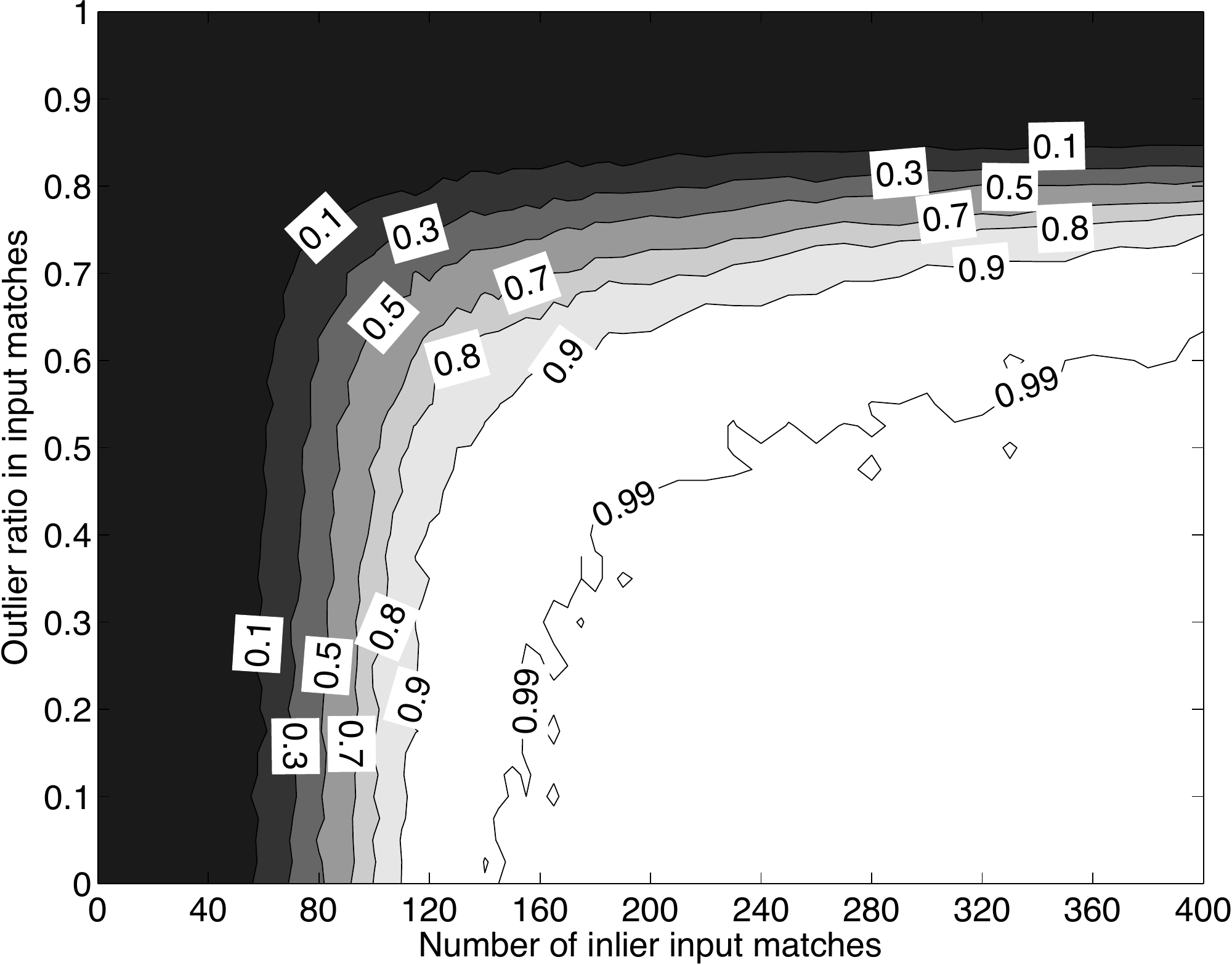} \hspace{-0.5cm} &
 \includegraphics[height=\robustcmph]{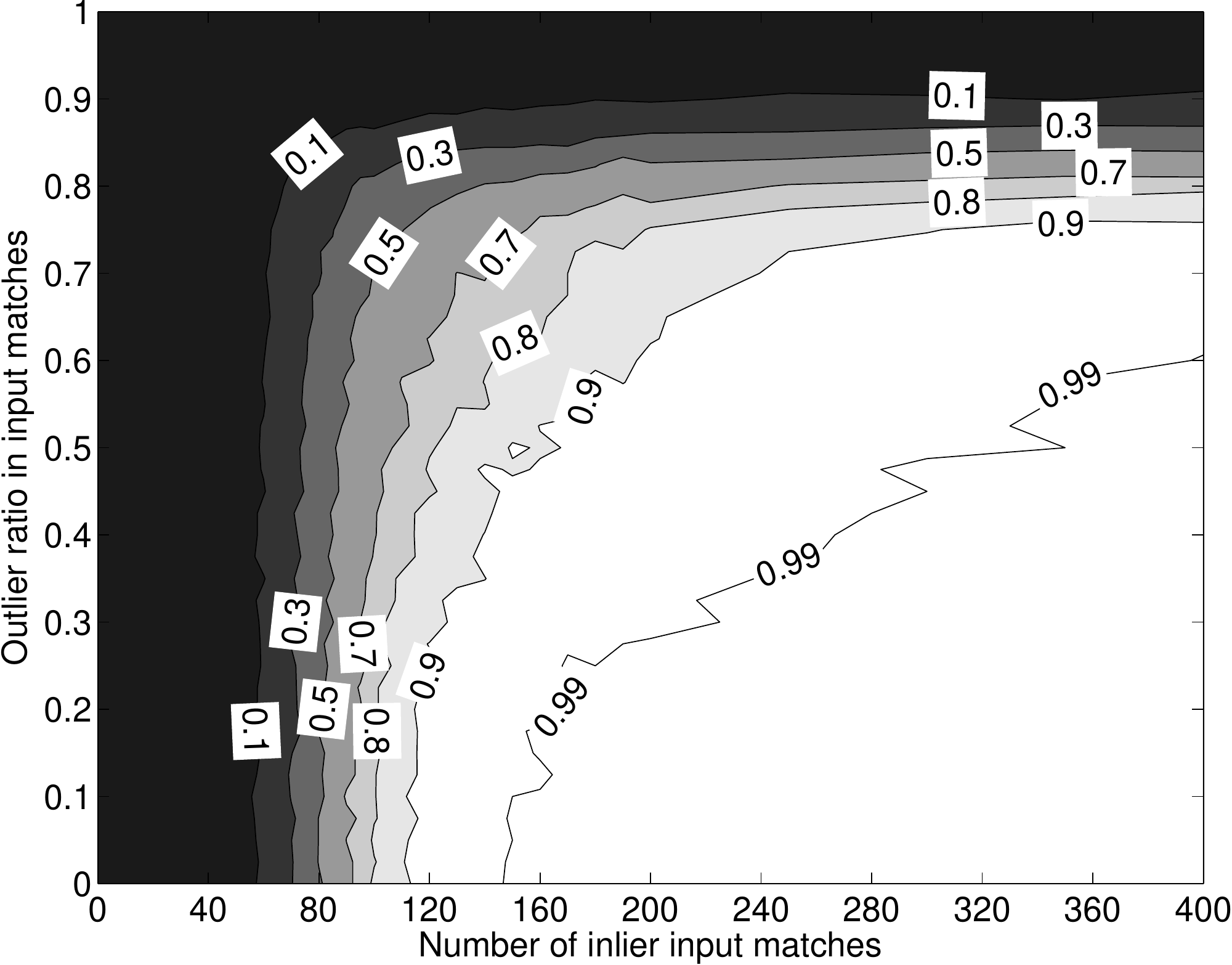} \hspace{-0.5cm} &
 \includegraphics[height=\robustcmph]{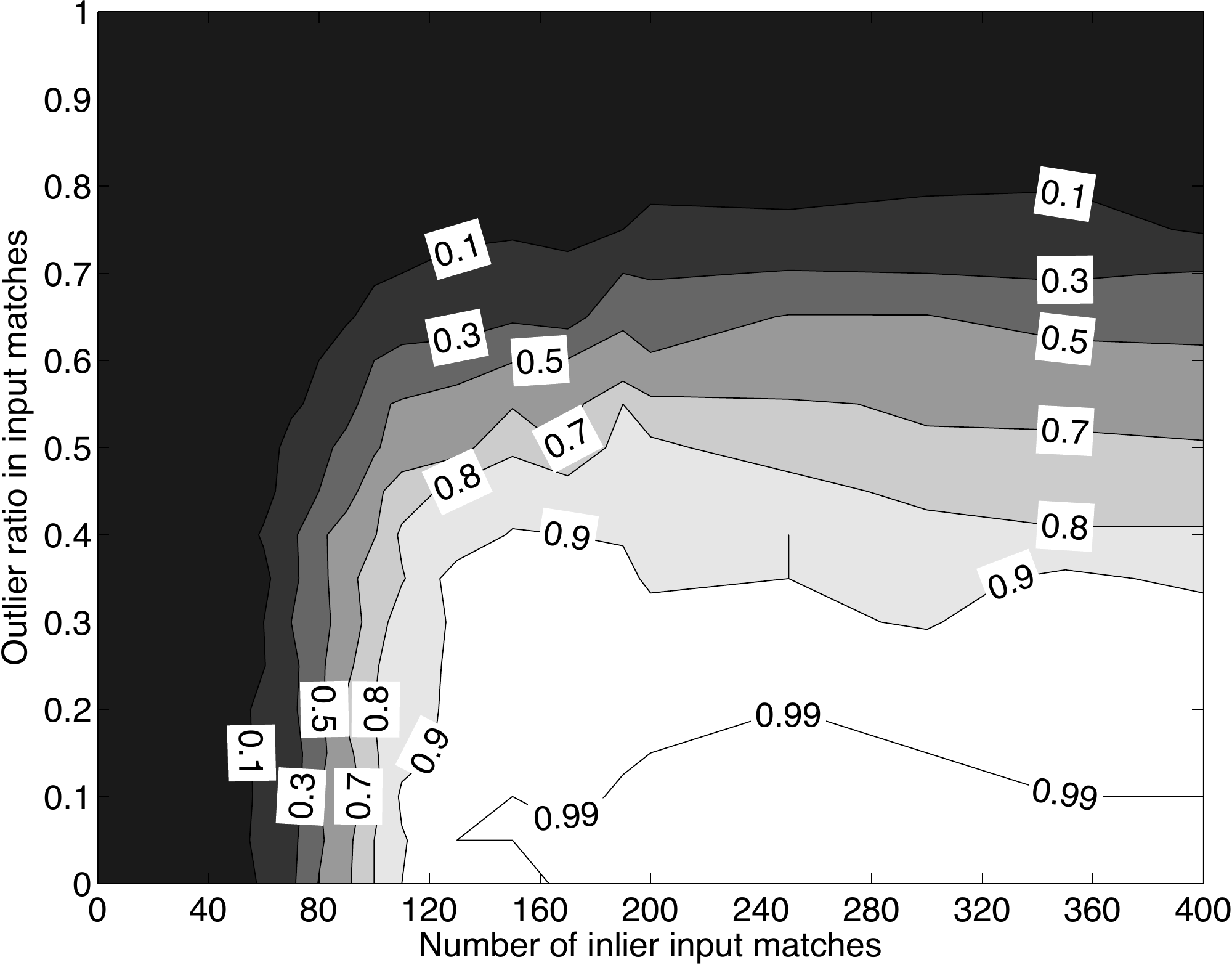} \\  
 (a) Ours & (b) Tran et al. ECCV 2012 \cite{Tran12} & (c) Pizarro et al. IJCV 2012 \cite{Pizarro12} \\ 
\end{tabular}
\end{center}
\vspace{-0.3cm}
\caption{Probability of success as a function of the number of inlier matches
  and proportion of outliers, on the x-axis and y-axis respectively. The lines
  are level lines of the probability of having at least $90\%$ mesh vertices  
  reprojected within 2 pixels of the solution. {\bf Top row:} paper dataset. {\bf Bottom row:} cushion dataset.}
\label{fig:robustnesscompare}
\vspace{-0.4cm}
\end{figure*}

%% file: figs_controlpts.tex
\begin{figure*}
\centering
\input{figs_ctrlselsrnd_table.tex}
\vspace{-.1cm}
\caption{{\bf Templates and  control vertices.} Each row depicts  the template and
  control vertex configuration used for  the paper, apron, cushion, banana leaf,
  and  sail sequences.   The number  below each  figures denotes  the  number of
  control vertices  used in  each case.  {\bf  Leftmost four  columns:} Manually
  chosen control  vertices. {\bf  Fifth  column:} All vertices as control
  vertices. {\bf Rightmost four columns:}  Examples of randomly
  chosen control vertices.}
\label{fig:ctrlsels}
\vspace{-.4cm}
\end{figure*}

%% file: figs_ctrlselsrnd_table.tex
\newcommand{\ctrlselsrndwidth}{0.08\linewidth}
\newcommand{\ctrlselspace}{0.02cm}
\begin{tabular}{ccccccccc}
\includegraphics[width=\ctrlselsrndwidth]{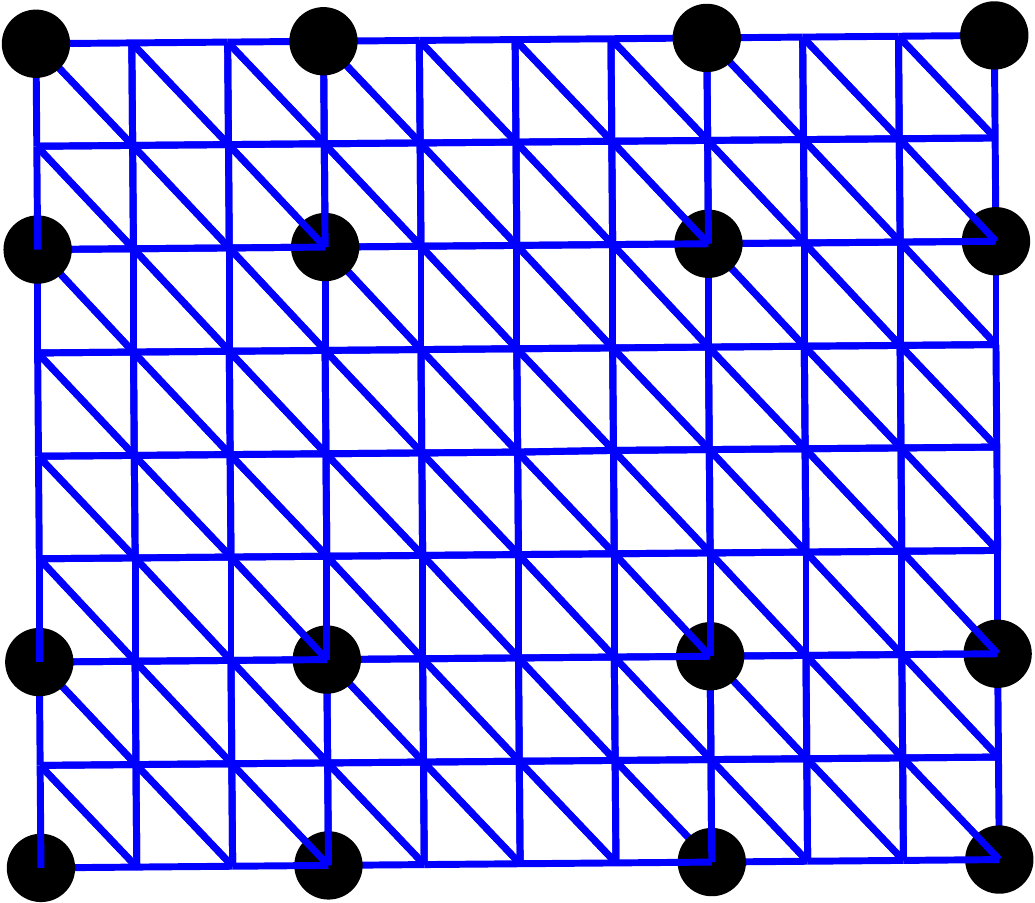} \hspace{\ctrlselspace} & \includegraphics[width=\ctrlselsrndwidth]{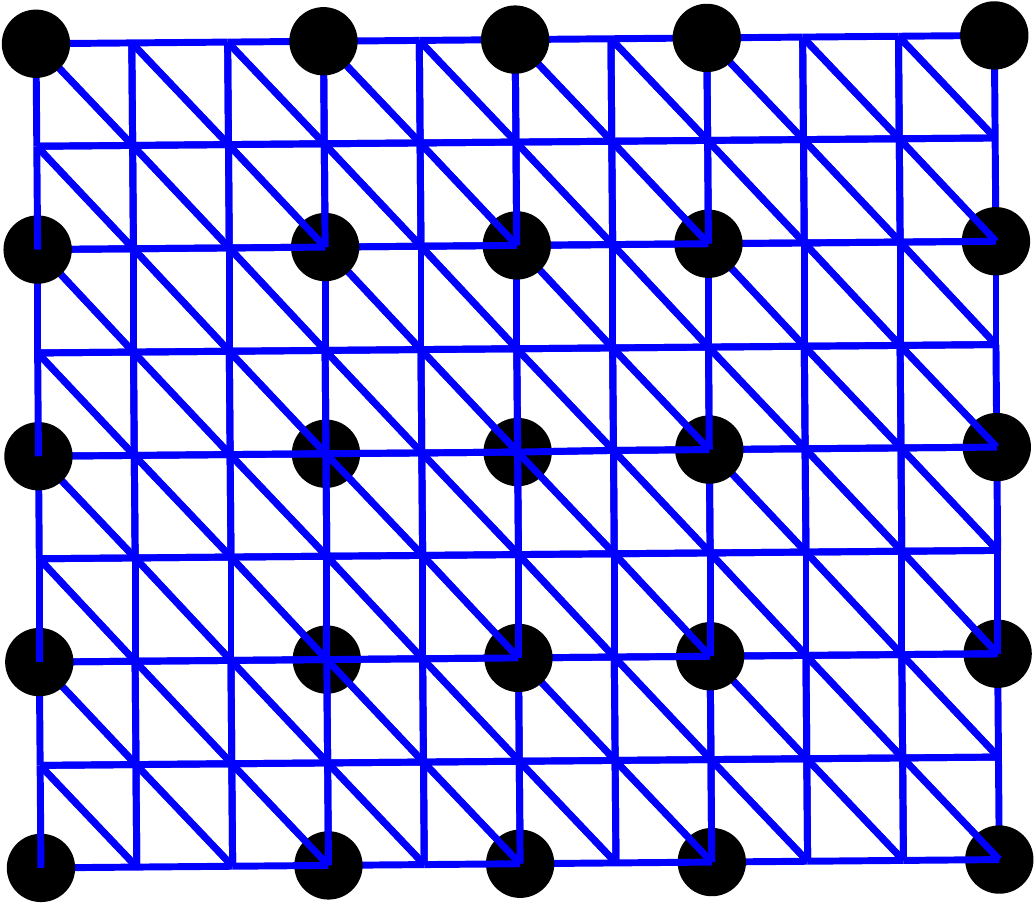} \hspace{\ctrlselspace} & \includegraphics[width=\ctrlselsrndwidth]{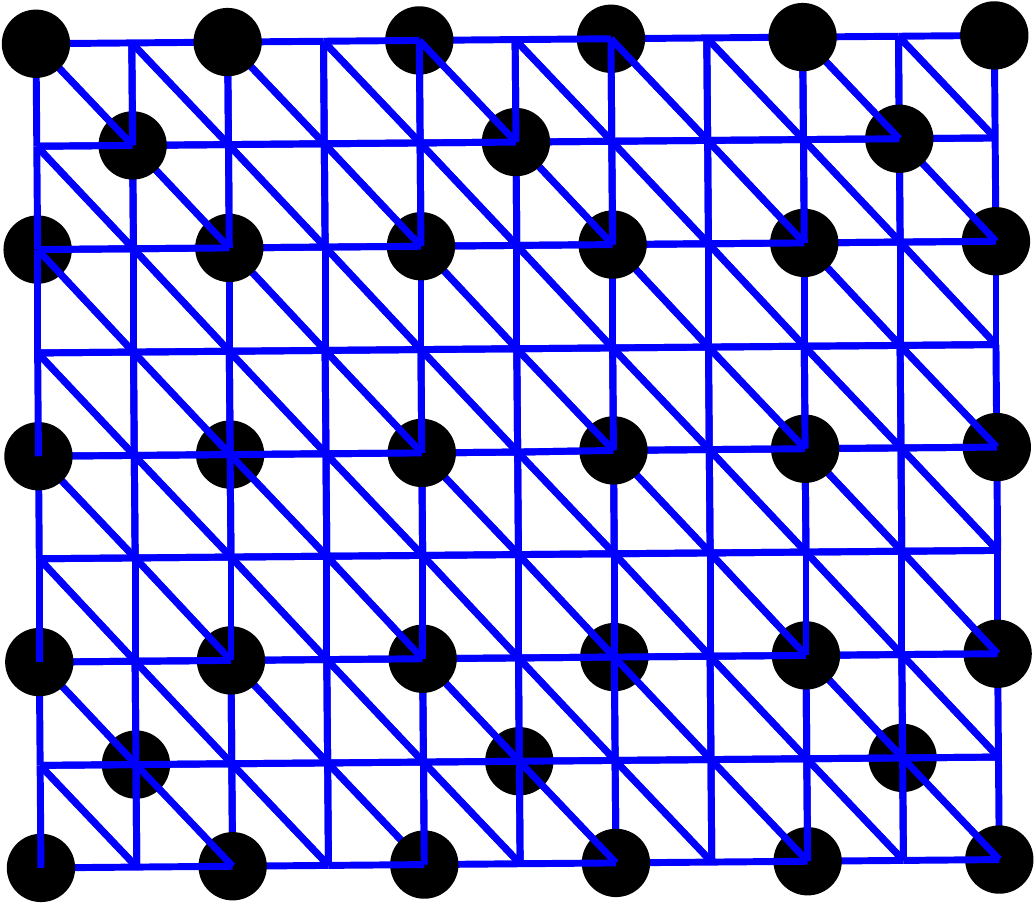} \hspace{\ctrlselspace} & \includegraphics[width=\ctrlselsrndwidth]{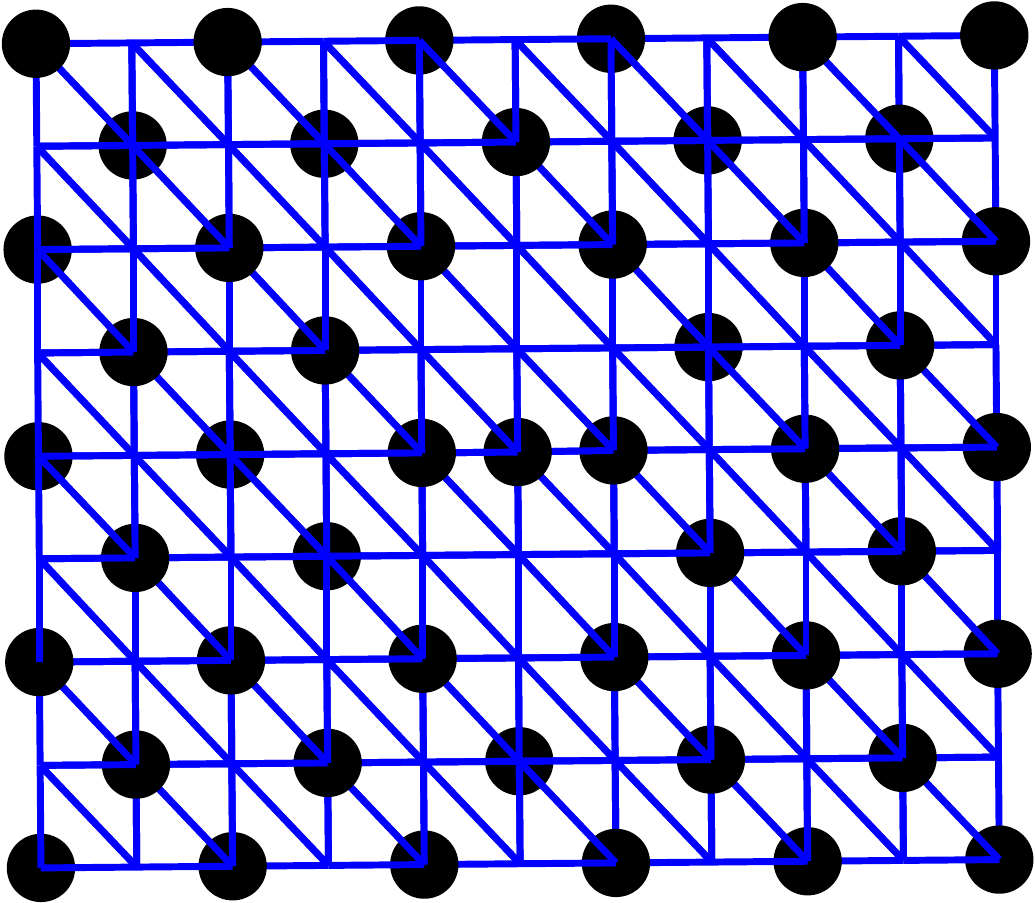} \hspace{\ctrlselspace} & \includegraphics[width=\ctrlselsrndwidth]{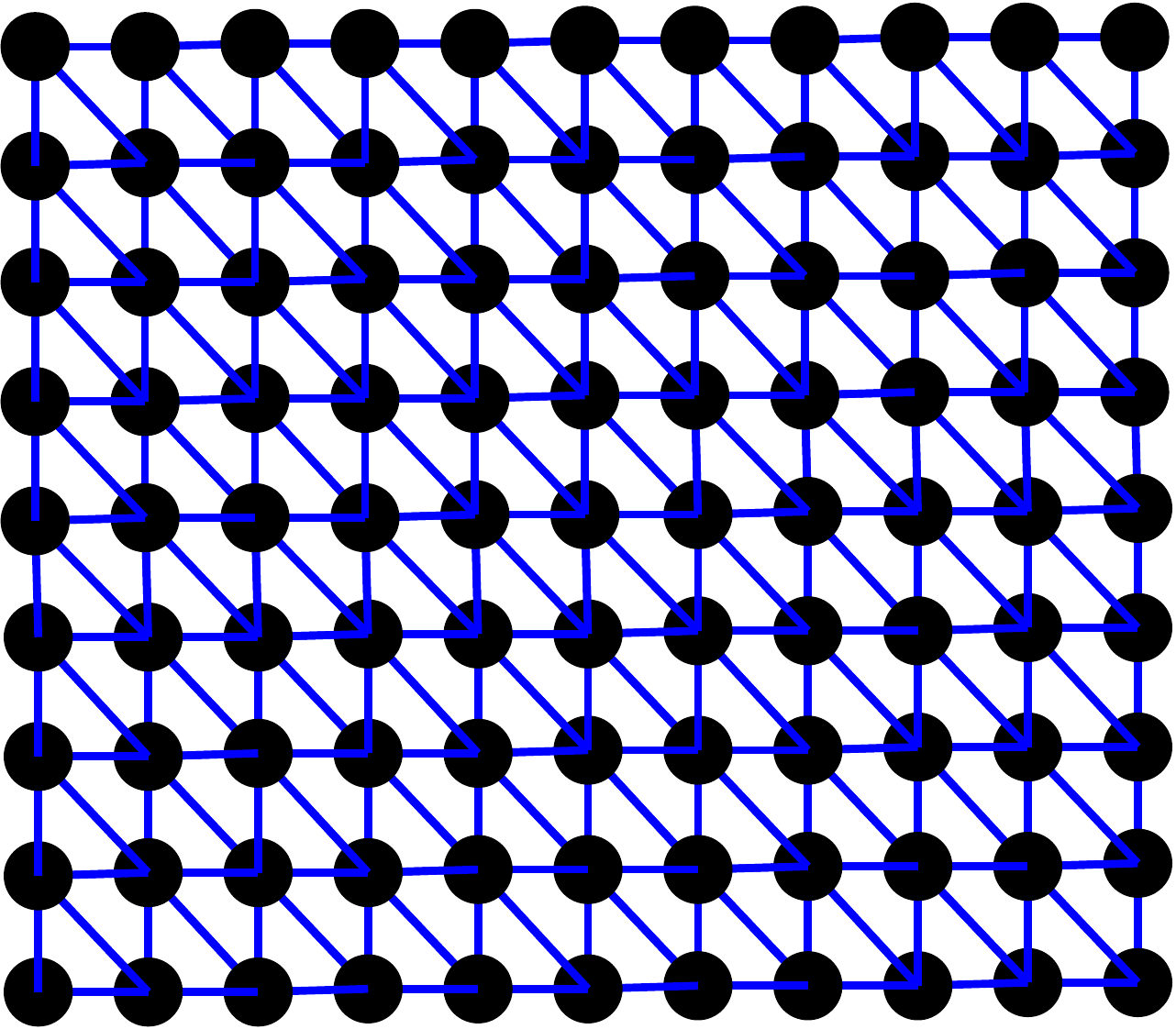} \hspace{\ctrlselspace} & \includegraphics[width=\ctrlselsrndwidth]{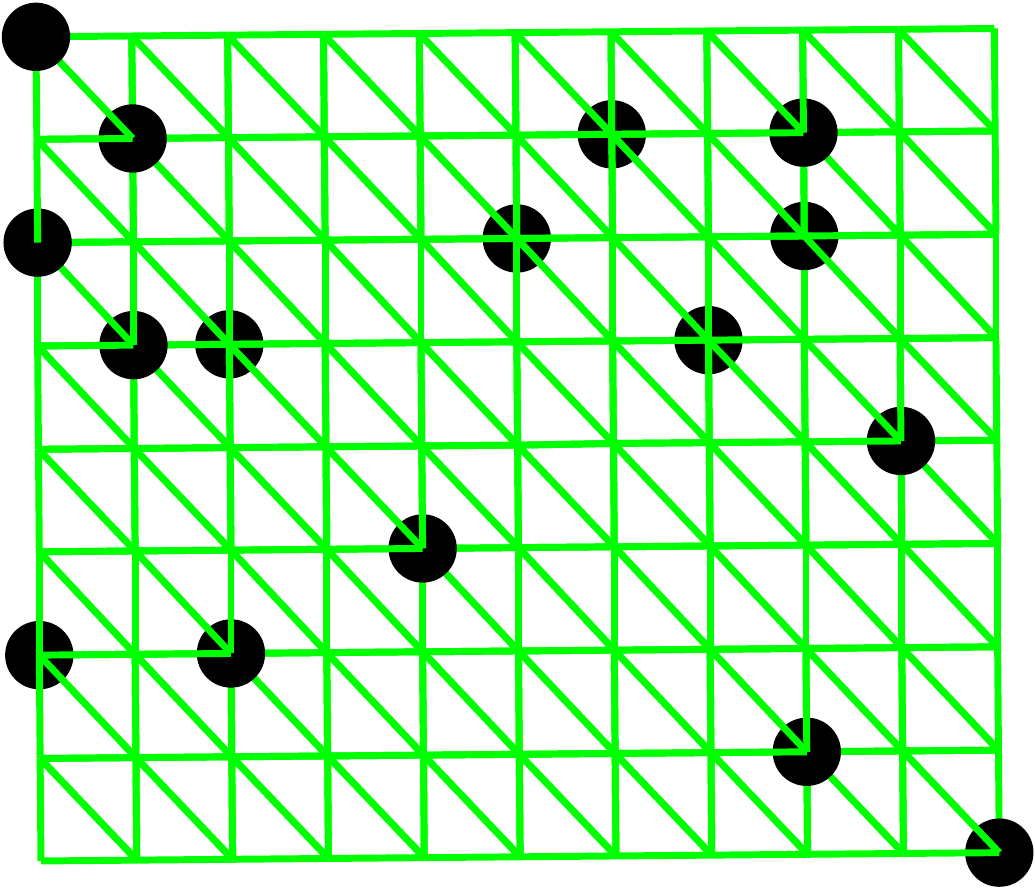} \hspace{\ctrlselspace} & \includegraphics[width=\ctrlselsrndwidth]{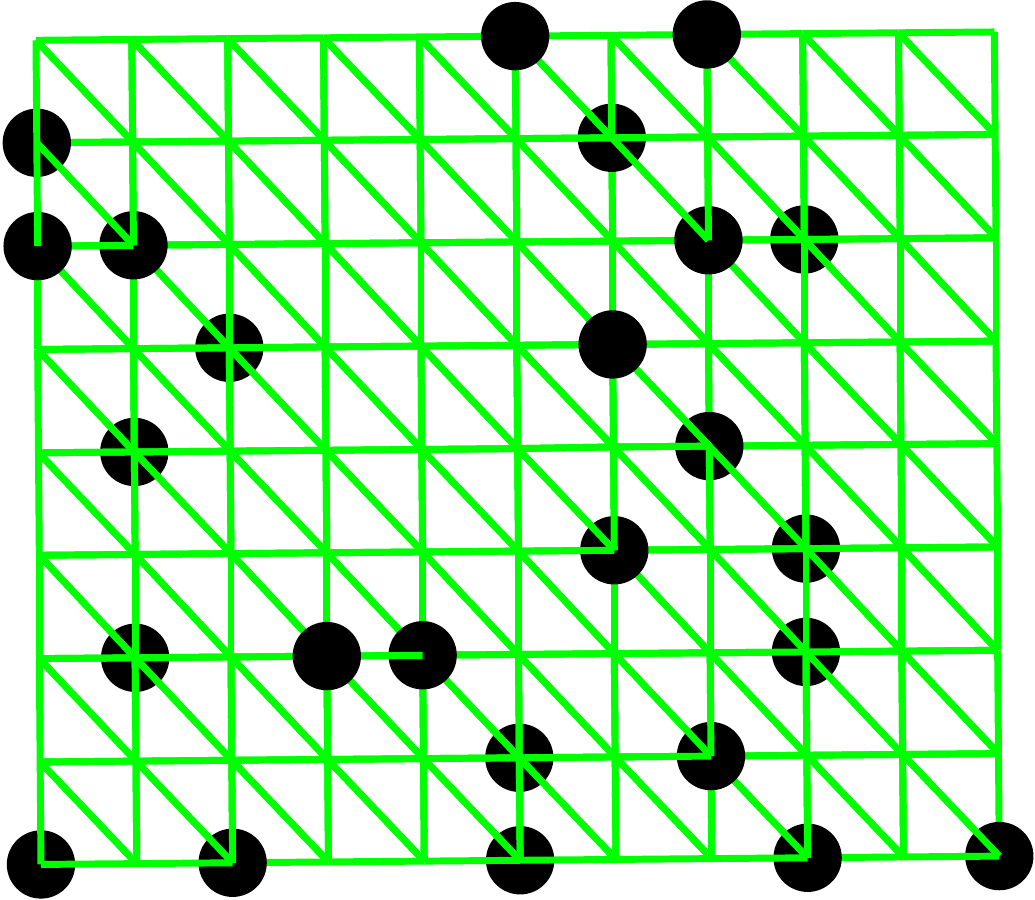} \hspace{\ctrlselspace} & \includegraphics[width=\ctrlselsrndwidth]{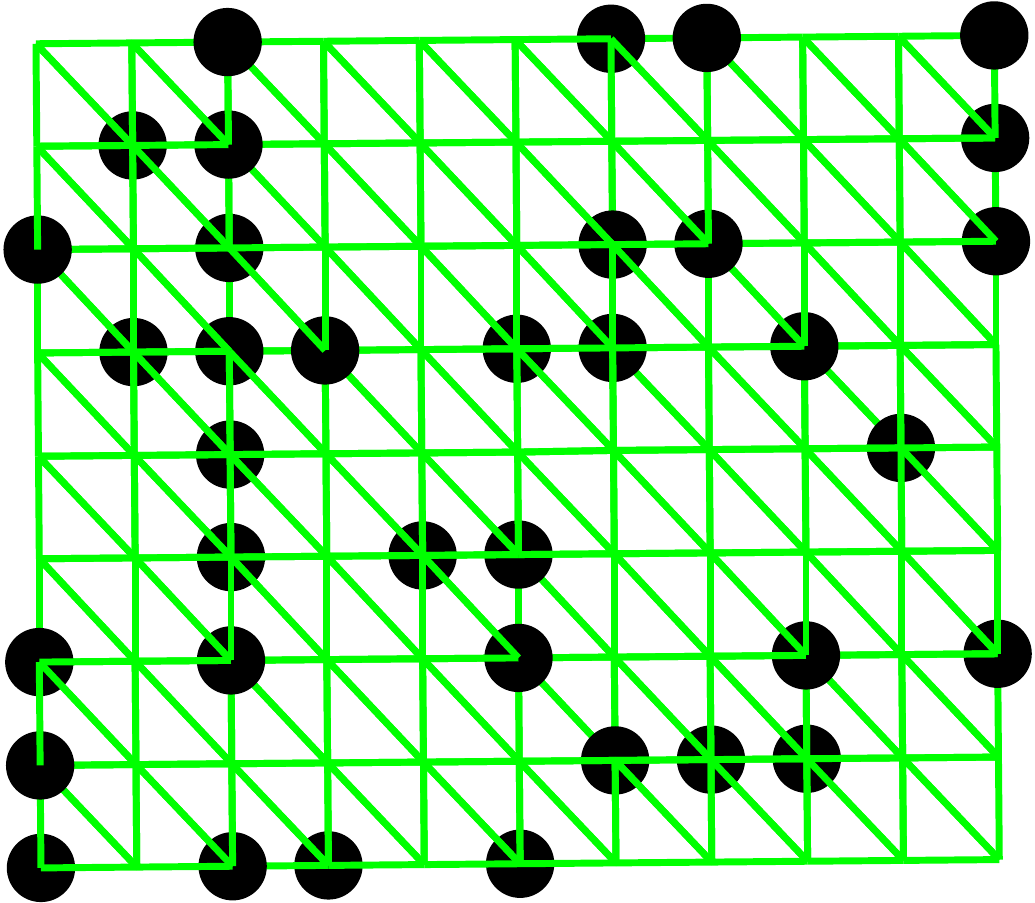} \hspace{\ctrlselspace} & \includegraphics[width=\ctrlselsrndwidth]{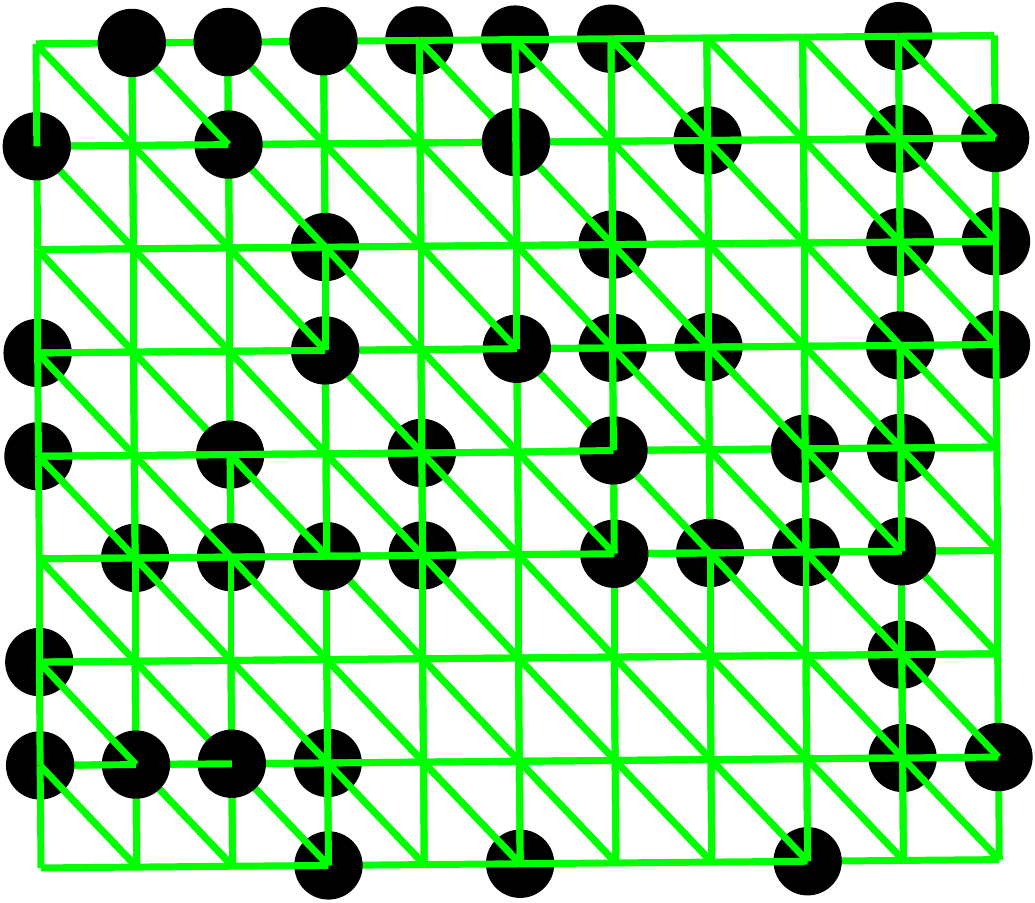}\\
(16) \hspace{\ctrlselspace} & (25) \hspace{\ctrlselspace} & (36) \hspace{\ctrlselspace} & (49) \hspace{\ctrlselspace} & (99) \hspace{\ctrlselspace} & (16) \hspace{\ctrlselspace} & (25) \hspace{\ctrlselspace} & (36) \hspace{\ctrlselspace} & (49)\\
\includegraphics[width=\ctrlselsrndwidth]{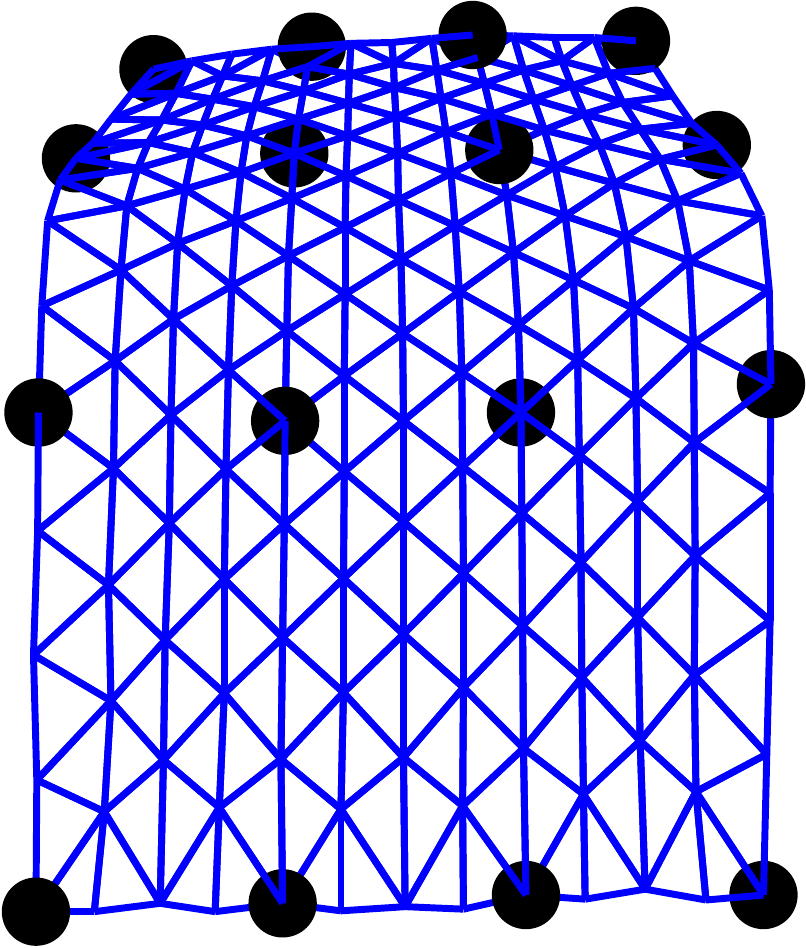} \hspace{\ctrlselspace} & \includegraphics[width=\ctrlselsrndwidth]{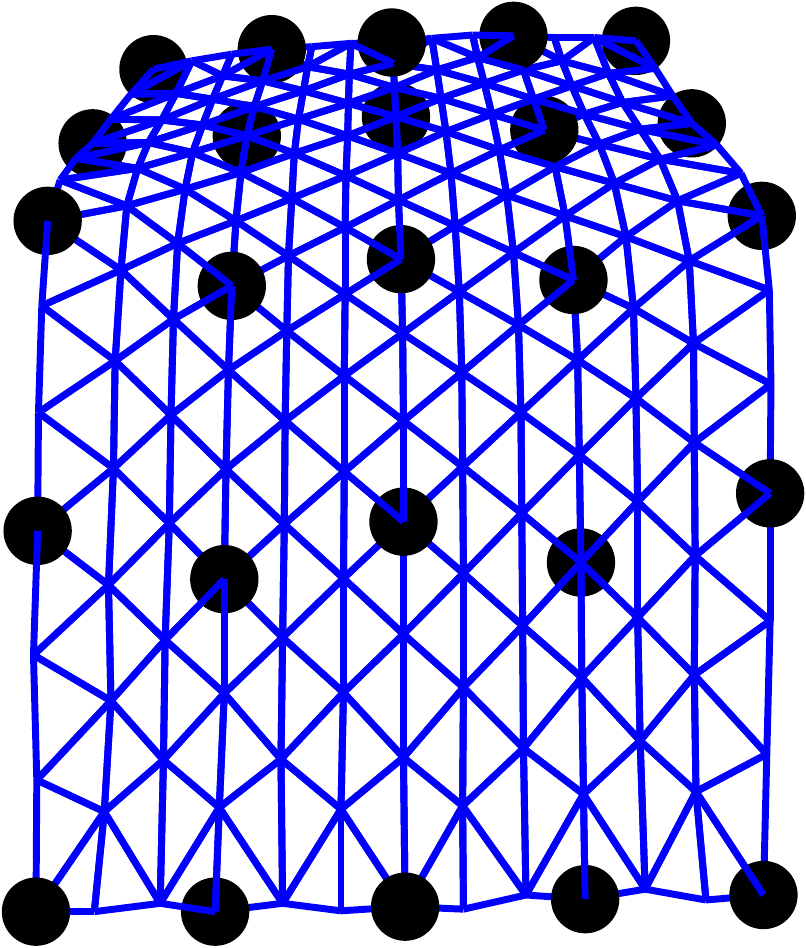} \hspace{\ctrlselspace} & \includegraphics[width=\ctrlselsrndwidth]{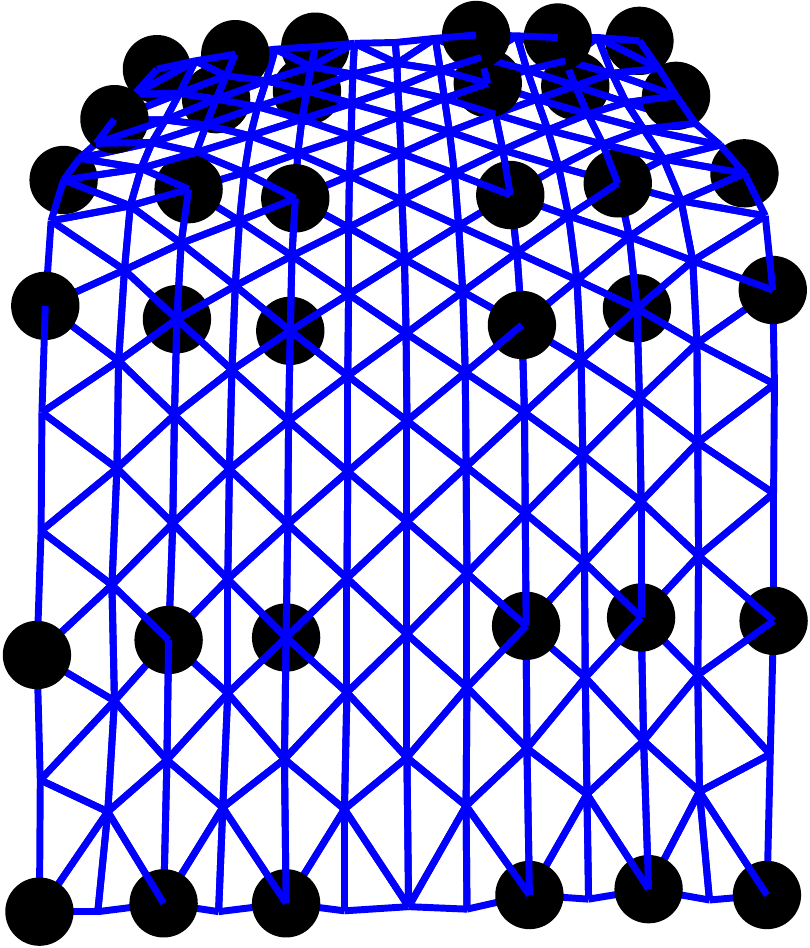} \hspace{\ctrlselspace} & \includegraphics[width=\ctrlselsrndwidth]{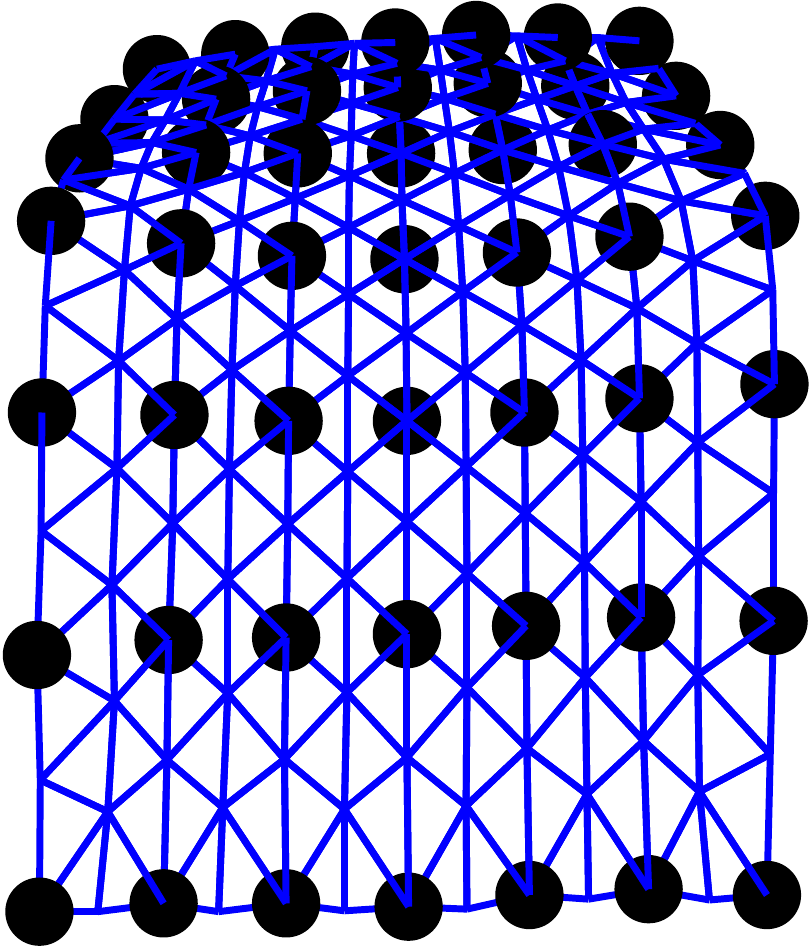} \hspace{\ctrlselspace} & \includegraphics[width=\ctrlselsrndwidth]{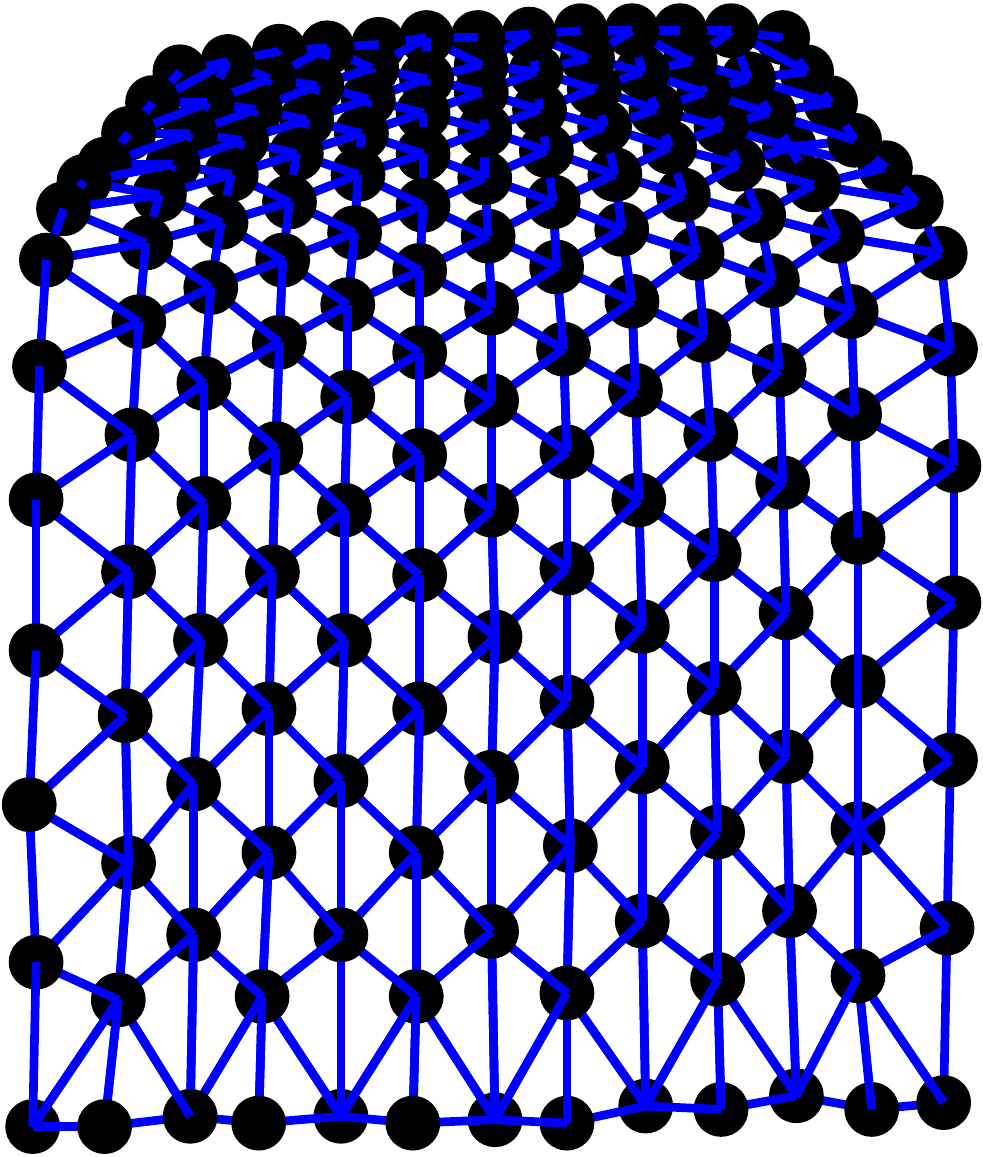} \hspace{\ctrlselspace} & \includegraphics[width=\ctrlselsrndwidth]{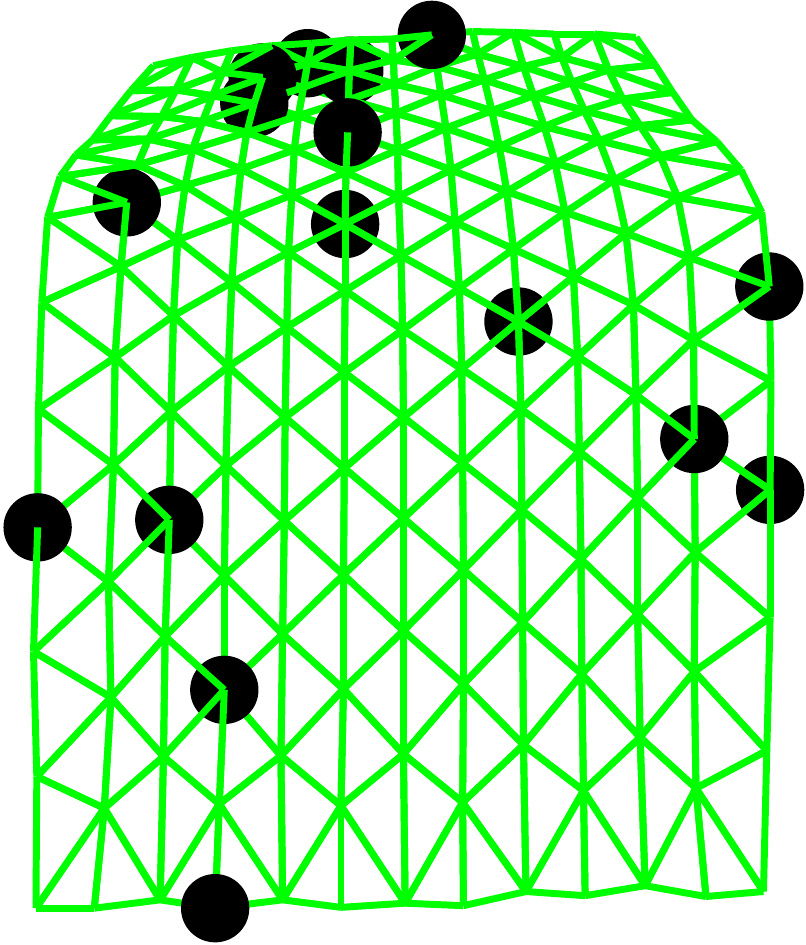} \hspace{\ctrlselspace} & \includegraphics[width=\ctrlselsrndwidth]{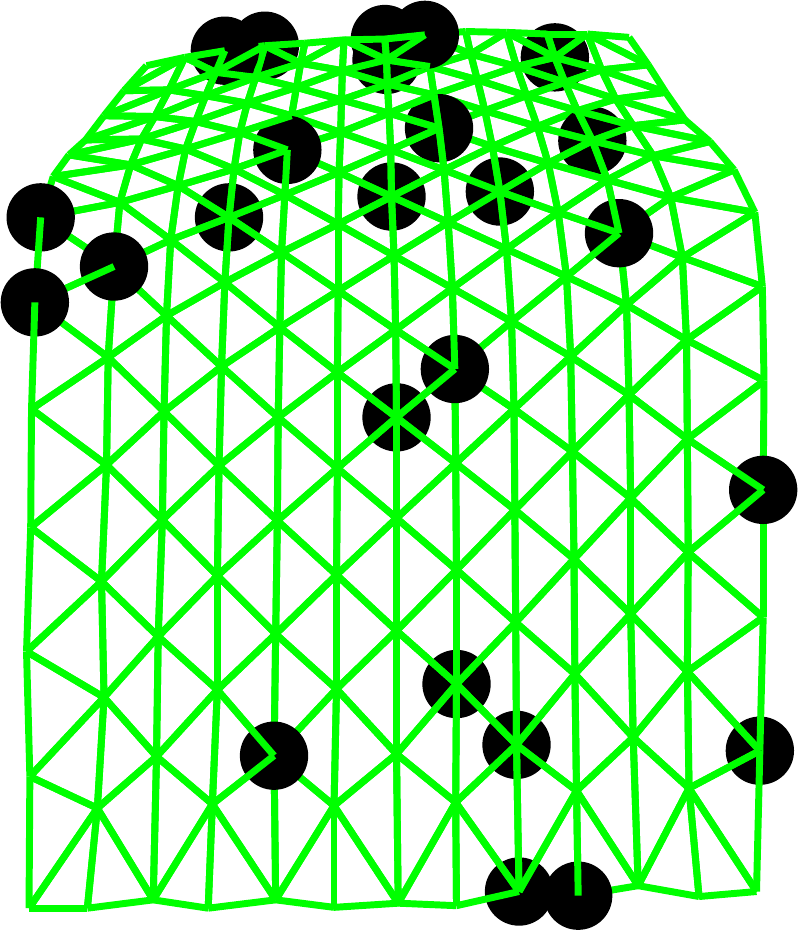} \hspace{\ctrlselspace} & \includegraphics[width=\ctrlselsrndwidth]{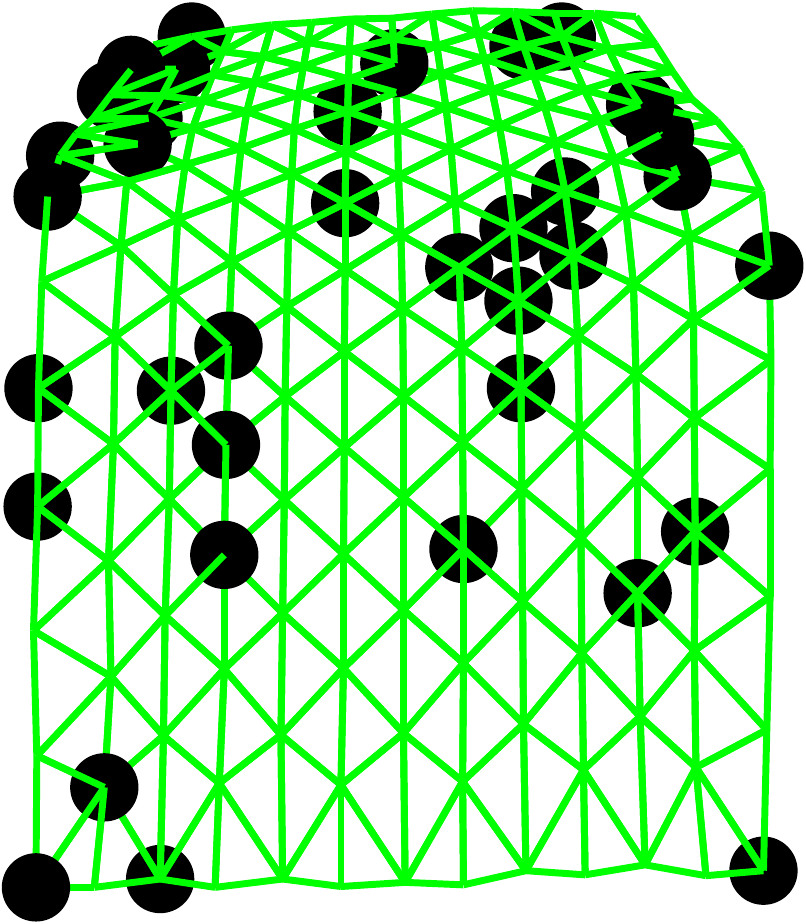} \hspace{\ctrlselspace} & \includegraphics[width=\ctrlselsrndwidth]{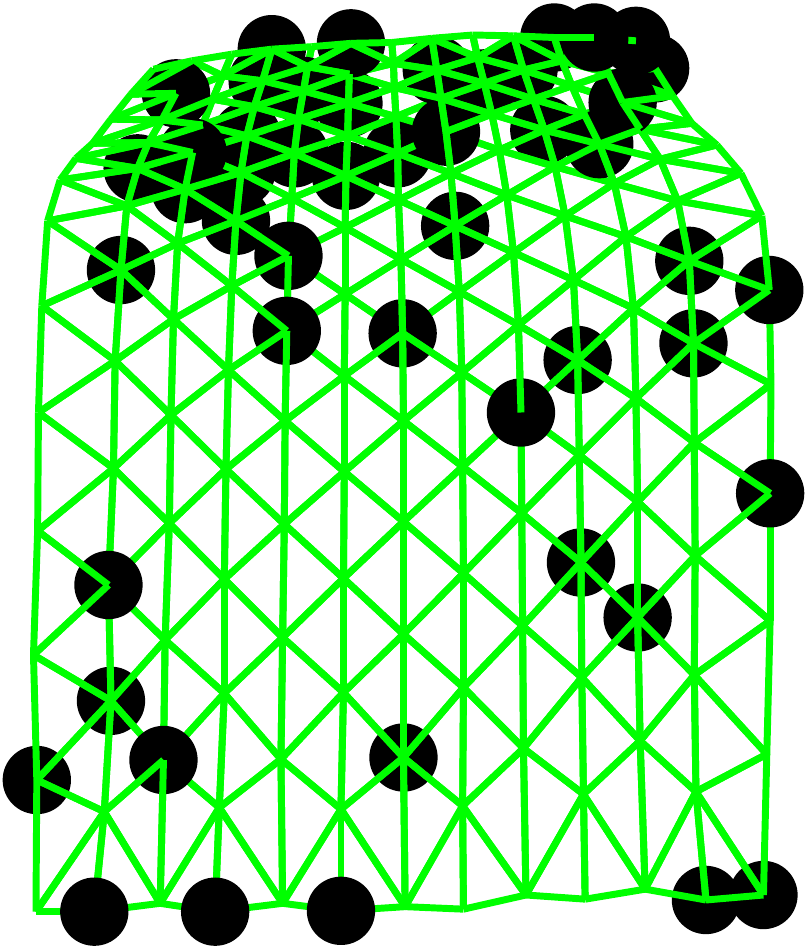}\\
(16) \hspace{\ctrlselspace} & (25) \hspace{\ctrlselspace} & (36) \hspace{\ctrlselspace} & (49) \hspace{\ctrlselspace} & (169) \hspace{\ctrlselspace} & (16) \hspace{\ctrlselspace} & (25) \hspace{\ctrlselspace} & (36) \hspace{\ctrlselspace} & (49)\\
\includegraphics[width=\ctrlselsrndwidth]{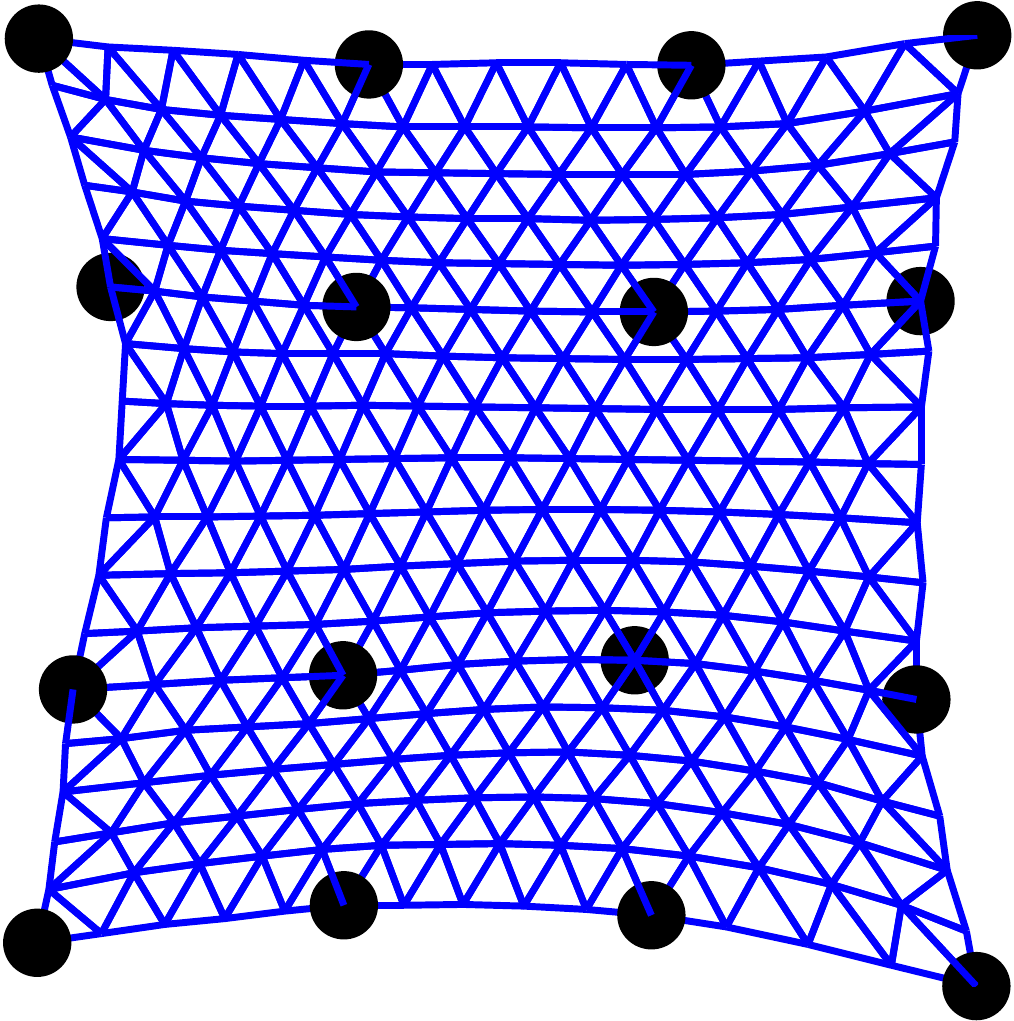} \hspace{\ctrlselspace} & \includegraphics[width=\ctrlselsrndwidth]{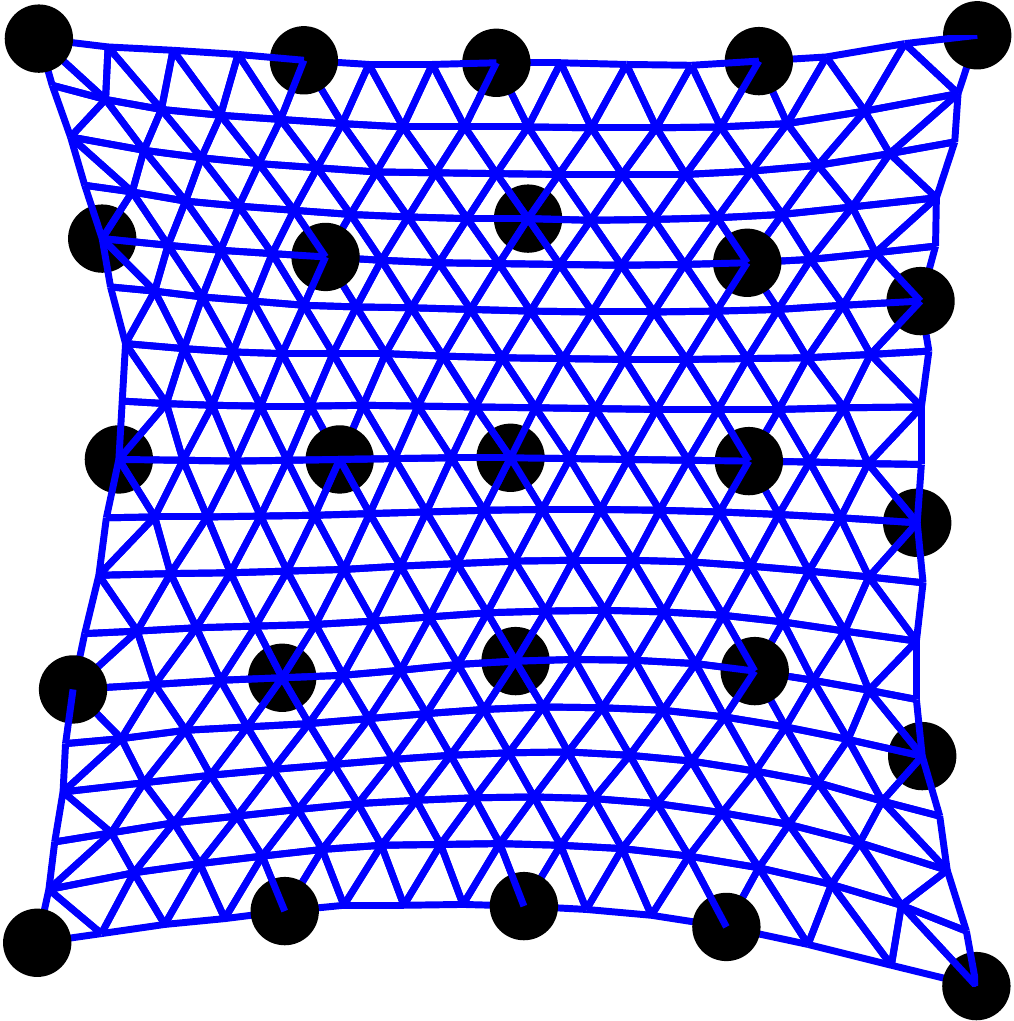} \hspace{\ctrlselspace} & \includegraphics[width=\ctrlselsrndwidth]{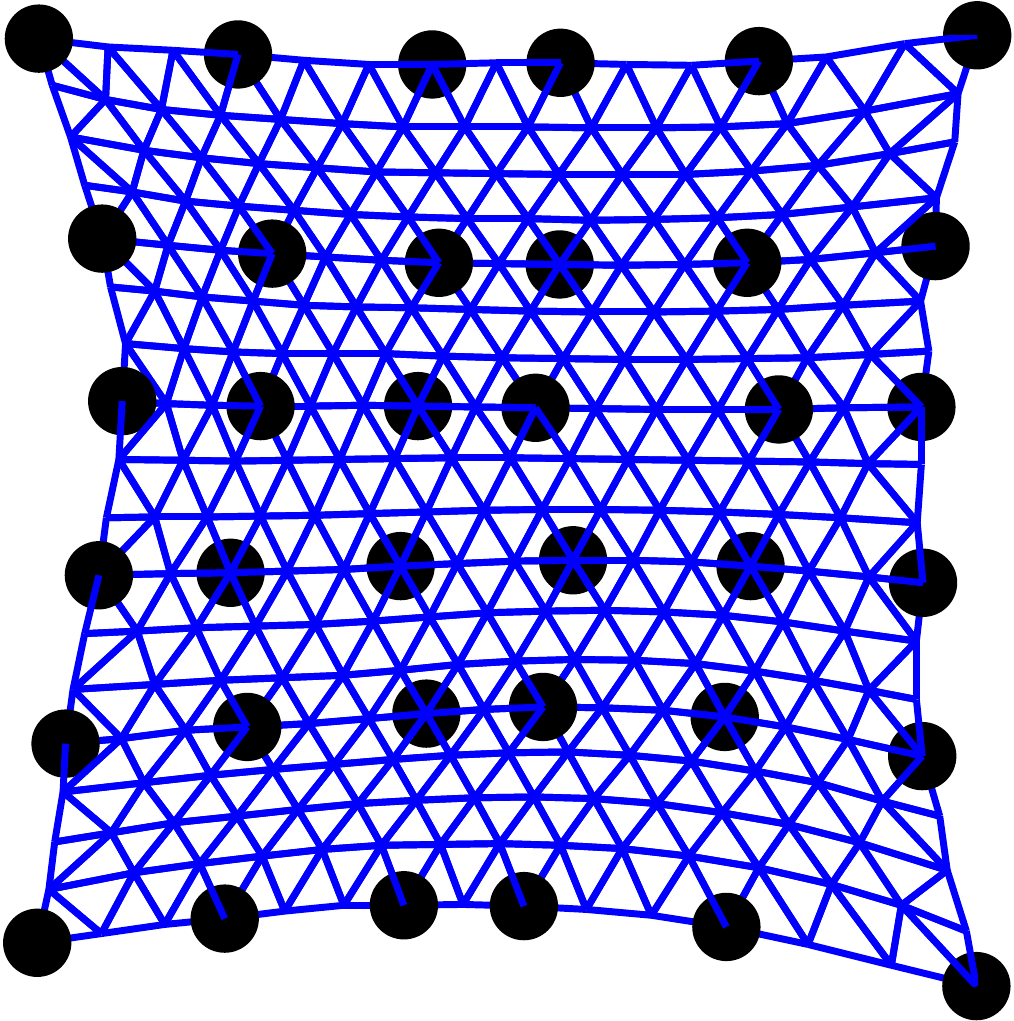} \hspace{\ctrlselspace} & \includegraphics[width=\ctrlselsrndwidth]{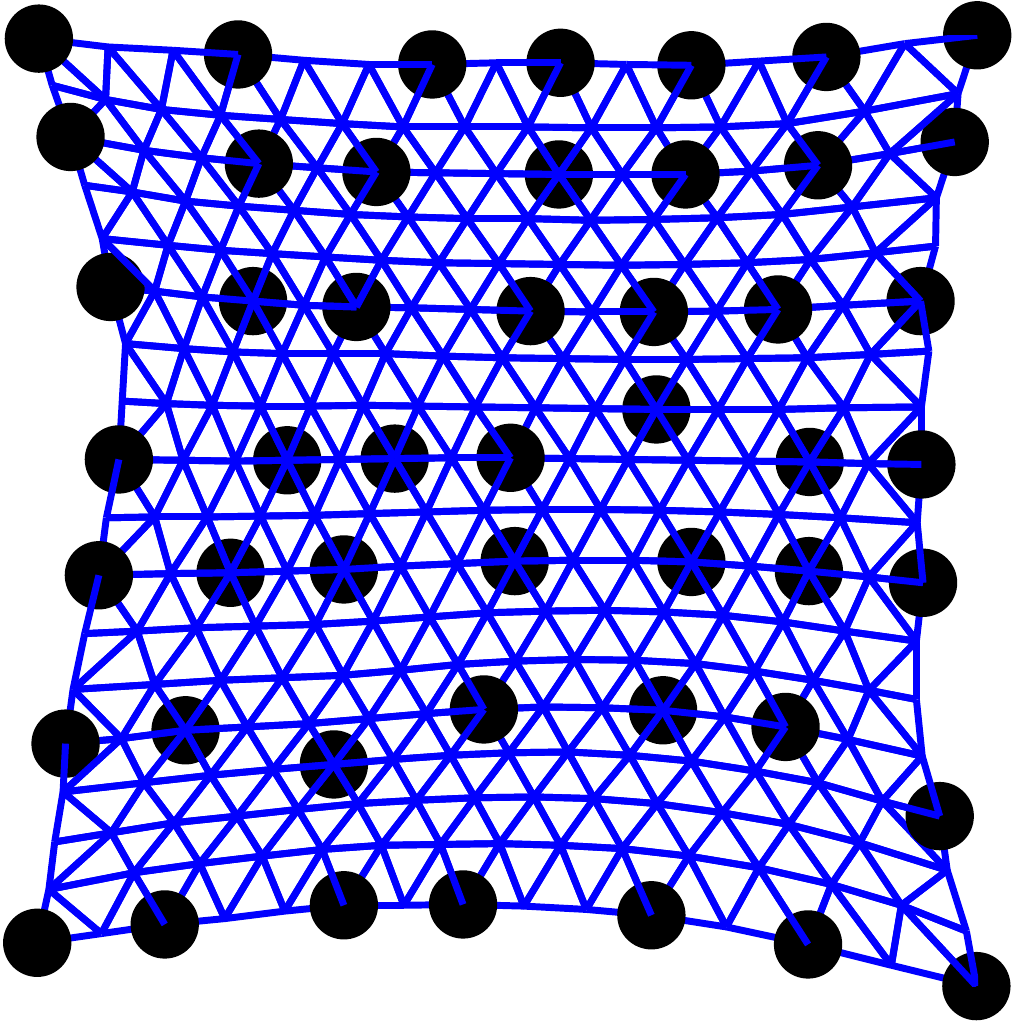} \hspace{\ctrlselspace} & \includegraphics[width=\ctrlselsrndwidth]{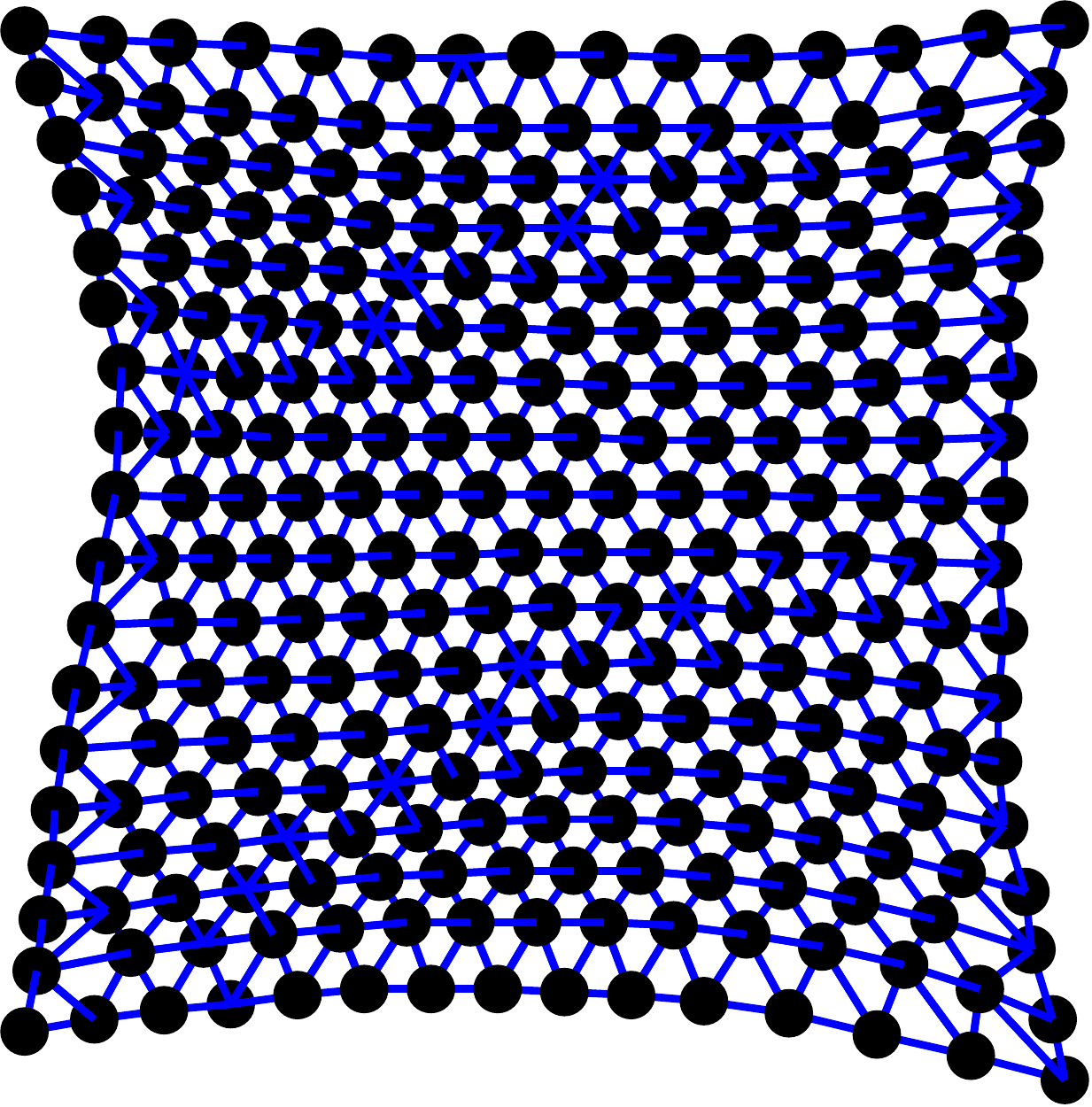} \hspace{\ctrlselspace} & \includegraphics[width=\ctrlselsrndwidth]{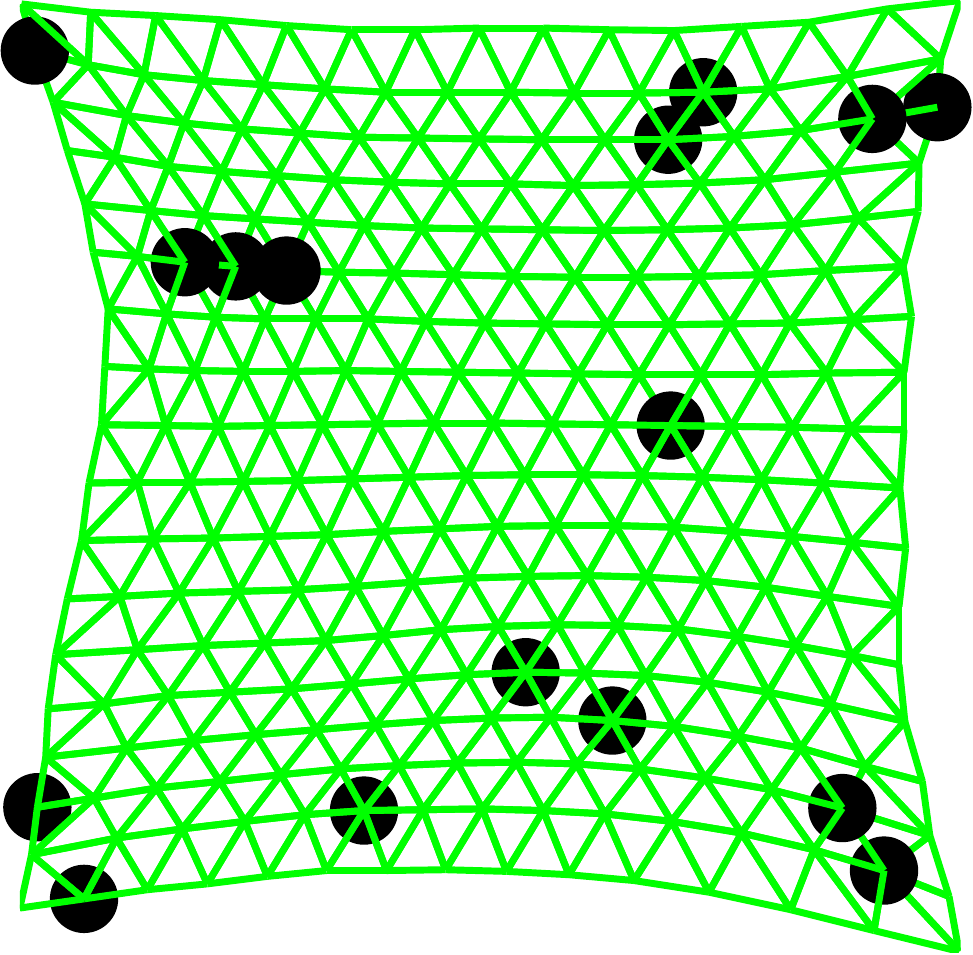} \hspace{\ctrlselspace} & \includegraphics[width=\ctrlselsrndwidth]{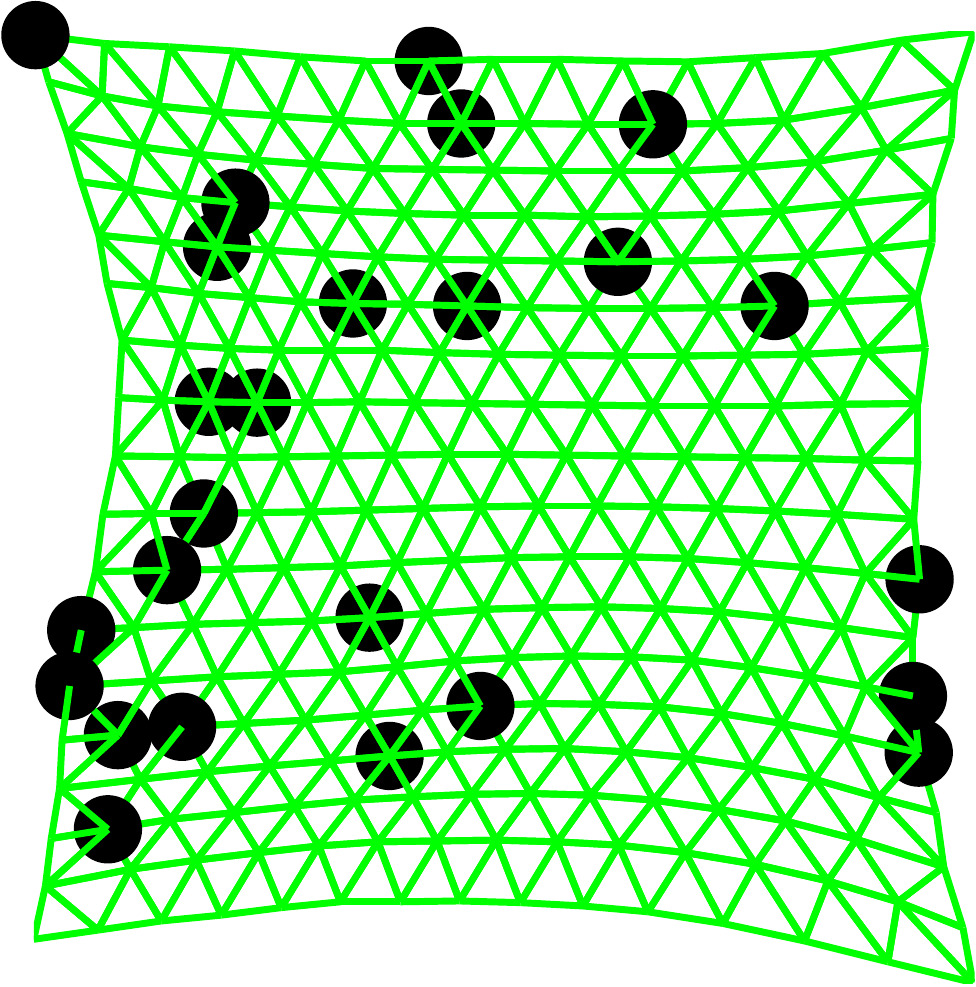} \hspace{\ctrlselspace} & \includegraphics[width=\ctrlselsrndwidth]{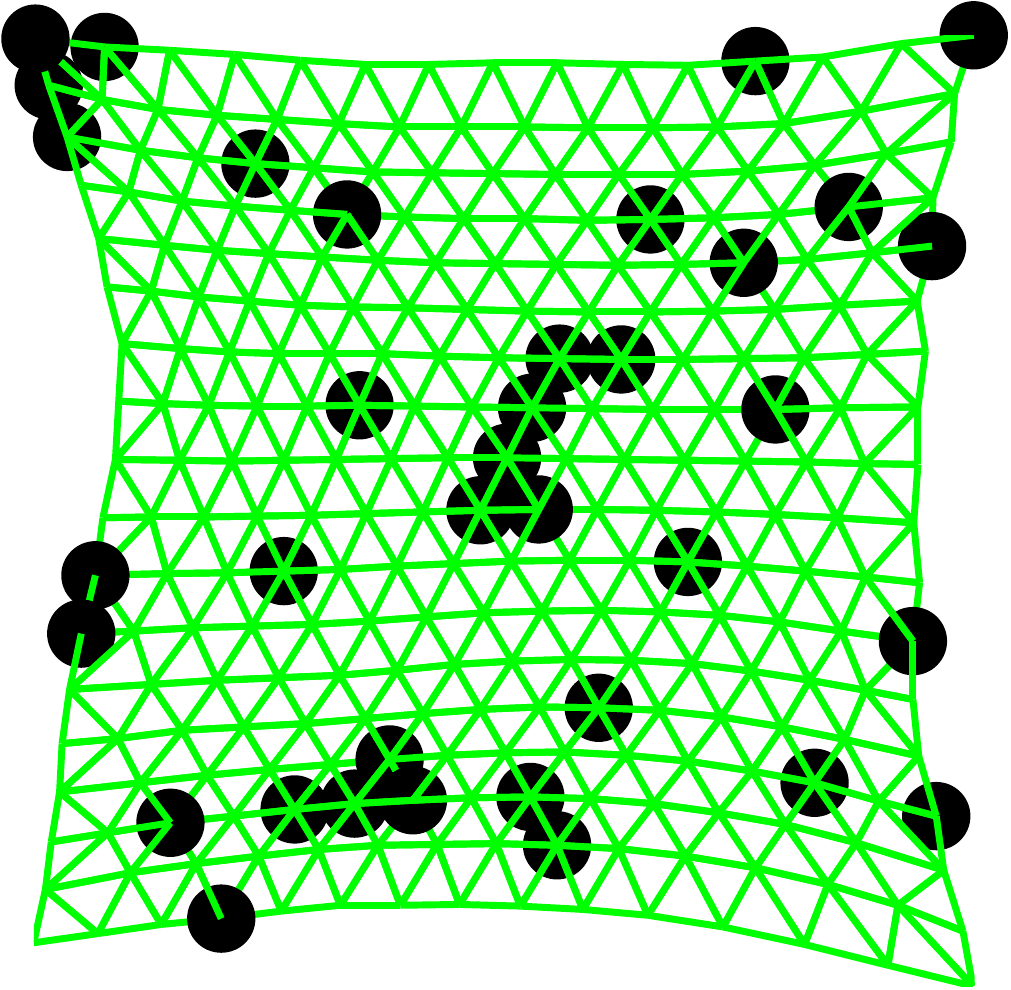} \hspace{\ctrlselspace} & \includegraphics[width=\ctrlselsrndwidth]{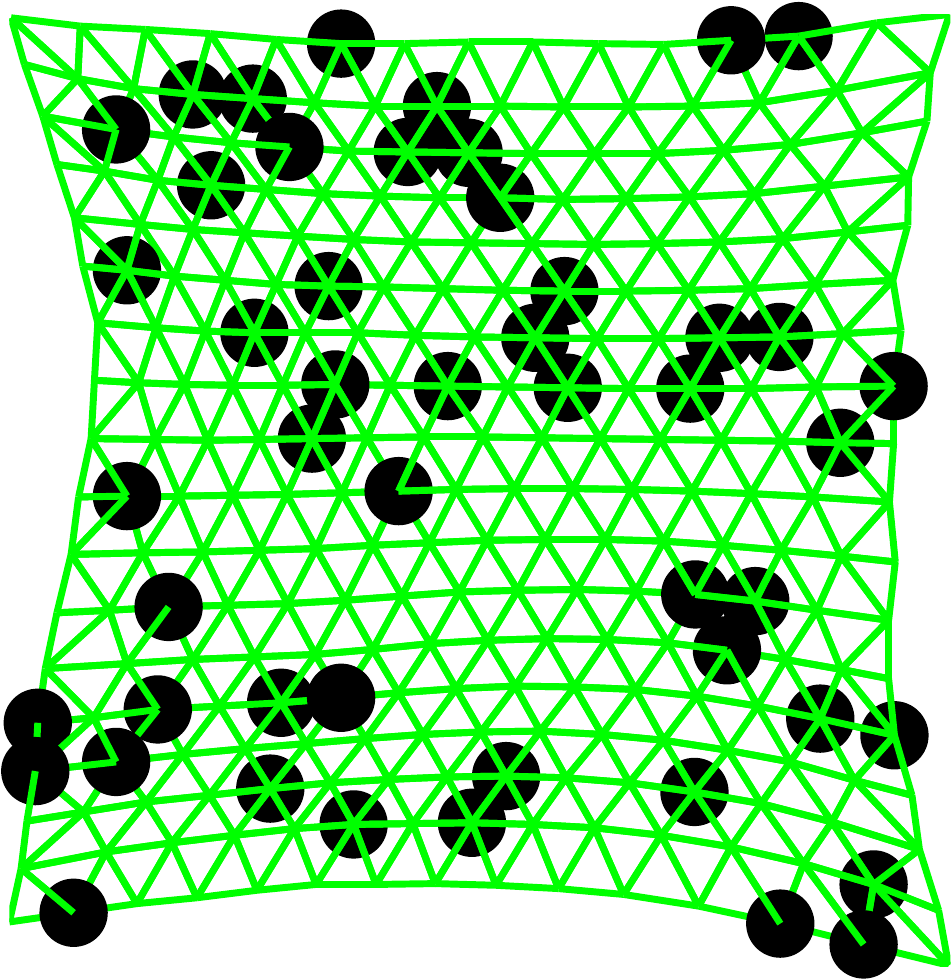}\\
(16) \hspace{\ctrlselspace} & (25) \hspace{\ctrlselspace} & (36) \hspace{\ctrlselspace} & (49) \hspace{\ctrlselspace} & (270) \hspace{\ctrlselspace} & (16) \hspace{\ctrlselspace} & (25) \hspace{\ctrlselspace} & (36) \hspace{\ctrlselspace} & (49)\\
\includegraphics[width=\ctrlselsrndwidth]{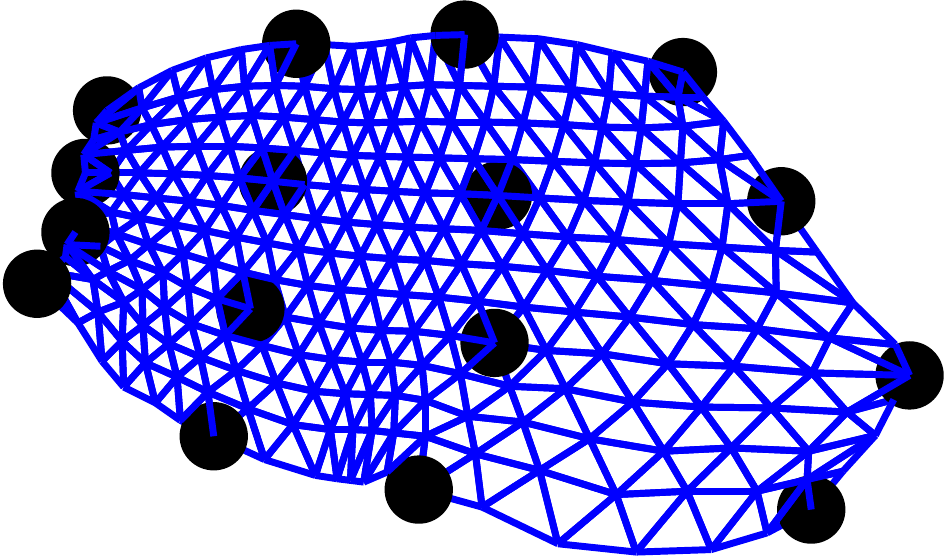} \hspace{\ctrlselspace} & \includegraphics[width=\ctrlselsrndwidth]{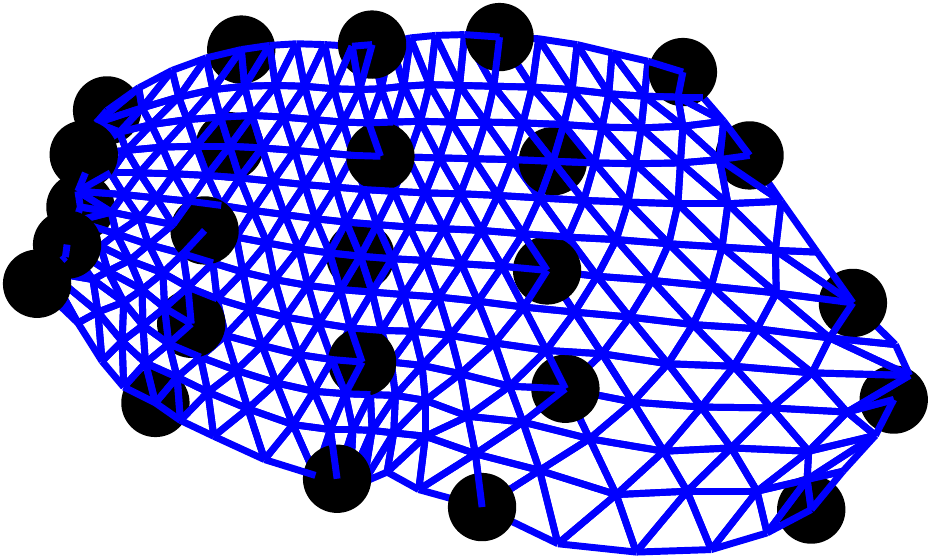} \hspace{\ctrlselspace} & \includegraphics[width=\ctrlselsrndwidth]{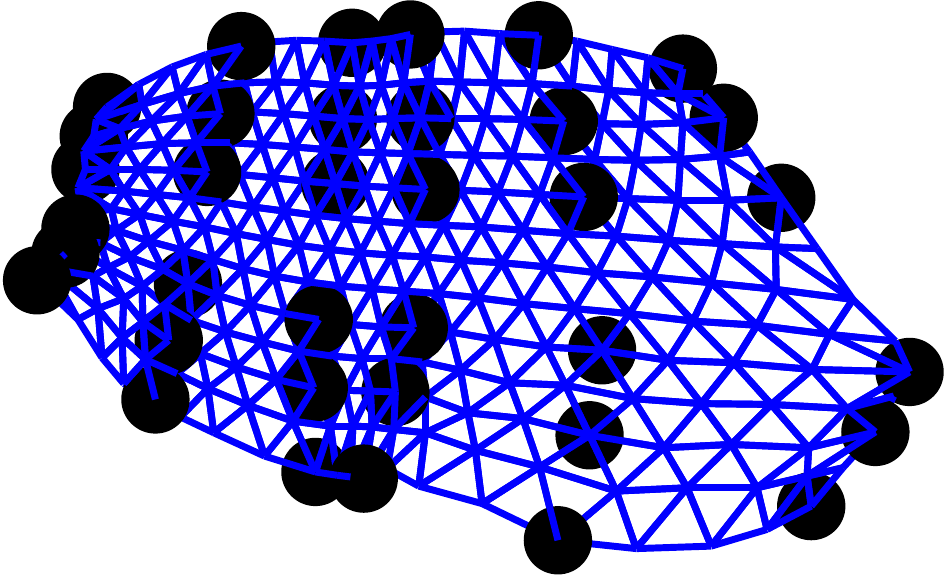} \hspace{\ctrlselspace} & \includegraphics[width=\ctrlselsrndwidth]{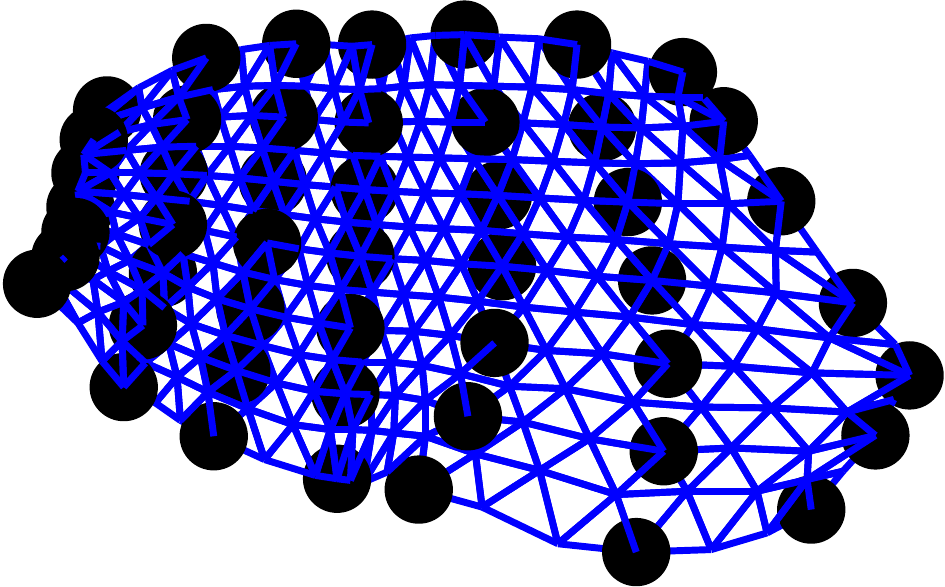} \hspace{\ctrlselspace} & \includegraphics[width=\ctrlselsrndwidth]{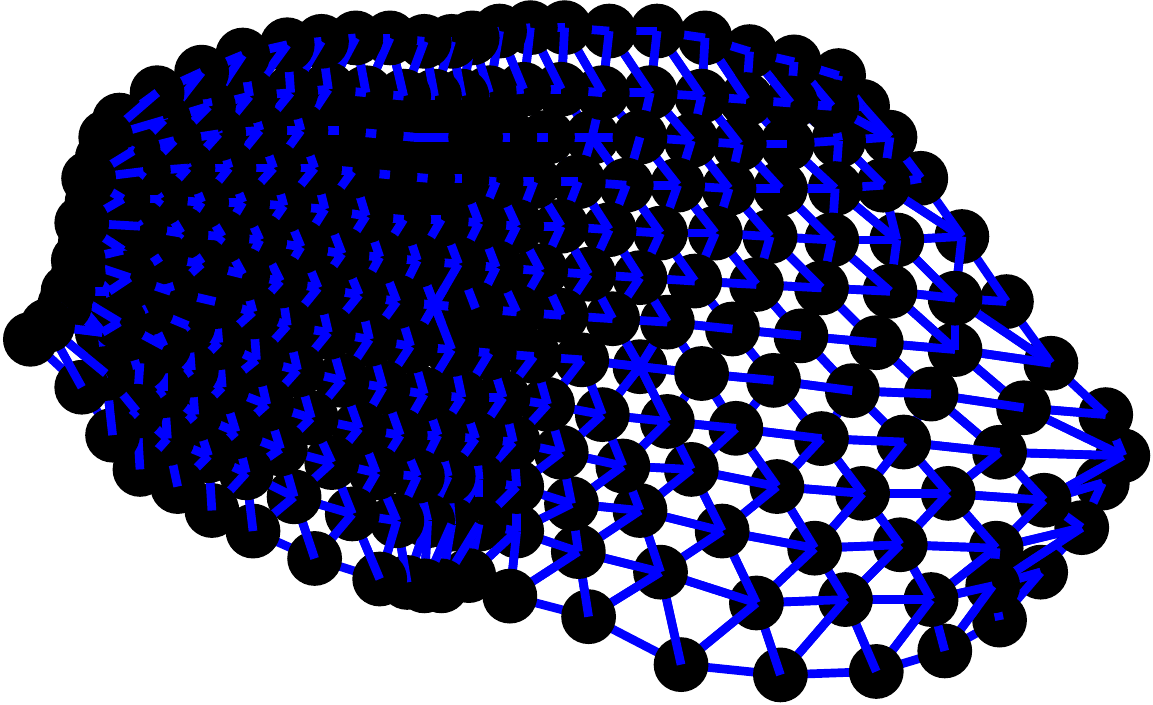} \hspace{\ctrlselspace} & \includegraphics[width=\ctrlselsrndwidth]{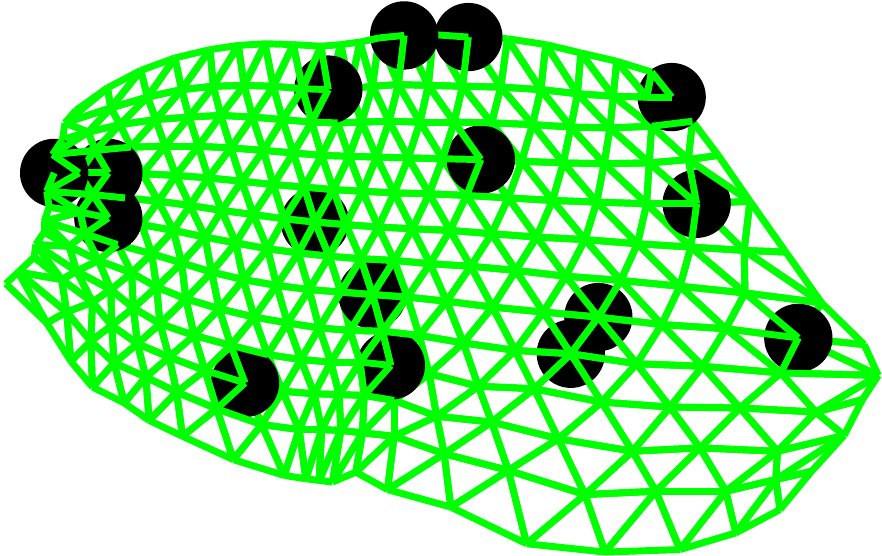} \hspace{\ctrlselspace} & \includegraphics[width=\ctrlselsrndwidth]{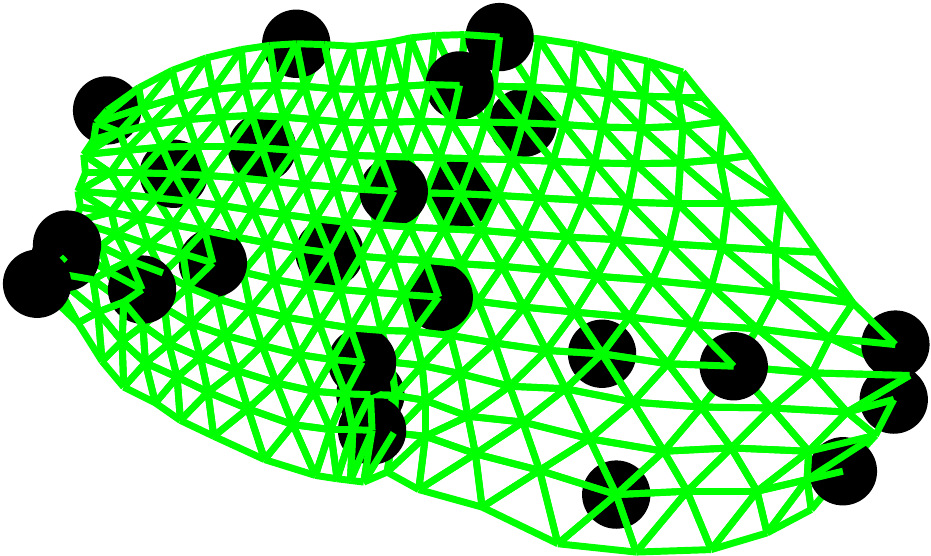} \hspace{\ctrlselspace} & \includegraphics[width=\ctrlselsrndwidth]{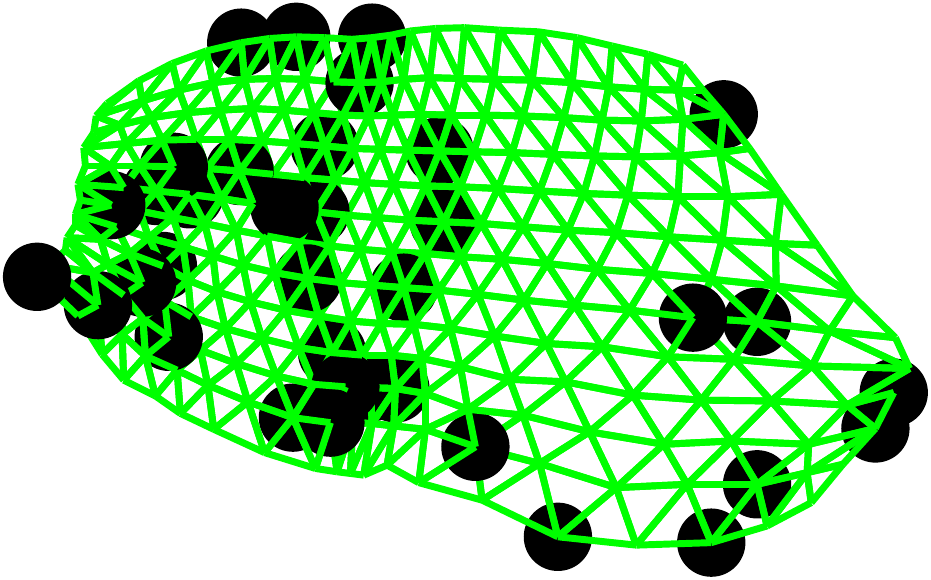} \hspace{\ctrlselspace} & \includegraphics[width=\ctrlselsrndwidth]{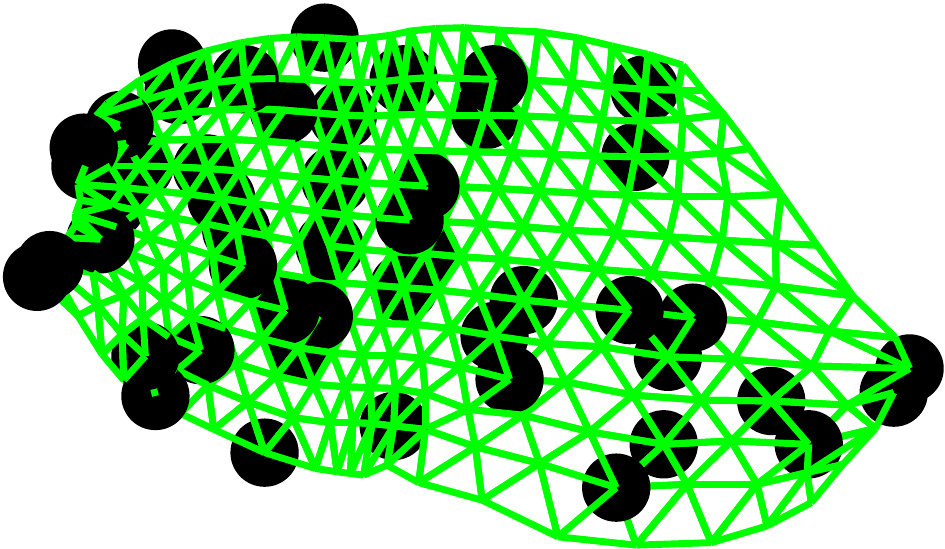}\\
(16) \hspace{\ctrlselspace} & (25) \hspace{\ctrlselspace} & (36) \hspace{\ctrlselspace} & (49) \hspace{\ctrlselspace} & (260) \hspace{\ctrlselspace} & (16) \hspace{\ctrlselspace} & (25) \hspace{\ctrlselspace} & (36) \hspace{\ctrlselspace} & (49)\\
\includegraphics[width=\ctrlselsrndwidth]{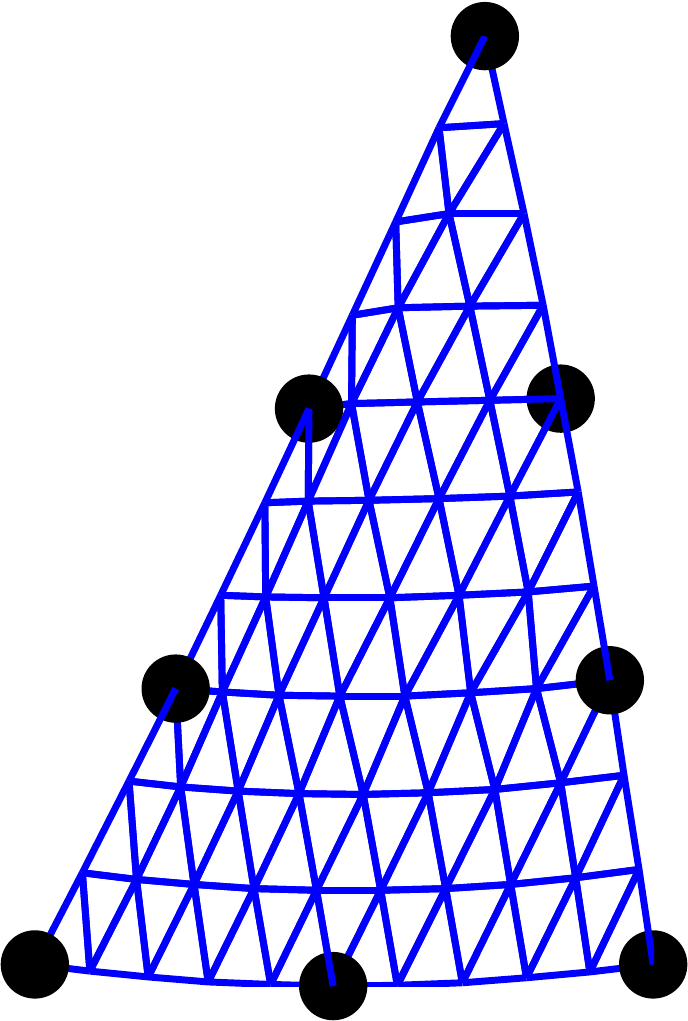} \hspace{\ctrlselspace} & \includegraphics[width=\ctrlselsrndwidth]{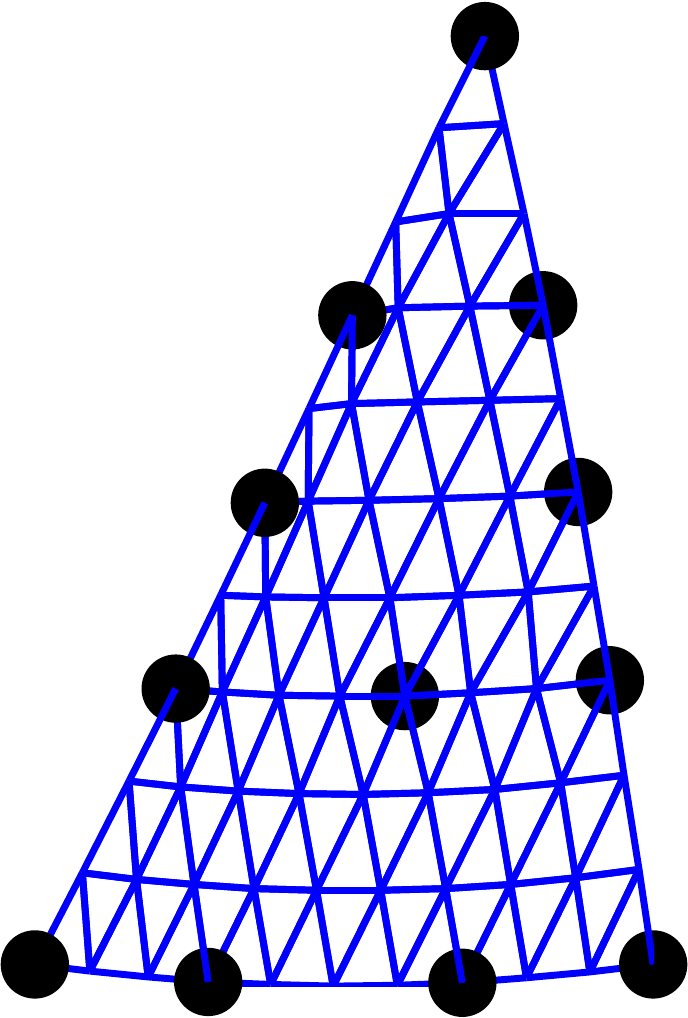} \hspace{\ctrlselspace} & \includegraphics[width=\ctrlselsrndwidth]{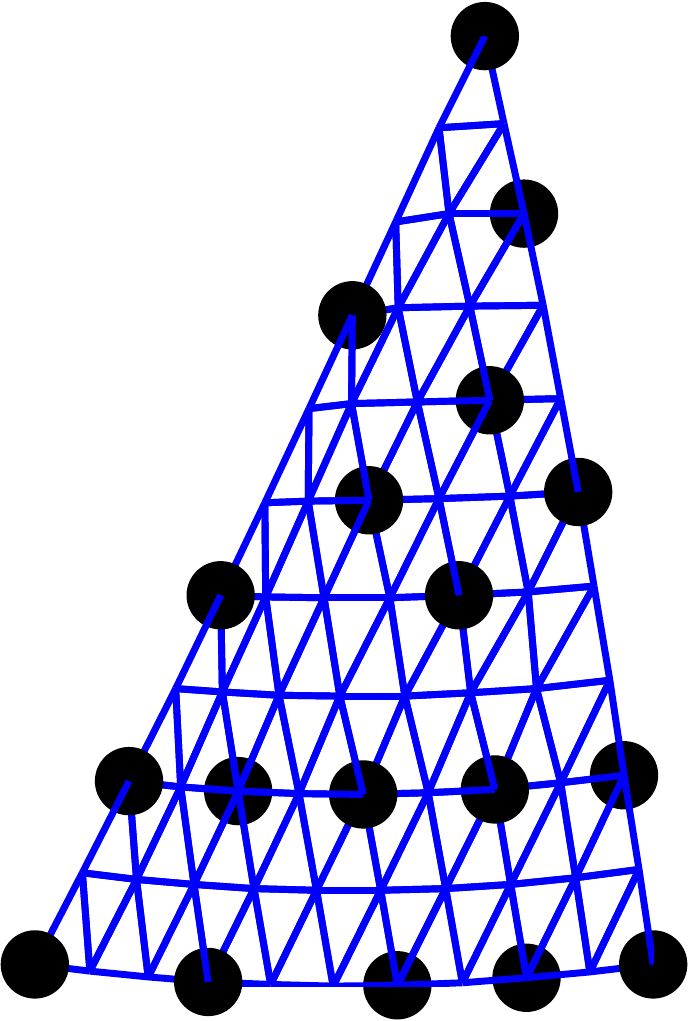} \hspace{\ctrlselspace} & \includegraphics[width=\ctrlselsrndwidth]{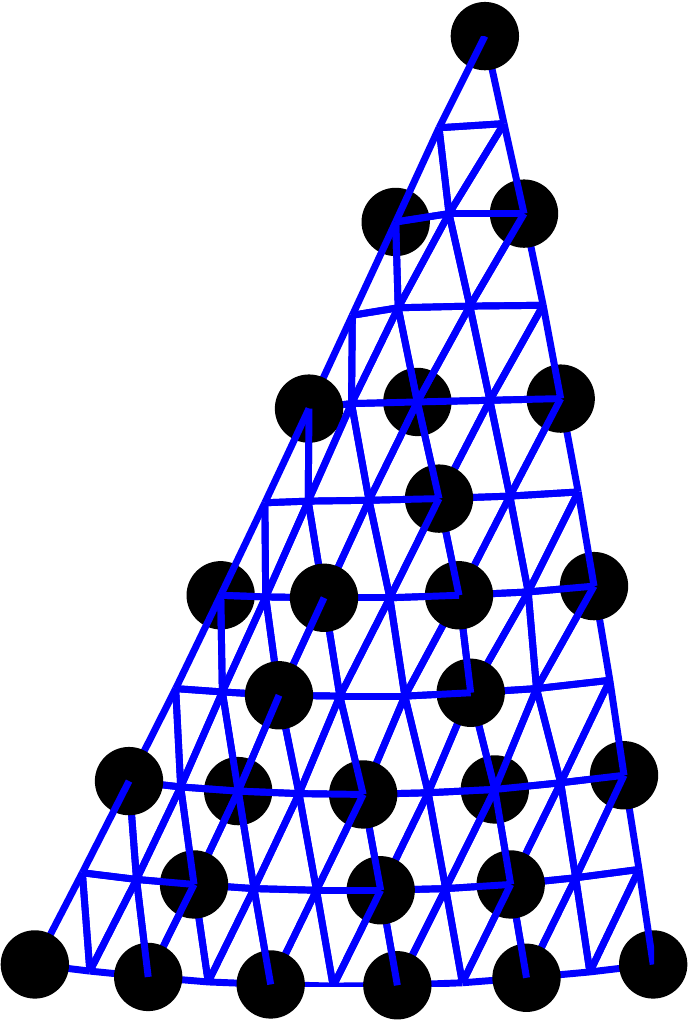} \hspace{\ctrlselspace} & \includegraphics[width=\ctrlselsrndwidth]{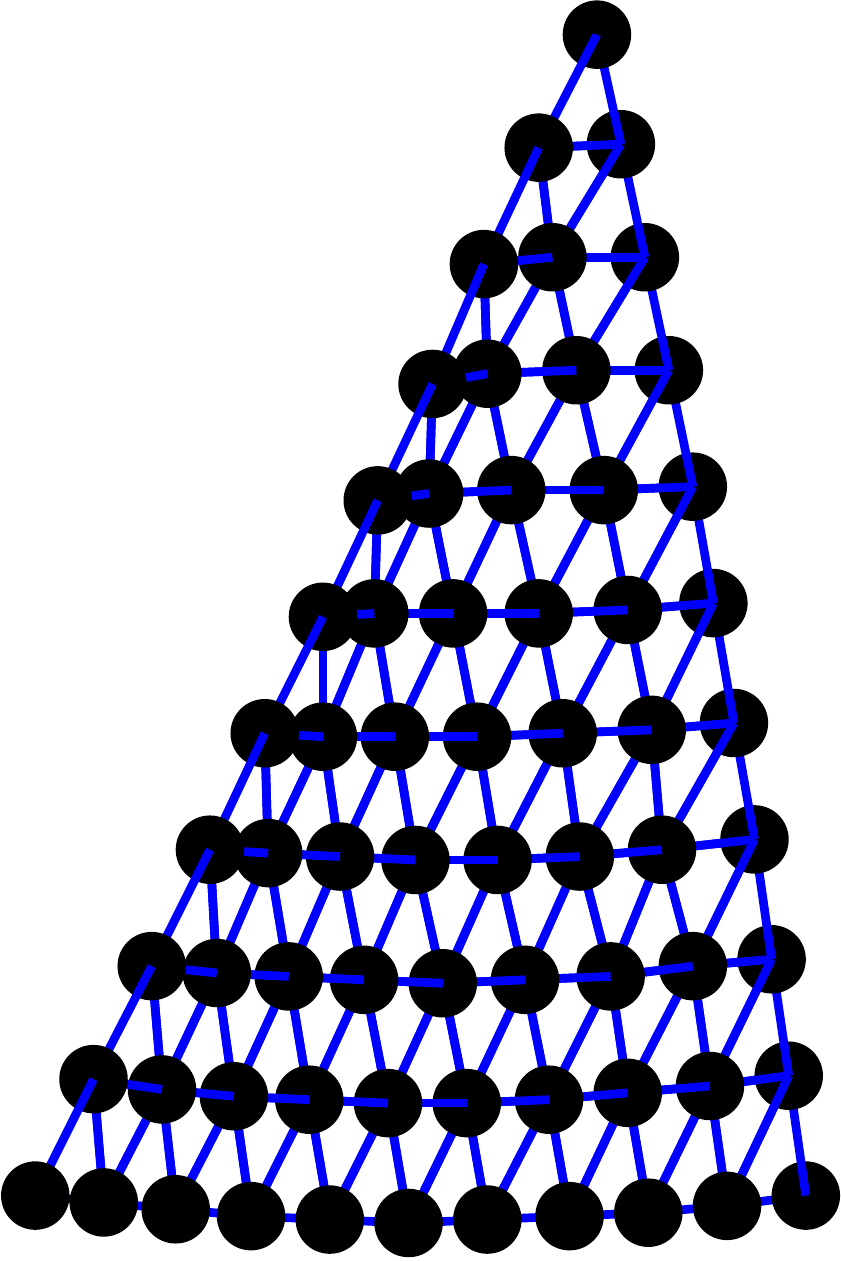} \hspace{\ctrlselspace} & \includegraphics[width=\ctrlselsrndwidth]{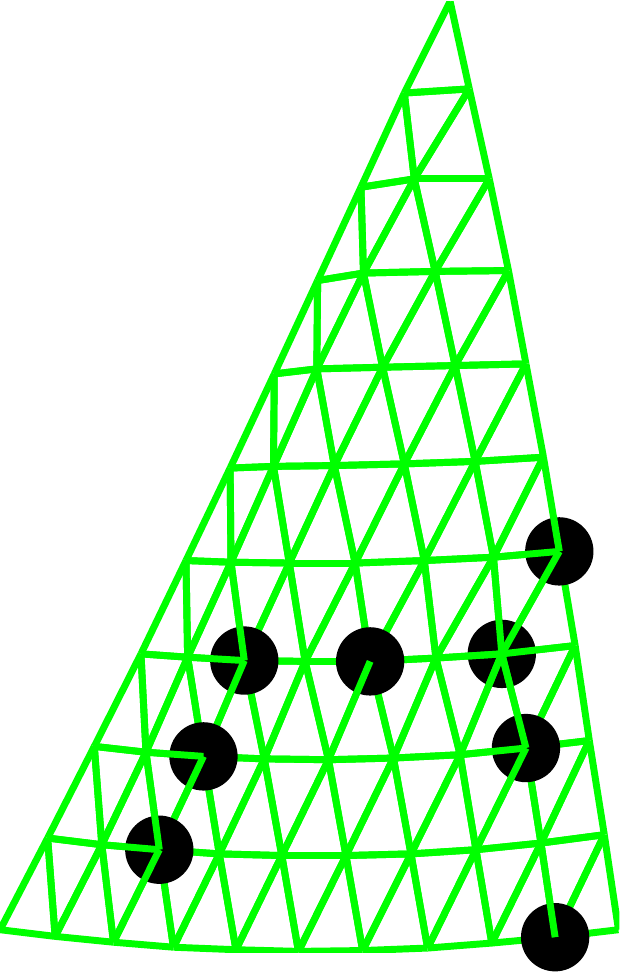} \hspace{\ctrlselspace} & \includegraphics[width=\ctrlselsrndwidth]{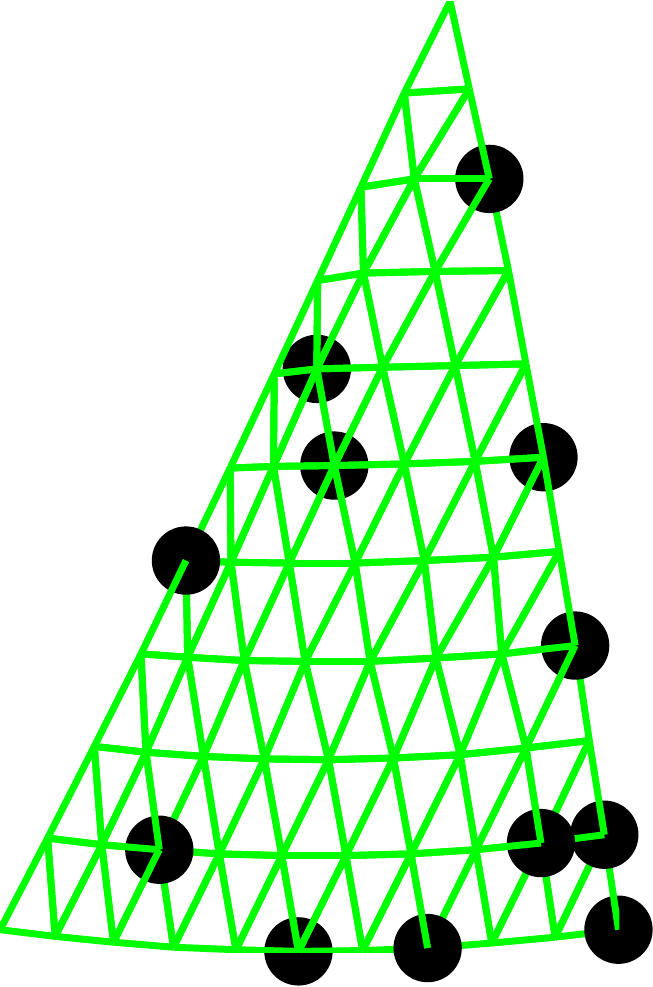} \hspace{\ctrlselspace} & \includegraphics[width=\ctrlselsrndwidth]{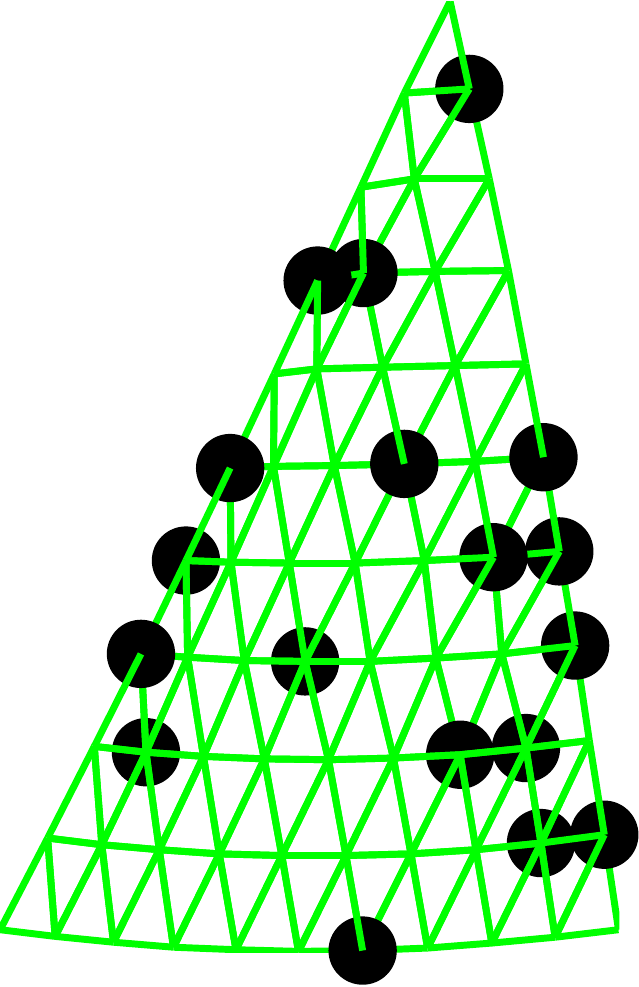} \hspace{\ctrlselspace} & \includegraphics[width=\ctrlselsrndwidth]{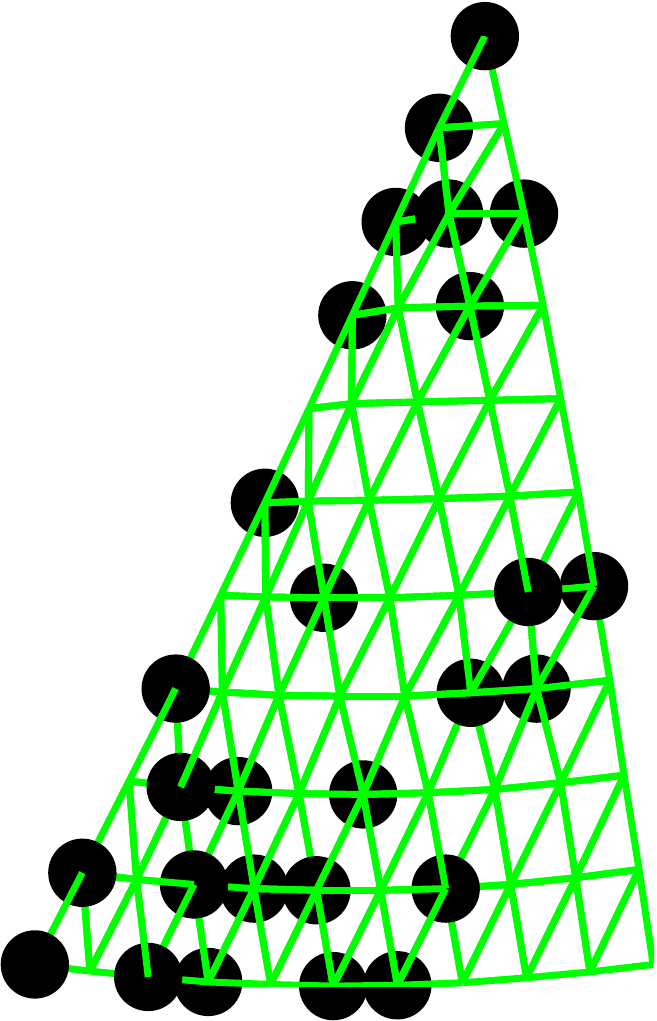}\\
(8) \hspace{\ctrlselspace} & (12) \hspace{\ctrlselspace} & (18) \hspace{\ctrlselspace} & (27) \hspace{\ctrlselspace} & (66) \hspace{\ctrlselspace} & (8) \hspace{\ctrlselspace} & (12) \hspace{\ctrlselspace} & (18) \hspace{\ctrlselspace} & (27)
\end{tabular}

%% file: figs_ctest022results_table.tex
\newcommand{\ctestaccwidth}{0.2\linewidth}
\newcommand{\ctestaccheight}{2.6cm}
\begin{tabular}{ccccc}
\includegraphics[height=\ctestaccheight]{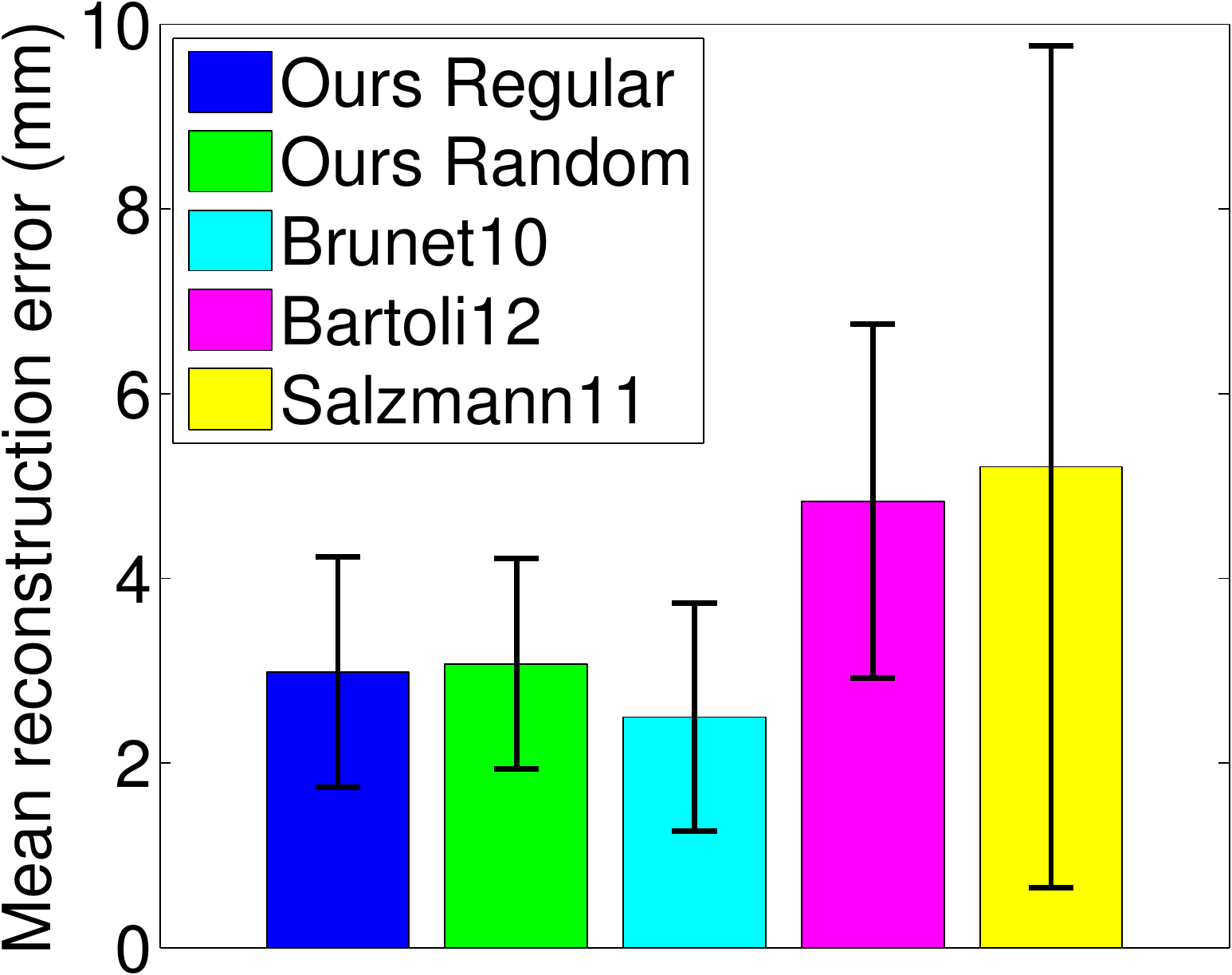} & \includegraphics[height=\ctestaccheight]{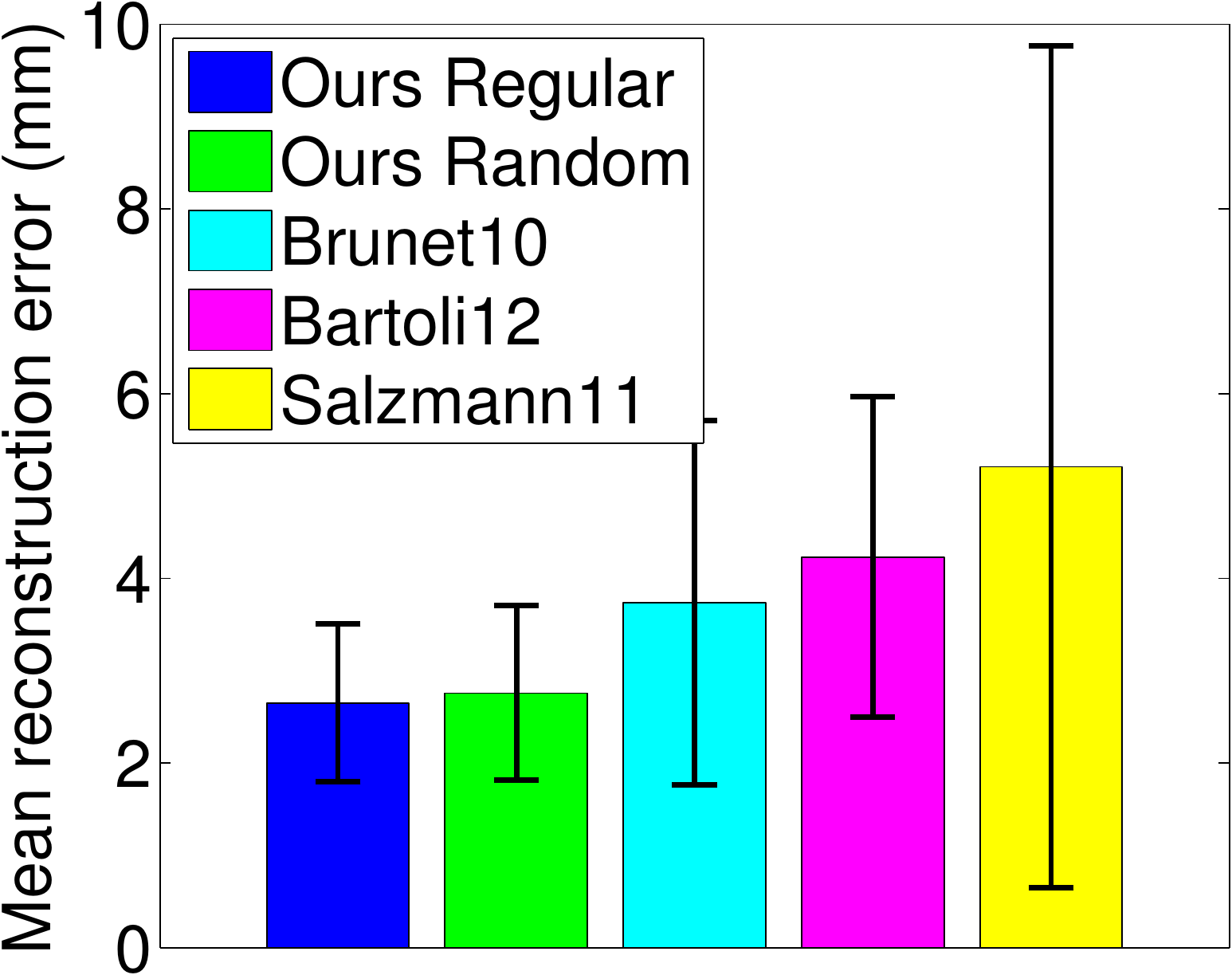} & \includegraphics[height=\ctestaccheight]{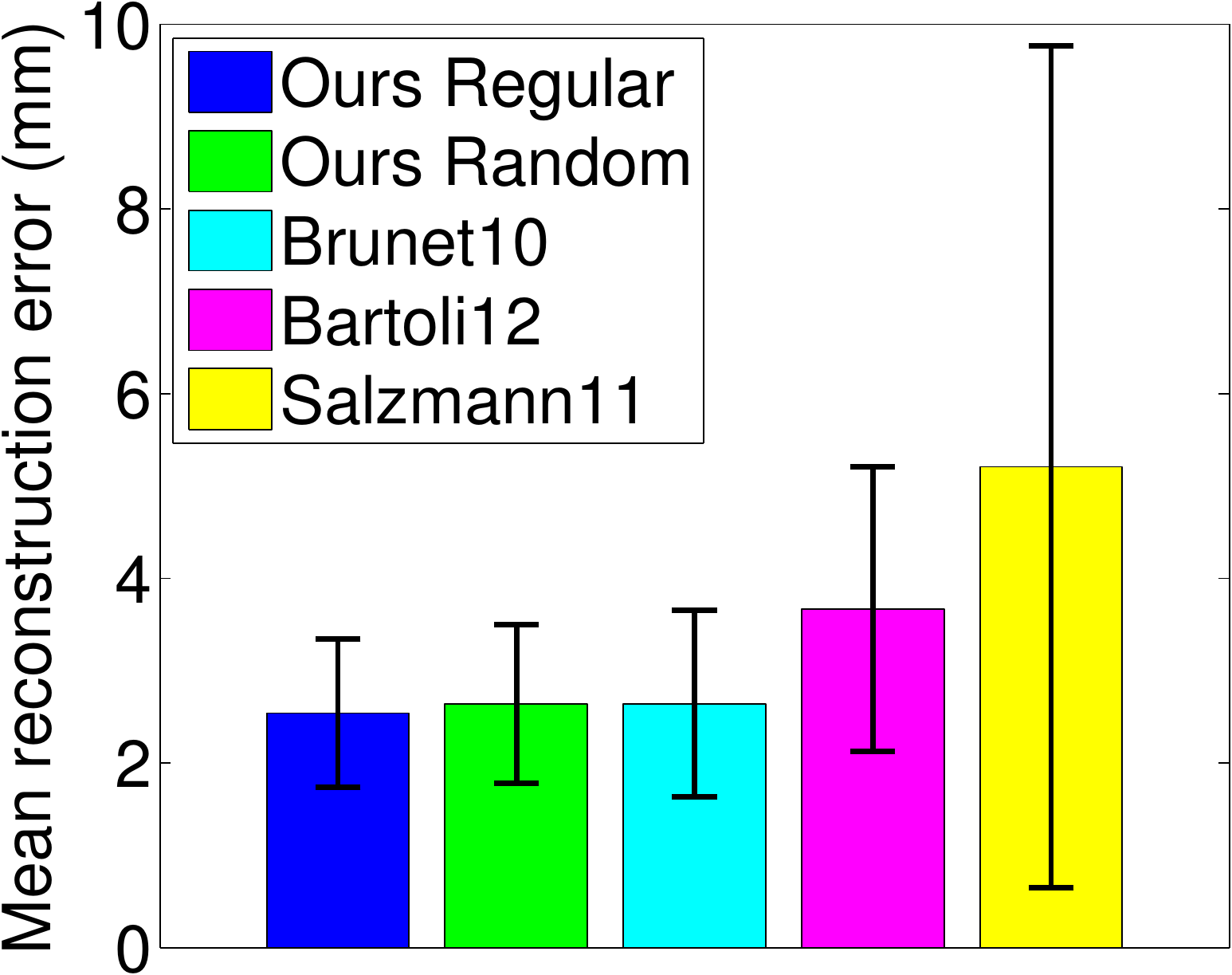} & \includegraphics[height=\ctestaccheight]{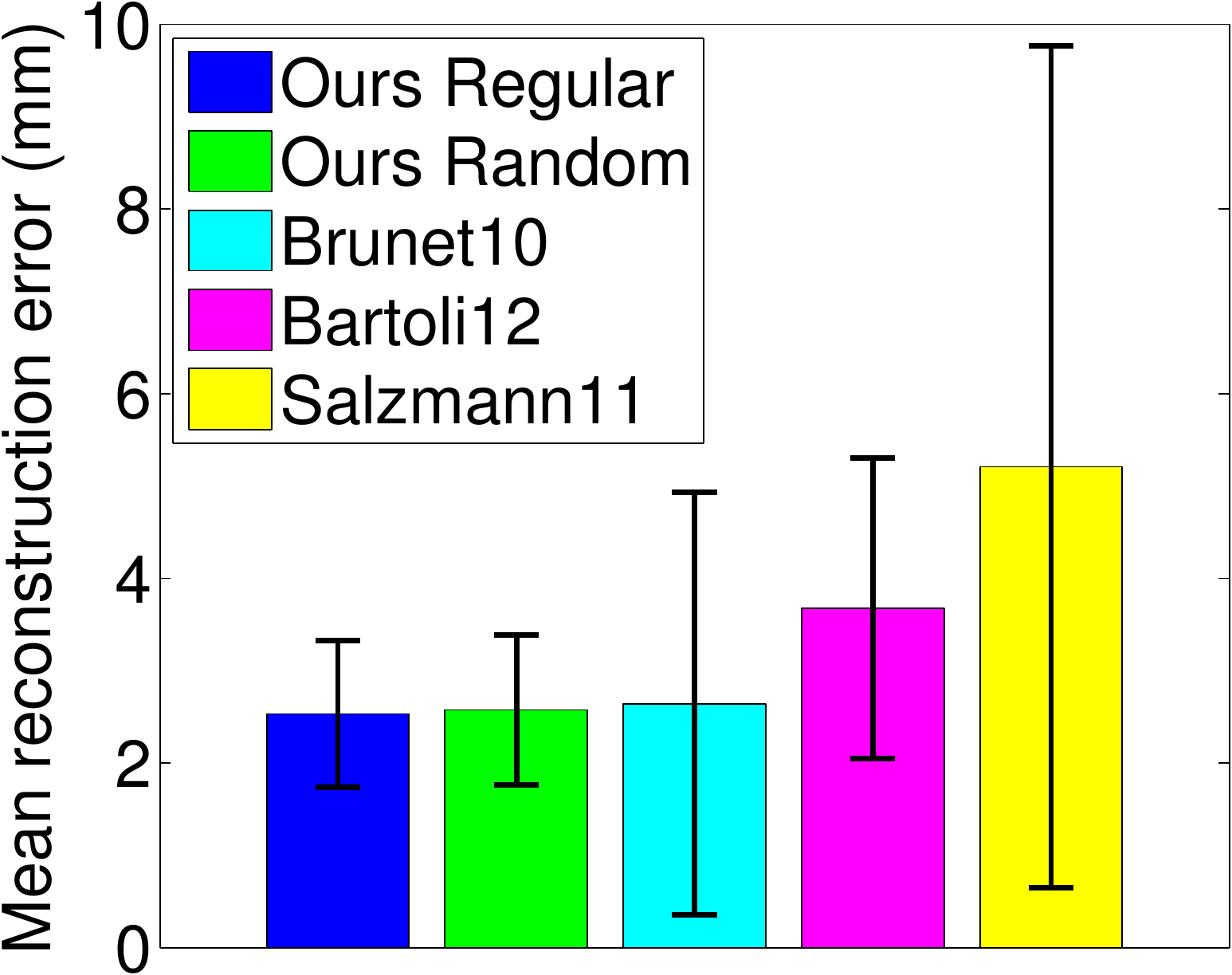} & \includegraphics[height=\ctestaccheight]{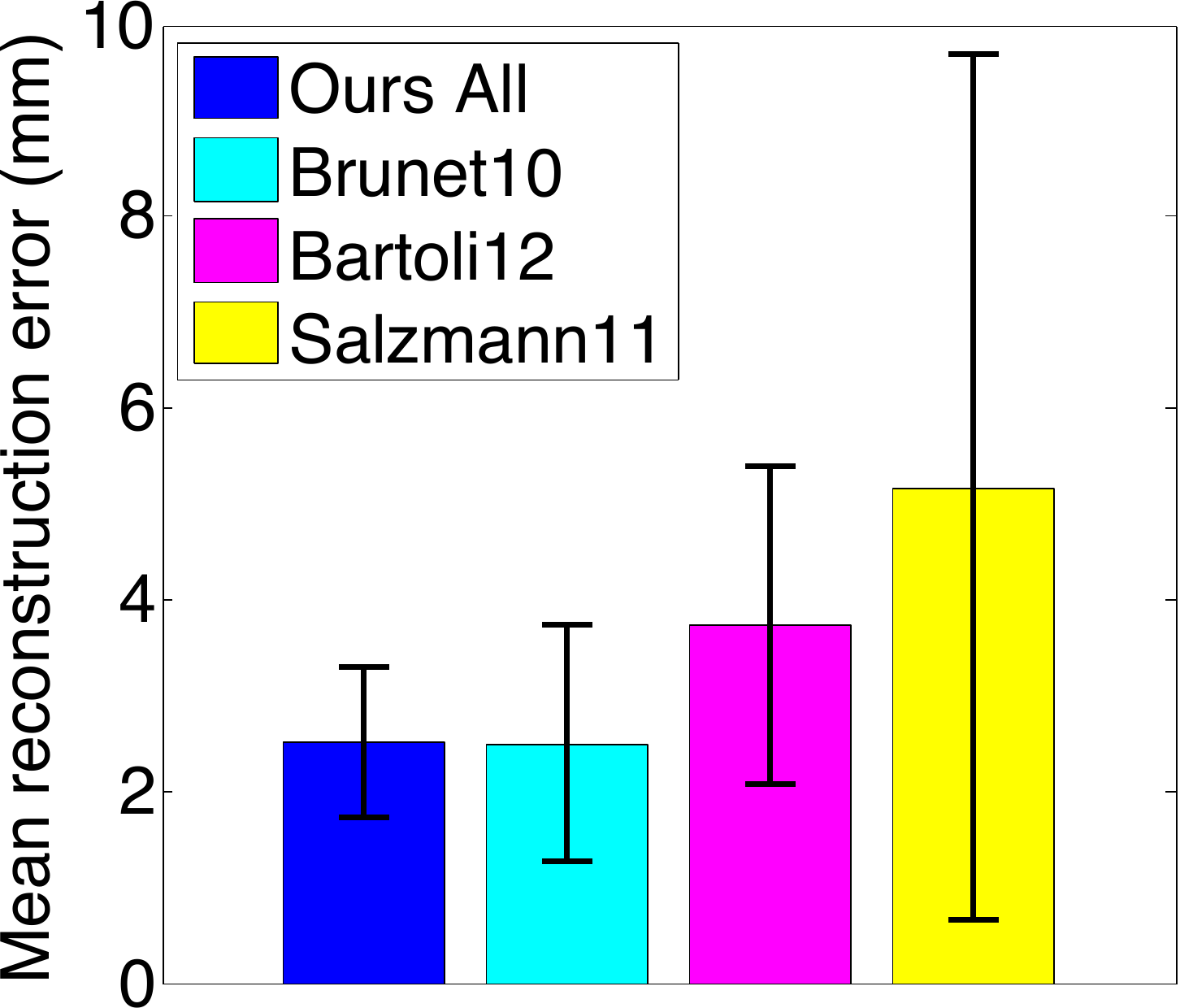} \\
(16) & (25) & (36) & (49) & (99)
\end{tabular}

%% file: figs_ctest030results_table.tex
\newcommand{\ctestadawidth}{0.225\linewidth}
\newcommand{\ctestadaheight}{2.475cm}
\begin{tabular}{ccccc}
\includegraphics[height=\ctestadaheight]{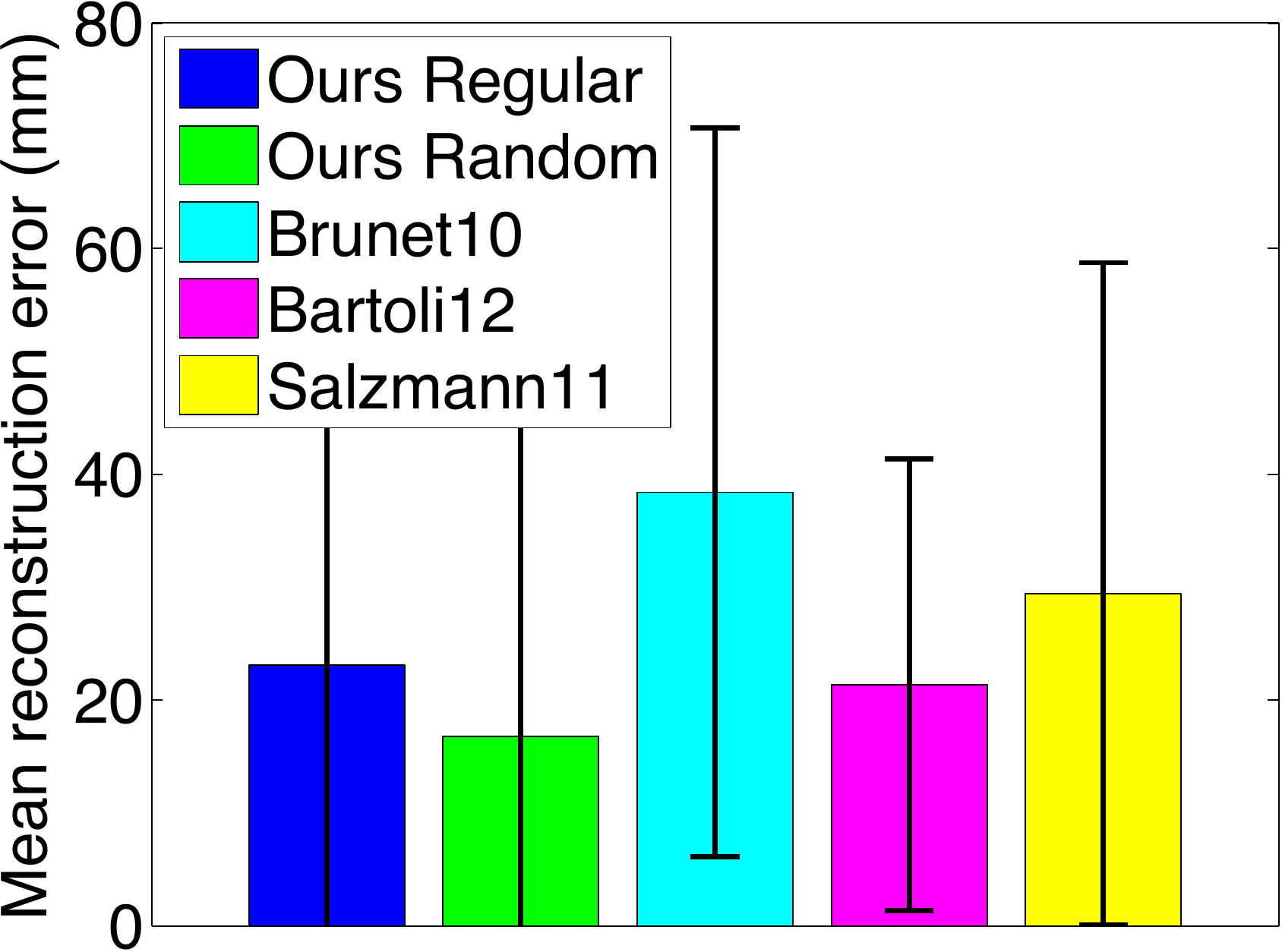} & \includegraphics[height=\ctestadaheight]{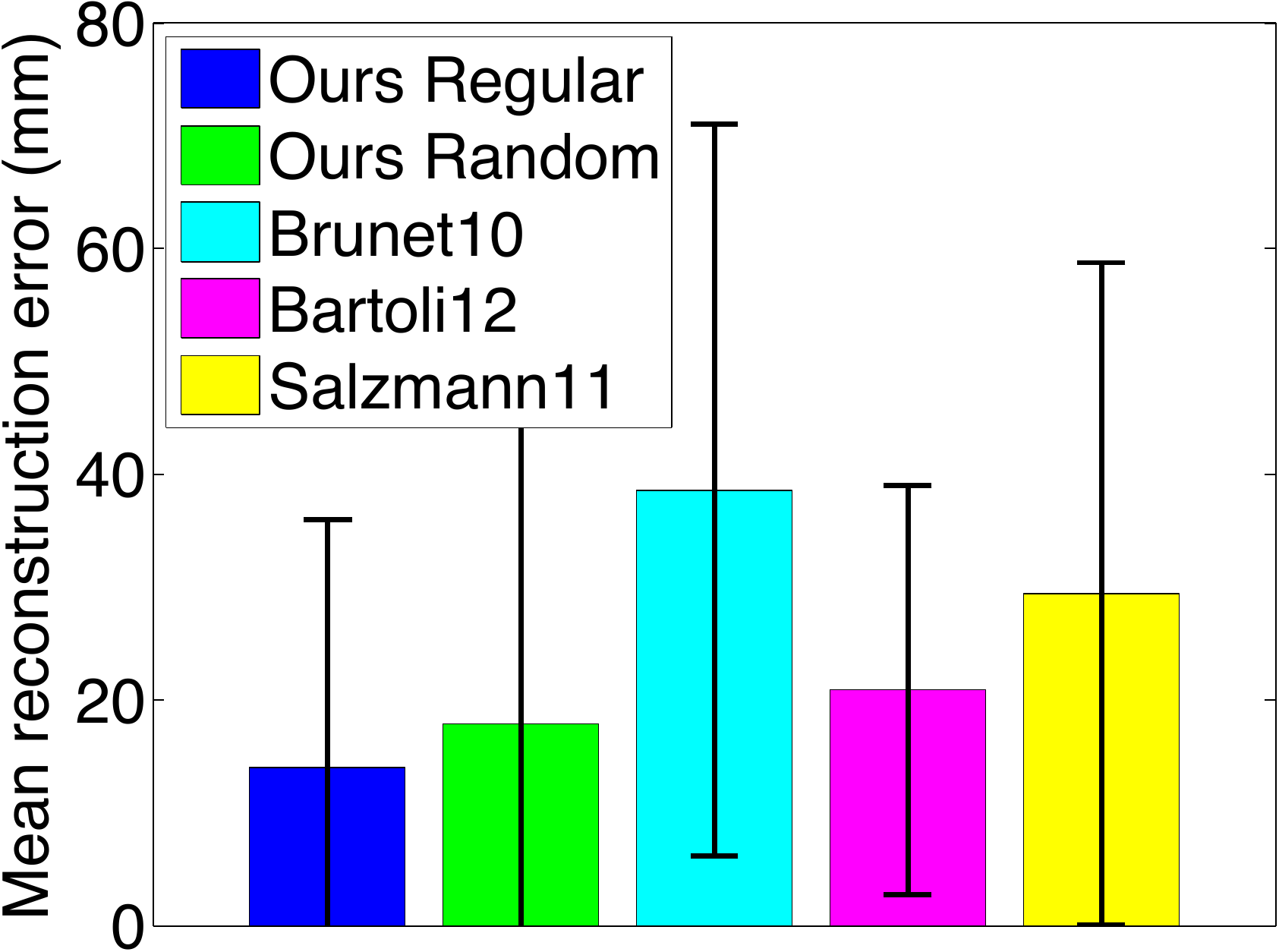} & \includegraphics[height=\ctestadaheight]{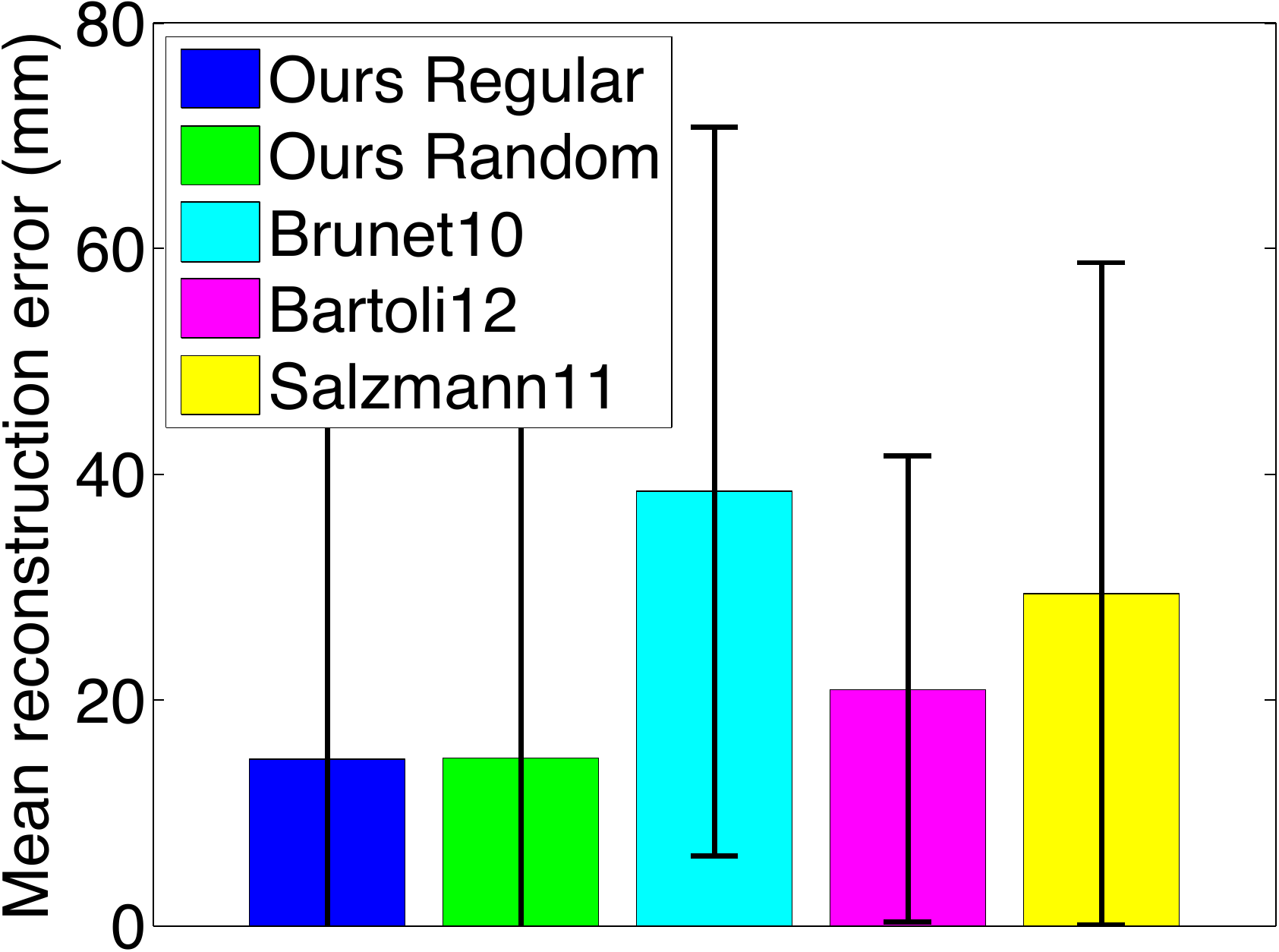} & \includegraphics[height=\ctestadaheight]{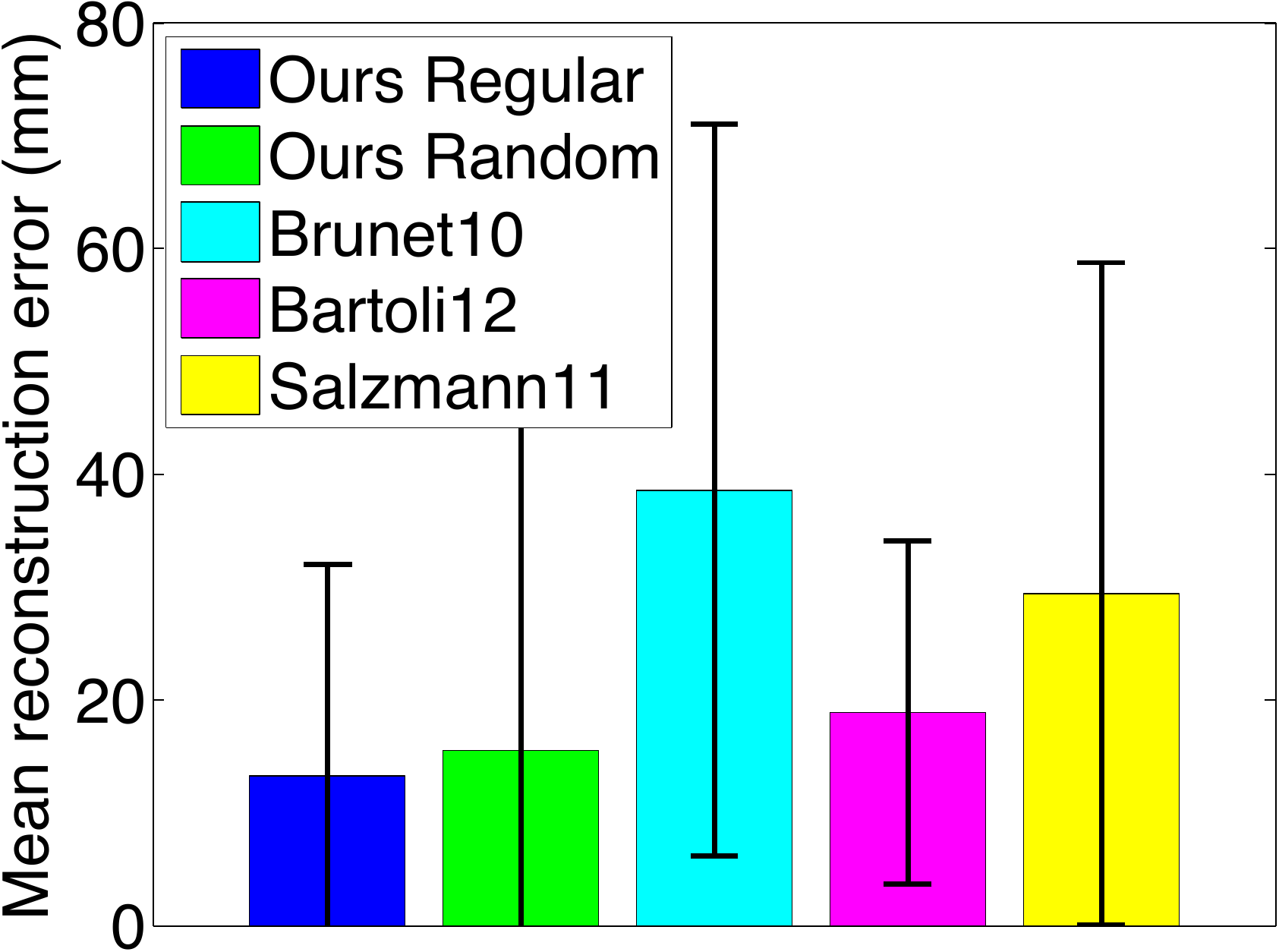} & \includegraphics[height=\ctestadaheight]{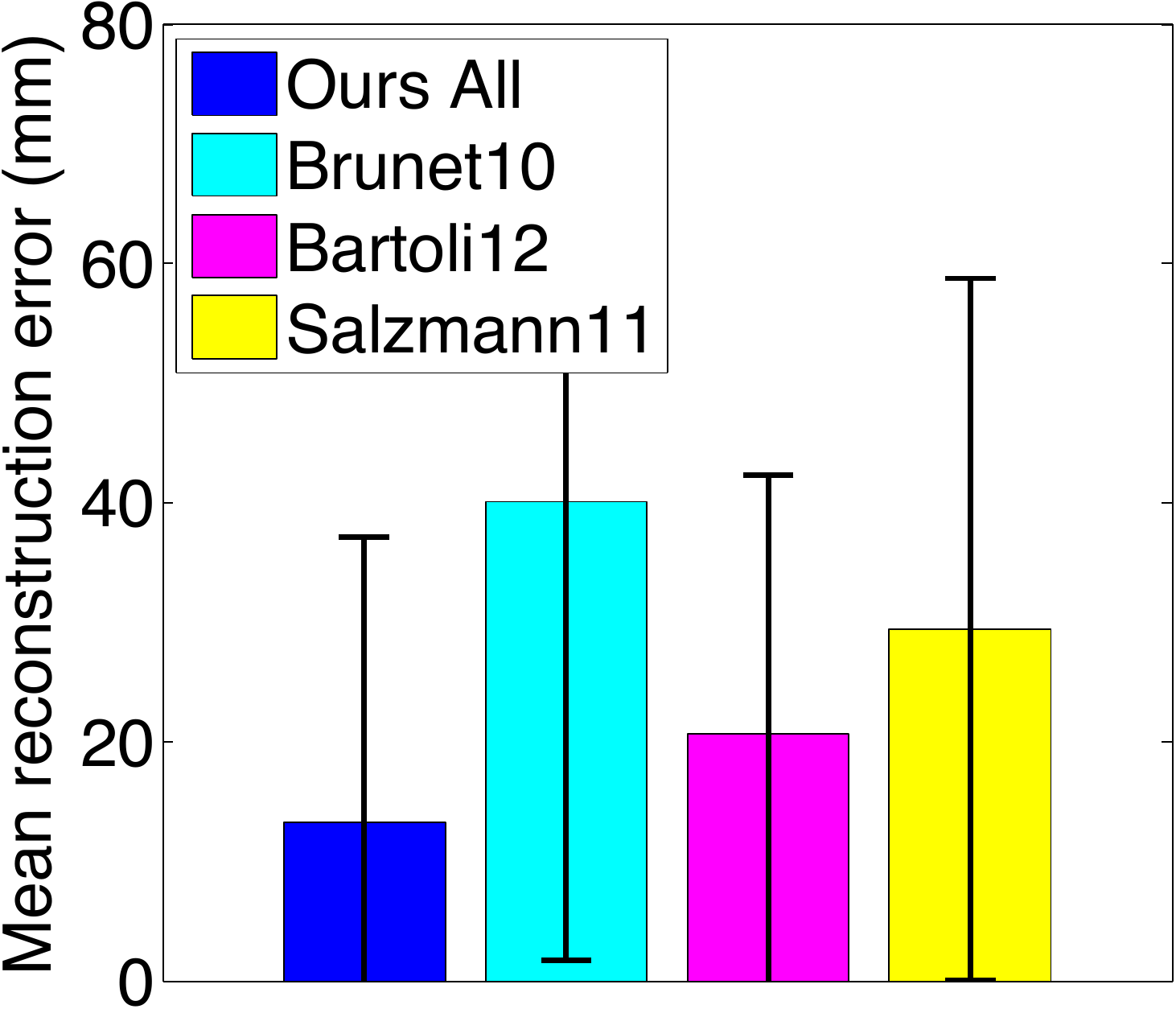}\\
(16) & (25) & (36) & (49) & (169)
\end{tabular}

%% file: figs_ctest026results_table.tex
\newcommand{\ctestacgwidth}{0.185\linewidth}
\newcommand{\ctestacgheight}{2.67cm}
\begin{tabular}{ccccc}
\includegraphics[height=\ctestacgheight]{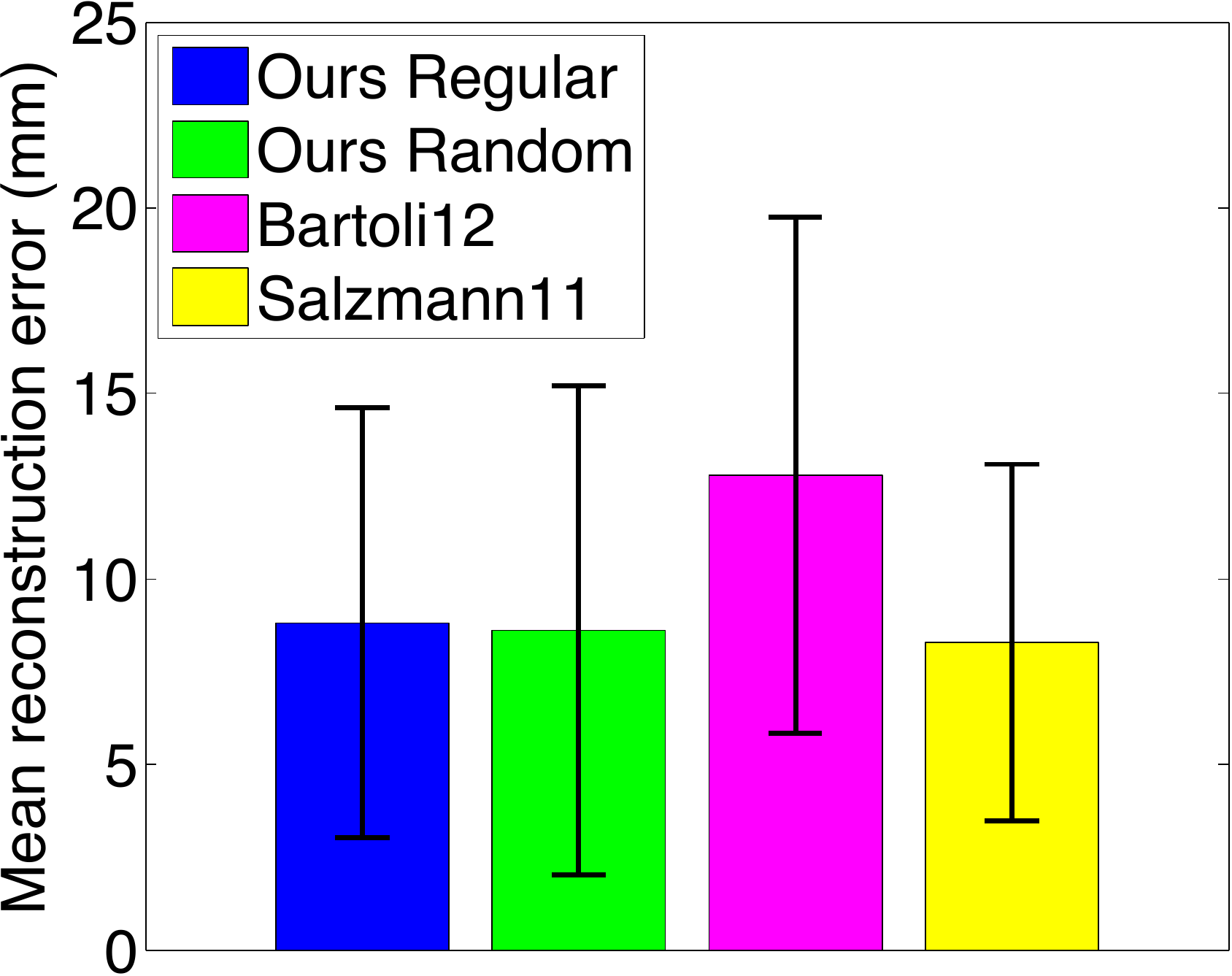} & \includegraphics[height=\ctestacgheight]{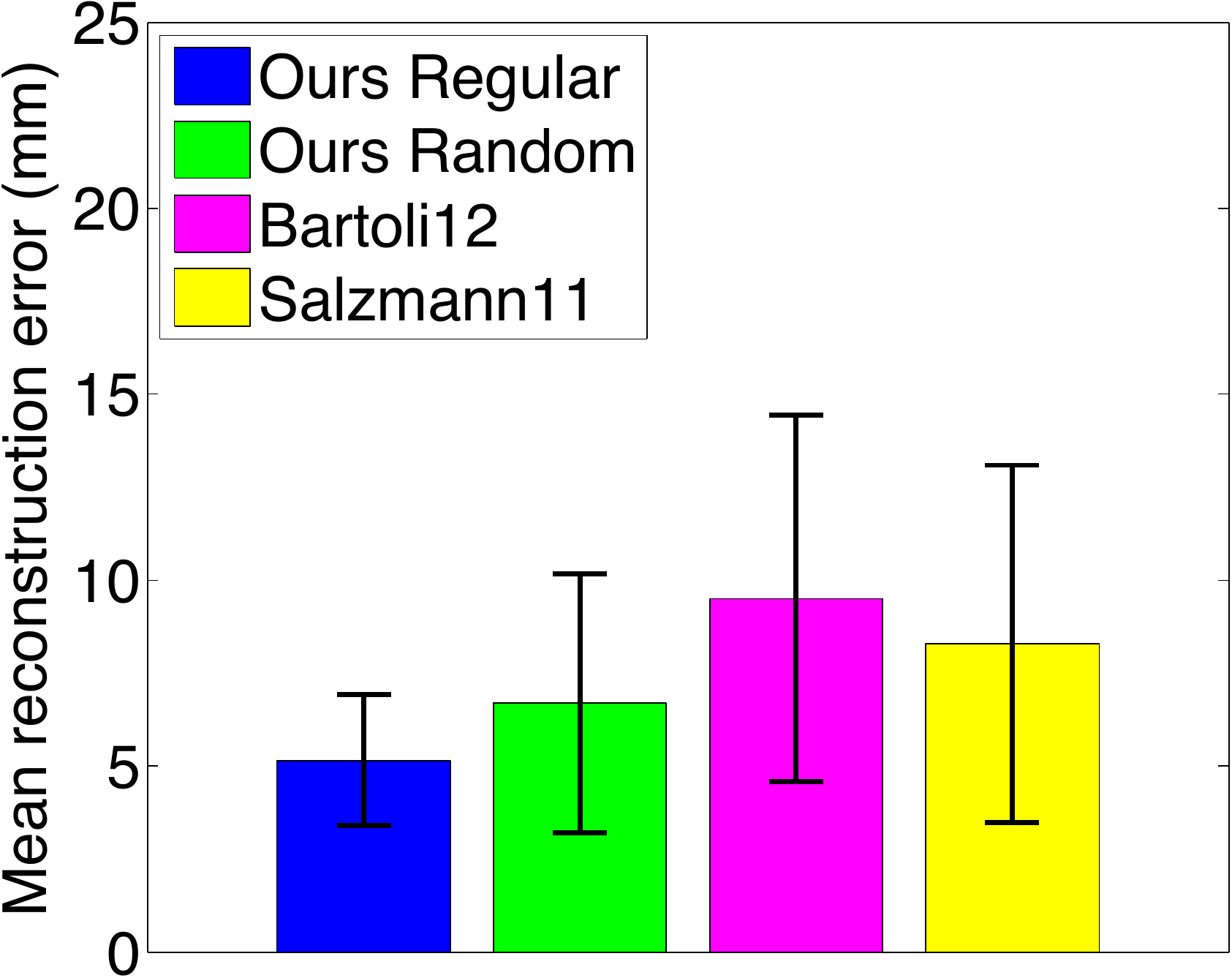} & \includegraphics[height=\ctestacgheight]{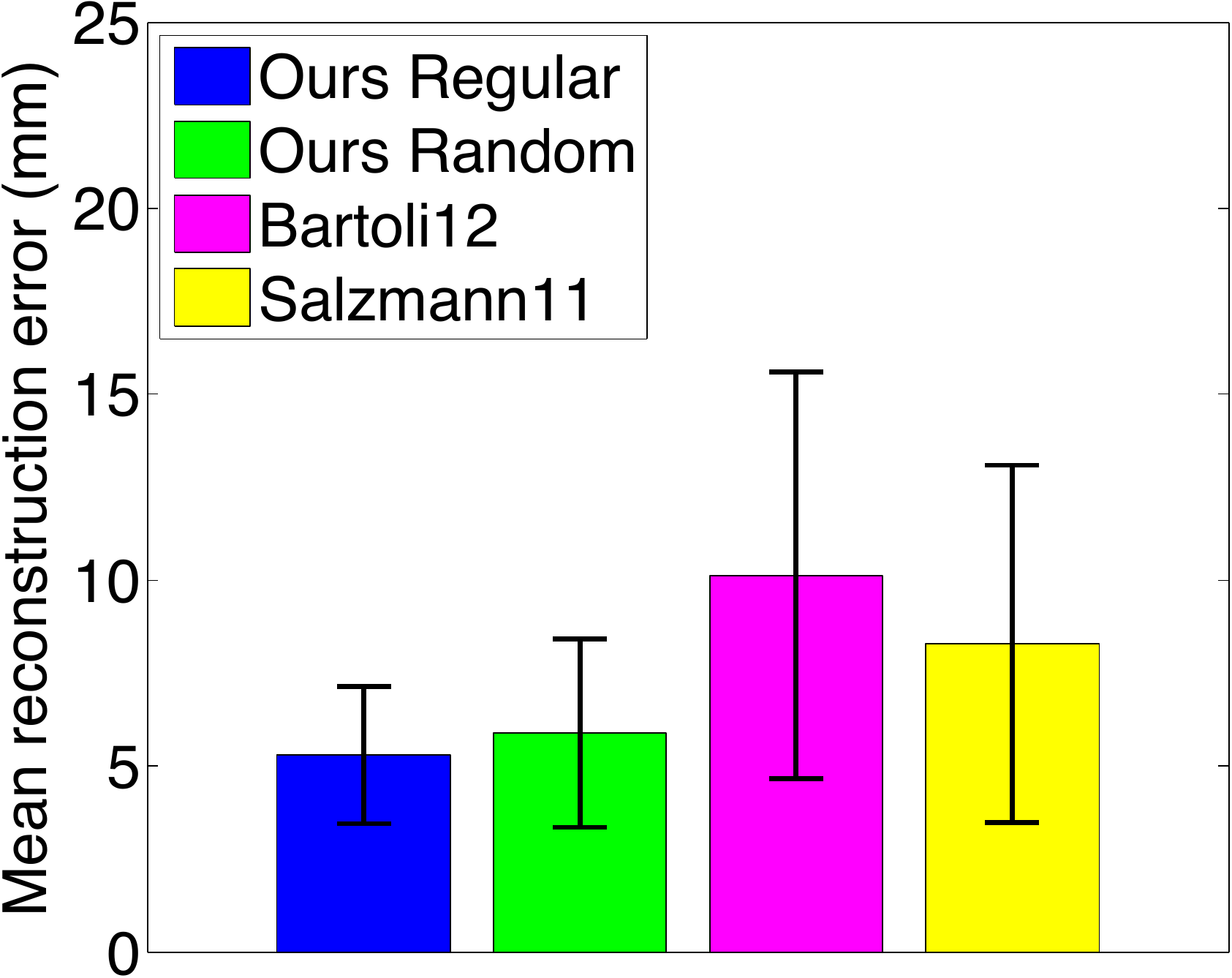} & \includegraphics[height=\ctestacgheight]{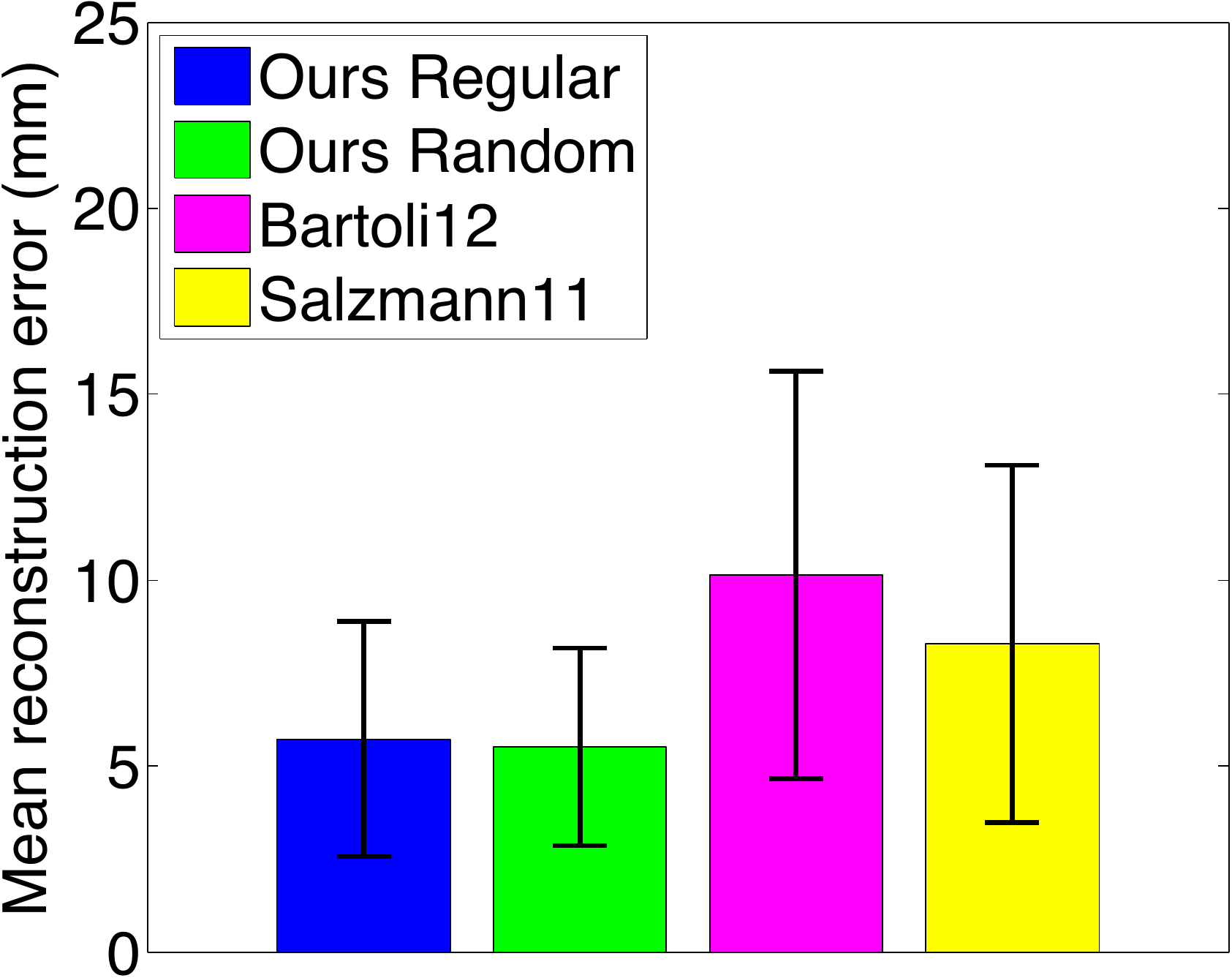} & \includegraphics[height=\ctestacgheight]{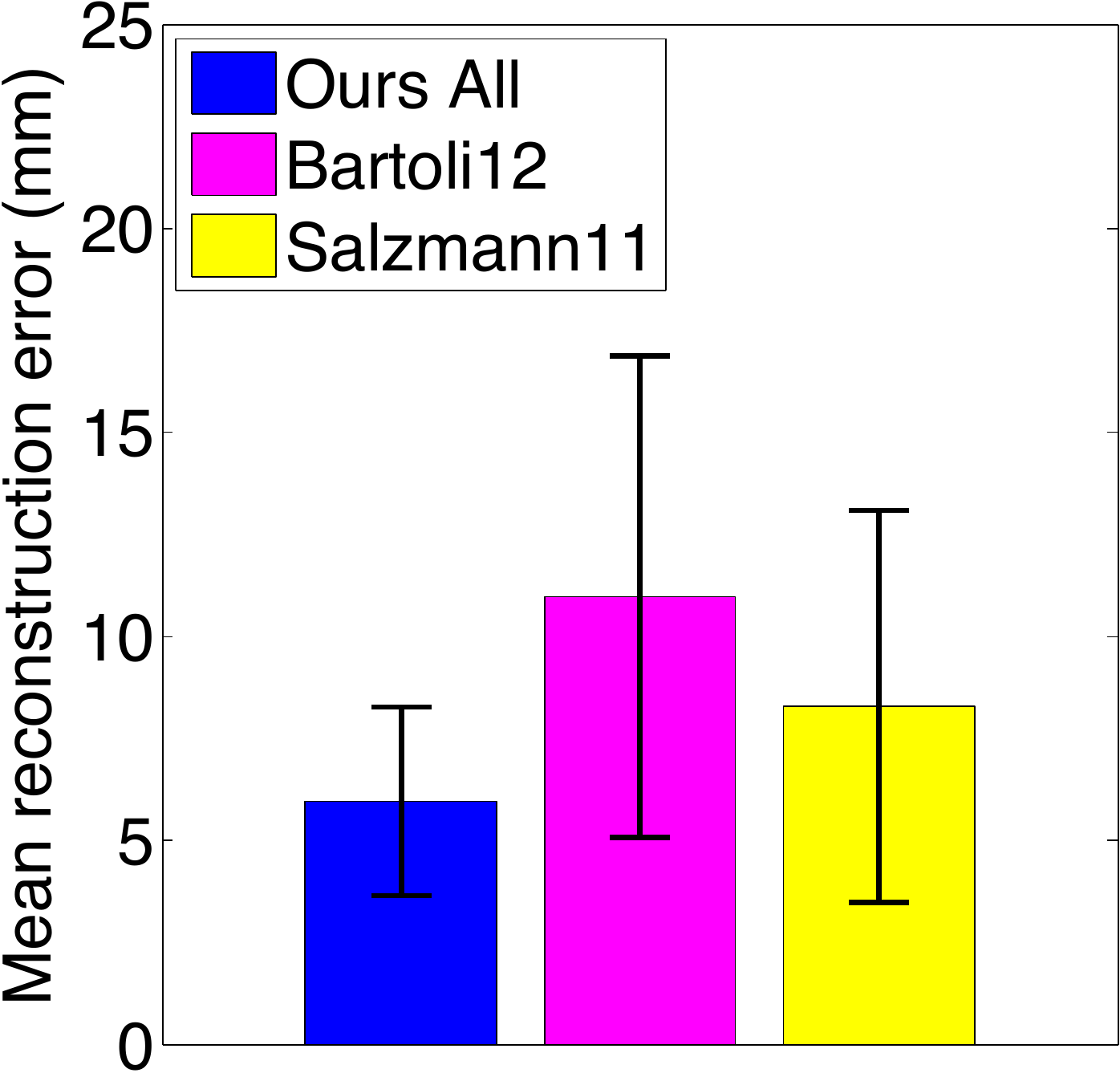} \\
(16) & (25) & (36) & (49) & (270)
\end{tabular}

%% file: figs_ctest031results_table.tex
\newcommand{\ctestadbwidth}{0.185\linewidth}
\newcommand{\ctestadbheight}{3.13cm}
\begin{tabular}{ccccc}
\includegraphics[height=\ctestadbheight]{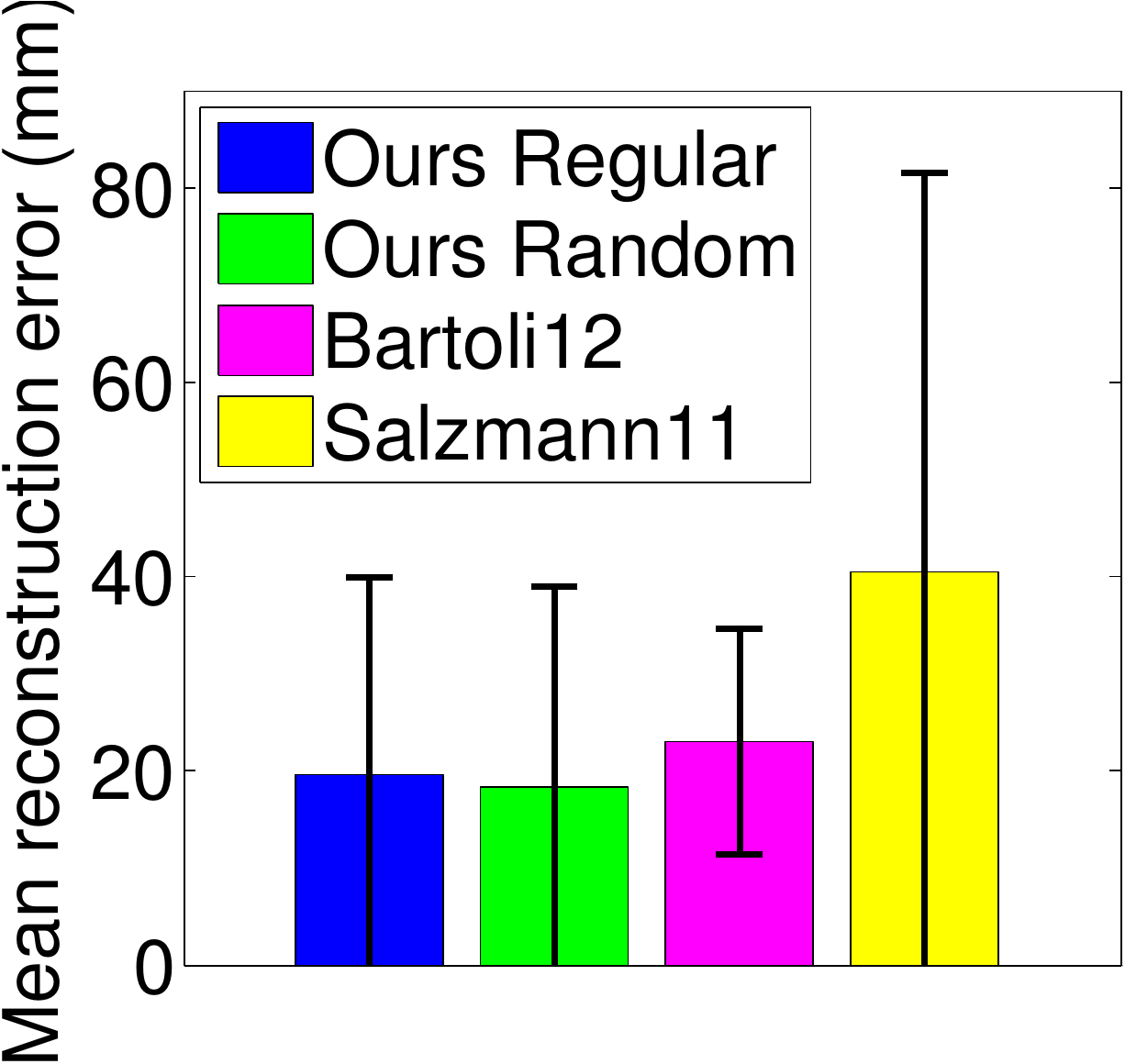} & \includegraphics[height=\ctestadbheight]{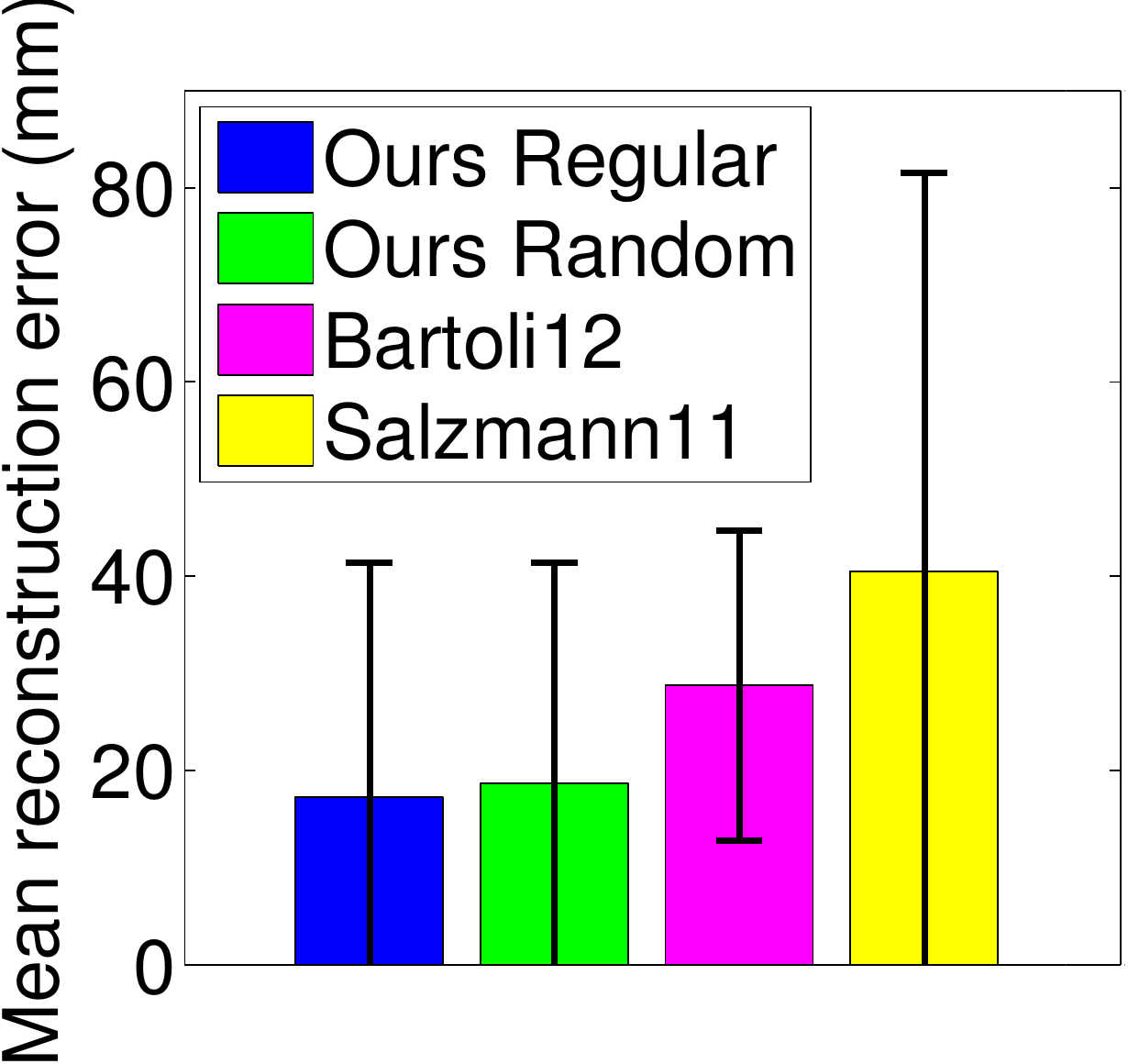} & \includegraphics[height=\ctestadbheight]{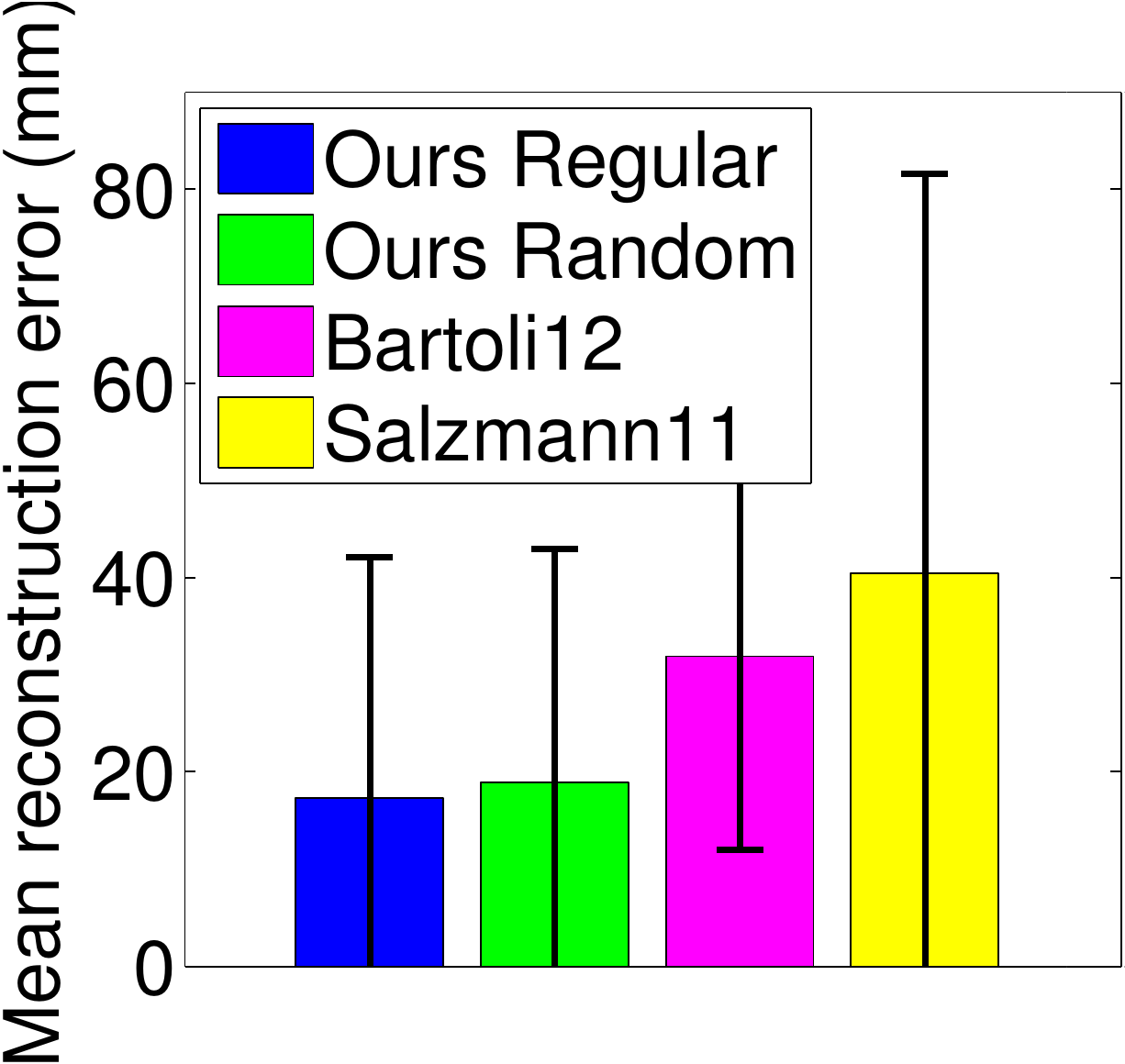} & \includegraphics[height=\ctestadbheight]{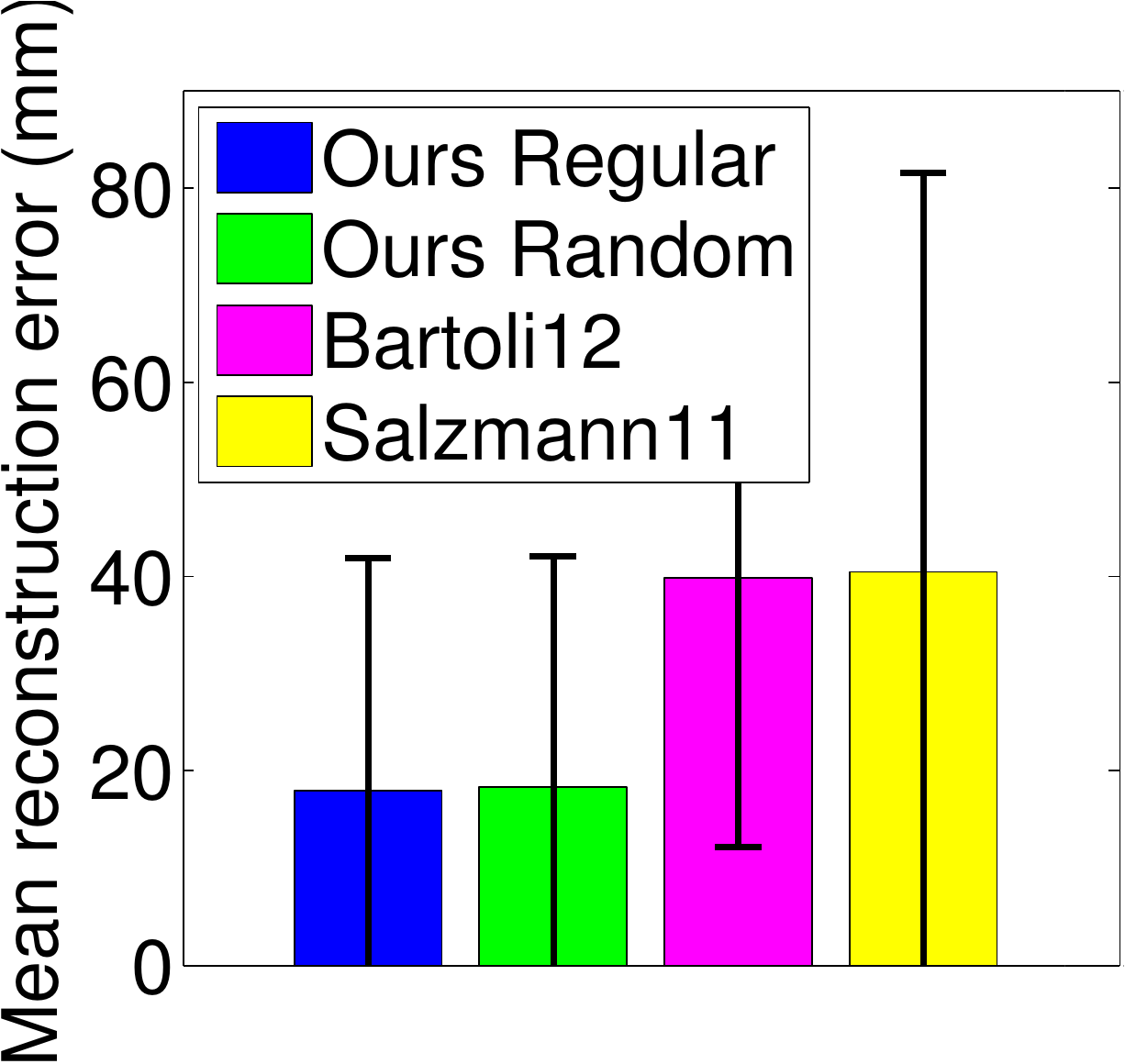} & \includegraphics[height=\ctestadbheight]{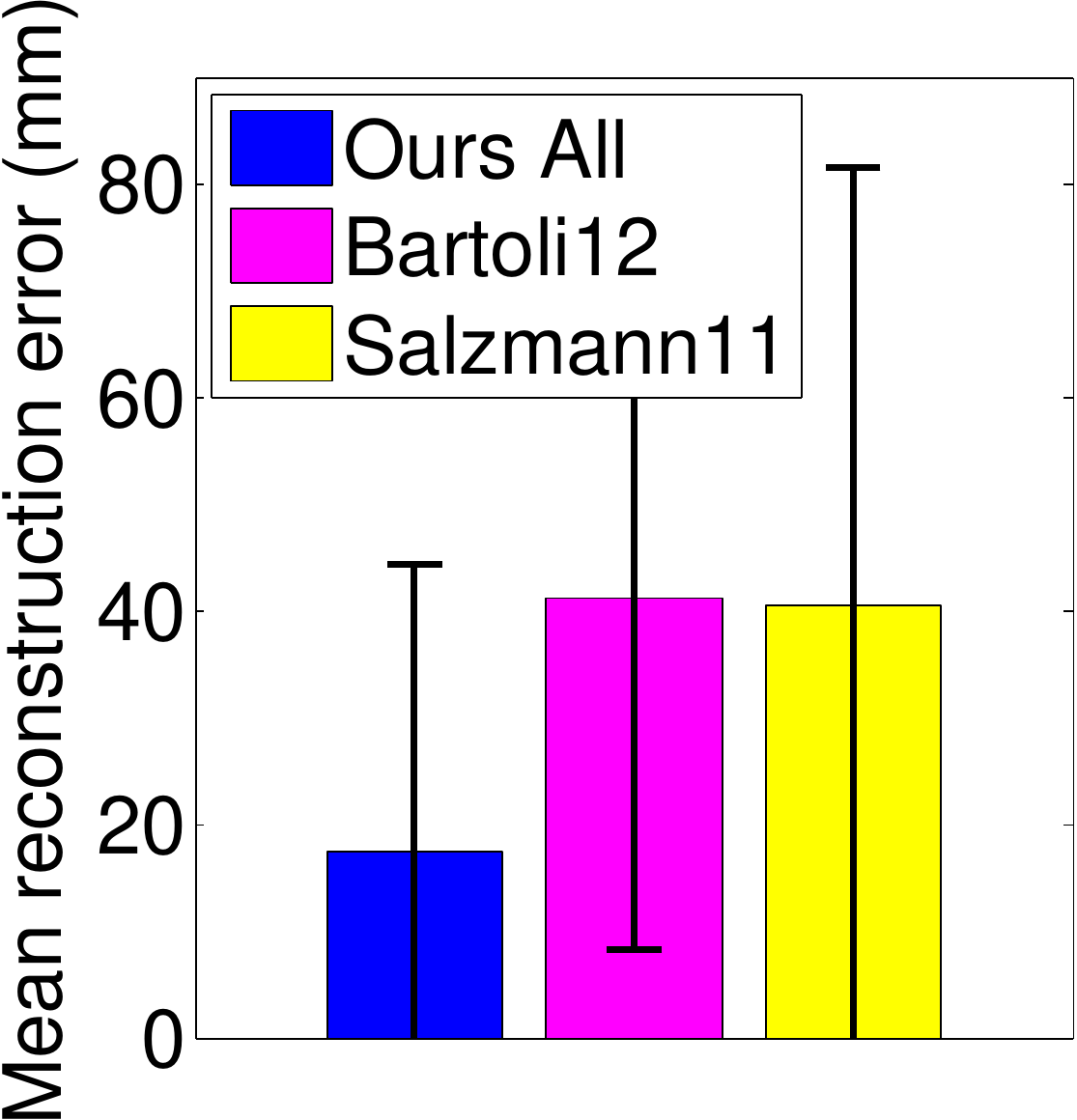} \\
(16) & (25) & (36) & (49) & (260)
\end{tabular}

%% file: figs_ctest019results_table.tex
\newcommand{\ctestabjwidth}{0.185\linewidth}
\newcommand{\ctestabjheight}{2.5cm}
\begin{tabular}{ccccc}
\includegraphics[height=\ctestabjheight]{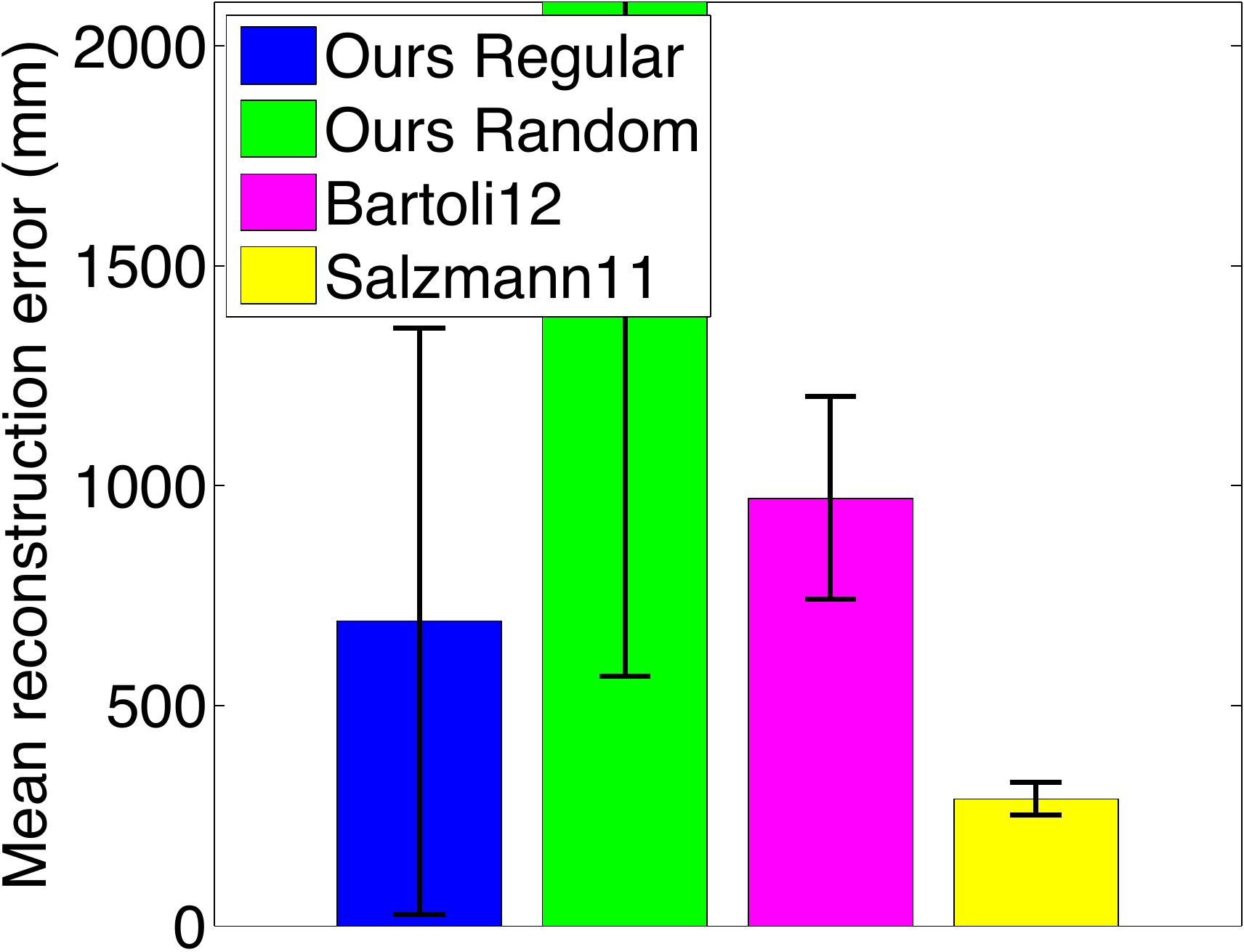} & \includegraphics[height=\ctestabjheight]{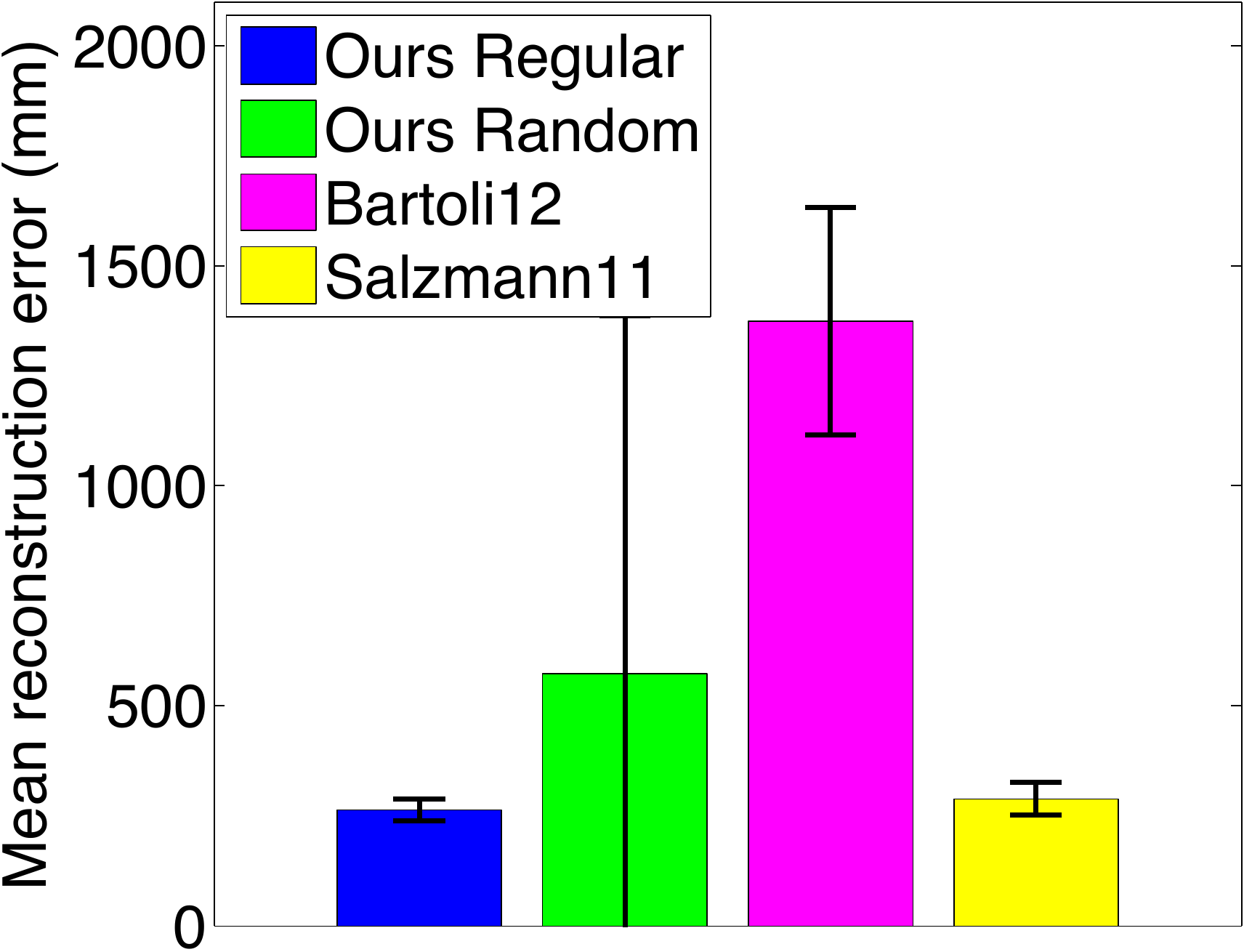} & \includegraphics[height=\ctestabjheight]{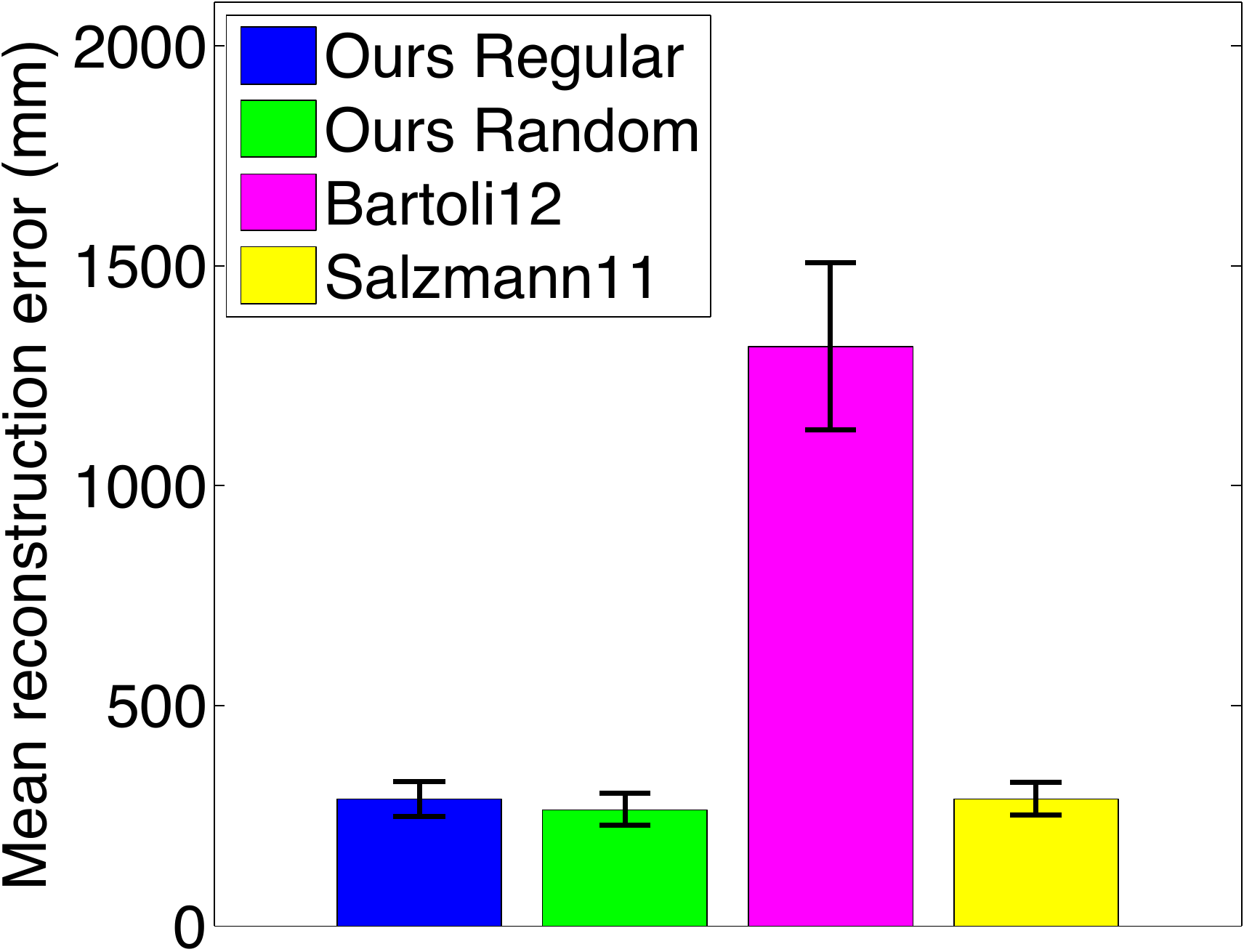} & \includegraphics[height=\ctestabjheight]{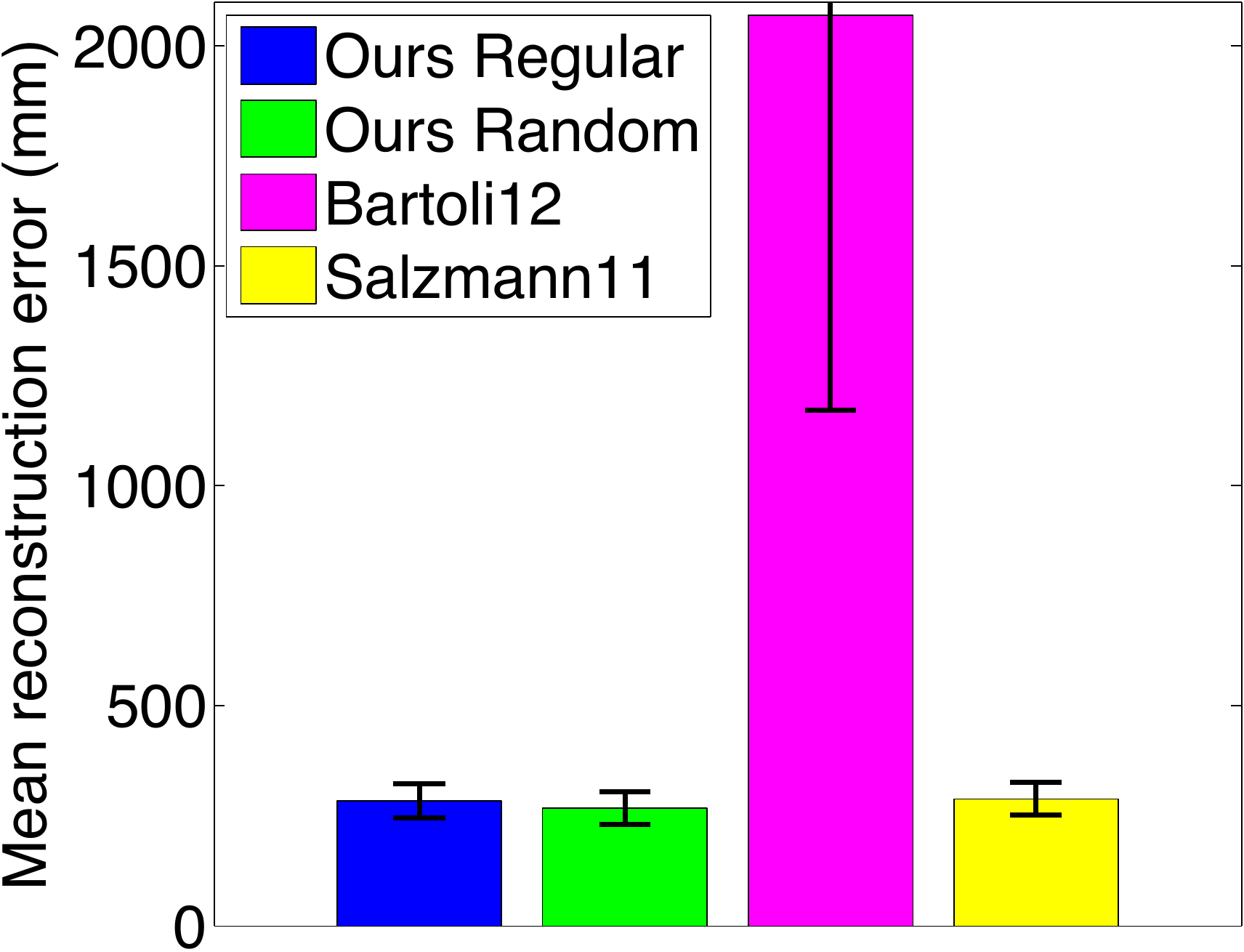} & \includegraphics[height=\ctestabjheight]{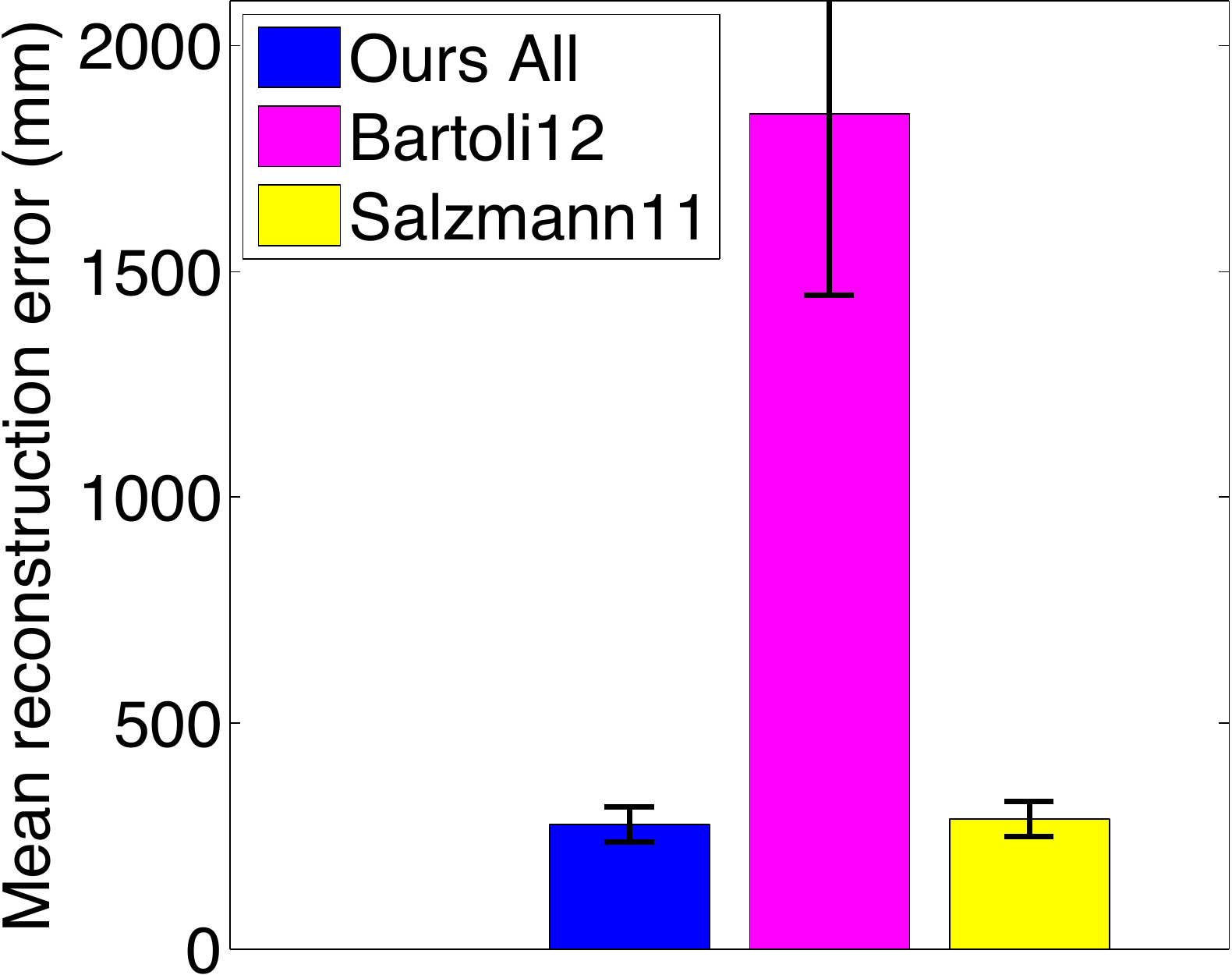} \\
(8) & (12) & (18) & (27) & (66)
\end{tabular}

%% file: figs_wellposed.tex
\newcommand{\withavgvals}{many04}

\begin{figure}
\centering
 \begin{tabular}{ccc}
 \includegraphics[width=0.46\linewidth, height=0.43\linewidth]{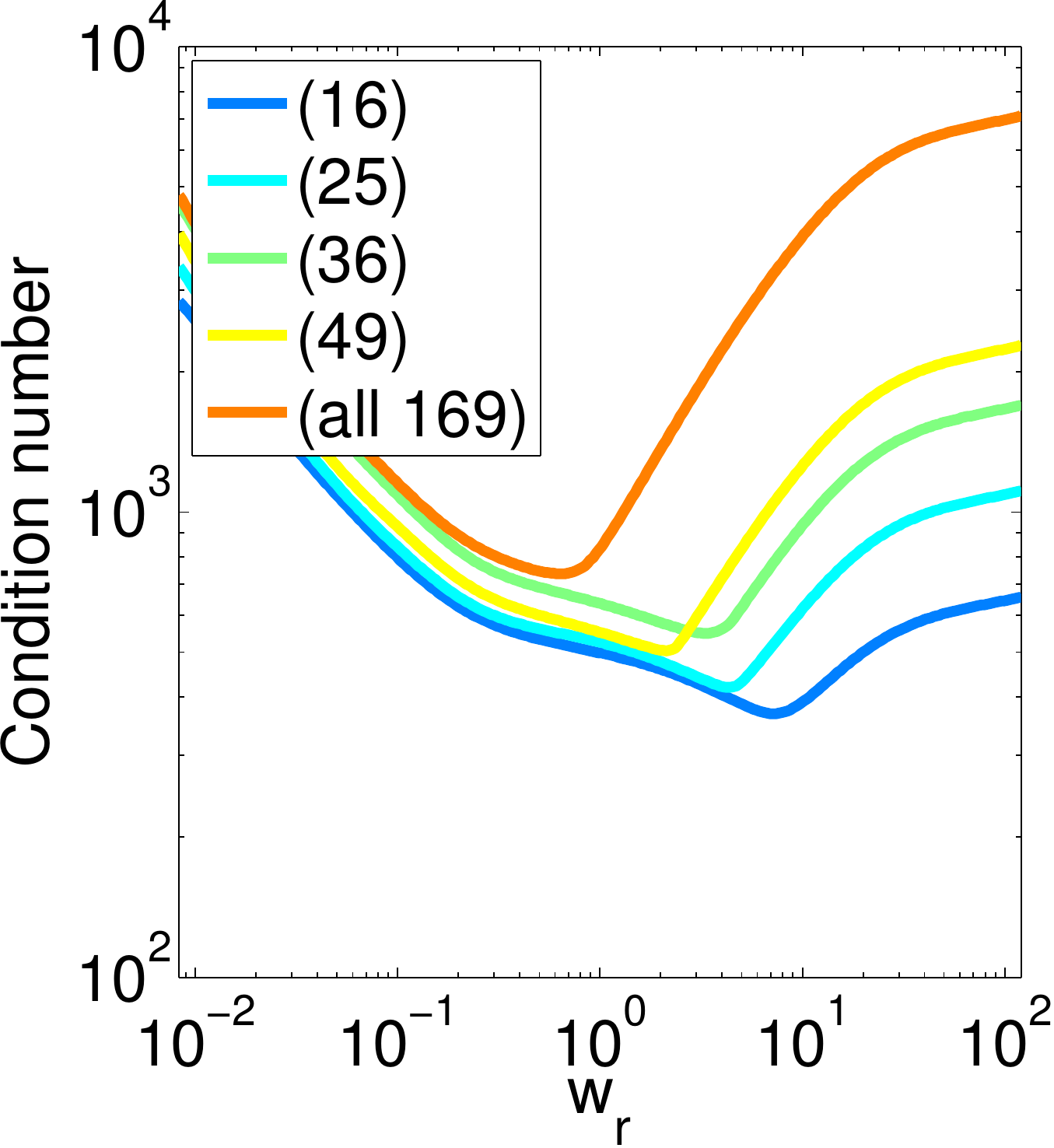}&
 \includegraphics[width=0.46\linewidth, height=0.43\linewidth]{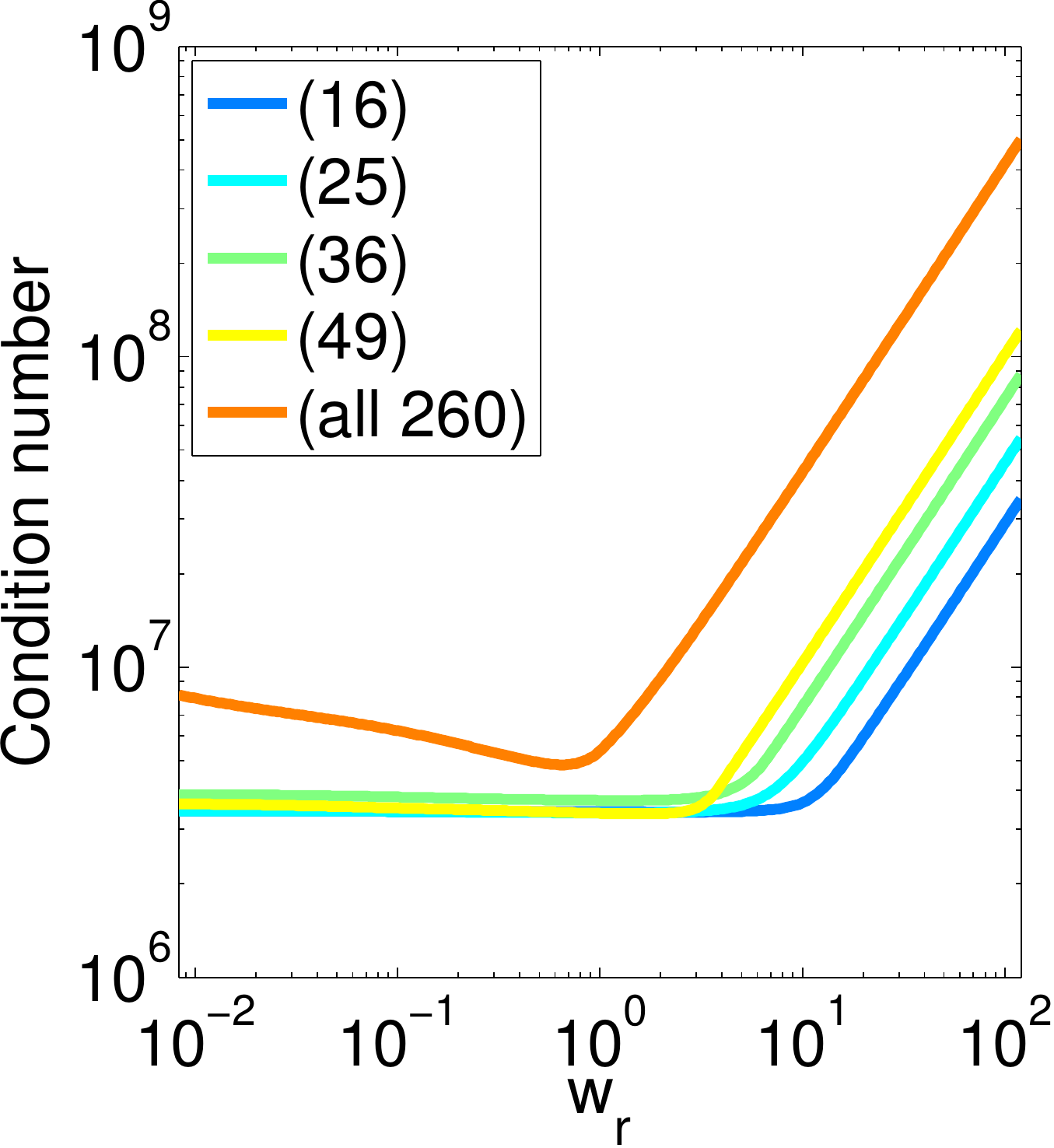}\\ 
 \includegraphics[width=0.46\linewidth, height=0.43\linewidth]{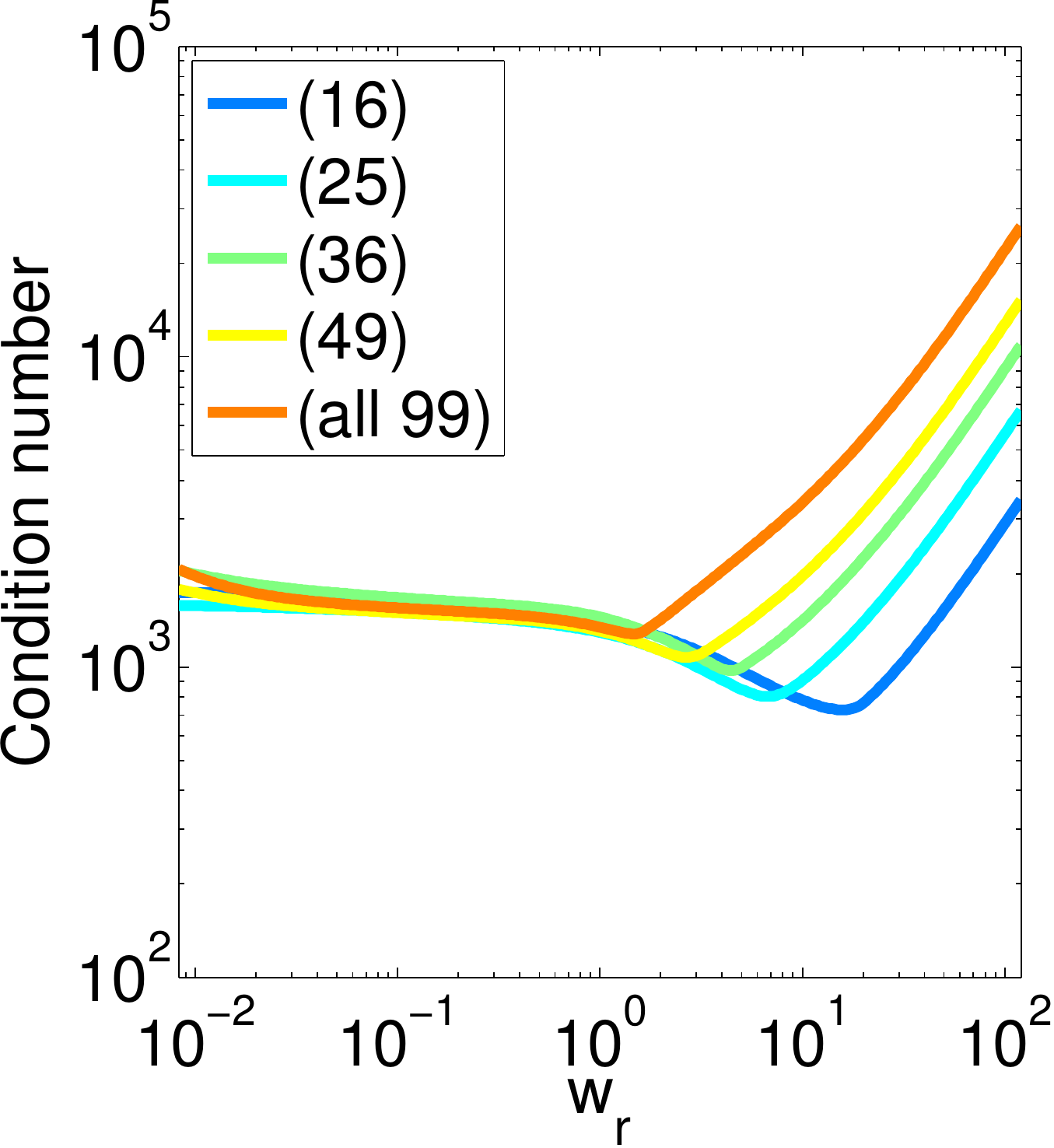}&
 \includegraphics[width=0.46\linewidth, height=0.43\linewidth]{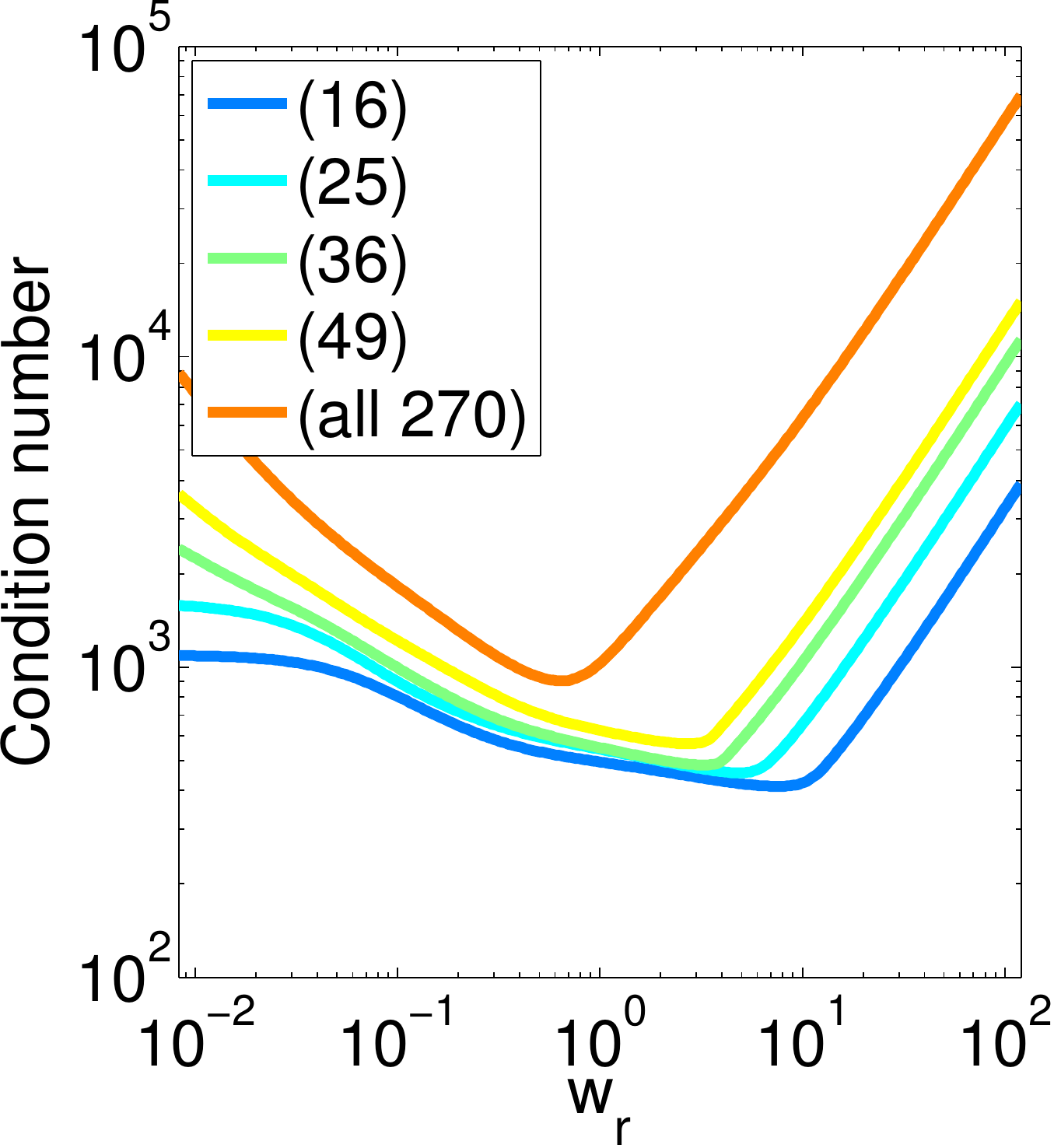}\\
 \includegraphics[width=0.46\linewidth, height=0.28\linewidth]{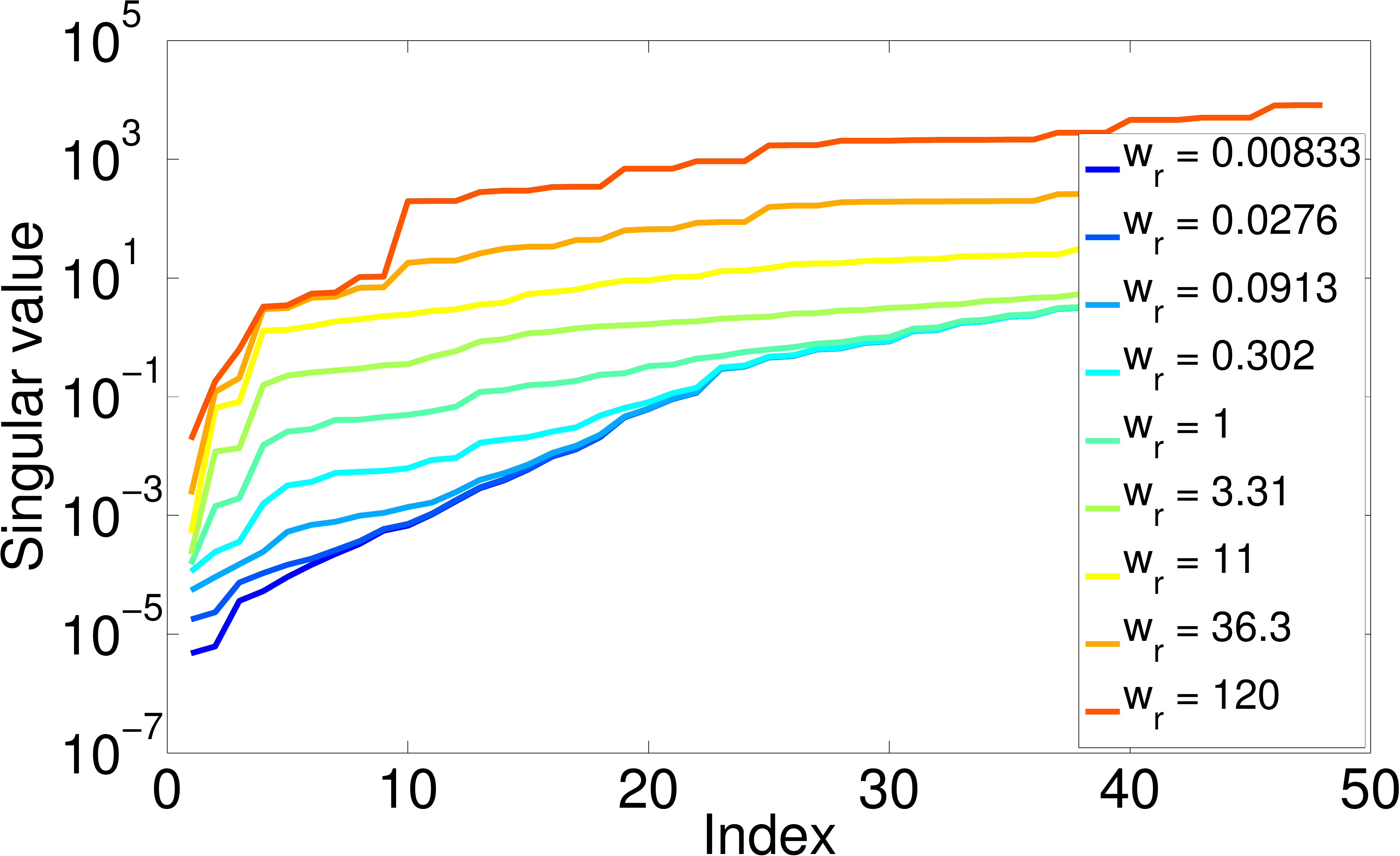}&
 \includegraphics[width=0.46\linewidth, height=0.27\linewidth]{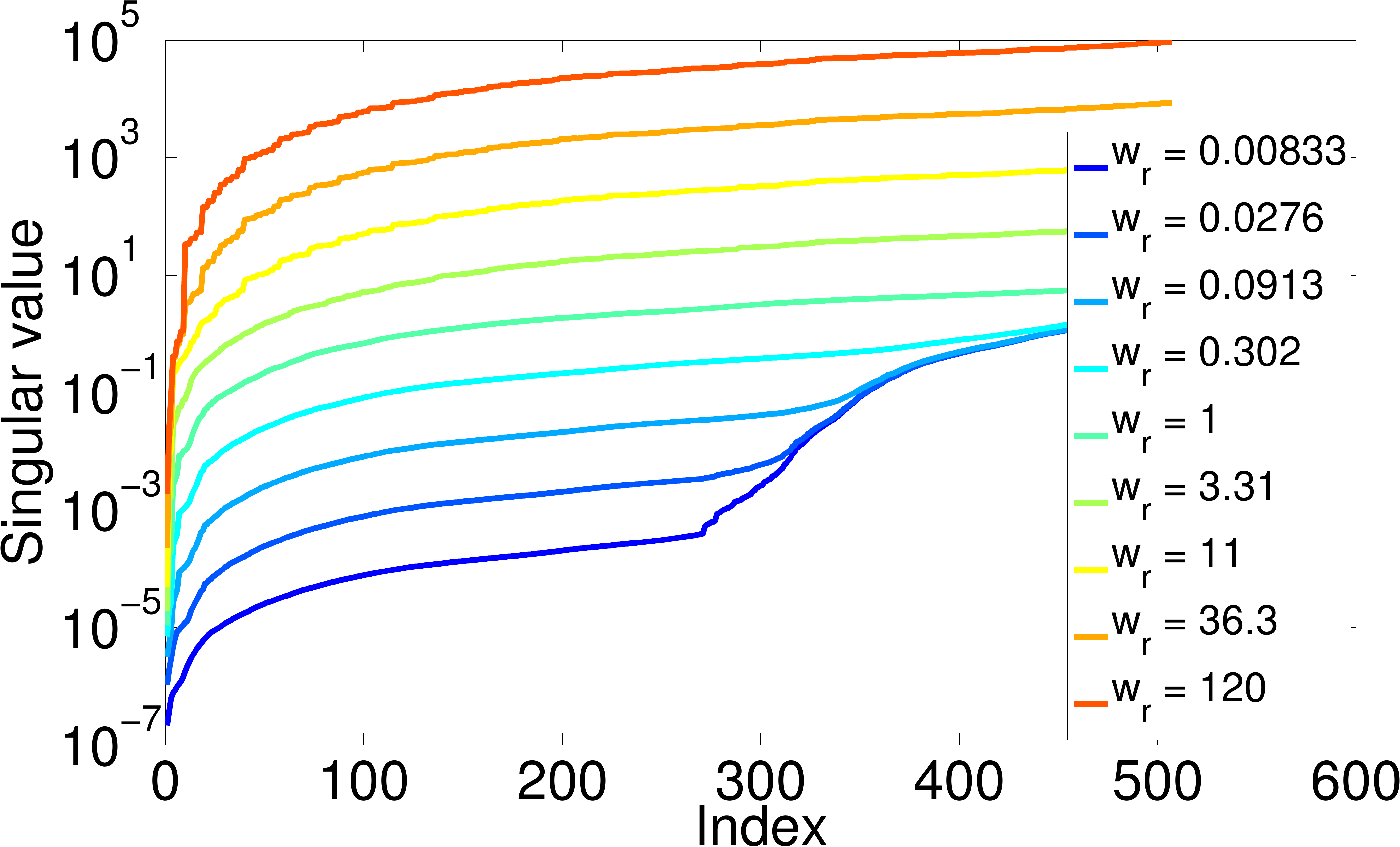}
 \end{tabular}
\vspace{-.1cm}
\caption{\comment{\bf  Condition numbers and singular values.} {\bf  First two rows:}
    Plots   of   the  condition number   of   the   matrix  ${\bM_{w_r}}$   of
    Eq.~\ref{eq:Mwrcat} as  a function of  $w_r$ for each one  of the four   
    kinect   sequences    used   to    perform   the    experiments   of
    Section~\ref{sec:comparative}.  In each plot, we show one curve for each
    regular control  vertex configuration of Fig.~\ref{fig:ctrlsels}. Their
    respective labels denote  the number of control vertices  used. \comment{We also
    show the curve that depicts what happens when every vertex is treated as a
    control vertex, which means the matrix ${\bP}$ is the identity.}{\bf Bottom
      row:} Singular values of ${\bM_{w_r}}$ in the apron case for two different
    configurations  of the  control vertices  and several  values of  $w_r$. The
    curves for the three other cases look very similar.}
\label{fig:wellposed}
\vspace{-.5cm}
\end{figure}

%% file: conc.tex

\section{Conclusion}

We have presented a novel approach to parameterizing the vertex coordinates of a
mesh as a  linear combination of a  subset of them. In addition  to reducing the
dimensionality  of  the  monocular  3D  shape  recovery  problem,  it  yields  a
rotation-invariant curvature-preserving  regularization term that  produces good
results without training data or having to explicitly handle global rotations.

In our current implementation we applied constraints on every single edge of the
mesh. In  future work,  we will  explore other types  of surface  constraints in
order to further decrease computational complexity without sacrificing accuracy.
Another direction of research would be to take advantage of temporal consistency
as has been  done in many recent works~\cite{Torresani08a,Russell11,Salzmann11a}
and of additional sources of shape information, such as shading and contours, to
increase robustness and accuracy.

%% file: bio.tex
\begin{IEEEbiography}
[{\includegraphics[width=1in,height=1.25in,clip,keepaspectratio]{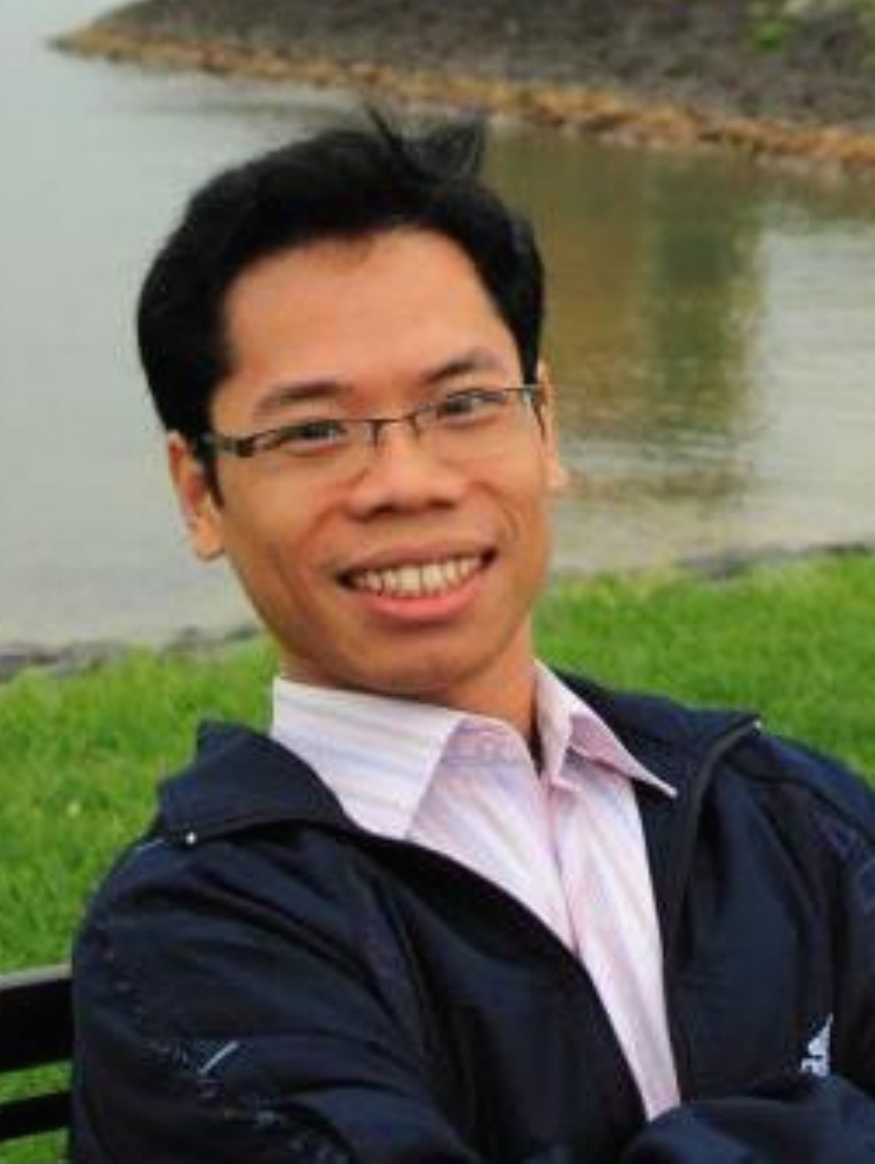}}]{Dat Tien Ngo}
is pursuing his Ph.D. at the computer vision laboratory, EPFL, under the
supervision of Prof. Pascal Fua. He received his B.Sc. in Computer Science from
Vietnam National University Hanoi with double title {\em Ranked first in both
entrance examination and graduation} and his M.Sc. in Artificial Intelligence
from VU University Amsterdam. His main research interests include deformable
surface reconstruction, object tracking, augmented reality.
\end{IEEEbiography}
\vspace{-0.7cm}
\begin{IEEEbiography}
[{\includegraphics[width=1in,height=1.25in,clip,keepaspectratio]{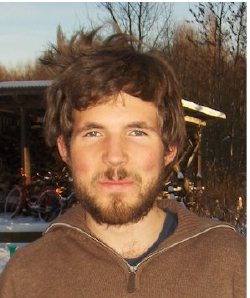}}]{Jonas \"Ostlund}
received his M.Sc. degree in computer science from Linköping University in 2010,
along with the Tryggve Holm medal for outstanding academic achievements. From
2010 to 2013 he worked as a scientific assistant in the computer vision
laboratory at EPFL, where his main research interest was monocular 3D deformable
surface reconstruction with applications for sail shape modeling. Currently, he
is co-founding the Lausanne-based startup Anemomind developing data mining
algorithms for sailing applications.
\end{IEEEbiography}
\vspace{-0.7cm}
\begin{IEEEbiography}
[{\includegraphics[width=1in,height=1.25in,clip,keepaspectratio]{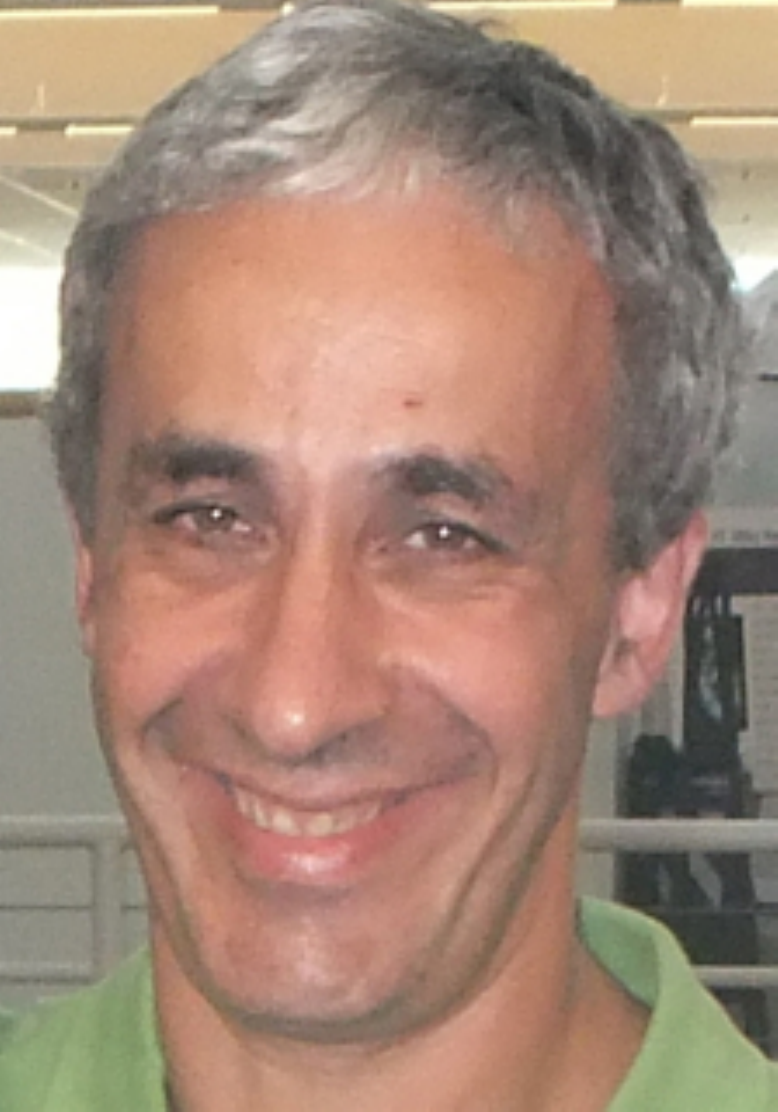}}]{Pascal Fua} 
received an  engineering degree from Ecole Polytechnique,  Paris, in 1984
and  the Ph.D.  degree  in Computer  Science  from the  University  of Orsay  in
1989.  He joined EPFL  in 1996  where he  is now  a Professor  in the  School of
Computer and Communication Science. Before  that, he worked at SRI International
and at  INRIA Sophia-Antipolis as  a Computer Scientist. His  research interests
include shape modeling  and motion recovery from images,  analysis of microscopy
images,  and Augmented  Reality. He  has (co)authored  over 150  publications in
refereed journals  and conferences.  He is an  IEEE fellow  and has been  a PAMI
associate editor.  He often serves as  program committee member,  area chair, or
program chair of major vision conferences.
\end{IEEEbiography}
\vfill

%% file: appendix.tex
\appendices

\section{Properties of the regularization term}
\label{app:proof}

Here we prove the claim made in Section~\ref{sec:PlanarReg} that $\mathbf{Ax=0}$
when $\bx$  represents an affine transformed  version of the reference  mesh and
that $\sqnorm{\bA \bx}$ is invariant to rotations and translations.

The  first equation  in Eq.~\ref{eq:planar}  applies for  each of  three spatial
coordinates $x,y,z$ and can be rewritten in a matrix form as
\begin{equation}
 \left [ w_1 ~~ w_2 ~~ w_3 ~~ w_4 \right ] 
 \begin{bmatrix}
  {\vtr{1}}_x & {\vtr{1}}_y & {\vtr{1}}_z \\ 
  {\vtr{2}}_x & {\vtr{2}}_y & {\vtr{2}}_z \\ 
  {\vtr{3}}_x & {\vtr{3}}_y & {\vtr{3}}_z \\ 
  {\vtr{4}}_x & {\vtr{4}}_y & {\vtr{4}}_z
 \end{bmatrix} = \mathbf{0}^T \; .
 \label{eq:planar_matrix}
\end{equation}
It can be seen from Eq.~\ref{eq:planar_matrix} and the second equation in
Eq.~\ref{eq:planar} that the three vectors of the $x,y,z$ components of the reference mesh
and the vector of all 1s lie in the kernel of the matrix $\bA^\prime$. It means
our regularization term $\sqnorm{\bA \bx}$ does not penalize affine
transformation of the reference mesh. Hence, $\mathbf{Ax=0}$ as long as $\bx$
represents an affine transformed version of the reference mesh.

Let $\vt{i}' = \bR \vt{i} + \bt$ be the new location of vertex $\vt{i}$ under a
rigid transformation of the mesh, where $\bR$ is the rotation matrix and $\bt$
is the translation vector. Since $\bR^T\bR = \bI_3$ and $w_1 + w_2 + w_3  + w_4
= 0$, we have
\begin{eqnarray}
  &   & \sqnorm{w_1\vt{i_1}' + w_2\vt{i_2}' + w_3\vt{i_3}' + w_4{\vt{i_4}'}} \nonumber \\ 
  & = & \bigl\| \bR (w_1\vt{i_1} + w_2 \vt{i_2} + w_3 \vt{i_3} + w_4 {\vt{i_4}}) \nonumber \\ 
  &   & \hspace{2cm} + \bt (w_1+w_2+w_3+w_4) \bigr\|^2 \\
  & = & \sqnorm{w_1\vt{i_1} + w_2\vt{i_2} + w_3\vt{i_3} + w_4{\vt{i_4}}} \nonumber \; .
\end{eqnarray}
Hence, $\sqnorm{\bA \bx^\prime} = \sqnorm{\bA \bx}$ when $\bx\any$ is a rigidly
transformed version of $\bx$. In other words, $\sqnorm{\bA \bx}$ is invariant to
rotations and translations.

\section{Importance of Non-Linear Minimization}
\label{app:constrained}

\input{figs_paper.tex}

Fig.~\ref{fig:capaperlincst}  illustrates  the   importance  of  the  non-linear
constrained minimization step of  Eq.~\ref{eq:ConstrOpt} that refines the result
obtained by solving the linear least-squares problemof Eq.~\ref{eq:UnconstrOpt}.
In the left column  we show the solution of the  linear least-squares problem of
Eq.~\ref{eq:UnconstrOpt}.   It  projects  correctly  but, as  evidenced  by  its
distance to the ground-truth gray dots, its 3D shape is incorrect.  By contrast,
the  surface  obtained  by  solving  the  constrained  optimization  problem  of
Eq.~\ref{eq:ConstrOpt} still  reprojects correctly  while also  being metrically
accurate.  Fig.~\ref{fig:kinect_paper_seq} depicts  similar situations  in other
frames of the sequence.

\comment{
\section{Robustness to Erroneous Correspondences}
\label{app:inlierrate}

\input{figs_robustness_comparison2.tex}

Fig.~\ref{fig:robustnesscompareinlier} depicts the success rates according to
the second criteria, i.e. at least $90\%$ of the inlier matches are correctly
labeled and retrieved, as a function of the total number of inliers and the
proportion of outliers. 
}

%% file: figs_paper.tex

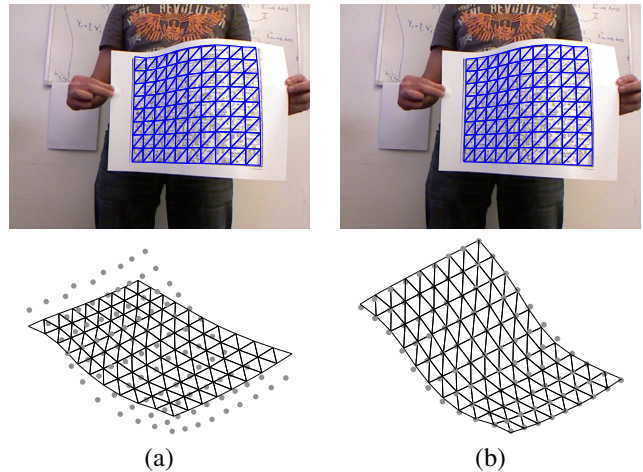
\begin{figure}
\centering
\input{figs_capaperlincst_table.tex}
\caption{{\bf  Unconstrained vs  constrained optimization  results.}  (a)
    The surface  obtained by solving  the unconstrained minimization  problem of
    Eq.~\ref{eq:UnconstrOpt} and  rescaling the result.  It is  projected on the
    original image  at the top and shown  from a different angle  at the bottom.
    (b) The surface obtained by  solving the constrained minimization problem of
    Eq.~\ref{eq:ConstrOpt}.   The  projections   of  both  surfaces  are  almost
    identical but only the second has the correct 3D shape.}
\label{fig:capaperlincst}
\end{figure}

\begin{figure*}
\centering
\begin{tabular}{ccccc}
\includegraphics[width=0.18\linewidth]{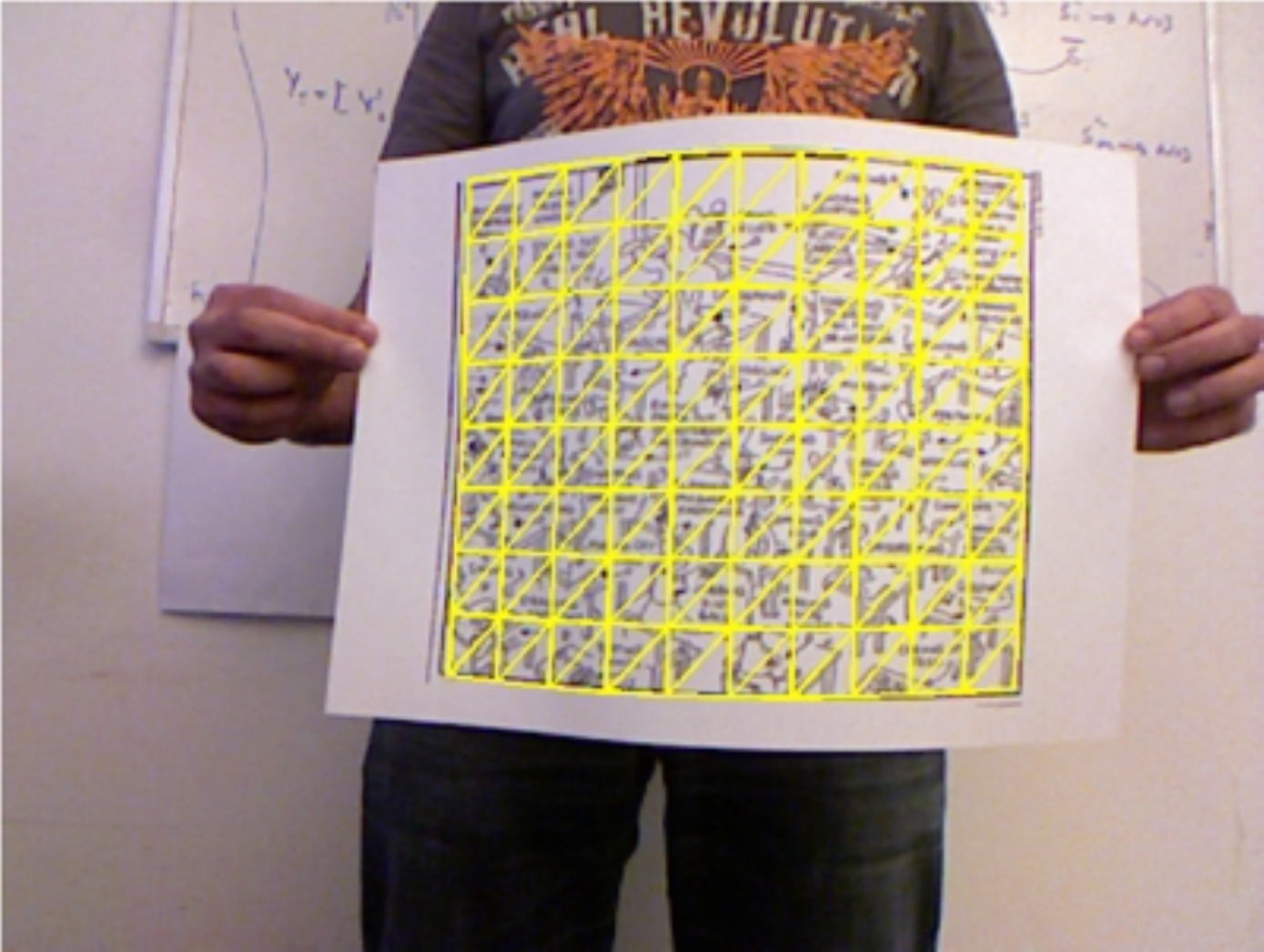}       &
\includegraphics[width=0.18\linewidth]{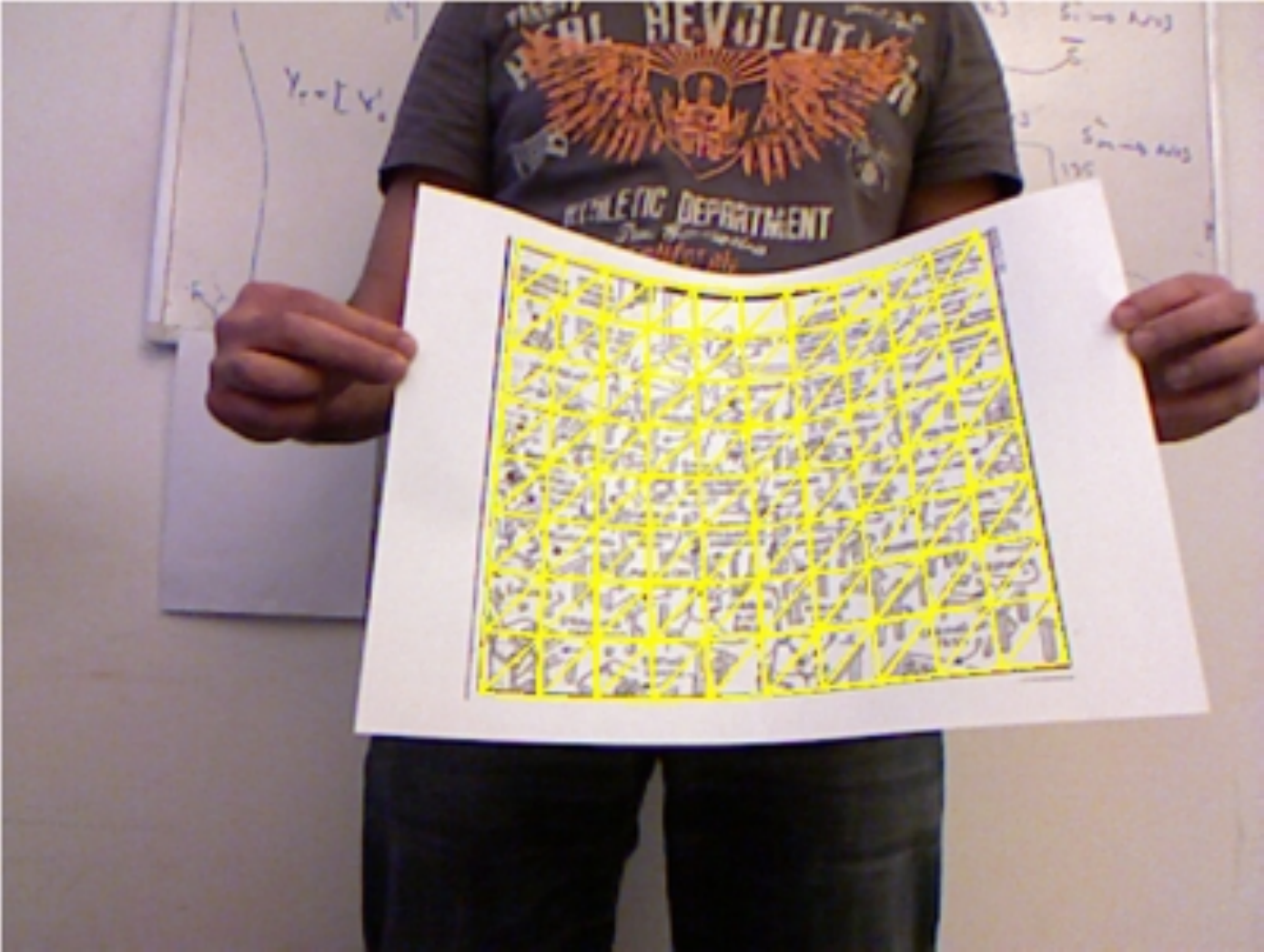}       &
\includegraphics[width=0.18\linewidth]{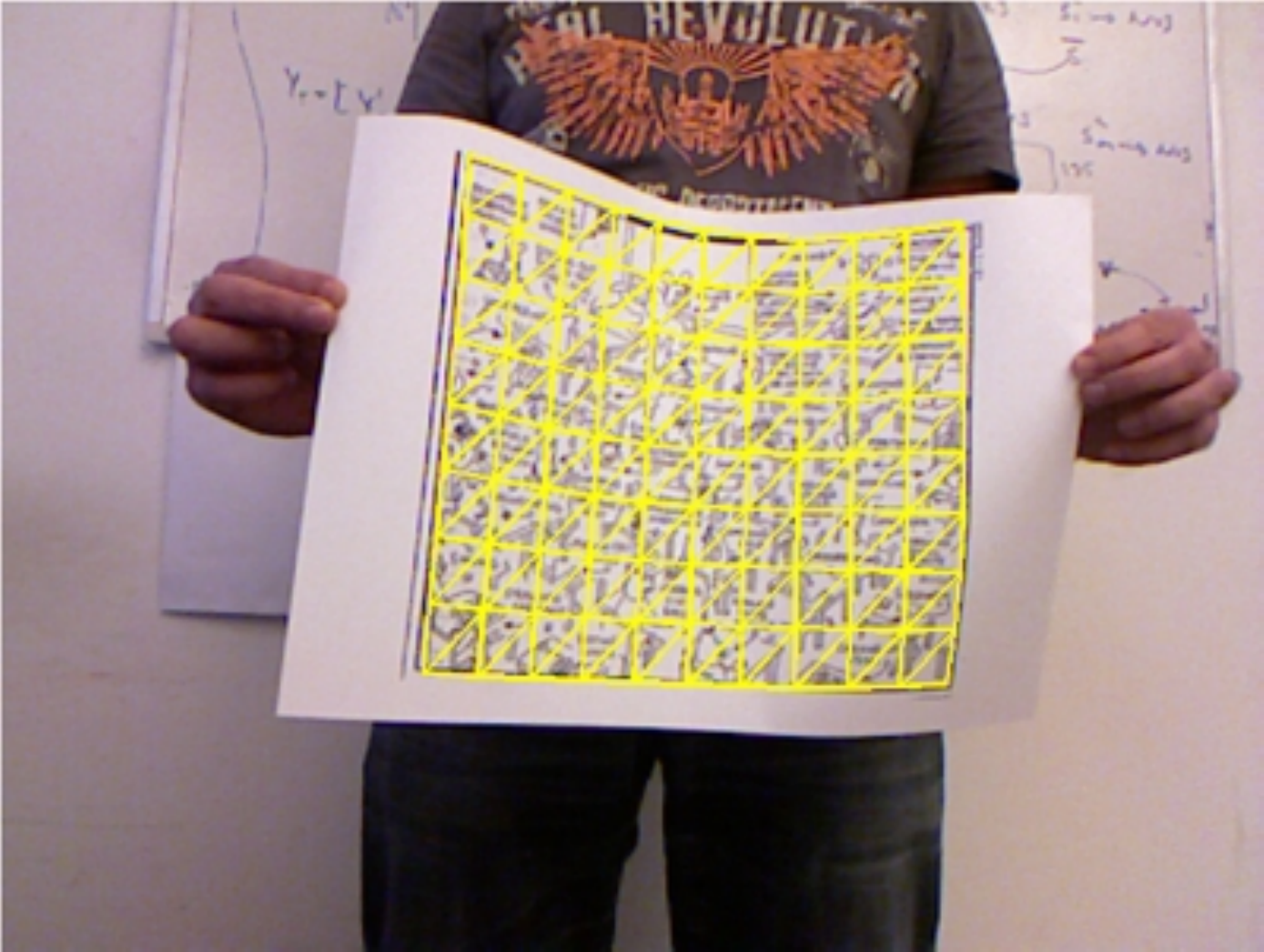}       &
\includegraphics[width=0.18\linewidth]{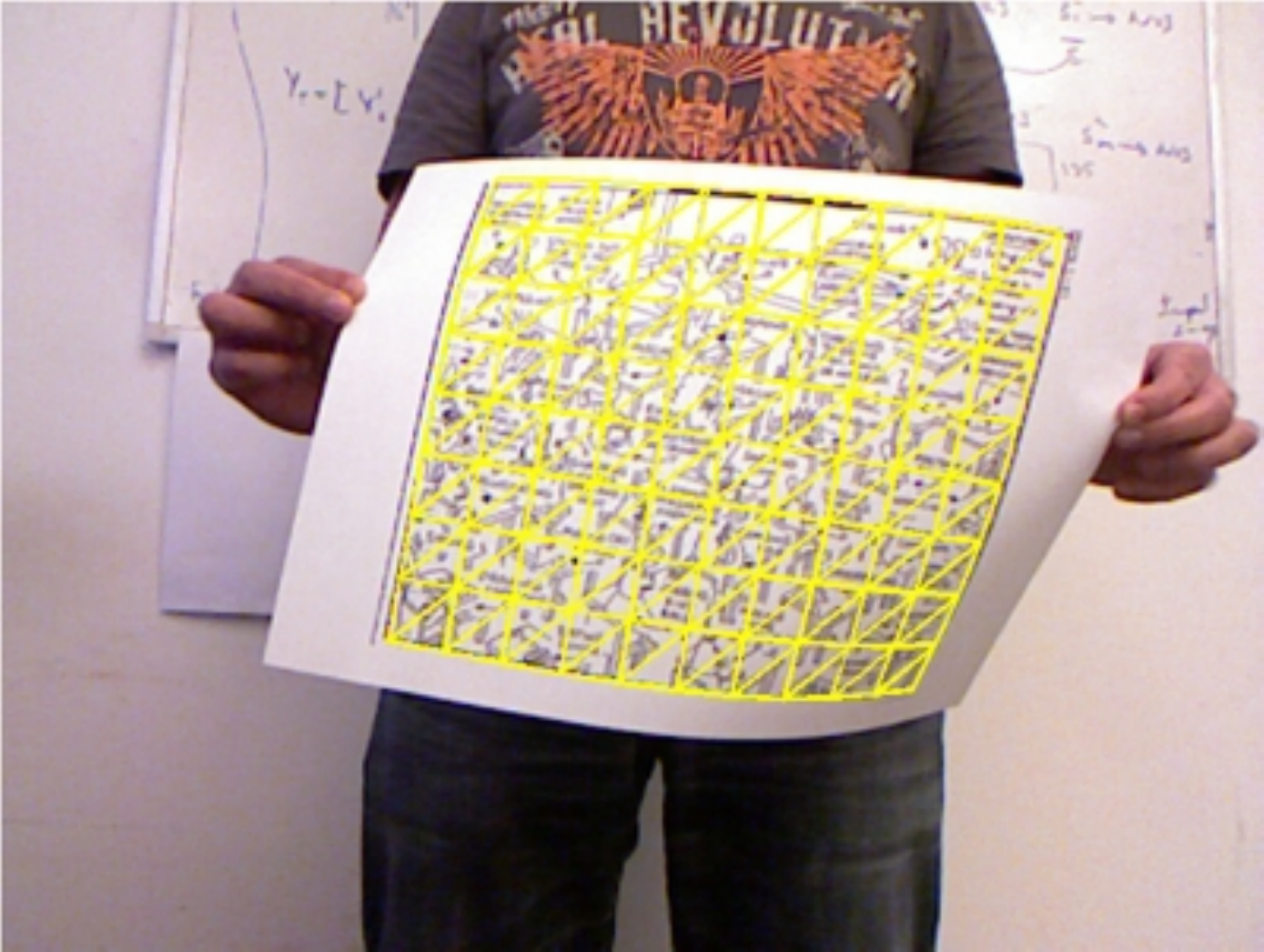}       &
\includegraphics[width=0.18\linewidth]{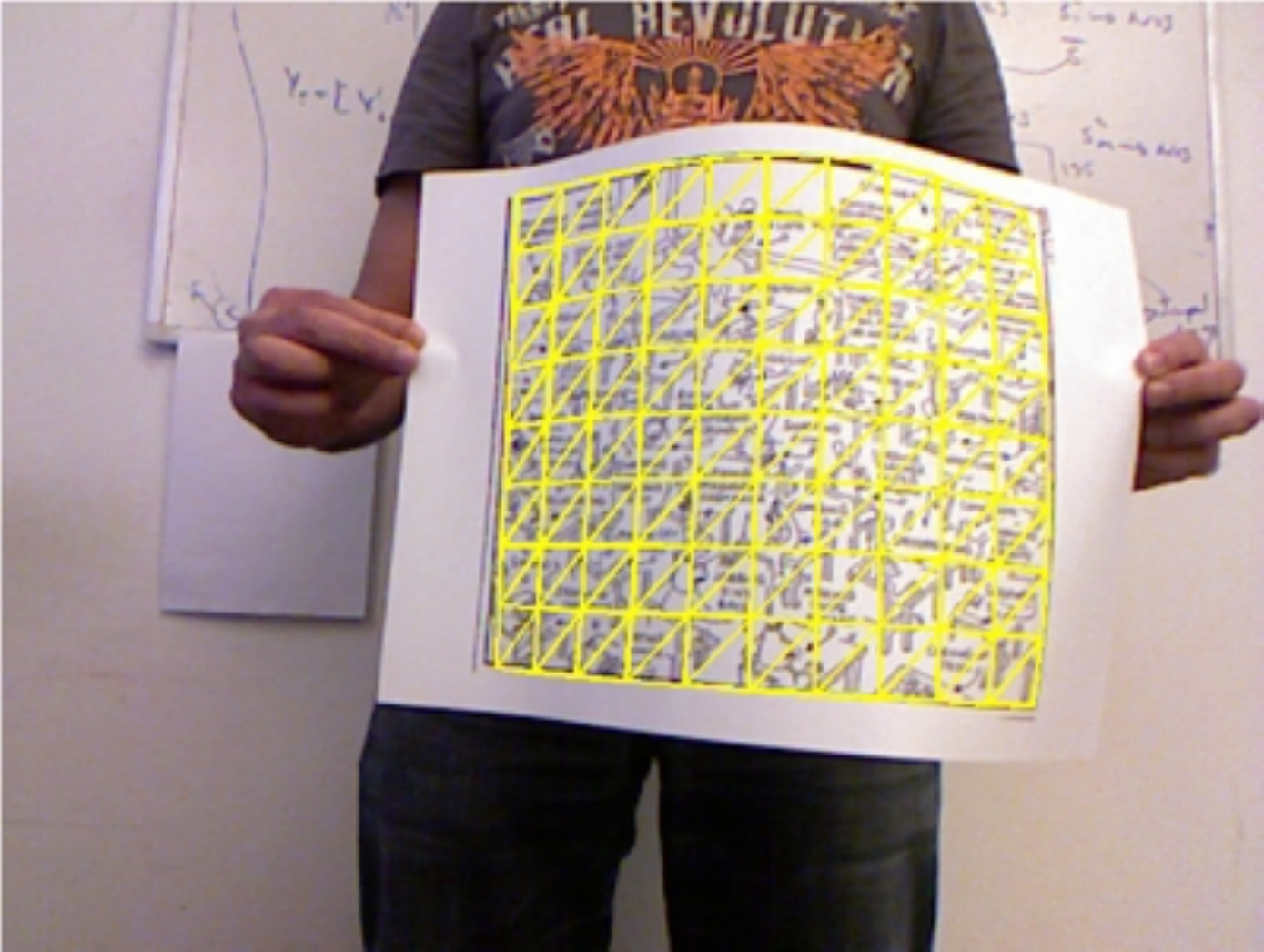}
\\
\includegraphics[width=0.18\linewidth]{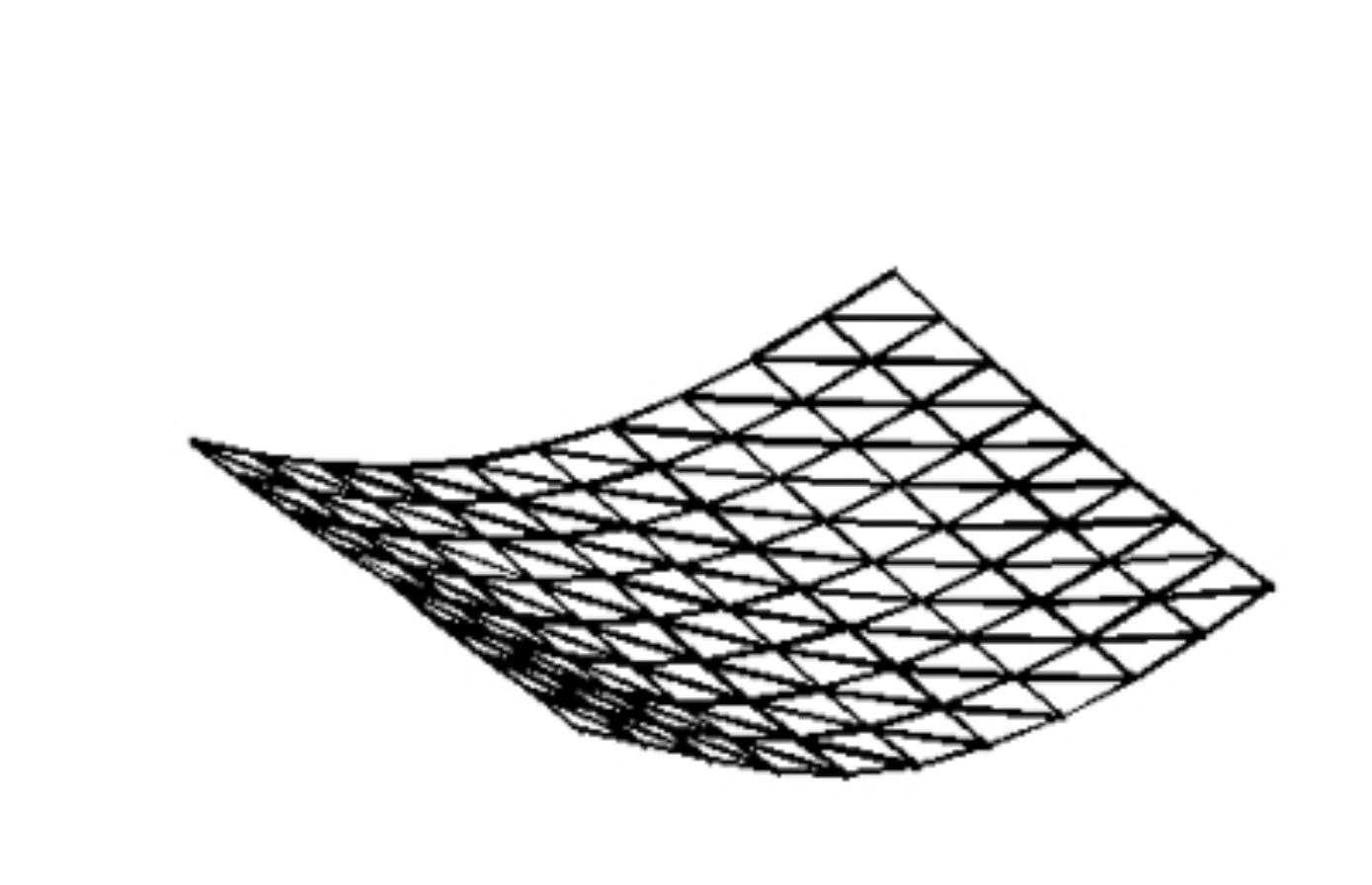}  &
\includegraphics[width=0.18\linewidth]{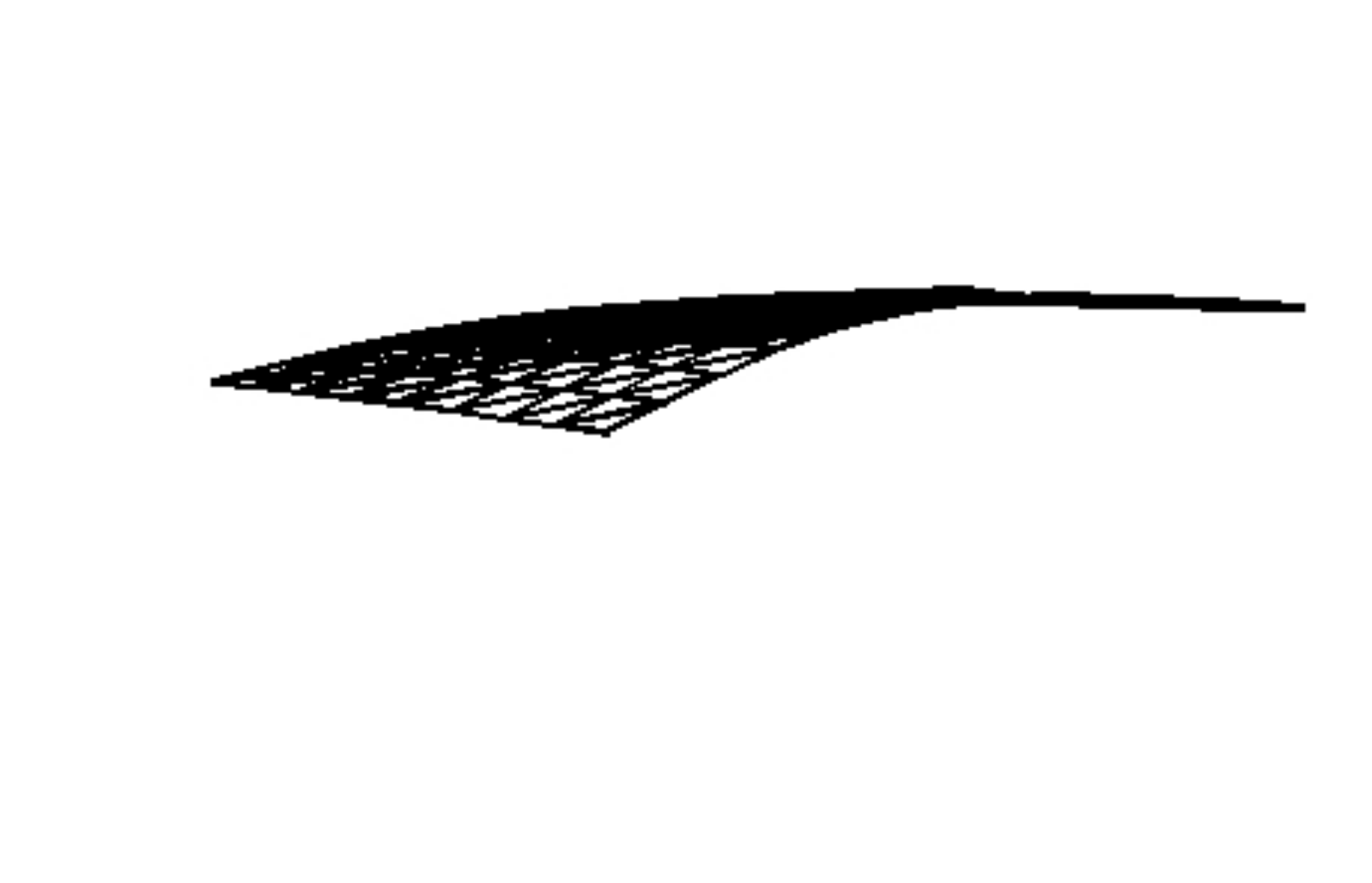}   &
\includegraphics[width=0.18\linewidth]{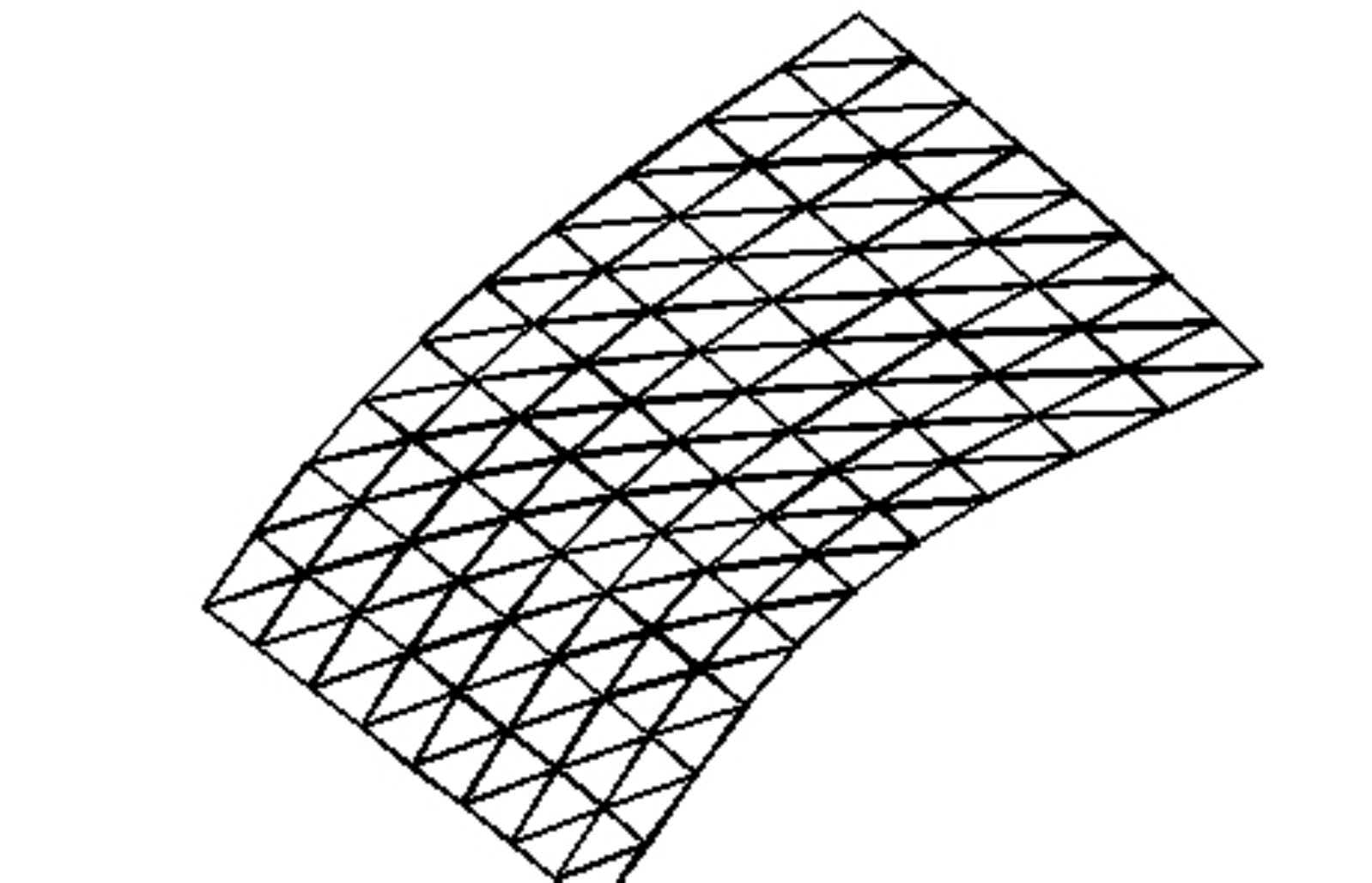}   &
\includegraphics[width=0.18\linewidth]{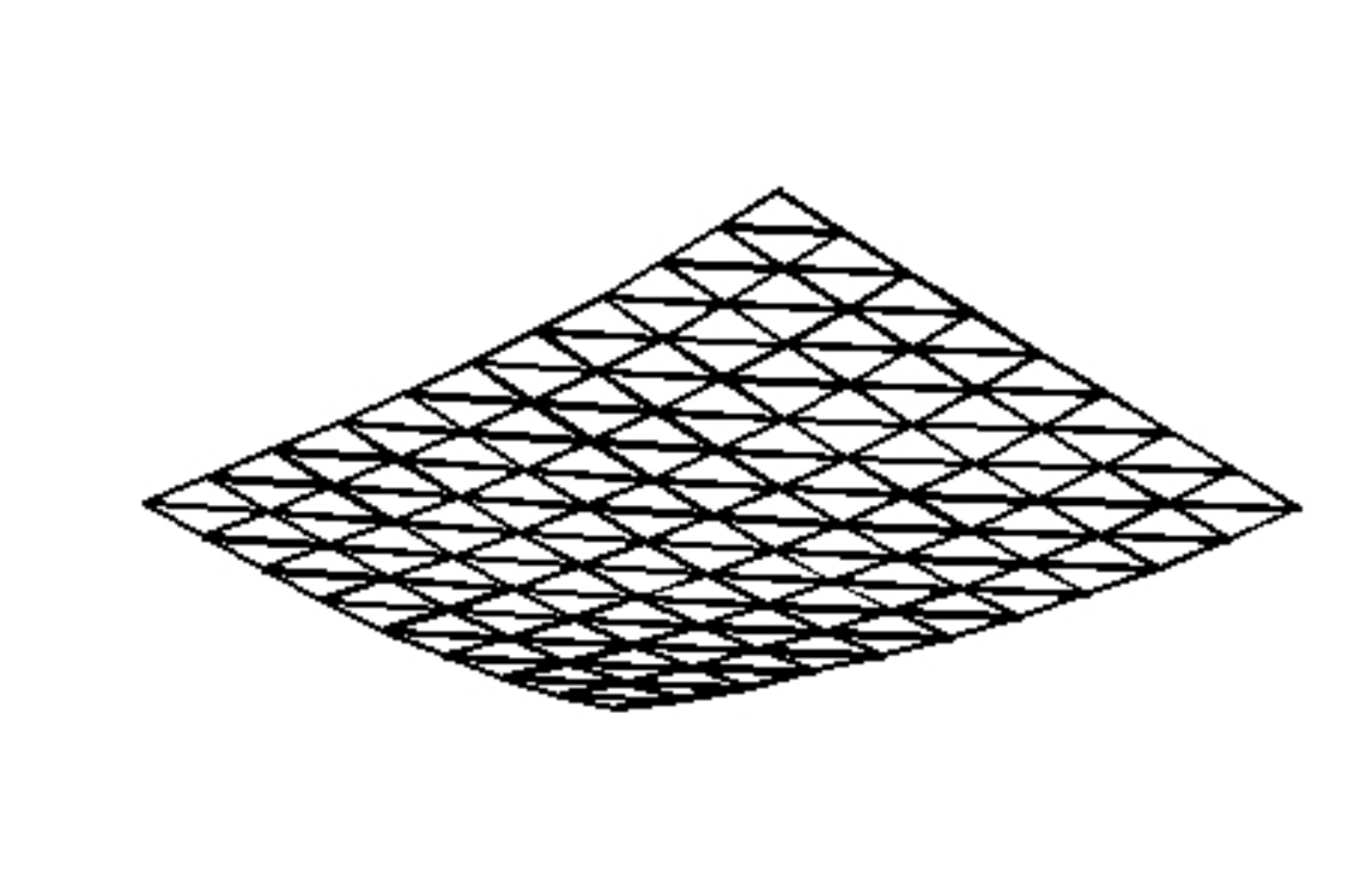}   &
\includegraphics[width=0.18\linewidth]{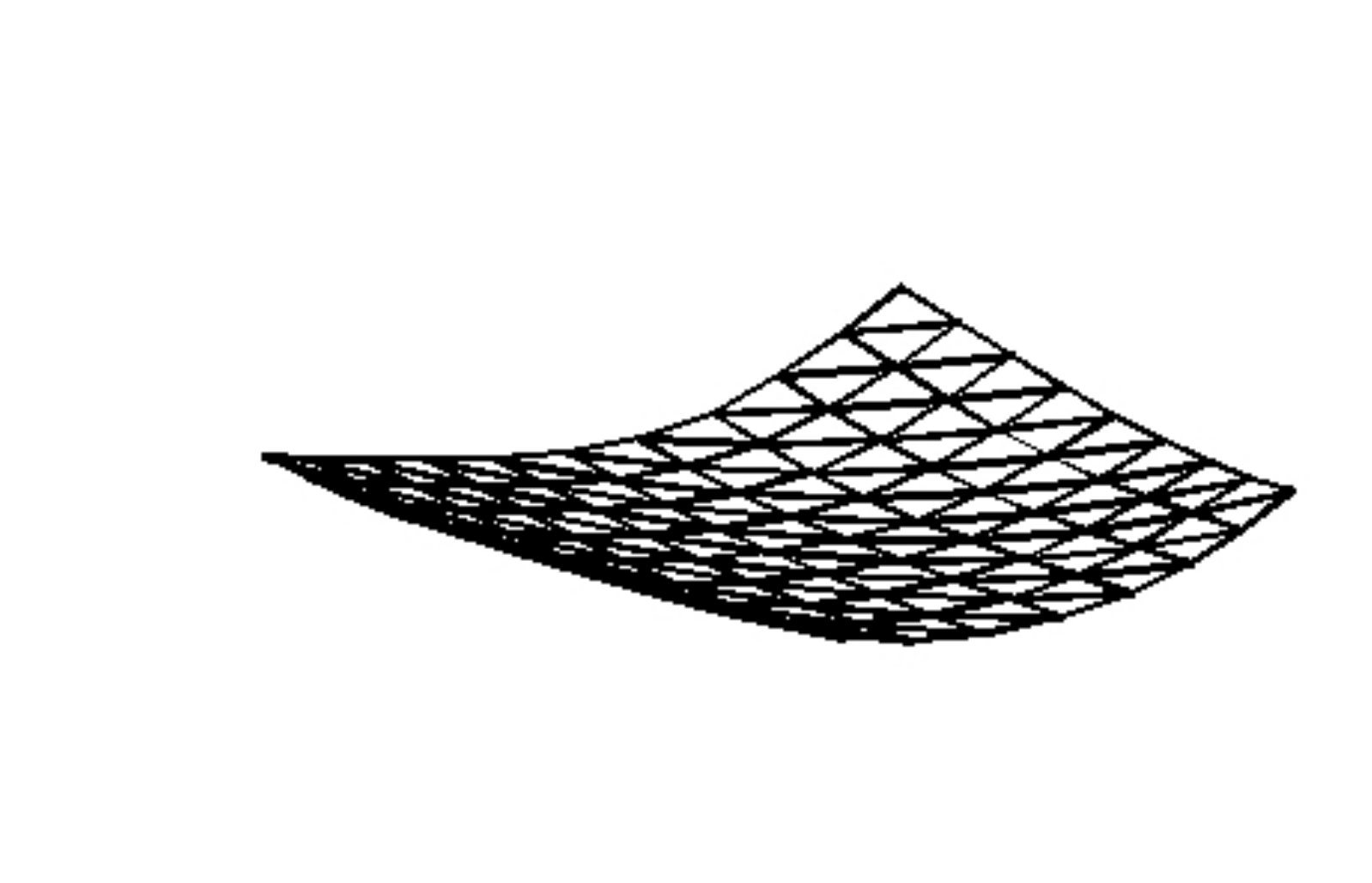}
\\
\includegraphics[width=0.18\linewidth]{figs_kinect_paper_side_000041}  &
\includegraphics[width=0.18\linewidth]{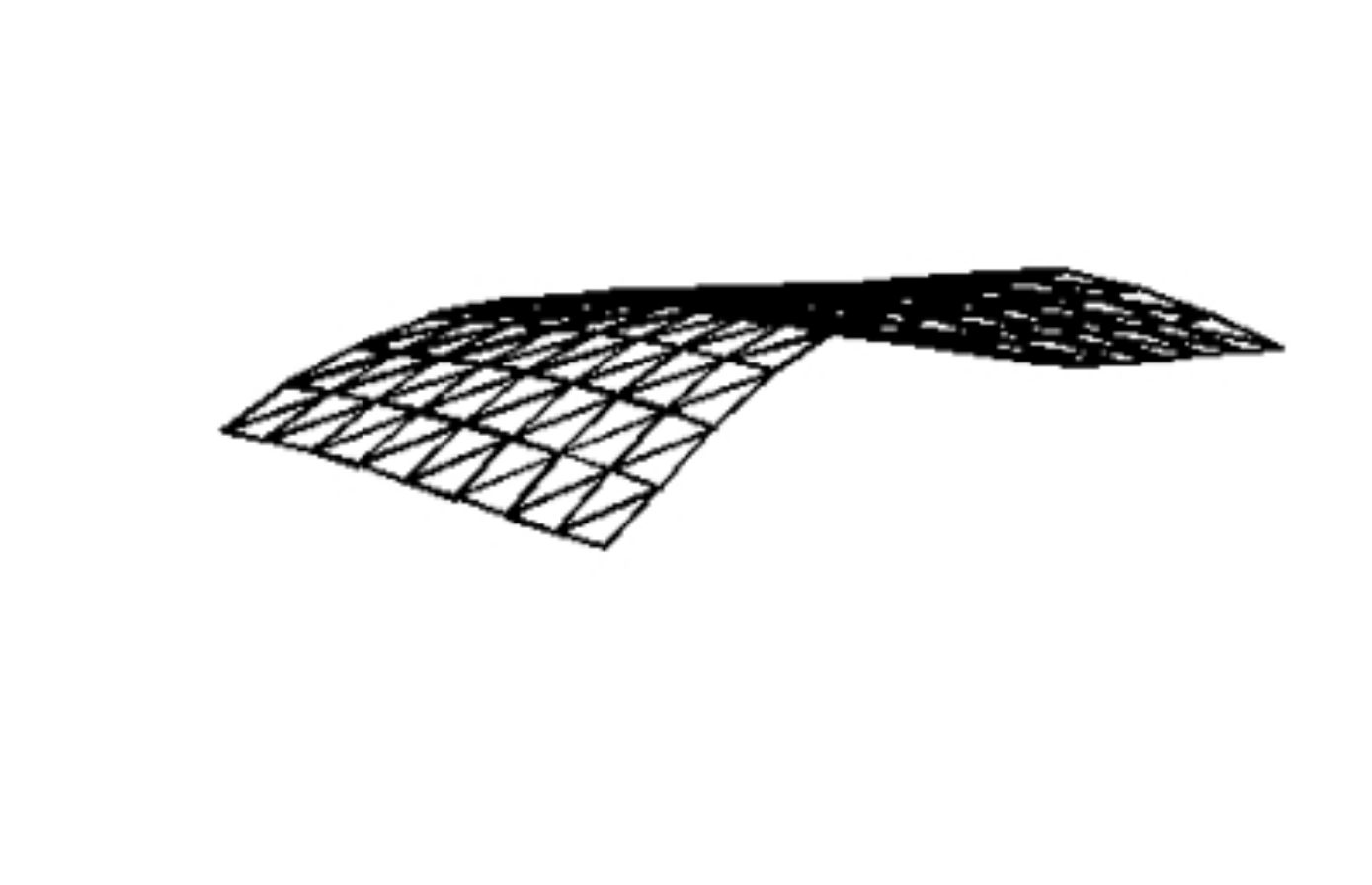}       &
\includegraphics[width=0.18\linewidth]{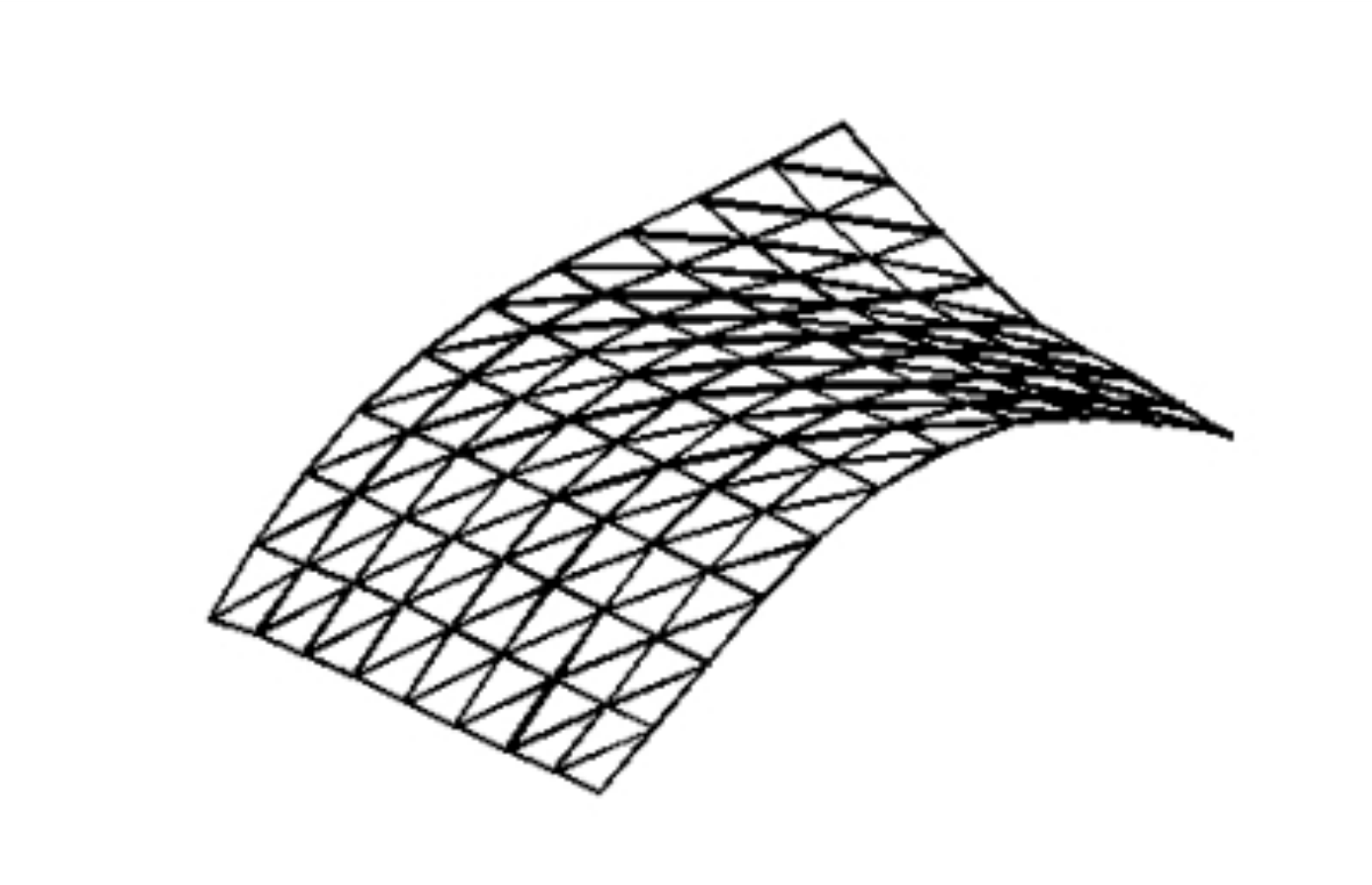}       &
\includegraphics[width=0.18\linewidth]{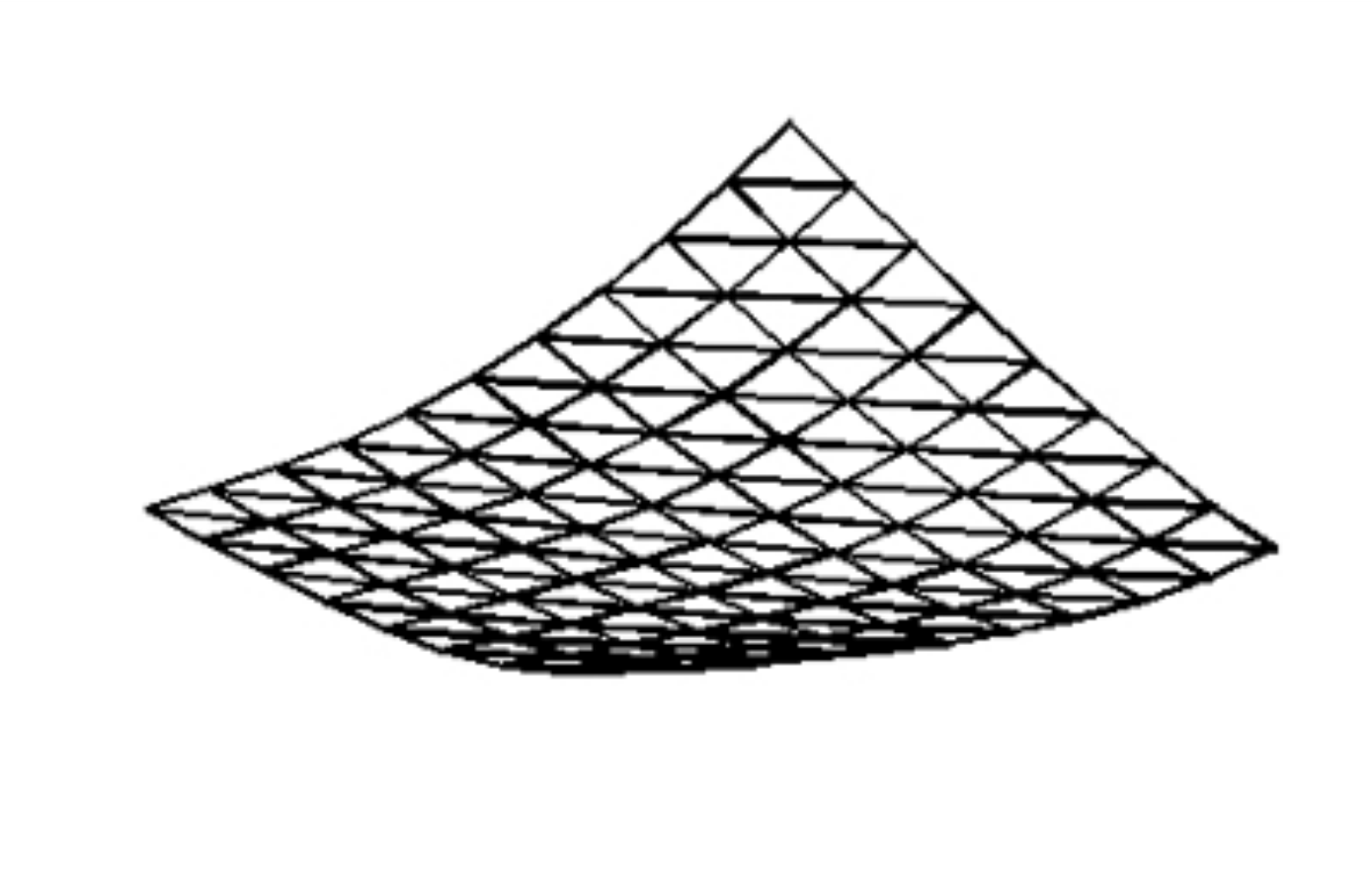}       &
\includegraphics[width=0.18\linewidth]{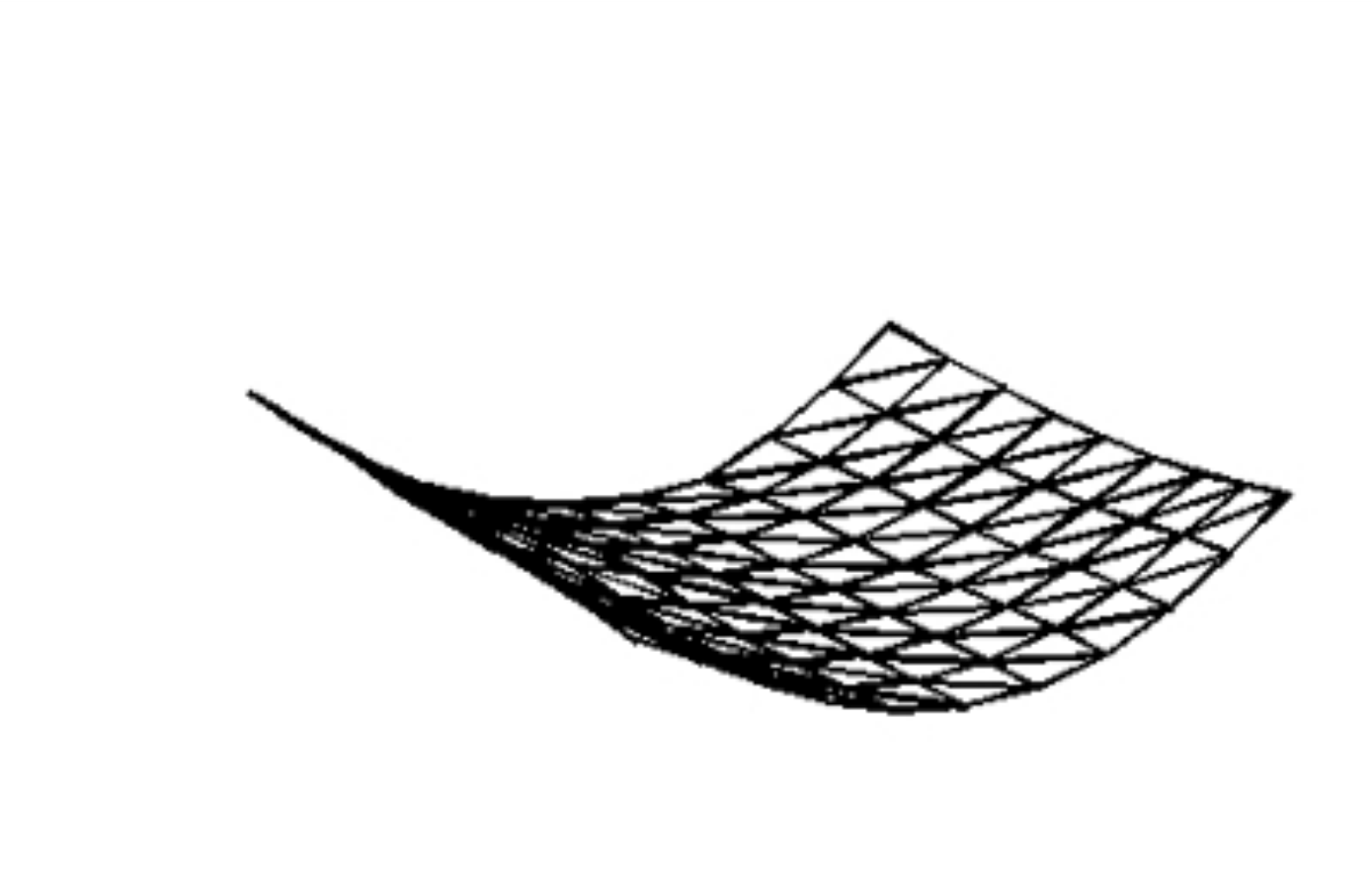}
\end{tabular}
\caption{{\bf  Additional unconstrained  and constrained  results for  the paper
    sequence.}  {\bf  Top row:}  Reprojection on the  input images of  the final
  reconstructions obtained  by solving  the constrained minimization  problem of
  Eq.~\ref{eq:ConstrOpt}.    {\bf  Middle  row:}   Intermediate  reconstructions
  obtained    by   solving   the    unconstrained   minimization    problem   of
  Eq.~\ref{eq:UnconstrOpt} and  seen from  a different view-point.   {\bf Bottom
    row:}  Final   reconstructions  seen  from   the  same  view-point   as  the
  intermediate  ones.  Note  that the  3D shapes  can be  quite  different, even
  though their image projections are very similar. }
\label{fig:kinect_paper_seq}
\end{figure*}

%% file: figs_capaperlincst_table.tex
\newcommand{\capaperlincstwidth}{0.45\linewidth}
\newcommand{\capaperlincstwidthtwo}{0.40\linewidth}
\begin{tabular}{cc}
\includegraphics[width=\capaperlincstwidth]{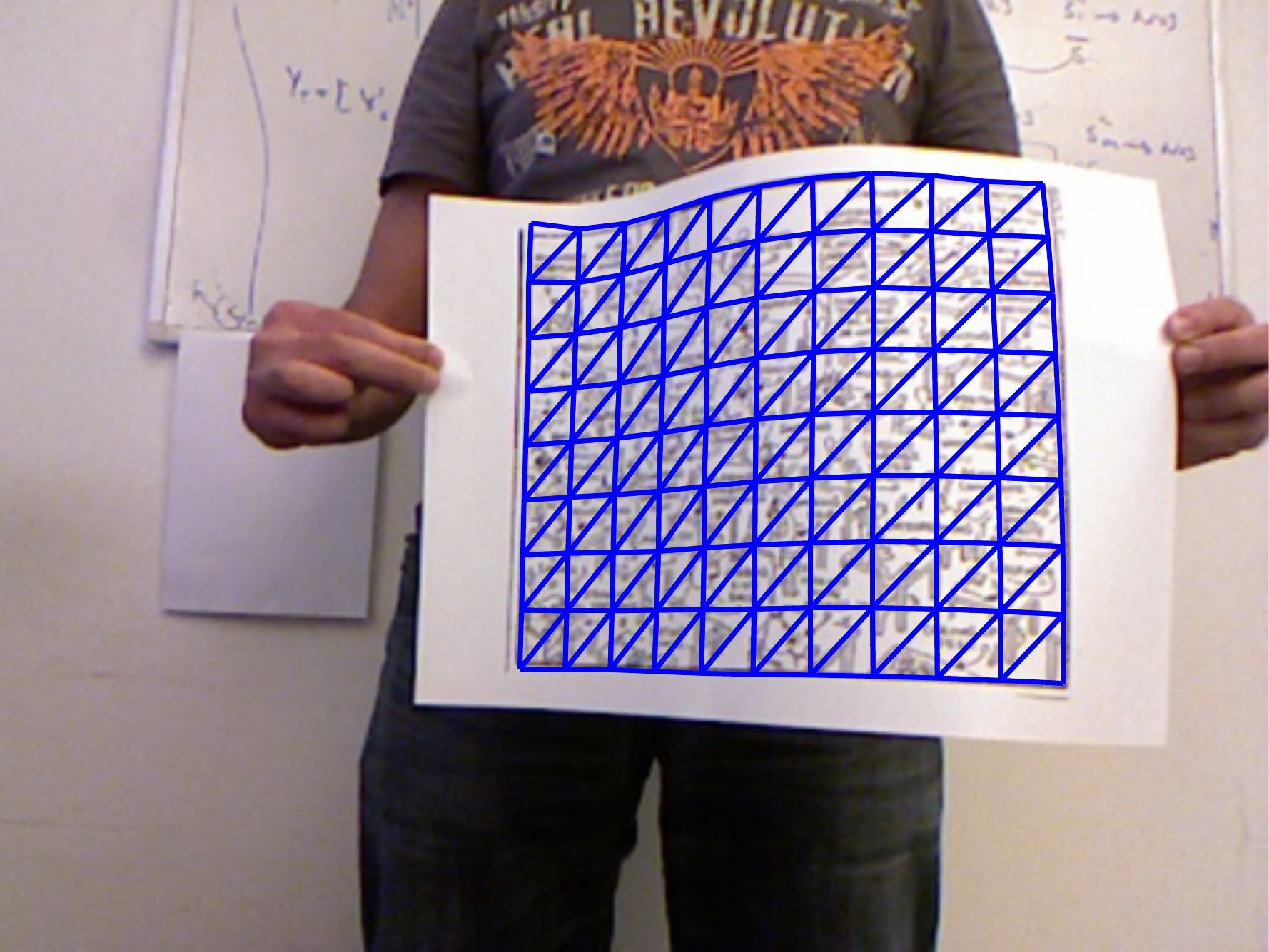} & \includegraphics[width=\capaperlincstwidth]{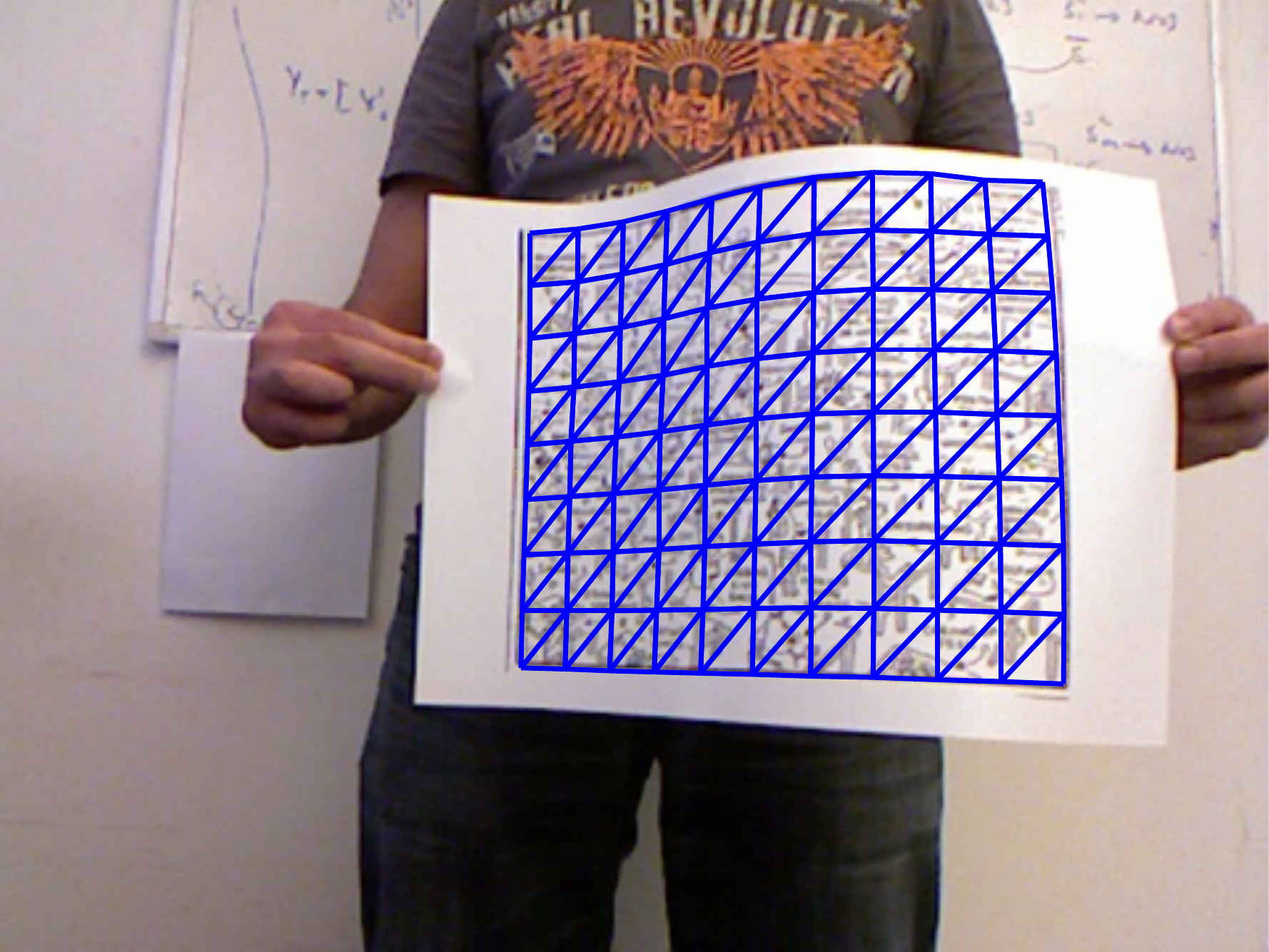}\\
\includegraphics[width=\capaperlincstwidthtwo]{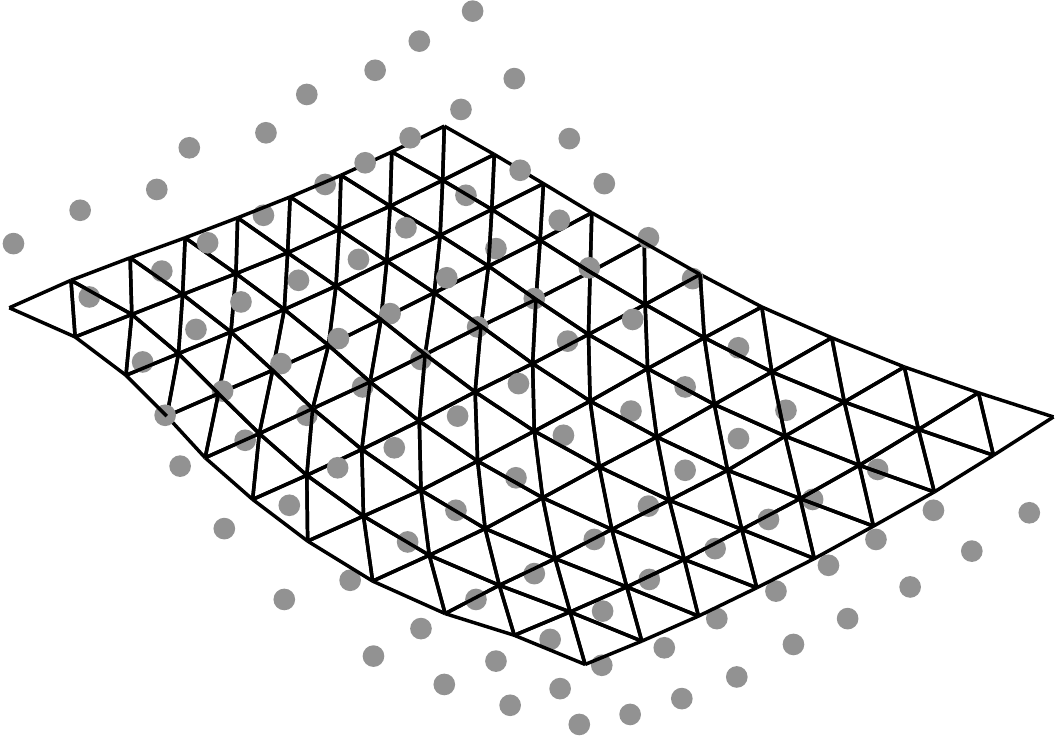} & \includegraphics[width=\capaperlincstwidthtwo]{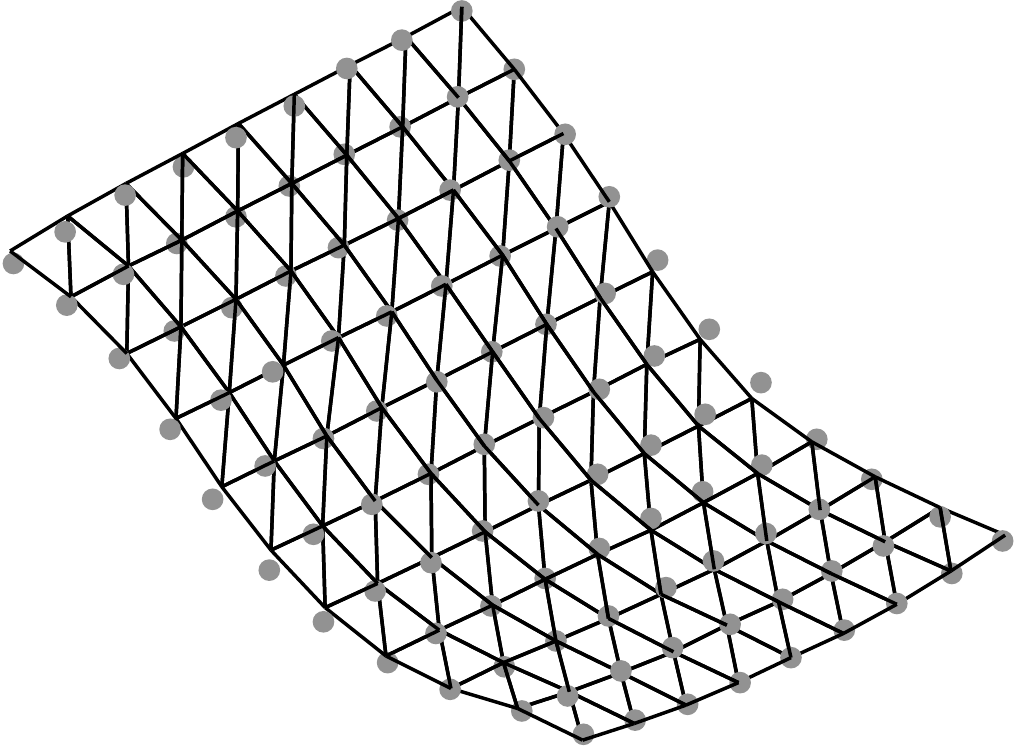}\\
(a) & (b)
\end{tabular}

%% file: figs_robustness_comparison2.tex
\newcommand{\robustcmpwinlier}{0.40\linewidth}
\newcommand{\robustcmphinlier}{0.258\linewidth}
\begin{figure*}[t]
\begin{center}
\begin{tabular}{ccc}
 \includegraphics[height=\robustcmphinlier]{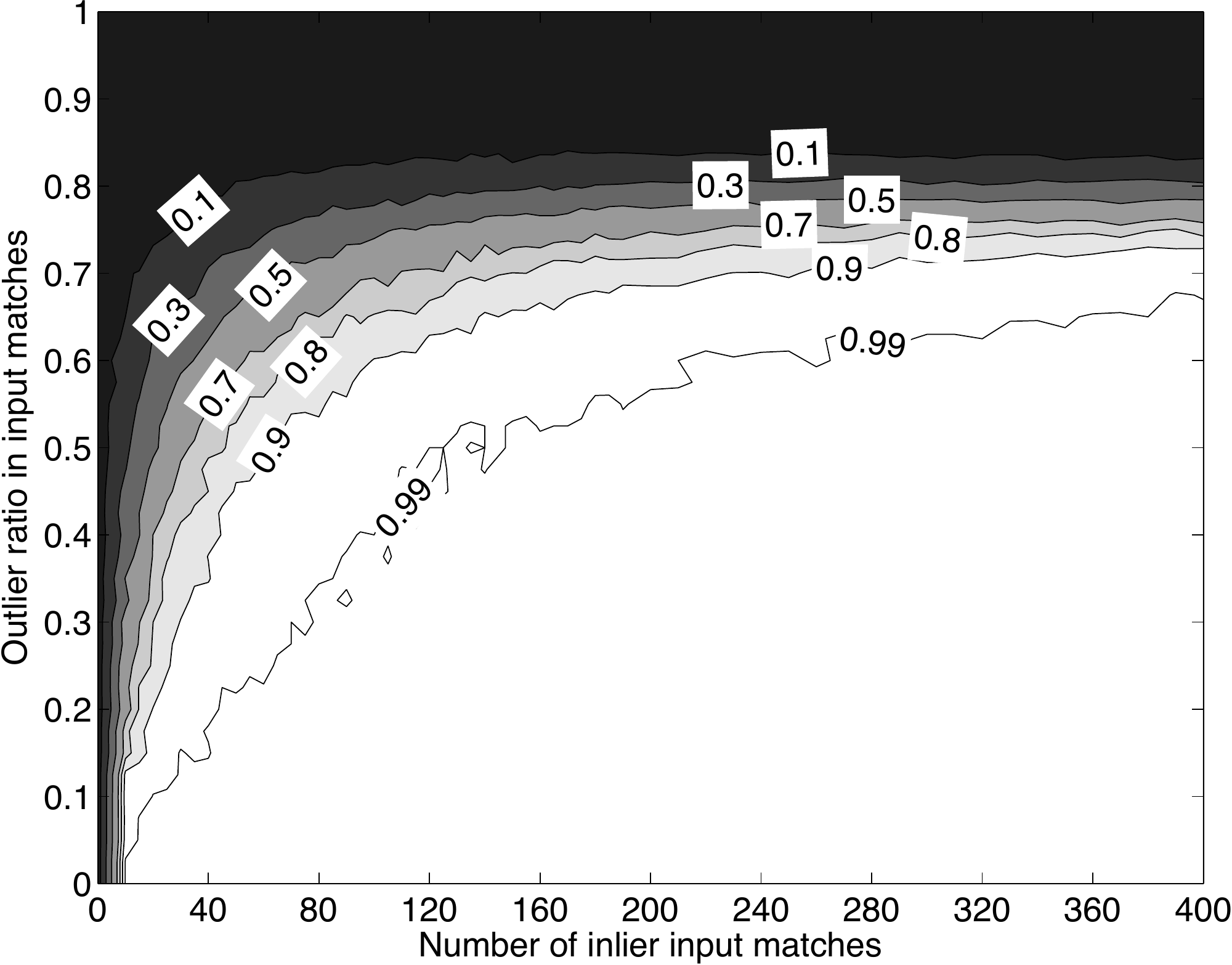} \hspace{-0.5cm} &
 \includegraphics[height=\robustcmphinlier]{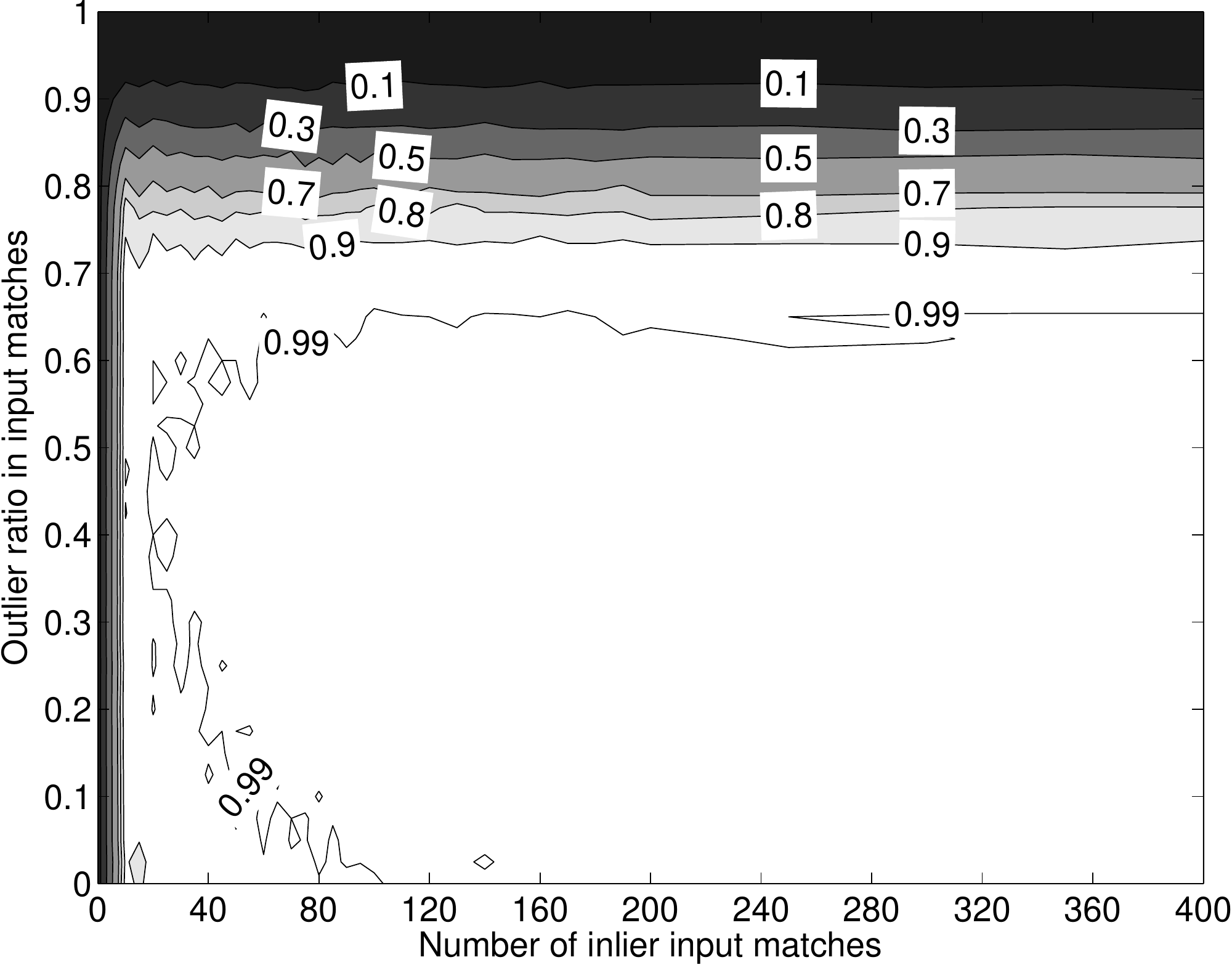} \hspace{-0.5cm} &
 \includegraphics[height=\robustcmphinlier]{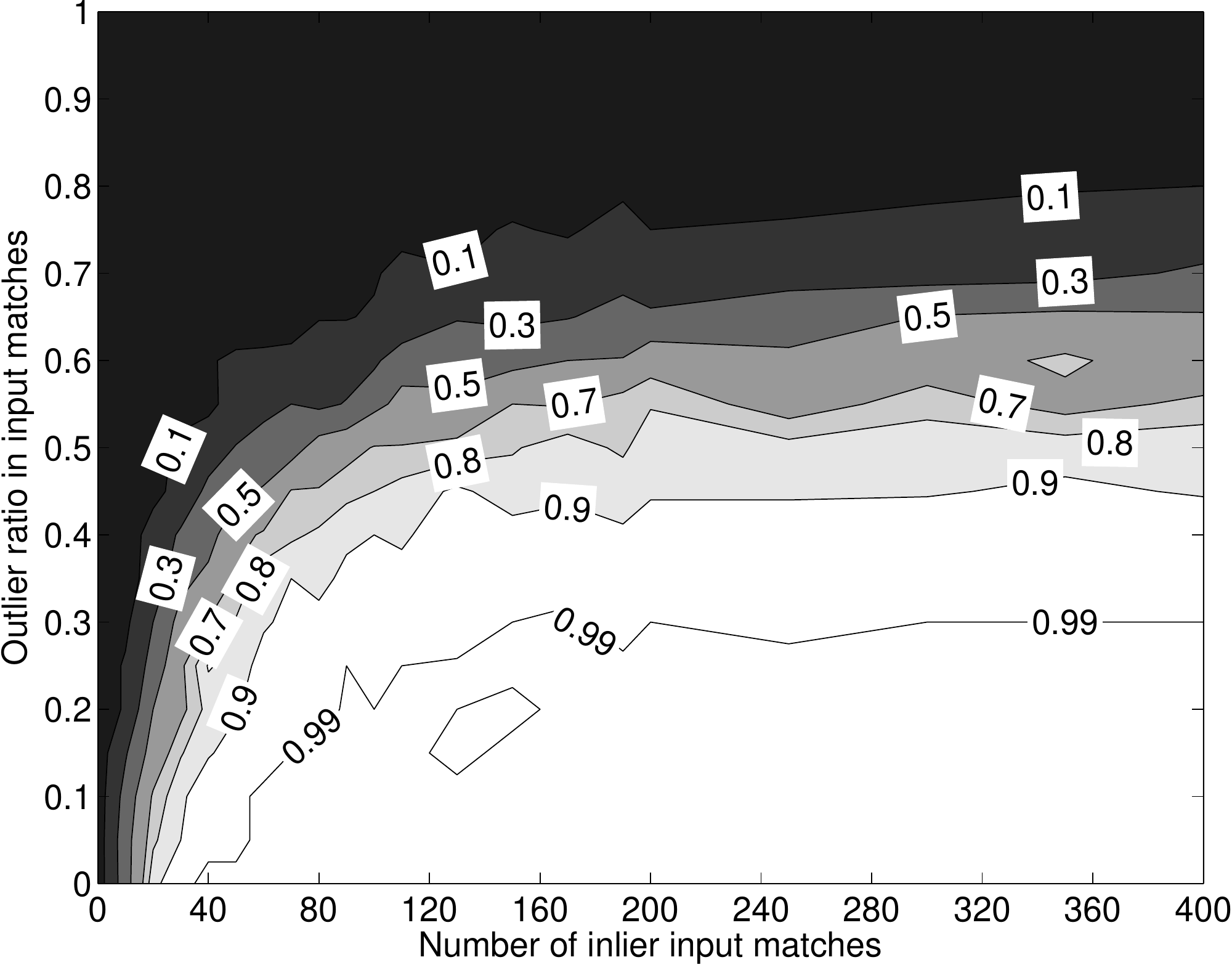} \\ 
 \includegraphics[height=\robustcmphinlier]{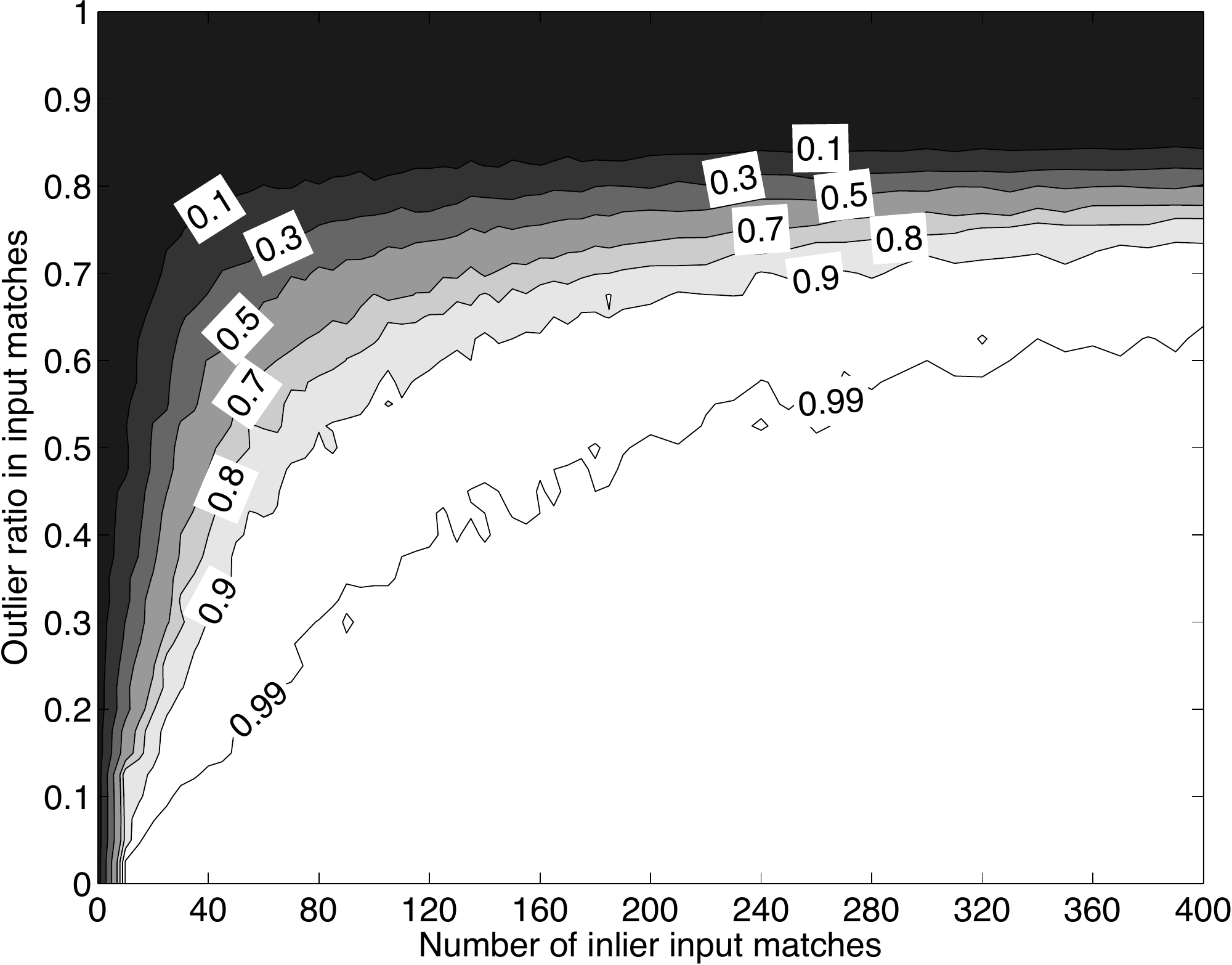} \hspace{-0.5cm} &
 \includegraphics[height=\robustcmphinlier]{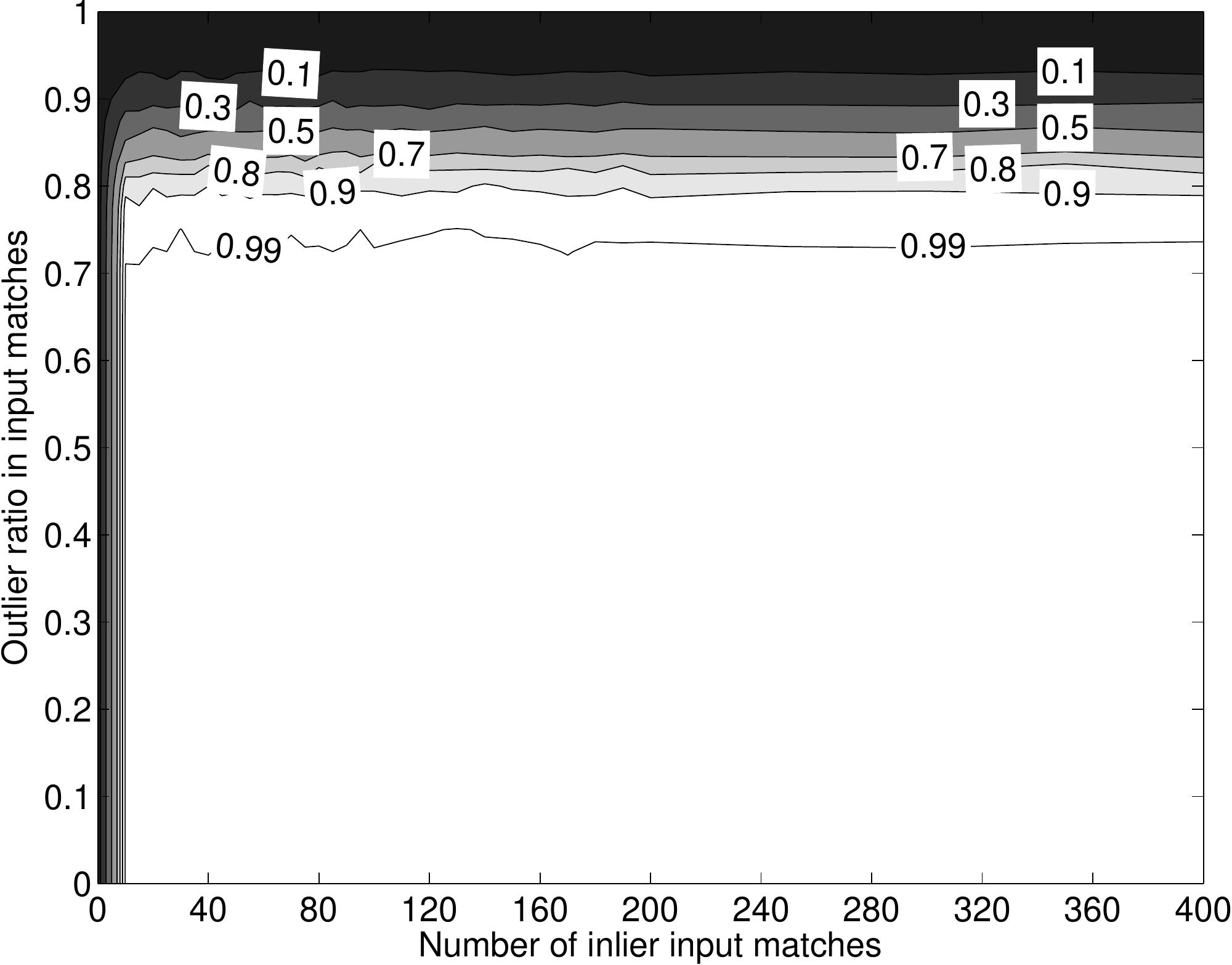} \hspace{-0.5cm} &
 \includegraphics[height=\robustcmphinlier]{figs_robustness_pizarro_cushion_mesh.pdf} \\  
 (a) Ours & (b) Tran et al. ECCV 2012 \cite{Tran12} & (c) Pizarro et al. IJCV 2012 \cite{Pizarro12} \\ 
\end{tabular}
\end{center}
\caption{Probability of success as a function of the number of inlier matches
  and proportion of outliers, on the x-axis and y-axis respectively. The lines
  are level lines of the probability of success according to the second criterion (see Sec.~\ref{sec:robust}), i.e. at least $90\%$
  of the inlier matches are correctly labeled and retrieved. {\bf Top row:} paper dataset. {\bf Bottom row:} cushion dataset.}
\label{fig:robustnesscompareinlier}
\end{figure*}

%% file: top.bbl
\begin{thebibliography}{10}
\providecommand{\url}[1]{#1}
\csname url@samestyle\endcsname
\providecommand{\newblock}{\relax}
\providecommand{\bibinfo}[2]{#2}
\providecommand{\BIBentrySTDinterwordspacing}{\spaceskip=0pt\relax}
\providecommand{\BIBentryALTinterwordstretchfactor}{4}
\providecommand{\BIBentryALTinterwordspacing}{\spaceskip=\fontdimen2\font plus
\BIBentryALTinterwordstretchfactor\fontdimen3\font minus
  \fontdimen4\font\relax}
\providecommand{\BIBforeignlanguage}[2]{{%
\expandafter\ifx\csname l@#1\endcsname\relax
\typeout{** WARNING: IEEEtran.bst: No hyphenation pattern has been}%
\typeout{** loaded for the language `#1'. Using the pattern for}%
\typeout{** the default language instead.}%
\else
\language=\csname l@#1\endcsname
\fi
#2}}
\providecommand{\BIBdecl}{\relax}
\BIBdecl

\bibitem{Bartoli12b}
A.~Bartoli, Y.~G{\'e}rard, F.~Chadebecq, and T.~Collins, ``{On Template-Based
  Reconstruction from a Single View: Analytical Solutions and Proofs of
  Well-Posedness for Developable, Isometric and Conformal Surfaces},'' in
  \emph{Conference on Computer Vision and Pattern Recognition}, 2012.

\bibitem{Salzmann10b}
M.~Salzmann and P.~Fua, \emph{{Deformable Surface 3D Reconstruction from
  Monocular Images}}.\hskip 1em plus 0.5em minus 0.4em\relax Morgan-Claypool,
  2010.

\bibitem{Blanz99}
V.~Blanz and T.~Vetter, ``{A Morphable Model for the Synthesis of 3D Faces},''
  in \emph{ACM SIGGRAPH}, August 1999, pp. 187--194.

\bibitem{Salzmann11a}
M.~Salzmann and P.~Fua, ``{Linear Local Models for Monocular Reconstruction of
  Deformable Surfaces},'' \emph{IEEE Transactions on Pattern Analysis and
  Machine Intelligence}, vol.~33, no.~5, pp. 931--944, 2011.

\bibitem{Gumerov04}
N.~Gumerov, A.~Zandifar, R.~Duraiswami, and L.~Davis, ``{Structure of
  Applicable Surfaces from Single Views},'' in \emph{European Conference on
  Computer Vision}, May 2004.

\bibitem{Liang05}
J.~Liang, D.~Dementhon, and D.~Doermann, ``{Flattening Curved Documents in
  Images},'' in \emph{Conference on Computer Vision and Pattern Recognition},
  2005, pp. 338--345.

\bibitem{Perriollat12}
M.~Perriollat and A.~Bartoli, ``{A Computational Model of Bounded Developable
  Surfaces with Application to Image-Based Three-Dimensional Reconstruction},''
  \emph{Computer Animation and Virtual Worlds}, vol.~24, no.~5, 2013.

\bibitem{Ecker08}
A.~Ecker, A.~Jepson, and K.~Kutulakos, ``{Semidefinite Programming Heuristics
  for Surface Reconstruction Ambiguities},'' in \emph{European Conference on
  Computer Vision}, October 2008.

\bibitem{Shen09}
S.~Shen, W.~Shi, and Y.~Liu, ``{Monocular 3D Tracking of Inextensible
  Deformable Surfaces Under L2-Norm},'' in \emph{Asian Conference on Computer
  Vision}, 2009.

\bibitem{Brunet10}
F.~Brunet, R.~Hartley, A.~Bartoli, N.~Navab, and R.~Malgouyres, ``{Monocular
  Template-Based Reconstruction of Smooth and Inextensible Surfaces},'' in
  \emph{Asian Conference on Computer Vision}, 2010.

\bibitem{White06b}
R.~White and D.~Forsyth, ``{Combining Cues: Shape from Shading and Texture},''
  in \emph{Conference on Computer Vision and Pattern Recognition}, 2006.

\bibitem{Moreno09b}
F.~Moreno-noguer, M.~Salzmann, V.~Lepetit, and P.~Fua, ``{Capturing 3D
  Stretchable Surfaces from Single Images in Closed Form},'' in
  \emph{Conference on Computer Vision and Pattern Recognition}, June 2009.

\bibitem{Malti13}
A.~Malti, R.~Hartley, A.~Bartoli, and J.-H. Kim, ``{Monocular Template-Based 3D
  Reconstruction of Extensible Surfaces with Local Linear Elasticity},'' in
  \emph{Conference on Computer Vision and Pattern Recognition}, 2013.

\bibitem{Moreno10}
F.~Moreno-noguer, J.~Porta, and P.~Fua, ``{Exploring Ambiguities for Monocular
  Non-Rigid Shape Estimation},'' in \emph{European Conference on Computer
  Vision}, September 2010.

\bibitem{Perriollat11}
M.~Perriollat, R.~Hartley, and A.~Bartoli, ``{Monocular Template-Based
  Reconstruction of Inextensible Surfaces},'' \emph{International Journal of
  Computer Vision}, vol.~95, 2011.

\bibitem{Salzmann08b}
M.~Salzmann, F.~Moreno-noguer, V.~Lepetit, and P.~Fua, ``{Closed-Form Solution
  to Non-Rigid 3D Surface Registration},'' in \emph{European Conference on
  Computer Vision}, October 2008.

\bibitem{Sorkine04}
O.~Sorkine, D.~Cohen-Or, Y.~Lipman, M.~Alexa, C.~R\"{o}ssl, and H.-P. Seidel,
  ``{Laplacian Surface Editing},'' in \emph{Symposium on Geometry Processing},
  2004, pp. 175--184.

\bibitem{Summer04}
R.~Sumner and J.~Popovic, ``{Deformation Transfer for Triangle Meshes},''
  \emph{ACM Transactions on Graphics}, pp. 399--405, 2004.

\bibitem{Botsch06}
M.~Botsch, R.~Sumner, M.~Pauly, and M.~Gross, ``{Deformation Transfer for
  Detail-Preserving Surface Editing},'' in \emph{Vision Modeling and
  Visualization}, 2006, pp. 357--364.

\bibitem{Ostlund12}
J.~Ostlund, A.~Varol, D.~Ngo, and P.~Fua, ``{Laplacian Meshes for Monocular 3D
  Shape Recovery},'' in \emph{European Conference on Computer Vision}, 2012.

\bibitem{Fayad10}
J.~Fayad, L.~Agapito, and A.~Delbue, ``{Piecewise Quadratic Reconstruction of
  Non-Rigid Surfaces from Monocular Sequences},'' in \emph{European Conference
  on Computer Vision}, 2010.

\bibitem{Garg13}
R.~Garg, A.~Roussos, and L.~Agapito, ``{Dense Variational Reconstruction of
  Non-Rigid Surfaces from Monocular Video},'' in \emph{Conference on Computer
  Vision and Pattern Recognition}, 2013.

\bibitem{Salzmann07b}
M.~Salzmann, R.~Hartley, and P.~Fua, ``{Convex Optimization for Deformable
  Surface 3D Tracking},'' in \emph{International Conference on Computer
  Vision}, October 2007.

\bibitem{Alcantarilla12}
P.~Alcantarilla, P.~Fern{\'a}ndez, and A.~Bartoli, ``{Deformable 3D
  Reconstruction with an Object Database},'' in \emph{British Machine Vision
  Conference}, 2012.

\bibitem{Cootes98}
T.~Cootes, G.~Edwards, and C.~Taylor, ``{Active Appearance Models},'' in
  \emph{European Conference on Computer Vision}, June 1998, pp. 484--498.

\bibitem{Dimitrijevic04}
M.~Dimitrijevi\'{c}, S.~Ili\'{c}, and P.~Fua, ``{Accurate Face Models from
  Uncalibrated and Ill-Lit Video Sequences},'' in \emph{Conference on Computer
  Vision and Pattern Recognition}, June 2004.

\bibitem{DelBue11}
A.~Delbue and A.~Bartoli, ``{Multiview 3D Warps},'' in \emph{International
  Conference on Computer Vision}, 2011.

\bibitem{Ilic02}
S.~Ili\'{c} and P.~Fua, ``{Using Dirichlet Free Form Deformation to Fit
  Deformable Models to Noisy 3D Data},'' in \emph{European Conference on
  Computer Vision}, May 2002.

\bibitem{Jacobson11}
A.~Jacobson, I.~Baran, J.~Popovic, and O.~Sorkine, ``{Bounded Biharmonic
  Weights for Real-Time Deformation},'' in \emph{ACM SIGGRAPH}, 2011.

\bibitem{Fua10}
P.~Fua, A.~Varol, R.~Urtasun, and M.~Salzmann, ``{Least-Squares Minimization
  Under Constraints},'' EPFL, Tech. Rep., 2010.

\bibitem{Lowe04}
D.~Lowe, ``{Distinctive Image Features from Scale-Invariant Keypoints},''
  \emph{International Journal of Computer Vision}, vol.~20, no.~2, 2004.

\bibitem{Ozuysal10}
M.~Ozuysal, M.~Calonder, V.~Lepetit, and P.~Fua, ``{Fast Keypoint Recognition
  Using Random Ferns},'' \emph{IEEE Transactions on Pattern Analysis and
  Machine Intelligence}, vol.~32, no.~3, pp. 448--461, 2010.

\bibitem{Tran12}
Q.-H. Tran, T.-J. Chin, G.~Carneiro, M.~S. Brown, and D.~Suter, ``In defence of
  ransac for outlier rejection in deformable registration,'' in \emph{European
  Conference on Computer Vision}.\hskip 1em plus 0.5em minus 0.4em\relax
  Springer, 2012, pp. 274--287.

\bibitem{Pizarro12}
D.~Pizarro and A.~Bartoli, ``{Feature-Based Deformable Surface Detection with
  Self-Occlusion Reasoning},'' \emph{International Journal of Computer Vision},
  March 2012.

\bibitem{Zhu07}
J.~Zhu and M.~Lyu, ``{Progressive Finit Newton Approach to Real-Time Nonrigid
  Surface Detection},'' in \emph{International Conference on Computer Vision},
  October 2007.

\bibitem{Pilet08a}
J.~Pilet, V.~Lepetit, and P.~Fua, ``{Fast Non-Rigid Surface Detection,
  Registration and Realistic Augmentation},'' \emph{International Journal of
  Computer Vision}, vol.~76, no.~2, February 2008.

\bibitem{Collins14b}
T.~Collins and A.~Bartoli, ``Using isometry to classify correct/incorrect
  {3D-2D} correspondences,'' in \emph{European Conference on Computer Vision},
  2014, pp. 325--340.

\bibitem{Furukawa09b}
Y.~Furukawa and J.~Ponce, ``{Accurate, Dense, and Robust Multi-View
  Stereopsis},'' \emph{IEEE Transactions on Pattern Analysis and Machine
  Intelligence}, vol.~99, 2009.

\bibitem{Torresani08a}
L.~Torresani, A.~Hertzmann, and C.~Bregler, ``{Nonrigid Structure-From-Motion:
  Estimating Shape and Motion with Hierarchical Priors},'' \emph{IEEE
  Transactions on Pattern Analysis and Machine Intelligence}, vol.~30, no.~5,
  pp. 878--892, 2008.

\bibitem{Russell11}
C.~Russell, J.~Fayad, and L.~Agapito, ``{Energy Based Multiple Model Fitting
  for Non-Rigid Structure from Motion},'' in \emph{Conference on Computer
  Vision and Pattern Recognition}, 2011.

\end{thebibliography}
